\documentclass[a4paper,12pt,twoside,openright,figuresright,dvipsnames,table,x11name]{report}
\usepackage[left=3cm,right=3cm,top=3cm,bottom=3cm]{geometry}
\usepackage[dvipsnames]{xcolor}
\usepackage[T1]{fontenc}
\usepackage{tgbonum}
\usepackage[english]{babel} 
\usepackage[english=american]{csquotes} 

\PassOptionsToPackage{hyphens}{url}\usepackage{hyperref}

\usepackage{hyperref}
\usepackage{multirow}
\usepackage[utf8]{inputenc}

\usepackage{graphicx}
\usepackage{verbatim}
\usepackage{latexsym}
\usepackage{mathchars}
\usepackage{setspace}
\usepackage[toc]{appendix}
\usepackage{lmodern}
\usepackage[explicit]{titlesec}
\usepackage{microtype}
\usepackage{tikz}
\usepackage[dmyyyy]{datetime}
\usepackage{setspace}

\usepackage[hang,flushmargin]{footmisc}

\input{blocked.sty}
\input{uhead.sty}
\input{boxit.sty}
\input{thesis.sty}

\newcommand{\NN}{{\sf I\kern-0.14emN}}   
\newcommand{\ZZ}{{\sf Z\kern-0.45emZ}}   
\newcommand{\QQQ}{{\sf C\kern-0.48emQ}}   
\newcommand{\RR}{{\sf I\kern-0.14emR}}   

\newcommand{\normallinespacing}{\renewcommand{\baselinestretch}{1.5} \normalsize}

\newcommand{\syncc}{~\stackrel{\textstyle \rhd\kern-0.57em\lhd}{\scriptstyle L}~}

\titleformat%
{\chapter}[hang]%
{\bfseries}{%
	\begin{minipage}[t]{0.25\linewidth}  
		\vspace{0pt}
		\begin{tikzpicture}
		\node[
		outer sep=0pt,
		text width=2.5cm,
		minimum height=2.5cm,
		fill=black,
		font=\color{white}\fontsize{80}{90}\selectfont,
		align=center
		] (num) {\thechapter};
		\node[
		outer sep=0pt,
		inner sep=0pt,
		anchor=south,
		font=\color{black}\Large\normalfont
		] at ([yshift=5pt]num.north) {\textls[180]{\textsc{\chaptertitlename}}};
		\end{tikzpicture}  
	\end{minipage}%
}
{0pt}%
{%
	\begin{minipage}[t]{.75\linewidth}%
		\vspace{2pt} 
		\rule{\linewidth}{2.5pt}\\\vskip -1.75\baselineskip%
			\rule{\linewidth}{.7pt}\vskip 5pt
		{\LARGE\raggedright\textsf{#1}}
	\end{minipage}%
}

\titleformat%
{name=\chapter,numberless}[hang]%
{\bfseries}{%
}
{0pt}%
{%
	\begin{minipage}[t]{1.0\linewidth}%
		{\LARGE\raggedright\textsf{#1}}
	\end{minipage}%
}

\makeatletter
\newcommand{\address}[1]{\newcommand{\@address}{#1}}
\newcommand{\institute}[1]{\newcommand{\@institute}{#1}}

\renewcommand{\maketitle}{
	\begin{titlepage}
		{\flushright
			\includegraphics[width=0.3\textwidth]{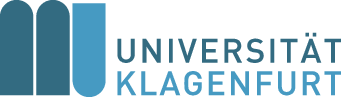}\par\vspace{0.1cm}
		}
		
		\noindent\rule{\textwidth}{1pt}%
		\vspace{0.2\baselineskip}
		{%
			\LARGE%
                                \begin{spacing}{1.1}
			\@title%
                                \end{spacing}
		}
		\vspace{0.2\baselineskip}
		\noindent\rule{\textwidth}{1pt}

                      \vskip 2em

                      \centering
		{\Large\noindent\ignorespaces\@author\par}\vskip 1.5em

		\vskip 2.5em
		
		{\LARGE\noindent\ignorespaces \textsc{Dissertation}\par}\vskip 1.5em
		{\large\noindent\ignorespaces zur Erlangung des akademischen Grades\\Doktor der Technischen Wissenschaften\par}
		
		\vskip 2em
		
		{\Large{\vspace{0.5\baselineskip}Doktoratsstudium der Technischen Wissenschaften\\im Dissertationsgebiet Informatik}\par}
		
		\vskip 2em
		
		{\large\noindent\ignorespaces\@institute\par}
		
		\vskip 3em
		
		\begin{minipage}{0.54\textwidth}
			\flushleft\large
			\textsc{Erstbetreuer}\\
			\normalsize
			Assoc. Prof. Dr. Klaus Schöffmann\\
                                \footnotesize
			Institut für Informationstechnologie\\
			Alpen-Adria-Universität Klagenfurt
		\end{minipage}
		\begin{minipage}{0.45\textwidth}
			\flushright\large
			\textsc{Zweitbetreuer}\\
			\normalsize
			Assoc. Prof. Dr. Christian Timmerer\\
                                \footnotesize
			Institut für Informationstechnologie\\
			Alpen-Adria-Universität Klagenfurt
		\end{minipage}
                     \vskip 2em
                     \begin{minipage}{0.53\textwidth}
			\flushleft\large
			\textsc{Erstgutachter}\\
			\normalsize
			Prof. Dr. Henning Müller\\
                                \footnotesize
                                Institute of Information Systems\\
			University of Applied Sciences Western Switzerland\\
                                Department of Radiology and Medical Informatics\\
			University of Geneva
		\end{minipage}
                     \begin{minipage}{0.46\textwidth}
			\flushright\large
			\textsc{Zweitgutachter}\\
			\normalsize
			Prof. Dr. Raphael Sznitman\\
                                \footnotesize
			ARTORG Center for Biomedical Engineering\\
			Faculty of Medicine\\
                                University of Bern
		\end{minipage}
		
		\vskip 2em
		
		{\normalsize\noindent\ignorespaces\@address, \@date}
	\end{titlepage}
}
\makeatother

\newcommand{\chapterintro}[1]{
	\begin{minipage}{\columnwidth}
		\begin{minipage}{0.15\columnwidth}
			~
		\end{minipage}
		\begin{minipage}{0.85\columnwidth}
			\vspace{-2em}
			\noindent\rule{\columnwidth}{0.5pt}
			\vspace{-2.1em}
			\singlespacing\noindent\textit{Chapter overview~---}
			#1
			\noindent\rule{\columnwidth}{0.5pt}
		\end{minipage}
	\end{minipage}
}

\usepackage{tabularx,booktabs}

\newcolumntype{Y}{>{\centering\arraybackslash}X}
\newcolumntype{R}{>{\raggedleft\arraybackslash}X}
\usepackage{arydshln}
\usepackage{tabu}
\let\olddegree\degree  
\let\degree\relax      
\usepackage{gensymb}
\let\degree\olddegree  
\usepackage{ctable}
\usepackage{enumerate,multicol}
\usepackage{adjustbox}
\usepackage{multirow}
\usepackage{multicol}
\usepackage[graphicx]{realboxes}
\usepackage{enumitem}
\usepackage{amsmath,array,graphicx}
\usepackage{algorithm}
\usepackage{algorithmic}
\usepackage{float}
\usepackage{caption}
\usepackage{subcaption}
\usepackage{balance}
\usepackage{xspace}
\usepackage[]{rotating}

\newcolumntype{P}[1]{>{\centering\arraybackslash}p{#1}}
\newcommand{\ie}{\textit{i.e.}, }
\newcommand{\eg}{\textit{e.g.}, }
\newcommand{\etc}{\emph{etc.\xspace}}
\newcommand{\etal}{\emph{et al.\xspace}}
\newcommand{\HEVC}{\emph{High Efficiency Video Coding }}
\definecolor{shadecolor}{rgb}{0.8,0.8,0.8}
\usepackage{pgfplots}
\usepackage{pgfplotstable}
\usetikzlibrary{pgfplots.groupplots}
\usepgfplotslibrary{colorbrewer}
\pgfplotsset{/pgfplots/error bars/error bar style={black,thick}}

\pgfplotsset{compat=newest,
        /pgfplots/ybar legend/.style={
        /pgfplots/legend image code/.code={%
        \draw[##1,/tikz/.cd,bar width=3pt,yshift=-0.2em,bar shift=0pt]
                plot coordinates {(0cm,0.8em)};},
},}
\usepackage{amssymb}
\usepackage{xfrac}
\usepackage{pifont}
\newcommand{\xmark}{\ding{55}}
\newcommand{\avsum}{\mathop{\mathpalette\avsuminner\relax}\displaylimits}
\makeatletter
\newcommand\avsuminner[2]{%
  {\sbox0{$\m@th#1\sum$}%
   \vphantom{\usebox0}%
   \ooalign{%
     \hidewidth
     \smash{\vrule height\dimexpr\ht0+1pt\relax depth\dimexpr\dp0+1pt\relax}%
     \hidewidth\cr
     $\m@th#1\sum$\cr
   }%
  }%
}

\newenvironment{customlegend}[1][]{%
    \begingroup
    \csname pgfplots@init@cleared@structures\endcsname
    \pgfplotsset{#1}%
}{%
    \csname pgfplots@createlegend\endcsname
    \endgroup
}%

\usepackage{scalerel}
\usepackage{stackengine}
\newcounter{iloop}
\newcommand\openbigstar[1][0.7]{%
  \scalerel*{%
    \stackinset{c}{-.125pt}{c}{}{\scalebox{#1}{\color{white}{$\bigstar$}}}{%
      $\bigstar$}%
  }{\bigstar}
}
\newcommand{\Stars}[1]{\ensuremath{%
\pgfmathtruncatemacro{\imax}{ifthenelse(int(#1)==#1,#1-1,#1)}%
\pgfmathsetmacro{\xrest}{0.9*(1-#1+\imax)}%
\setcounter{iloop}{0}%
\loop\stepcounter{iloop}\ifnum\value{iloop}<\the\numexpr\imax+1
\bigstar\repeat
\openbigstar[\xrest]%
\setcounter{iloop}{0}%
\loop\stepcounter{iloop}\ifnum\value{iloop}<\the\numexpr5-\imax\relax
\openbigstar[.9]\repeat}}
\usepackage{hyperref}
\hypersetup{
    colorlinks,
    linkcolor={blue!40!black},
    citecolor={green!40!black},
    urlcolor={black}
}

\begin{document}


\title{Deep-Learning-Assisted Analysis \\ of Cataract Surgery Videos}
\author{\textbf{Negin Ghamsarian}}
\institute{
	Alpen-Adria-Universität Klagenfurt\\
	Fakultät für Technische Wissenschaften
	
}
\address{Klagenfurt am Wörthersee}
\date{27.9.2021}

\normallinespacing
\maketitle
\thispagestyle{empty}

\preface


\begin{minipage}{\columnwidth}
\begin{minipage}{0.49\columnwidth}
	Negin Ghamsarian
\end{minipage} 
\begin{minipage}{0.49\columnwidth}
	\flushright Klagenfurt am Wörthersee, 27.9.2021
\end{minipage}
\end{minipage}
	
\chapter*{Eidesstattliche Erklärung}

{
	\singlespacing

	Ich versichere an Eides statt, dass ich
	
	\begin{itemize}
		\item[--] die eingereichte wissenschaftliche Arbeit selbstständig verfasst und keine anderen als die ange­gebenen Hilfsmittel benutzt habe,
		
		\item[--] die während des Arbeitsvorganges von dritter Seite erfahrene Unterstützung, ein­schließ­lich signifikanter Betreuungshinweise, vollständig offengelegt habe,
		
		\item[--] die Inhalte, die ich aus Werken Dritter oder eigenen Werken wortwörtlich oder sinn­gemäß übernommen habe, in geeigneter Form gekennzeichnet und den Ursprung der Infor­mation durch möglichst exakte Quellenangaben (z.B. in Fußnoten) ersichtlich gemacht habe,
		
		\item[--] die eingereichte wissenschaftliche Arbeit bisher weder im Inland noch im Ausland einer Prüfungsbehörde vorgelegt habe und
		
		\item[--] bei der Weitergabe jedes Exemplars (z.B. in gebundener, gedruckter oder digitaler Form) der wissenschaftlichen Arbeit sicherstelle, dass diese mit der eingereichten digitalen Version übereinstimmt.
	\end{itemize}
	
	Mir ist bekannt, dass die digitale Version der eingereichten wissenschaftlichen Arbeit zur Plagiatskontrolle herangezogen wird.
	
	Ich bin mir bewusst, dass eine tatsachenwidrige Erklärung rechtliche Folgen haben wird.
	
}

\vskip 12em

\begin{minipage}{\columnwidth}
	\begin{minipage}{0.49\columnwidth}
		Negin Ghamsarian e. h.
	\end{minipage} 
	\begin{minipage}{0.49\columnwidth}
		\flushright Klagenfurt am Wörthersee, 27.9.2021
	\end{minipage}
\end{minipage}

\chapter*{Affidavit}
{	
	\singlespacing
		
	I hereby declare in lieu of an oath that
	
	\begin{itemize}
		\item[--] the submitted academic paper is entirely my own work and that no auxiliary materials have been used other than those indicated,
		
		\item[--] I have fully disclosed all assistance received from third parties during the process of writing the thesis, including any significant advice from supervisors,
		
		\item[--] any contents taken from the works of third parties or my own works that have been included either literally or in spirit have been appropriately marked and the respective source of the information has been clearly identified with precise bibliographical references (e.g. in footnotes),
		
		\item[--] to date, I have not submitted this paper to an examining authority either in Austria or abroad and that
		
		\item[--] when passing on copies of the academic thesis (e.g. in bound, printed or digital form), I will ensure that each copy is fully consistent with the submitted digital version.
	\end{itemize}
	
	I understand that the digital version of the academic thesis submitted will be used for the purpose of conducting a plagiarism assessment.
	
	I am aware that a declaration contrary to the facts will have legal consequences.
	
}

\vskip 12em

\begin{minipage}{\columnwidth}
	\begin{minipage}{0.49\columnwidth}
		Negin Ghamsarian e. h.
	\end{minipage} 
	\begin{minipage}{0.49\columnwidth}
		\flushright Klagenfurt am Wörthersee, 27.9.2021
	\end{minipage}
\end{minipage}

\chapter*{Acknowledgments}

I take pride in working under the supervision of professor Klaus Schöffmann and taking advantage of his invaluable guidance during my doctoral studies. I was very fortunate to work with an open-minded, kind-hearted, approachable, and knowledgeable supervisor, who taught me how to develop critical thinking, work independently, and think on my feet. I would like to express my sincere gratitude for his empathy and compassionate leadership, especially during the hard time of the COVID pandemic.

I would like to thank the examiners of my dissertation, professor Henning Müller and professor Raphael Sznitman, for their time and precious feedback.

My heartfelt thanks should also go to the administrative staff of the ITEC department not only for their prompt support but also for their kindness and companionship during my work at Klagenfurt University.

Finally, I would like to thank my mother for her endless moral support and believing in me when no one believes in me, and thank my father for always motivating me to go the extra mile and encouraging me not to give up. No words can convey my profound gratitude to them for consistently supporting me through thick and thin.

\begin{center}
	\textit{Thank you!}
\end{center}

\chapter*{Abstract}
Following the technological advancements in medicine, the operation rooms are evolving into intelligent environments. The context-aware systems (CAS) can comprehensively interpret the surgical state, enable real-time warning, and support decision-making, especially for novice surgeons. These systems can automatically analyze surgical videos and perform indexing, documentation, and post-operative report generation. The ever-increasing demand for such automatic systems has sparked machine-learning-based approaches for surgical video analysis. This thesis addresses the significant challenges in cataract surgery video analysis to pave the way for building efficient context-aware systems. The main contributions of this thesis are five folds: (1) This thesis demonstrates that spatio-temporal localization of the relevant content can considerably improve phase recognition accuracy. (2) This thesis proposes a novel deep-learning-based framework for relevance-based compression to enable real-time streaming and adaptive storage of cataract surgery videos. (3) Several convolutional modules are proposed to boost the networks' semantic interpretation performance in challenging conditions. These challenges include blur and reflection distortion, transparency, deformability, color and texture variation, blunt edges, and scale variation. (4) This thesis proposes and evaluates the first framework for automatic irregularity detection in cataract surgery videos. (5) To alleviate the requirement for manual pixel-based annotations, this thesis proposes novel strategies for self-supervised representation learning adapted to semantic segmentation.

\body

\listoffigures
\listoftables


\chapter{Introduction}

\chapterintro{
	This chapter starts with a brief description of cataract surgery. Afterwards, I delineate the rational behind computerized analysis of cataract surgery, which is the focus of this dissertation. The research questions are then introduced in this chapter. Finally, the last section gives a summary of the research subjects studied  in the course of this thesis using an initial pipeline of a cataract surgery exploration system.

}

{
	\singlespacing This chapter is an adapted version of:
	
	``Ghamsarian, N. Enabling relevance-based exploration of cataract videos.
In Proceedings of the 2020 International Conference on Multimedia Retrieval
(New York, NY, USA, 2020), ICMR ’20, Association for Computing Machinery,
p. 378–382.''
}

\section{Cataract Surgery}
\label{secIntro: Introduction}
\begin{figure}[!tb]
\centering
\includegraphics[width=0.7\textwidth]{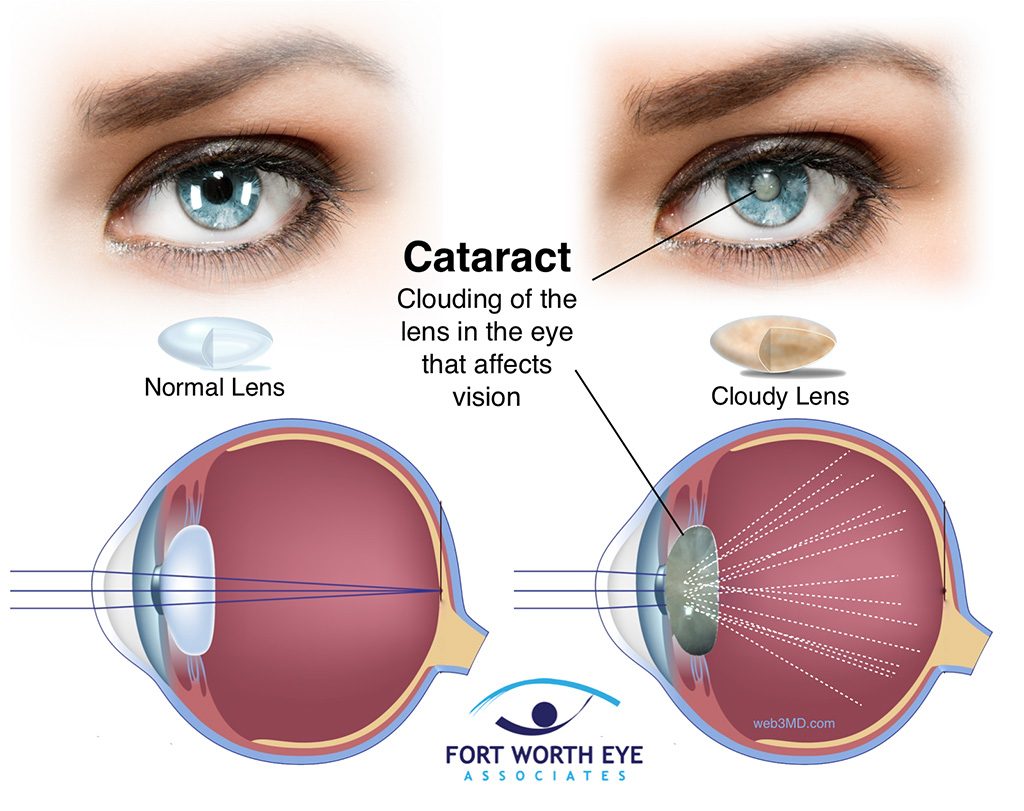}
\caption{Cataract as the eye's natural lens, having become cloudy and causing vision deterioration (picture from https://www.ranelle.com/cataract-surgery/).}
\label{fig_Intro: cataract}
\end{figure}

\begin{figure}[!tb]
\centering
\includegraphics[width=0.5\textwidth]{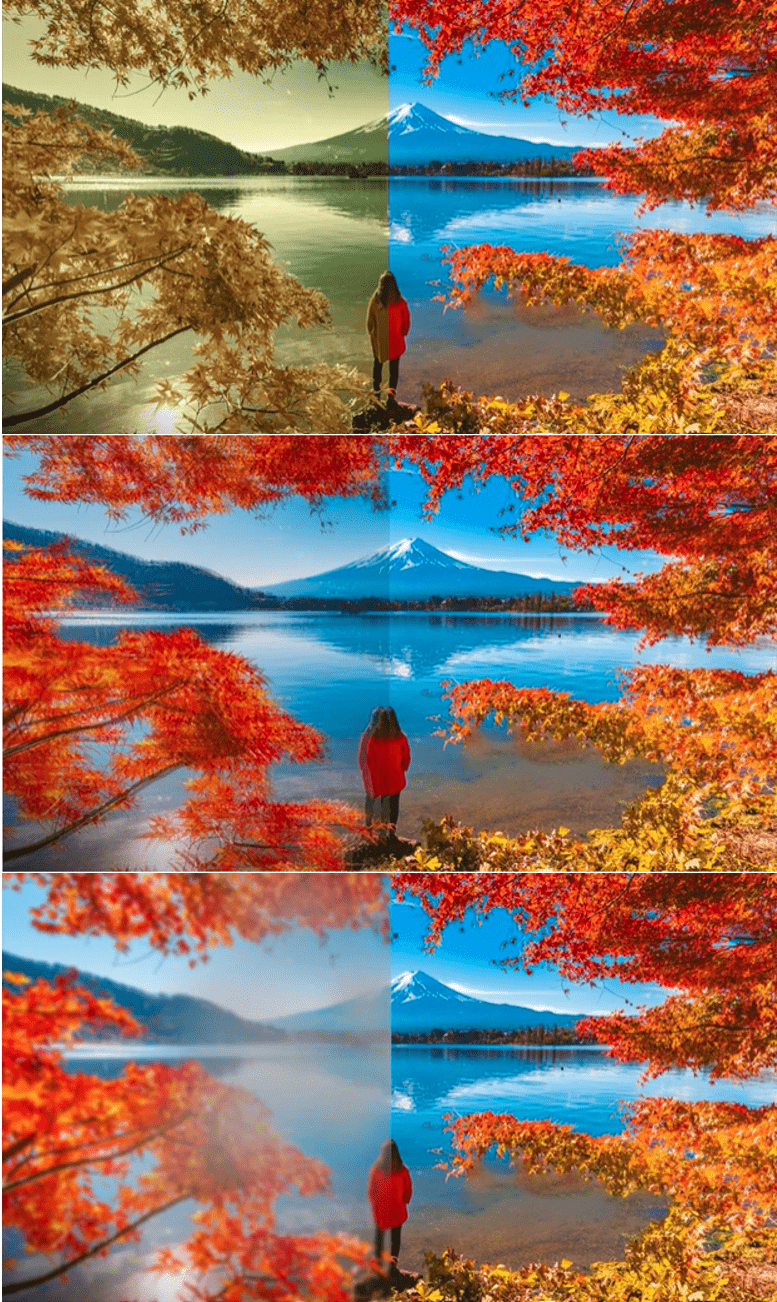}
\caption{Cataract symptoms including color perception distortion, double vision, and blurred vision (picture from \url{https://www.mathworks.com/company/mathworks-stories/}).}
\label{fig_Intro: cat_symp}
\end{figure}

Cataract refers to the opacity and occlusion of the eye's natural lens due to the process of aging, eye inflammation, congenital problems, and other issues. (Figure~\ref{fig_Intro: cataract}). This natural lens' cloudiness causes vision deterioration and transparency degradation, but also blindness in the case of dense occlusion. Color perception distortion due to the lens' yellowing, double vision (ghosting), and blurred vision are the most common symptoms of cataracts (Figure~\ref{fig_Intro: cat_symp}). Cataract surgery is the procedure of returning a clear vision to the eye by removing the occluded lens, followed by implanting an artificial lens named intraocular lens (IoL). Involving over 100 million people worldwide, cataract surgery is the causative factor of a substantial fraction of worldwide blindness~\cite{GVO}.

\begin{figure}[!tb]
\centering
\includegraphics[width=0.7\textwidth]{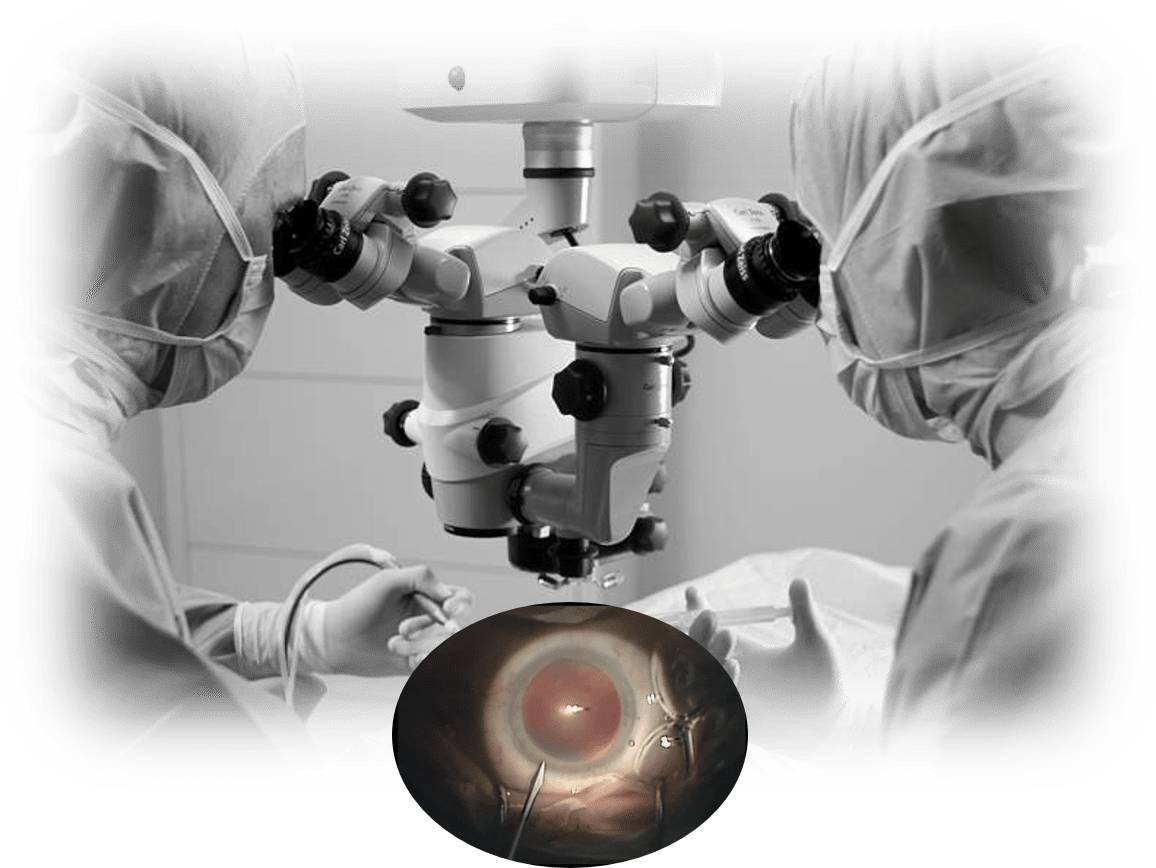}
\caption{The binocular microscope that is used during the surgery to enable accurately watching the patient's eye.}
\label{fig_Intro: microscope}
\end{figure}

Cataract surgery is not only the most frequent ophthalmic surgery~\cite{COC} but also one of the most common surgeries worldwide~\cite{JFCS}. This operation is conducted with the aid of a binocular microscope that provides a three-dimensional magnified and illuminated image of the eye for accurately watching the patient's eye (Figure~\ref{fig_Intro: microscope}). The microscope contains a mounted camera, which records and stores the whole surgery for several post-operative objectives.


\section{Motivation}

Due to the growing demand for cataract surgery, this surgery significantly impacts the patient's quality of life. Accordingly, a large body of research is devoted to computerized workflow analysis in this surgery to improve surgical outcomes and diminish potential surgical risks. This objective can be achieved through (1) alleviating the training procedure for junior surgeons via enabling content-aware explorations, (2) facilitating the storage and streaming of cataract surgery videos, (3) and detecting the unexplored risk factors and irregularities through computerized investigations. In the following, the importance of each contribution is justified.

\subsection{Training Procedure Alleviation } 
\begin{figure}[!tb]
    \centering
    \includegraphics[width=1\textwidth]{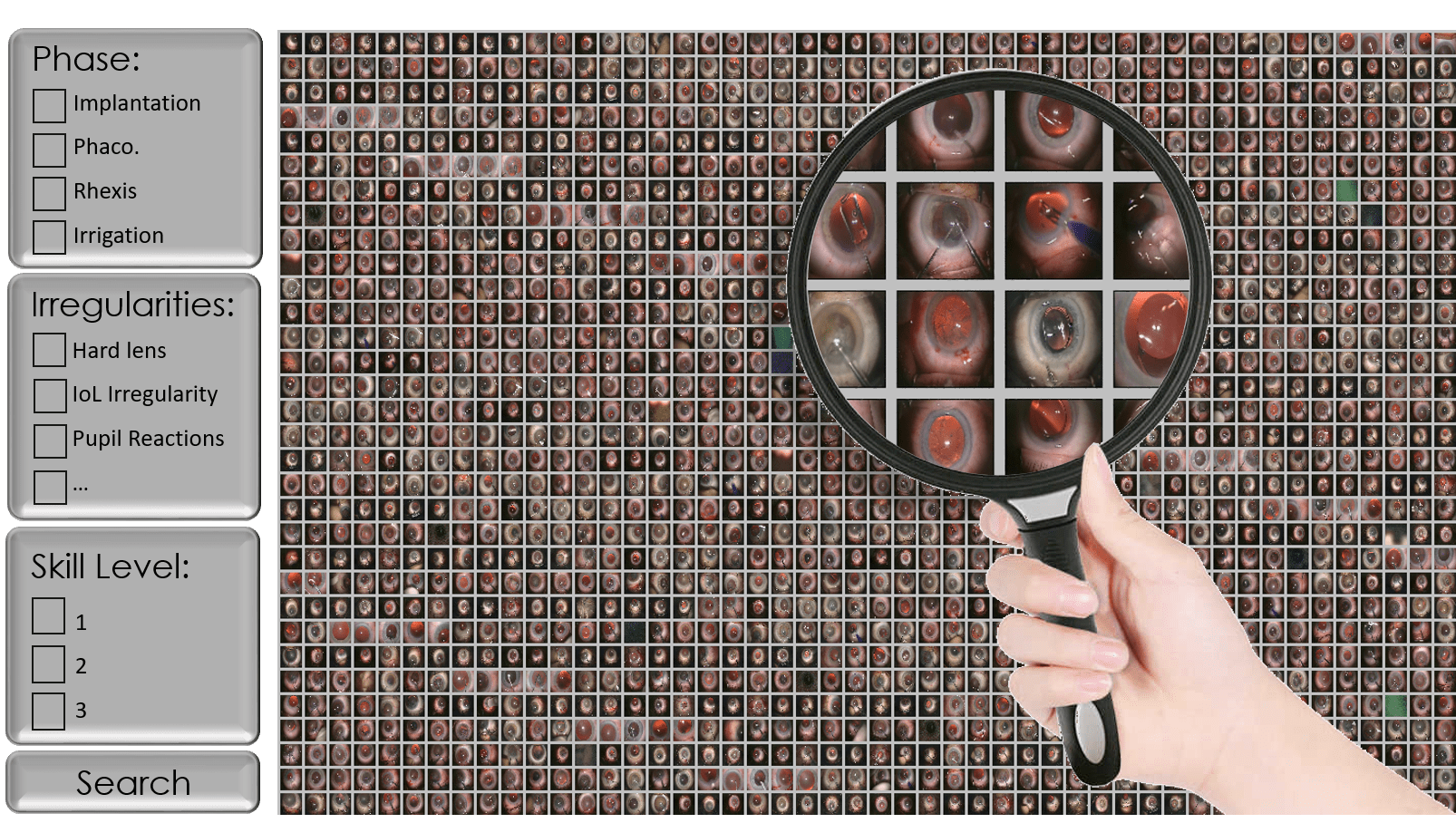}
    \caption{A cataract surgery exploration system can substantially optimize the training procedure by enabling relevance-based retrieval.}
    \label{fig_Intro: CSES}
\end{figure}

During the operation of cataract surgery, only two trainees can watch the real-time surgery throughout the teaching binoculars. Thus, the major part of the training procedure is conducted using the videos recordings of cataract surgery. A faster training process results in reducing complications during and after surgeries and diminishing the surgical risks for less-experienced surgeons. As a concrete example, the risk of developing \textit{reactive corneal edema} after surgery for novice surgeons is reported to be 1.6 times higher than that for experienced expert surgeons~\cite{IOSE2017}. Accordingly, it is vital to accelerate the teaching and training process for a surgical technique. Training new surgeons as one of the major duties of experienced expert surgeons demands a considerable supervisory investment. To expedite the training process and subsequently reduce the extra workload on their tight schedule, surgeons are seeking a surgical video retrieval system~\cite{SurgXplore} (Figure~\ref{fig_Intro: CSES}). Automatic workflow analysis approaches can optimize the training procedure by indexing the surgical video segments for online video exploration~\cite{ReIS, InViR, SDS}.

\subsection{Storage and Streaming Facilitation }

Recorded cataract surgery videos play a prominent role in training, investigating the surgery, and enhancing surgical outcomes. Due to storage limitations in hospitals, however, the recorded cataract surgeries are typically deleted after a short time, and this precious source of information cannot be fully utilized. Moreover, these videos can be exploited for knowledge exchange among hospitals and surgeons using remote online exploration. Online exploration usually imposes quality degradation due to limited bandwidth. Lowering the quality for reducing the required storage space or streaming is not advisable since the degraded visual quality results in the loss of relevant information that limits the usage of these videos. Relevance-based compression of cataract surgery videos is the best way to simultaneously provide high quality for the relevant content and low bitrate. This technique entails spatio-temporal relevance detection (\ie phase detection and semantic segmentation) in cataract surgery videos.

\subsection{Irregularity Detection }

Notwithstanding the numerous advancements in surgical tools and techniques, there are still some intra-operative and post-operative complications in cataract surgery requiring real-time or large-scale computerized workflow analysis. 

Irregularity detection in surgical videos can serve two major purposes:

\begin{enumerate}
\item  By studying the correlations between the intra-operative irregularities and post-operative complications, the surgeons will be able to assess risk factors associated with different complications. Such an advanced knowledge can result in achieving optimal post-operative results.  

\item Irregularities in different phases are of great importance for the surgeons in terms of teaching. Indeed, surgical training on irregularities plays a key role in surgical competency enhancement \footnote{Surgical competency is defined as the required skill level to perform a safe surgery indepently. Surgical competency entails not only the technical skill to accurately perform the regular surgical phases, but also the knowledge and judgment required to safely deal with irregularities during surgery~\cite{ObjAss2017}}.
\end{enumerate}

Pupil reactions and IoL instability, unfolding delay, and rotation are the major unexplored implications in cataract surgery.

\paragraph{Pupil reactions: }During the phacoemulification phase, where the occluded natural lens is corrupted and suctioned, the amount of light received by photoreceptors may suddenly increase. This increase in light reception affects the size of pupil, usually resulting in slow (gradual) pupil contraction. In some cases, however, the pupil unexpectedly reacts to the lighting changes and becomes quickly contracted. This sudden reactions in pupil size can lead to serious intra-operative implications.  Especially during phacoemulification phase where the instrument is deeply inserted inside the eye, sudden changes of pupil size may lead to injuries to the eye's tender tissues. Pupil reaction is regarded as a major intra-operative issue, leading to serious implications during surgery. Real-time automatic detection of  pupil reactions during phacoemulification phase can highly contribute to a safer surgical procedure as well as providing important insight for further post-operative investigations~\cite{PBIPS}. Detecting this irregularity involves phacoemulification phase recognition and pupil/iris segmentation.

\paragraph{IoL irregularity: }A critical complication after cataract surgery is the dislocation of the lens implant leading to vision deterioration and eye trauma. In order to reduce the risk of this complication, it is vital to discover the risk factors during the surgery. 
It is argued that there is a direct relationship between the lens irregularities during and after surgery.  Especially, the surgeons believe that lens unfolding delay, instability, and rotation during  surgery are some of the symptoms of lens dislocation after surgery. Besides, the surgeons claim that these irregularities may bound with some brands of the IoL.
However, studying the relationship between lens dislocation and its suspicious risk factors using numerous videos is a time-extensive procedure. Hence, the surgeons demand an automatic approach to enable large-scale investigations of this problem. Such study involves implantation phase detection and IoL/pupil segmentation.

\section{Research Questions}
\label{sec_intro: Research Questions}

Considering the major demands for enhancing the recent approaches or investigating unexplored research subjects, this thesis tries to address the following four research questions:

\begin{itemize}
\item Can the accuracy of automatic operation phase recognition be improved upon state-of-the-art techniques using deep learning approaches? 
\item How and to what extent can we reduce the required bitrate for cataract surgery videos while preserving high  quality of the relevant content?
\item How reliably can unspecified irregularities be automatically detected in ophthalmic surgery videos?
\item How can the semantic segmentation performance for the relevant objects in cataract surgery be improved?
\end{itemize}

\section{Research Overview}
\label{sec_Intro: Context-Aware Retrieval System}

Aiming to enhance the surgical outcomes and diminish the clinical risks, there is a great desire for context-aware retrieval systems. Such a system is particularly advantageous for training amateur and less-skillful surgeons by representing the relevant content from the surgeons' eyes.   
 One of the aims of this doctoral project is to provide the basis for a cataract video exploration system, that is able to automatically analyze and extract the relevant segments of videos from cataract surgery. Indeed, in this doctoral project, content analysis and retrieval methods for recorded cataract surgery videos will be investigated to meet the requirements for building a content exploration system. The system is going to be used by clinicians as a teaching means to communicate operation techniques and complications to trainee surgeons. Hence, such a system should be able to automatically filter the irrelevant video segments and classify the relevant content (e.g., specific operation tools, operation phases, and surgical actions). 
\begin{figure}[!tb]
    \centering
    \includegraphics[width=1\textwidth]{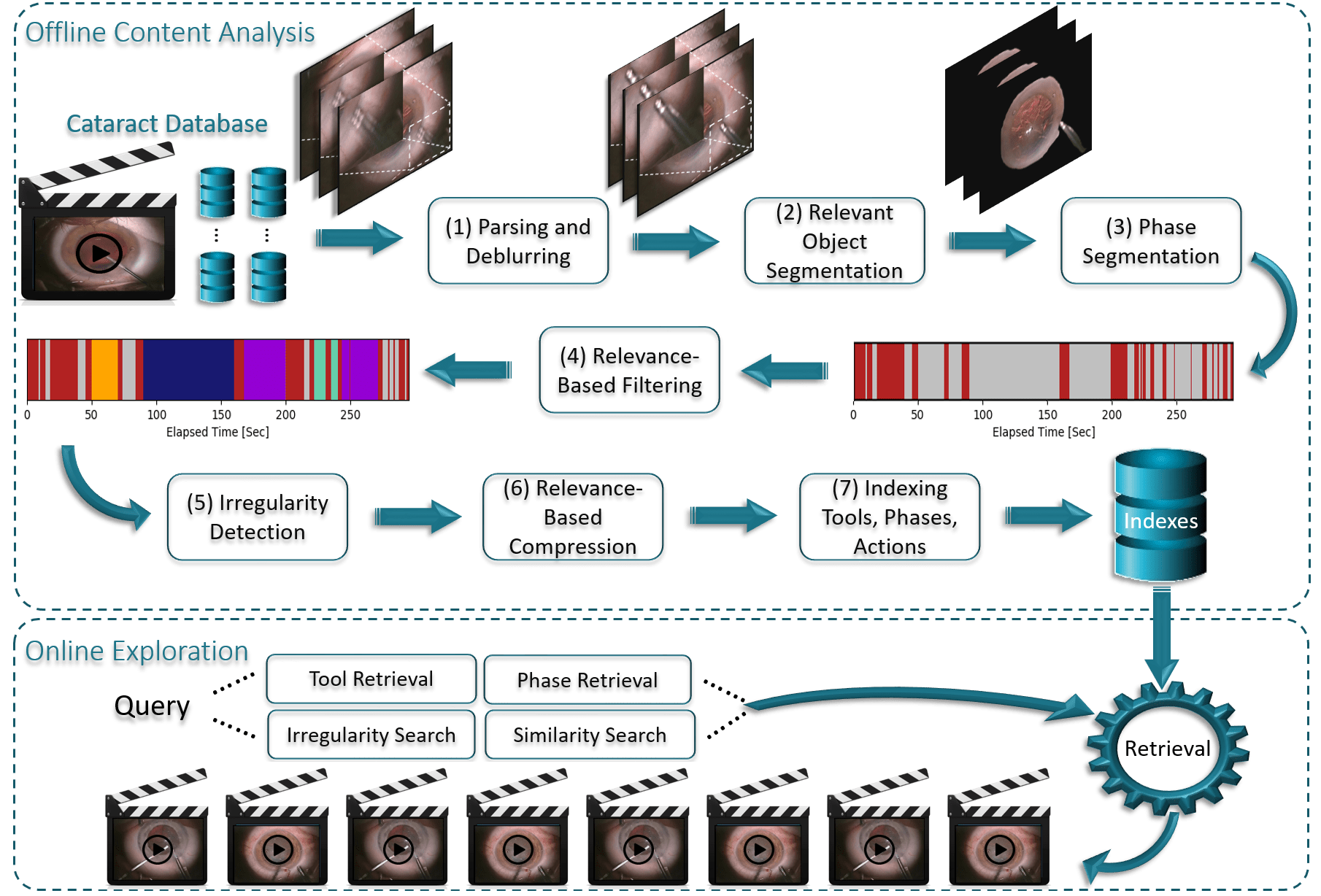}
    \caption{The pipeline of the proposed cataract surgery exploration system consisting of an offline content analysis module and an online exploration module.}
    \label{fig_Intro: BD_RBE}
\end{figure}

Typically, the recorded cataract videos do not contain any meta-data and are basically one-shot videos, without any editing. Moreover, the content of different operation phases is highly similar and suffers from the following problems: 
\begin{enumerate}
    \item Since the camera focus is manually adjusted by the surgeons as a pre-surgical step, the recorded videos may contain defocus blur.
    \item Unconscious eye movements, as well as fast instrument motions, may lead to harsh motion blur in the salient segments of each frame.
    \item The operation instruments used in different phases are highly similar in terms of visual appearance, resulting in a narrow inter-class deviation.
\end{enumerate}
Designing an approach to classify the relevant content in such videos, therefore, is quite challenging. Based on the mentioned challenges, we have designed the first pipeline of cataract surgery exploration system as shown in Figure~\ref{fig_Intro: BD_RBE}. The proposed pipeline consists of an offline content analysis module and an online exploration module. The resulting system allows the user to retrieve tools, phases, and irregulities, as well as performing content-based similarity search. However, the focus of this work is on the content analysis, since another companion doctoral project in our research group will focus on the indexing methods and the interface to use the results achieved herein.

The offline content analysis module consists of several steps. The cataract surgery videos being impaired by defocusing and motion blur are deblurred in the first stage, prior to feeding them into a neural network for our task of interest. The relevant segments of each frame are then extracted using pixel-based or region-based segmentation networks. 

In the next step, a CNN for idle frame recognition is employed to retain only the segments showing some actions (i.e., performed with some instruments) and completely ignore the content without any instruments.
Afterwards, the segmented frames are fed into the networks being accounted to detect different relevant phases or irregularities. The outputs of these stages will be further exploited for indexing the segments of cataract surgery videos and provided to the user of the online exploration system. The mentioned steps will be described in details in the following. 

\subsection{Parsing and Deblurring}
Neural networks recognize the edges in the first layers to combine them to complicated and meaningful content in the last layers. Thus, neural networks suffer from performance degradation when training on blurry inputs~\cite{IoB}. This is the reason why cataract surgery videos should be deblurred as a pre-processing step to retrieve the diluted high-frequency features and acquire the underlying sharp frames as the inputs for our neural networks. 

In Chapter~\ref{Chapter:Deblurring}, we address the low visual quality of surgical videos using a multi-scale deconvolutional neural network inspired by~\cite{MDW}. The network architecture is rooted in the cross-scale patch recurrence theory: ``the small patches in a high-resolution image usually recurr in its ideally downscaled versions'', a property that is frequently exploited for super-resolution. Cross-scale patch recurrence is also used for deblurring since the small patches in an ideally downsampled blurry image tend to be similar to the small patches of the original sharp image rather than the blurry one. 

The proposed network includes three subgraphs, each undertaking the task of deblurring the downsampled input image and outputting a residual frame. The first subgraph deblurs the input frame downsampled at $1/4$ of the initial resolution and outputs a residual frame to counteract the blurriness impairing the input. The sharp downsampled image is then fed to the next subgraph to perform superresolution and retrieve the underlying sharp image at scale $1/2$. Similarly, the last subgraph is responsible for outputting a sharp frame with the original resolution.

We prove that using skip connections in each subgraph thanks to defining same number of filters for the consecutive layers can effectively boost deblurring performance. Besides, we model defocus blur using gaussian filters and show that a model trained to deblur the gaussian-blur-degraded frames can deal with defocus blur. More details of this approach are described in \cite{DRNet}.



\subsection{Relevant Object Segmentation}
\begin{figure}[!tb]
    \centering
    \includegraphics[width=0.5\textwidth]{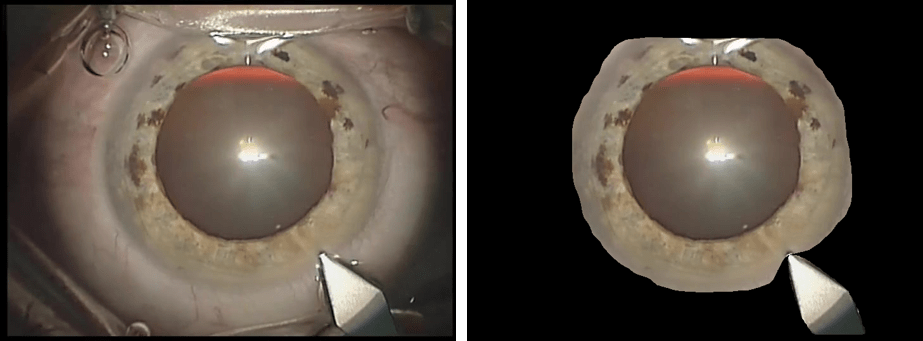}
    \caption{Detecting the spatially relevant segments and removing the substantial redundant information using semantic segmentation networks for a representative frame.}
    \label{fig_Intro: Semantic_Seg}
\end{figure}

Semantic segmentation in surgical videos is a  prerequisite for a broad range of applications towards improving surgical outcomes and surgical video analysis including but not limited to workflow analysis, compression, and irregularity detection.
Besides, neural networks generally suffer from redundant input information. In other words, the less redundant information we provide to the network, the better performance we can expect~\cite{UED}. Relevant object segmentation offers a solution to this problem. Depending on the task (relevance detection or a particular irregularity detection), the relevant segments of each frame, being informative for classification are extracted using pixel-based segmentation networks. In the case of phase recognition, the background being considered as the redundant part of each frame is then replaced with black pixels (Figure~\ref{fig_Intro: Semantic_Seg}).

In cataract surgery, various features of the relevant objects such as blunt edges, color and context variation, reflection, transparency, and motion blur pose a challenge for semantic segmentation. In particular, motion blur and reflection distortion in instruments, and color and texture variations in the other relevant objects lead to distant semantic respresentations for the same semantic labels, and close semantic representations for different semantic labels. 
In Chapter~\ref{Chapter: Recal-Net}, we propose a novel feature-map callibration module to enhance the semantic representation performance. The proposed module callibrates the feature maps considering multi-angle local features centering around each pixel position, and region-channel inter-dependencies.

In Chapter~\ref{Chapter:DeepPyram}, we propose a novel semantic segmentation network termed as ``DeepPyram'', which can deal with various semantic segmentation challenges in cataract surgery videos using three proposed modules: (i) Pyramid View Fusion, (ii) Deformable Pyramid Reception, and (iii) Pyramid Loss. The experimental evaluations confirm the superiority of the DeepPyram over state-of-the-art rival approaches. 

Due to many reasons, atificial intelligence in medical domain has lagged behind other domains. First and foremost, the supervised machine learning approaches depend heavily on manual annotations, which requires expert knowledge when it comes to medical images and videos. Thus supervised learning in medical domain is more time extensive and costly. Secondly, the pre-trained weights of convolutional neural networks using larg-scale image datasets are not adaptive initializations due to the large distribution gap between the medical and natural images. On top of that, the rapid advancements in technological tools and instruments inevitably entails expeditious distribution shift in raw data.  Self-supervised learning suggests ultimate solutions to alleviate the mentioned problems. In Chapter~\ref{Chapter:SSL}, we propose three self-supervised strategies to encourage representation learning adapted to semantic segmentation in cataract surgery videos. Specifically, we provide a moderate-to-hard representation learning task to bridge the gap between human and neural network semantic interpretation.

\begin{figure}[!tb]
    \centering
    \includegraphics[width=1\textwidth]{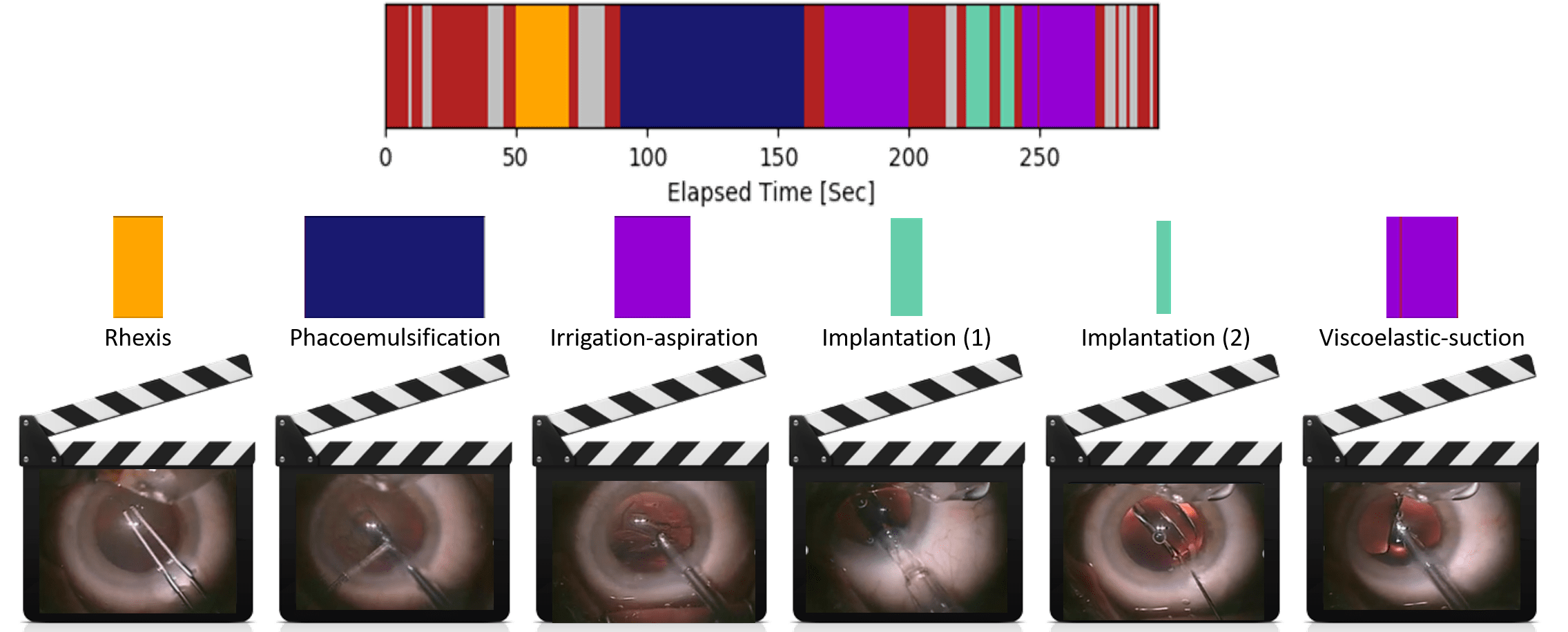}
    \caption{Relevence Detection Results of the proposed CNN-RNN framework for a representative cataract surgery video.}
    \label{fig_Intro: Temporal_Seg}
\end{figure}
\subsection{Phase Segmentation}
In this stage, a CNN networks is employed to perform two-class classification on the input frames and categorize them as \textit{idle} or \textit{action}. Idle frames refer to the frames in which no instrument is visible. This gives us a clue that the surgeon is changing the instrument and accordingly a new phase will begin. In other words, each action phase in a cataract surgery video is delimited by two idle phases. By having a whole segment of a phase, we can exploit and accumulate the features coming from the network for every single frame to boost the accuracy in phase recognition.

\subsection{Relevance-Based Filtering}
A regular cataract surgery video consists of eleven phases: incision, hydrodissection, rhexis, viscoelastic, phacoemulsification, irrigation-aspiration, capsule polishing, lens implantation, viscoelastic-suction, tonifying, and antibiotics. To investigate and teach the cataract surgery, however, not all these phases are relevant (rhexis, phacoemulsification, irrigation-aspiration, and lens implantation are considered relevant by clinicians). Hence, an automatic retrieval approach is required to segment the video and output the relevant segments. 

Chapter~\ref{Chapter:Relevance-Detection} is devoted to relevance detection in cataract surgery videos using CNN-RNNs. We firstly segment the distinct phases using a static CNN. The label of each segmented phase is then determined independently and disregarding the information from previous and following phases. Such an strategy is particularly useful in the case of irregularities in catarcat surgery videos, where the networks trained on the sequence of consecutive regular phases usually fail to predict the right label. We also prove that removing the redundant spatial information and increasing the resolution of the relevant spatial content can significantly enhance phase recognition accuracy, especially in the case of visually similar instruments. Our model is able to detect the relevant phases with very high temporal resolution compared to its counterparts.

\subsection{Irregularity Detection}

A surgical video exploration system should be capable of identifying the irregular phases to exclude many redundant regular surgeries, enable irregularity search, and provide scalable statistical analysis. 
Chapter~\ref{Chapter:Irregularity-Detection} introduces a framework to detect two suspicious sympthoms of lens relocation after surgery. Specifically, we focus on computing IoL unfolding delay, instability, and rotation. This requires detecting the lens implantation phase and segmenting the pupil and IoL with high accuracy. We, therefore, propose a CNN-RNN to detect the implantation phase, and a novel U-Net-Based architecture termed as Adapt-Net to segment the pupil and IoL. The proposed segmentation network can be adapted to scale and transparency of IoL by fusing the sequential convolutional-block response maps. Moreover, Adapt-Net can deal with shape variations and unpredictable deformations during unfolding by fusing the deformable-filters and structured-filters response maps. 

\subsection{Relevance-Based Compression}
In Chapter~\ref{Chapter:Compression}, we introduce a novel framework for relevance-based compression of cataract surgery videos. The proposed framework consists of two modules: (i) a relevance detection module, which is designed to detect the spatio-temporally relevant content based on five differenr scenarios using static and region-based CNNs, and (ii) a compression module that is responsible for assigning low bitrate to irrelevant content to enhance compression ratio, while preserving the high quality of the relevant content.

\section{Publications in the Context of the Dissertation Topic}

In the following, a list of all publications that are directly related to the topic of this thesis is given in chronological order:

\begin{itemize}
\item ``Ghamsarian, N. Enabling Relevance-Based Exploration of Cataract Videos.
In Proceedings of the 2020 International Conference on Multimedia Retrieval
(New York, NY, USA, 2020), ICMR ’20, Association for Computing Machinery,
p. 378–382.''
\item ``Ghamsarian, N., Taschwer, M., and Schoeffmann, K. Deblurring
Cataract Surgery Videos Using a Multi-Scale Deconvolutional Neural Network.
In 2020 IEEE 17th International Symposium on Biomedical Imaging (ISBI)
(2020), pp. 872–876.''
\item ``Ghamsarian, N., Amirpourazarian, H., Timmerer, C., Taschwer,
M., and Schoffmann, K. ¨ Relevance-Based Compression of Cataract Surgery
Videos using Convolutional Neural Networks. In Proceedings of the 28th ACM
International Conference on Multimedia (New York, NY, USA, 2020), MM ’20,
Association for Computing Machinery, p. 3577–3585.''
\item ``Ghamsarian, N., Taschwer, M., Putzgruber-Adamitsch, D., Sarny,
S., and Schoeffmann, K. Relevance Detection in Cataract Surgery Videos
by Spatio-Temporal Action Localization. In 2020 25th International Conference
on Pattern Recognition (ICPR) (2021), pp. 10720–10727.''
\item ``Ghamsarian, N., Taschwer, M., Putzgruber-Adamitsch, D., Sarny, S., El-Shabrawi, Y., and Schoeffmann, K. LensID: A CNN-RNN-Based Framework Towards Lens Irregularity Detection in Cataract Surgery Videos. In Medical Image Computing and Computer Assisted Intervention – MICCAI 2021 (Cham, 2021), M. de Bruijne, P. C. Cattin, S. Cotin, N. Padoy, S. Speidel, Y. Zheng, and C. Essert, Eds., Springer International Publishing, pp. 76–86.''
\item ``Ghamsarian, N., Taschwer, M., Putzgruber-Adamitsch, D., Sarny,
S., El-Shabrawi, Y., and Schoeffmann, K. Recal-Net: Joint Region-Channel-Wise Calibrated Network for Semantic Segmentation in Cataract Surgery Videos. In 28th International Conference on Neural Information Processing
(ICONIP) (2021), p. To Appear.''
\item ``Ghamsarian, N., Taschwer, and Schoeffmann, K. DeepPyram: Enabling Pyramid View and Deformable Pyramid Reception for Semantic Segmentation in Cataract Surgery Videos. Under Review.'' 
\end{itemize}

Table~\ref{tab_Intro: paper contribution} gives an  overview  of  the papers  related  to this  thesis, and my level of contribution in the proposed approach, implementation, evaluation, and writing in each paper.

\begin{sidewaystable}[thpb!]
\renewcommand{\arraystretch}{1.3}
\caption{Papers.}
\label{tab_Intro: paper contribution}
\centering
\begin{tabu}{lccccc}
\specialrule{.12em}{.05em}{.05em}
Subject & Reference & Proposed Approach & Implementation & Evaluation & Writing\\
\specialrule{.12em}{.05em}{.05em}
Relevance-Based Exploration & \cite{RBE} & \Stars{5} & \Stars{5} & \Stars{5} & \Stars{5}\\
Deblurring & \cite{DCS} & \Stars{5} & \Stars{5} & \Stars{5} & \Stars{5}\\
Relevance Detection & \cite{RDCSV} & \Stars{5} & \Stars{5} & \Stars{5} & \Stars{5}\\
Relevance-Based Compression & \cite{RBCCSV} & \Stars{5} & \Stars{3.5} & \Stars{3} & \Stars{5}\\
Lens Irregularity Detection & \cite{LensID} & \Stars{5} & \Stars{5} & \Stars{5} & \Stars{5}\\
Semantic Segmentation (DeepPyram) & \cite{DeepPyram} & \Stars{5} & \Stars{5} & \Stars{5} & \Stars{5}\\
Semantic Segmentation (ReCal-Net) & \cite{Recal-Net} & \Stars{5} & \Stars{5} & \Stars{5} & \Stars{5}\\
Self-Supervised Learning & To be submitted & \Stars{5} & \Stars{5} & \Stars{5} & \Stars{5}\\
\specialrule{.12em}{.05em}{.05em}
\end{tabu}
\end{sidewaystable}

\chapter{A Survey on Deep-Learning-Based Surgical Video Analysis \label{Chapter:Survey}}

\chapterintro{
	This chapter summarizes and compares state-of-the-art machine-learning-based approaches towards computer assisted surgery, with a major focus on deep-learning-based approaches.
}

{
	\singlespacing 
}

Computerized surgical workflow analysis and computer-assisted surgery (CAS) are becoming integral parts of medicine. In particular, state-of-the-art deep-learning-based approaches have impacted scheduling and operation planning, intra-operative assessment and postoperative evaluation, patient briefing~\cite{VCA}, and surgical outcome estimation~\cite{CASM}. These approaches can be broadly categorized into workflow analysis, instrument recognition and segmentation, and skill assessment. The following sections briefly review the main approaches related to each of the mentioned categories.







\section{Workflow Analysis}
The applications of automatic workflow analysis approaches include but are not limited to:

\begin{itemize}
\item Providing real-time guidance to support decision making in the case of training with surgery simulators,
\item Predicting the next required tool,
\item Determining remaining surgery duration to optimize staff scheduling and reduce patient waiting time,
\item Early warnings through anomaly or deviation detection~\cite{OISD},
\item Determining the amount of required anesthesia
\end{itemize}

The existing approaches belonging to workflow analysis can be categorized into two groups: phase recognition and remaining surgery duration.

\subsection{Phase Recognition}
Phase recognition in surgical videos can be broadly divided into classical (feature-extraction-based) and deep-learning-based methods. The first-generation approaches extract hand-engineered features to be further used as the inputs to the classical machine learning methods. Some of these methods exploit hand-crafted features such as texture information, color, and shape~\cite{FRHS}. The major discriminative features in different phases of surgical videos are the specific tools used in these phases. Accordingly, many approaches exploit binary instrument usage information~\cite{SMRSW}, RFID tags~\cite{PRDSP}, tool tracking equipment, or built-in sensors~\cite{FRTWS}. Some methods based on conditional random fields~\cite{RTS}, random forests~\cite{RFPD}, or Hidden Markov Models (HMMs)~\cite{RTA}, have taken advantage of tool presence information for phase recognition. Another method exploits Dynamic Time Warping (DTW), Hidden Markov Models (HMMs), and Conditional Random Fields (CRFs) along with the tool presence signals for surgical video analysis~\cite{RTM}.
Regarding gesture and action recognition, Zappella~\etal~\cite{SGC} use linear dynamic systems, bag of features (BoF), and multiple kernel learning (MKL) to classify the pre-segmented video clips.

Hand-crafted features may provide sub-optimal classification results or fail to provide robust classification. Besides, tool usage information relies on additional sensors and devices, which are not ubiquitous, or manual annotations being expensive and time-consuming. Automated feature extraction thanks to deep neural networks have proved to provide more optimal and robust classifiers for complicated problems. Hence, deep-learning-based approaches have drawn much attention from the medical imaging community in recent years, and
many studies in the next generation have been dedicated to automatic workflow recognition using deep-learning-based approaches.

Authors of~\cite{SV-RCNet} use an integrated CNN-RNN (end to end ResNet+LSTM) network to fully utilize the complementary spatio-temporal features for phase recognition in cholecystectomy videos. To reduce the demand for annotations or compensate for the lack of enough labeled data, another work~\cite{TCB} proposes three self-supervised pre-training methods using temporal coherence: (1) contrastive loss, (2) ranking loss, and (3) first and second-order contrastive loss. Their results provide evidence that a pre-trained model fine-tuned on fewer videos outperforms the baseline trained on more videos. Using Gated Recurrent Unit (GRU) as an alternative for LSTM is also suggested to improve the classification performance~\cite{UTL}. The authors of~\cite{LIM} proposed a ``semi-supervised'' approach based on ``self-supervised pre-training'' to predict Remaining Surgery Duration (RSD). EndoNet~\cite{EndoNet} employs a CNN architecture to carry out phase recognition and tool presence detection in a multi-task manner. It leverages AlexNet trained for tool recognition as the feature extractor for Hierarchical Hidden Markov Model (HHMM) to classify the phases in laparoscopic surgeries. EndoRCN~\cite{EndoRCN} uses the same technique with ResNet50 as the backbone network and recurrent layer instead of HHMM for phase recognition. DeepPhase~\cite{DPS} exploits tool features as well as tool classification results of ResNet152 as the input for RNNs to perform phase recognition. A comparison between the performance of stacked GRU layers~\cite{GRU} and LSTM~\cite{LSTM} in phase recognition revealed that GRU is more successful in inferring the phase based on binary tool presence. On the other hand, LSTM performs better when trained on tool features. Authors of~\cite{SV-RCNet} propose an end-to-end CNN-LSTM to exploit the correlated spatio-temporal features for phase recognition in cholecystectomy videos. To address transition-sensitivity in predictions, they propose the concept of Prior Knowledge Inference (PKI). In another study, the most informative region of each frame in laparoscopic videos is extracted using the static version of ``Adaptive Whitening Saliency'' (AWS) to be used as input to the CNN \cite{SPR}.

Besides the mentioned deep-learning-based approaches for phase recognition in surgical videos, there are many research efforts focusing on action recognition in regular videos, being customizable to the surgical videos. Regarding action recognition in regular videos, DevNet~\cite{DevNet} has achieved promising results by adopting a spatio-temporal saliency map. LSTA~\cite{LSTA} proposes an attention mechanism to smoothly track the spatially relevant segments in egocentric activities. R(2+1)D~\cite{R(2+1)D} introduces a novel spatiotemporal convolutional block to boost the performance in action recognition. To address 3D CNNs' under-performing in the case of insufficient training examples, a gate-shift module is introduced to turn a static lightweight CNN into a spatiotemporal feature extractor~\cite{GSM}. While the aforementioned approaches provide outstanding results, they are specifically designed for action recognition in case of static backgrounds.

\begin{sidewaystable}
\renewcommand{\arraystretch}{0.9}
\caption{Comparisons among the deep-learning-based workflow recognition approaches. In the ``Dataset'' column, ``NP'' refers to Nonpublic.}
\label{Tab-survey: workflow1}
\centering
\resizebox{\textwidth}{!}{%
\begin{tabular}{ l  P{4.5cm}P{4cm}P{4cm}P{4cm}P{4.5cm}P{4.5cm}P{1cm} }
\specialrule{.12em}{.05em}{.05em} 
Reference & Target Surgery & Proposed Approach & Learning Method & Target Result & Features & Dataset & Year\\\specialrule{.12em}{.05em}{.05em} 
\cite{EndoRCN} & Cholecystectomy & Evaluation & Supervised & Phase classification & CNN-LSTM & Cholec80~\cite{EndoNet} & 2016\\
\cite{EndoNet} & Cholecystectomy & Framework & Supervised & Phase Classification & CNN \& SVM \& HMM & Cholec80~\cite{EndoNet} & 2017\\
\cite{UTL} &  Laparoscopy & Framework & Self-Supervised & Pre-Training with Frame Ordering for Phase Recognition & CNN & NP & 2017\\
\cite{NNPRSD} & Cholecystectomy & Network & Self-Supervised & Remaining Surgery Duration & CNN-LSTM & Cholec80~\cite{EndoNet} & 2017\\
\cite{VSC2017} & Gynecology & Evaluation & Supervised & Shot Classification & CNN-LSTM & NP & 2017\\
\cite{SV-RCNet} & Cholecystectomy & Framework & Supervised & Phase Classification & CNN-LSTM & Cholec80~\cite{EndoNet} & 2018\\
\cite{SPR} & Cholecystectomy & Framework & Supervised & Phase Classification & CNN-LSTM & Cholec80~\cite{EndoNet} & 2018\\
\cite{ULfSM} & Robotic Surgery (da Vinci Surgical System) &  Framework & Self-Supervised & Action Classification &  LSTM Autoencoder & MISTIC-SL (NP) & 2018\\
\cite{LIM} & Cholecystectomy & Framework & Self-Supervised & Pre-training with Remaining Surgery Duration for Phase Classification & CNN-LSTM & Cholec120~\cite{Endovis2015-workflow} & 2018\\
\cite{DPS} & Cataract & Network & Supervised & Joint Instrument Classification \& Phase Recognition & CNN-LTSM \& CNN-GRU & \cite{CataractsChallenge} & 2018\\
\cite{TCBSSL} & Cholecystectomy & Evaluation & Self-Supervised & Pre-training with Contrastive and Ranking Loss for Phase Classification & CNN-LSTM & Cholec80~\cite{EndoNet} & 2018\\
\cite{RSDNet} & Cholecystectomy \& Bypass & Network & Self-Supervised & Remaining Surgery Duration & CNN-LSTM & Cholec120~\cite{Endovis2015-workflow} \& Bypass170 (NP) & 2018\\
\cite{RTE2019} & Cataract & Evaluation & Supervised & Phase Classification & CNN & NP & 2019\\
\cite{DBN2019} & Cholecystectomy & Framework & Active Learning & Instrument \& Phase Classification & Bayesian CNN-LSTM & Cholec80~\cite{EndoNet} & 2019\\
\cite{HFR2019} & Cholecystectomy \& Laparoscopy &  Framework & Semi-Supervised & Hard Frame Detection for Phase Classification & CNN & m2cai16-tool~\cite{m2cai16-tool} \& Cholec80~\cite{EndoNet} & 2019\\
\cite{AAI2019} & Cataract & Evaluation & Supervised & Phase Classification & CNN-RNN & NP & 2019\\

\specialrule{.12em}{.05em}{.05em} 
\end{tabular}}
\end{sidewaystable}



\begin{sidewaystable}
\renewcommand{\arraystretch}{0.9}
\caption{Comparisons among the deep-learning-based workflow recognition approaches. In the ``Dataset'' column, ``NP'' refers to Nonpublic.}
\label{Tab-survey: workflow2}
\centering
\resizebox{\textwidth}{!}{%
\begin{tabular}{ l  P{4.5cm}P{4cm}P{4cm}P{4cm}P{4.5cm}P{4.5cm}P{1cm} }
\specialrule{.12em}{.05em}{.05em} 
Reference & Target Surgery & Proposed Approach & Learning Method & Target Result & Features & Dataset & Year\\\specialrule{.12em}{.05em}{.05em} 
\cite{STR2019} & Laparoscopy & Network &   Supervised & Early Surgery Type Recognition & CNN-LSTM & Laparo425 (NP) & 2019\\
\cite{ASAR2019} & Robotic Surgery (da Vinci Surgical System) &  Framework & Self-Supervised & Action Classification & RNN-Based Generative Model & JIGSAWS~\cite{JIGSAWS, JIGSAWS1, SRG2017} \& MISTIC-SL (NP)  & 2019\\
\cite{RTA2019} & Laparoscopic Sigmoidectomy & Evaluation & Supervised & Phase Classification & CNN & NP & 2020\\
\cite{TeCNO2020} & Cholecystectomy & Network &  Supervised & Phase Classification & Multi-Stage Temporal CNN & Cholec80~\cite{EndoNet} \& Cholec51 (NP) & 2020\\
\cite{LTD2020} & Cholecystectomy & Framework & Semi-Supervised & Teacher/Student approach for Phase Classification & CNN-biLSTM-CRF \&  CNN-LSTM & Cholec120~\cite{Endovis2015-workflow} & 2020\\
\cite{MTRCN} & Cholecystectomy &  Network & Supervised & Joint Phase Classification \& Instrument Presence Detection & CNN-LSTM \& Relatedness Loss Base on Kullback-Leibler (KL) Divergence & Cholec80~\cite{EndoNet} & 2020\\
\cite{SC2020} & Orthopedic Surgery & Framework & Supervised & Phase Classification & CNN-LSTM \& Multimodal Training using Elapsed Time & NP & 2020\\
\cite{ALC2020} & Laparoscopy & Evaluation & Supervised & Phase Classification \& Instrument Segmentation & CNN \& U-Net & NP & 2020\\
\cite{TSM2020} & Laparoscopy & Network & Supervised & Phase Classification & Tow-Stream Mixed CNN (2D-3D CNN) & m2cai16-workflow~\cite{m2cai16-workflow} & 2020\\
\cite{AGR2020} & Robotic Surgery (da Vinci Surgical System) & Framework & Supervised & Action Classification & Reinforcement Learning & JIGSAWS~\cite{JIGSAWS, JIGSAWS1, SRG2017} & 2020\\
\cite{TSSS} & Robotic Surgery & Network & Supervised & Action Segmentation & CNN-RNN & JIGSAWS~\cite{JIGSAWS, JIGSAWS1, SRG2017} \& RIOUS~\cite{TSSS}  (NP) & 2020\\
\cite{CataNet} & Cataract & Framework & Supervised & Remaining Surgery Duration jointly with Phase Classification and Surgeon Experience Classification & CNN-LSTM & NP & 2021\\
\specialrule{.12em}{.05em}{.05em} 
\end{tabular}}
\end{sidewaystable}


\subsection{Remaining Surgery Duration}

High expenditure in the surgical departments of hospitals primarily comes from two problems: (i) underutilization of operation room (OR) resources due to overestimation of surgery duration, and (ii) high patient waiting time due to underestimation of surgery duration~\cite{REMSD}.
In addition to the financial turnover, imprecise OR occupancy estimation adversely affects the patient's comfort and surgical safety due to increasing the awaiting time or anesthesia and ventilation duration. Accurate estimation of surgery duration is a crucial factor towards precise OR occupancy estimation for optimizing OR scheduling and enabling full utilization of OR capacity.

Surgery duration estimation approaches can be broadly categorized into pre-operative and intra-operative approaches. The first-category approaches use operational factors such as assigned surgical team~\cite{REMSD,IPSD2012}, temporal factors~\cite{IPSD2012}, and patient identity information~\cite{ESMPSD}.

Since surgery duration depends on many factors, including surgeon skill level, patient conditions, and intra-operative irregularities, accurate pre-operative surgery duration estimation is impossible. As a concrete example, the duration of cataract surgery depends on many factors, including the surgeon's experience, patient's eye irregularity, surgical risks and complications~\cite{PCST,AAIA}, and cataract hardness. It is reported that even the surgeons and anesthesiologists generally underestimate the surgery duration~\cite{OpeThe}. Therefore, a possible way to enhance surgery duration estimation is through verbal communications with the surgeons~\cite{AUTR}. However, intra-operative communications among the surgeons and clinicians for optimizing OR scheduling is not feasible due to distracting the surgeons, impacting surgical workflow smoothness, and consequently exposing the patient's health to risk~\cite{DiSF}. Hence, modern operation rooms demand an automatic and real-time remaining surgery duration (RSD) estimation approach for OR scheduling and full utilization.

The intra-operative RSD estimation approaches can be split into two groups. The approaches belonging to the first group use sensor information such as right-hand signals~\cite{OTaRM2017}. As acquiring particular signals in every hospital is impossible, the second group's approaches estimate the RSD purely based on visual information acquired from the recorded videos. Since different surgical tasks usually occur in different stages of surgery, many approaches opt for RSD estimation based on phase recognition. Besides, different surgical phases usually need particular instruments. Hence, some approaches use instrument presence information as cues for RSD estimation.

Tables~\ref{Tab-survey: workflow1} and \ref{Tab-survey: workflow2} summarize the properties of state-of-the-art approaches for workflow recognition in surgical videos. We can infer from the tables that recurrent neural networks are commonly used for workflow analysis in surgical videos. Besides, majority of the approaches have focused on Laparoscopy and  Cholecystectomy videos, whereas scant attention has been devoted to the challenges in cataract surgery workflow analysis.




\section{Instrument Recognition and Segmentation}

Recognition, localization, pose-estimation, tracking, and segmentation of surgical instruments are the intermediate steps in many applications of computer-assisted surgery (CAS), ranging from surgical skill assessment and surgical phase estimation to automatic guidance for workflow optimization and decision making~\cite{CASM}. In particular, motion analysis of instruments is a prerequisite for many surgical-skill-assessment approaches~\cite{STMI2007}. The classical instrument recognition and segmentation approaches use the histogram and rotation-invariant hough transform~\cite{CTMI}. 

Tables~\ref{Tab-survey: skill1} and \ref{Tab-survey: skill2} include the configurations of state-of-the-art instrument recognition and segmentation approaches. As listed in the tables, many approaches exploit region-based CNNs such as Faster R-CNN~\cite{RPN} and Mask R-CNN~\cite{MRCNN} for instrument localization, segmentation, and tracking. Some recent approaches opt for pixel-level recognition using U-Net-based architectures and propose various convolutional modules to deal with different instrument segmentation challenges~\cite{RAUNet,BARNet,PAANet}.


\begin{sidewaystable}
\renewcommand{\arraystretch}{0.9}
\caption{Comparisons among the deep-learning-based instrument recognition and skill assessment approaches. In the ``Dataset'' column, ``NP'' refers to Nonpublic.}
\label{Tab-survey: skill1}
\centering
\resizebox{\textwidth}{!}{%
\begin{tabular}{ l  P{4.5cm}P{4cm}P{4cm}P{4cm}P{4.5cm}P{4.5cm}P{1cm} }
\specialrule{.12em}{.05em}{.05em} 
Reference & Target Surgery & Proposed Approach & Learning Method & Target Result & Features & Dataset & Year\\\specialrule{.12em}{.05em}{.05em} 
\cite{STD2017} & Cholecystectomy & Evaluation & Supervised & Instrument Localization & YOLO & Cholec80~\cite{EndoNet} & 2017\\
\cite{TbD2017} & Robotic Surgery & Network & Supervised & Instrument Detection \& Tracking & CNN for instrument-tip detection& \cite{SaRI} & 2017\\
\cite{DLRT} & Robotic Surgery & Evaluation & Supervised & Instrument Localization & Region Proposal Network & \cite{DLRT} & 2017\\
\cite{AMIPE,SRPE2017} & Endoscopy \& Retinal Microscopy & Network & Supervised & Articulated Pose Estimation& CNN +  Bipartite Graph Matching& RMIT~\cite{DDVTRM}, Endovis 2015~\cite{Endovis2015} & 2018\\
\cite{CEIS2018} & Endoscopy & Challenge Results & Supervised & Instrument Detection & Static CNN & \cite{Endovis2015} & 2018\\
\cite{AIS2018} & Robotic Surgery & Comparative Evaluation & Supervised & Instrument Segmentation & U-Net-Based & \cite{RIS2017} & 2018\\
\cite{TDOSA2018} & Robotic Surgery & Framework & Supervised & Instrument Localization \& Skill assessment & Faster R-CNN & Modified \cite{DLRT} & 2018\\
\cite{TDOSA2018} & Cholecystectomy & Evaluation & Supervised & Instrument Localization \& Skill Assessment & Faster R-CNN & \cite{m2cai16-tool} & 2018\\
\cite{MTU} & Cholecystectomy & Network & Supervised & Instrument Detection & Boosted NN-RNN & Cholec80~\cite{EndoNet} \& CATARACTS~\cite{CATARACTS2020} & 2018\\
\cite{DLwCNN2018} & Robotic Surgery & Framework & Supervised & Skill Assessment & CNN & JIGSAWS~\cite{JIGSAWS, JIGSAWS1, SRG2017} & 2018\\
\cite{ESSkd2018} & Robotic Surgery & Network & Supervised & Skill Assessment & CNN & JIGSAWS~\cite{JIGSAWS, JIGSAWS1, SRG2017} & 2018 \\
\cite{DRL2019} & Robotic Surgery & Network & Supervised & Instrument Segmentation & ResNet18-Based &\cite{RIS2017} & 2019 \\
\cite{DBN2019} & Cholecystectomy & Network & Active Learning & Instrument Detection & Bayesian CNN-LSTM & Cholec80~\cite{EndoNet} & 2019\\
\cite{SID2019} & Robotic Surgery & Network & Supervised & Localization & Heatmap-Based Bounding-Box Regression & Endovis 2015~\cite{Endovis2015} \& ATLAS Dione~\cite{DLRT, ATLAS}& 2019\\
\cite{CATARACTS2020} & Cataract Surgery & Challenge Results & Supervised & Instrument Classification & Static CNNs & \cite{CATARACTS2020} & 2019 \\
\cite{LWL2019} & Robotic Surgery & Network &  Supervised & Instrument Segmentation \& Saliency Prediction & Multi-Branch CNN & \cite{RIS2017} & 2019 \\
\cite{ATD2019} & Biportal Endoscopic Spine Surgery & Framework & Supervised &  Instrument's Tip Detection & RetinaNet~\cite{RetinaNet,YOLOv2} & NP & 2019 \\
\cite{EFCNN2019} & Cataract Surgery & Network & Supervised & Instrument Classification and Localization & YOLOv2 Inspired & CaSToL~\cite{SID2019} & 2019 \\
\specialrule{.12em}{.05em}{.05em} 
\end{tabular}}
\end{sidewaystable}



\begin{sidewaystable}
\renewcommand{\arraystretch}{0.9}
\caption{Comparisons among the deep-learning-based instrument recognition and skill assessment approaches. In the ``Dataset'' column, ``NP'' refers to Nonpublic.}
\label{Tab-survey: skill2}
\centering
\resizebox{\textwidth}{!}{%
\begin{tabular}{ l  P{4.5cm}P{4cm}P{4cm}P{4cm}P{4.5cm}P{4.5cm}P{1cm} }
\specialrule{.12em}{.05em}{.05em} 
Reference & Target Surgery & Proposed Approach & Learning Method & Target Result & Features & Dataset & Year\\\specialrule{.12em}{.05em}{.05em} 
\cite{RAUNet} & Cataract Surgery & Network & Supervised & Instrument Segmentation & Dealing with Specular Reflection &  Cata7 (NP) & 2019 \\
\cite{ITP2019} & Robotic Surgery & Network & Supervised & Instrument Segmentation & Temporal-Prior Guided U-Net & \cite{RIS2017} & 2019 \\
\cite{SA3D2019} &  Robotic Surgery & Evaluation & Supervised & Skill Assessment & 3D-CNN~\cite{QuoVadis} \& Temporal Segment Network~\cite{TSN} & JIGSAWS~\cite{JIGSAWS, JIGSAWS1, SRG2017}  & 2019 \\
\cite{ESS2020} & Robotic Surgery & Evaluation & Supervised & Instrument Segmentation/Tracking \& Skill Assessment & Mask R-CNN & \cite{RIS2017} & 2020 \\
\cite{ASS2020} & Endoscopy & Evaluation & Supervised & Instrument Segmentation/Tracking \& Skill Assessment & NP & FCN & 2020 \\
\cite{SSL2020} & Gastrectomy & Framework & Semi-Supervised & Localization & Addressing Class-Imbalance & NP & 2020\\
\cite{ISINet2020} & Endoscopy & Framework & Supervised & Instance-Based Segmentation & Mask R-CNN~\cite{MRCNN} + FlowNet2~\cite{FlowNet2}& \cite{RIS2017,RSS2018} & 2020\\
\cite{TUL2020} & Endoscopy & Network  & Unsupervised & Semantic Segmentation & Cycle GAN & \cite{RIS2017} & 2020\\
\cite{daVinciNet2020} & Robotic Surgery & Framework & Supervised & Instrument Trajectory \& Surgical State & +LSTM& JIGSAWS~\cite{JHU-ISI} and RIOUS~\cite{TSSS} (NP) & 2020\\
\cite{ESS2020} & Robotic Surgery & Framework & Supervised & Instrument Trajectory \& Skill Assessment & Mask R-CNN~\cite{MRCNN} + deep SORT~\cite{DeepSORT} & \cite{RIS2017} & 2020\\
\cite{IRL2020} & Endoscopy & Framework & Supervised & Skill Assessment & Mask R-CNN~\cite{MRCNN} &\cite{Endovis2015, RIS2017}& 2020\\
\cite{BARNet} & Endoscopy, Cataract Surgery & Network & Supervised & Instrument Segmentation & Dealing with Illumination and Scale Variation & \cite{RIS2017}, Cata7 (NP) & 2020 \\
\cite{PAANet} & Endoscopy, Cataract Surgery & Network & Supervised & Instrument Segmentation & Dealing with Specular Reflection and Scale Variation & \cite{RIS2017}, Cata7 (NP) & 2020 \\
\cite{GP2020} & Cataract Surgery & Framework & Supervised & Instrument Classification \& Generalization Evaluation & Static CNNs & \cite{CAT101,CATARACTS2020} & 2020 \\
\cite{PaI-Net2021} & Robotic Surgery & Network & Supervised & Instrument Segmentation & U-Net-Based & \cite{RIS2017} & 2021\\
\specialrule{.12em}{.05em}{.05em} 
\end{tabular}}
\end{sidewaystable}


\section{Skill Assessment}

High-quality surgical procedure is a determining factor in global public health and a contributing factor in healthcare cost reduction. Computer-aided surgical skill assessment can effectively contribute to the quality of surgical procedures through individualized feedback and automatic coaching.
The computer-aided surgical skill assessment approaches can be split into two sub-problems: (1) how to extract the skill-related features, and (2) how to classify the skills based on the extracted features. Accordingly, skill assessment approaches can be broadly categorized into three groups: (i) hand-engineered features plus classical machine-learning-based classification, (ii) hand-engineered features plus deep-learning-based classification, and (iii) deep-learning-based feature extraction and classification.

The approaches belonging to the first category utilize the collected sensor data such as robot kinematics and tool motions with some hand-engineered features for skill assessment. Fard~\etal~\cite{AOPE2019} extract several features such as motion smoothness from the motion trajectories of the instruments collected from sensors. These features are used for support vector machine (SVM) classification, k-nearest-neighbors, and logistic regression to finally classify the surgical skill level as expert or novice. Zia~\etal~\cite{ASSA2018} exploit three different types of features: (1) frequency-based features using discrete Cosine transform (DCT) and discrete Fourier transform (DFT), (2) entropy-based features using approximate entropy (ApEn), and (3) texture features using sequential motion texture (SMT). The dimensionality of these features is then reduced using principal component analysis before being fed into a nearest-neighbor classifier.
Tao~\etal~\cite{SHMM2012} propose to model the surgical skills as time-series using Sparse Hidden Markov Models (SHMM). In the training stage, one SHMM is trained per each skill level. In the testing stage, the corresponding class to the generated model with the highest likelihood is selected as the class of the surgeon's skill. Zia~\etal~\cite{VAMA2018} exploit the combination of accelerometer data and spatio-temporal interest points extracted from the video frames for skill assessment. In another study~\cite{OATS2020}, the duration of different phases manually annotated by a surgeon is used to predict the surgeon's skill.

In the second group, Kim~\etal~\cite{OAITS2019} utilize the motion trajectories computed from the manual annotations of the tip of instruments in the \textit{Rhexis} phase. These trajectories are fed into a two-dimensional CNN to assess the intra-operative skills in this phase. Some recent methods suggest using instance detection and segmentation approaches to extract the position of the instrument using raw video~\cite{IRL2020,TDOSA2018,EFCNN2019}. Jin~\etal~\cite{TDOSA2018} propose to use a region-based CNN to track the instruments in laparoscopic videos. The bounding box information is then used to compute motion trajectories, heat maps, and a timeline of the instrument used for skill assessment.

Hand-engineered feature extraction as a preprocessing step is not only a burdensome process but also may lead to suboptimal results when it comes to complex problems. In contrast, convolutional neural networks, which can provide hierarchical representation from the input data, can perform feature extraction and classification simultaneously. Hence, CNNs are regarded as superior alternatives to hand-engineered-features-based methods. On the other hand, these approaches require a very big dataset which may not be usually available. As an example of the methods belonging to the third group, Fawaz~\etal~\cite{ESSkd2018} utilize a one-dimensional CNN containing three hidden layers to directly transfer 76-D kinematic data to low-dimensional latent variables. Ziheng~\etal~\cite{DLwCNN2018} exploit a deeper architecture consisting of five hidden layers to classify the surgeons' skill level in minimally invasive surgery. Besides, Funke~\etal~\cite{SA3D2019} propose a fully automatic skill assessment approach using a 3-D CNN~\cite{AR3D} for action recognition from video snippets and a temporal segment network~\cite{TSN} for optimal aggregation of the spatio-temporal features of several snippets.

Tables~\ref{Tab-survey: skill1} and \ref{Tab-survey: skill2} include the configurations of state-of-the-art skill assessment approaches. We can deduce from the table that a majority of instrument recognition and skill assessment approaches are based on supervised learning. There are, however, few approaches focus on active~\cite{DBN2019}, semi-supervised~\cite{SSL2020}, and unsupervised~\cite{TUL2020} learning. We can also infer from the tables that skill assessment is mainly performed using instrument recognition. Thus enhancing skill assessment accuracy necessitates accurate pixel-wise instrument recognition.

\chapter{Deblurring Using a Multi-Scale Deconvolutional Neural Network \label{Chapter:Deblurring}}

\chapterintro{
	A common quality impairment observed in surgery videos is blur, caused by object motion or a defocused camera. Degraded image quality hampers the progress of machine-learning-based approaches in learning and recognizing semantic information in surgical video frames like instruments, phases, and surgical actions. This problem can be mitigated by automatically deblurring video frames as a preprocessing method for any subsequent video analysis task. In this chapter, we propose and evaluate a multi-scale deconvolutional neural network to deblur cataract surgery videos. Experimental results confirm the effectiveness of the proposed approach in terms of the visual quality of frames as well as PSNR improvement.
}

{
	\singlespacing This chapter is an adapted version of: 
	
	``Ghamsarian, N., Taschwer, M., and Schoeffmann, K. Deblurring
cataract surgery videos using a multi-scale deconvolutional neural network.
In 2020 IEEE 17th International Symposium on Biomedical Imaging (ISBI)
(2020), pp. 872–876.
''
}

\section{Introduction}
Machine learning algorithms generally suffer from inadequate quality of training data~\cite{UED}.
Comparing the performance of various object detectors trained on quality degraded images via JPEG compression, motion blur, and defocus blur reveals that quality degradation negatively affects object detection performance~\cite{PECNN}. Besides, it is proved that increasing image or video compression ratio (in particular, JPEG and H264) for the test images decreases the semantic segmentation performance of both networks trained with low quality and networks trained with high-quality images~\cite{ILIVC}. A study on the effect of different quality degradation types on OCT image boundary segmentation suggests that U-Nets are vulnerable to noise, contrast reduction, and gamma correction~\cite{OCTIQ}.
In particular, neural networks' performance degrades when trained on blurry images~\cite{IoB} due to the fact that high-frequency components of visual signals are diluted\footnote{CNNs extract high-frequency features in their first layers such as sharp edges, and combine them to more complicated patterns in subsequent layers.} and the network has to undertake two tasks simultaneously: deblurring and the target task.
Moreover, it is known that a CNN trained on high-quality images is often not generalizable to blurry images~\cite{EIB}. 
Hence, we desire sharp images for both training and evaluation purposes.

Taking into account that acquiring surgical videos is a cost-intensive procedure that typically interferes with the main medical purpose of surgery, high-quality video acquisition is often not possible and making use of already captured videos with degraded visual quality is a necessity in practice. We therefore propose to use a separate neural network trained to deblur surgery videos as a preprocessing step, prior to feeding video frames to a machine learning algorithm for the target task. In this study, we focus on deblurring cataract surgery videos to further leverage them for video analysis tasks like instrument detection, operation phase recognition, and surgical action recognition. 

In cataract surgery, the video signal captured by the microscope is recorded and stored for postoperative analysis. Since the focus of the video camera and of the ocular tube used by surgeons need to be adjusted separately, the resulting videos are often very blurry. 

In this study, we aim to address quality degradation in cataract surgery videos resulting from a defocused camera. Since we need sharp ground-truth video frames to measure the differences between blurred and deblurred frames on a meaningful scale, we use Gaussian convolutional filters to simulate focus degradation. Our proposed multi-scale deblurring network is composed of convolutional, deconvolutional, and residual layers and termed as Defocused image Restoration Network (\textit{DRNet}).

In the following section, we review the state-of-the-art image and video deblurring approaches. Section~\ref{Sec-ISBI: Proposed Method} describes the proposed neural network architecture and implementation details. In Section~\ref{Sec-ISBI: Experiments}, the experimental setup is explained and experimental results are presented. The paper concludes with a short discussion in Section~\ref{Sec-ISBI: Discussion}.

\section{Related Work}
\label{Sec-ISBI: Related Work}
Image and video deblurring approaches can be divided into two classes: geometry-based approaches and deep-learning-based approaches~\cite{ASTL1}. Following the revolution of deep-learning-based image processing approaches, several neural network architectures have been proposed to address various blur-related tasks such as blur type deduction (e.g. linear motions, Gaussian blur, out-of-focused, etc.~\cite{BI}), blur modeling~\cite{BIIR}, blur kernel estimation~\cite{WLMD}, and specific deblurring \cite{NBID}). 
With respect to deblurring types, deblurring schemes can be divided into two categories: specific (targeted) deblurring, and blind (universal) deblurring. Targeted methods have been designed to deblur images blurred by a specific kernel, while blind methods assume that no information about the causes of blurring is available.
Since obtaining a dataset containing blurry images and their corresponding natural sharp images is costly, a lot of synthetic blur creation functions have been proposed~\cite{DVDHH, MDW, DVDLAB}. 

Synthetic blurry images can be obtained in two ways: convolution of sharp images with a blur kernel, or averaging a set of consecutive sharp frames of a given video (motion blur simulation). 

In~\cite{DeblurGAN}, a deblurring technique based on generative adversarial networks (GANs)~\cite{URLGAN} is proposed. The authors have suggested a novel blurring kernel based on different types of shaking, making the blurred version of each sample more complex and realistic.
In~\cite{DGFMD}, a global skip connection in the convolutional neural network (CNN) is proposed to make the network more compatible with the deblurring task. 
Not only helps this skip connection mitigate the vanishing gradient problem when training the CNN, but it also feeds gradients from the last layer directly to the first convolutional layer during backpropagation, enabling the network to learn more informative features and converge faster~\cite{VaU}.
It has been shown empirically~\cite{SRSI} that most of the small patches (namely $5 \times 5$ or $7 \times 7$) in a  naturally sharp image recur in its \textit{ideally} downsampled versions. This property is frequently exploited for blind super-resolution. In~\cite{BDIPR}, cross-scale recurrence is employed for deblurring since the small patches in a downsampled version of a blurry image tend to be similar to that of the sharp image. Also, it has been demonstrated in~\cite{MDW} that a multi-scale scheme facilitates the convergence of deblurring networks. Their proposed neural network architecture for deblurring consists of three subgraphs designed to ease the deblurring procedure. Since the amount of blur decreases with downsampling, the first subgraph undertakes the easiest task of deblurring the input image downsampled by a factor of 4 and outputs a residual frame. The second graph uses this deblurred image along with the input image downsampled at $1/2$ of its original scale to produce a deblurred half-size frame. Likewise, using the deblurred output frame at scale $1/2$, the last subgraph outputs the deblurred frame with the same resolution as the input.

\section{Proposed Method}
\label{Sec-ISBI: Proposed Method}
\paragraph*{\textbf{Overview.}}
\begin{figure*}[htbp!]
\centering
\includegraphics[width=1\textwidth]{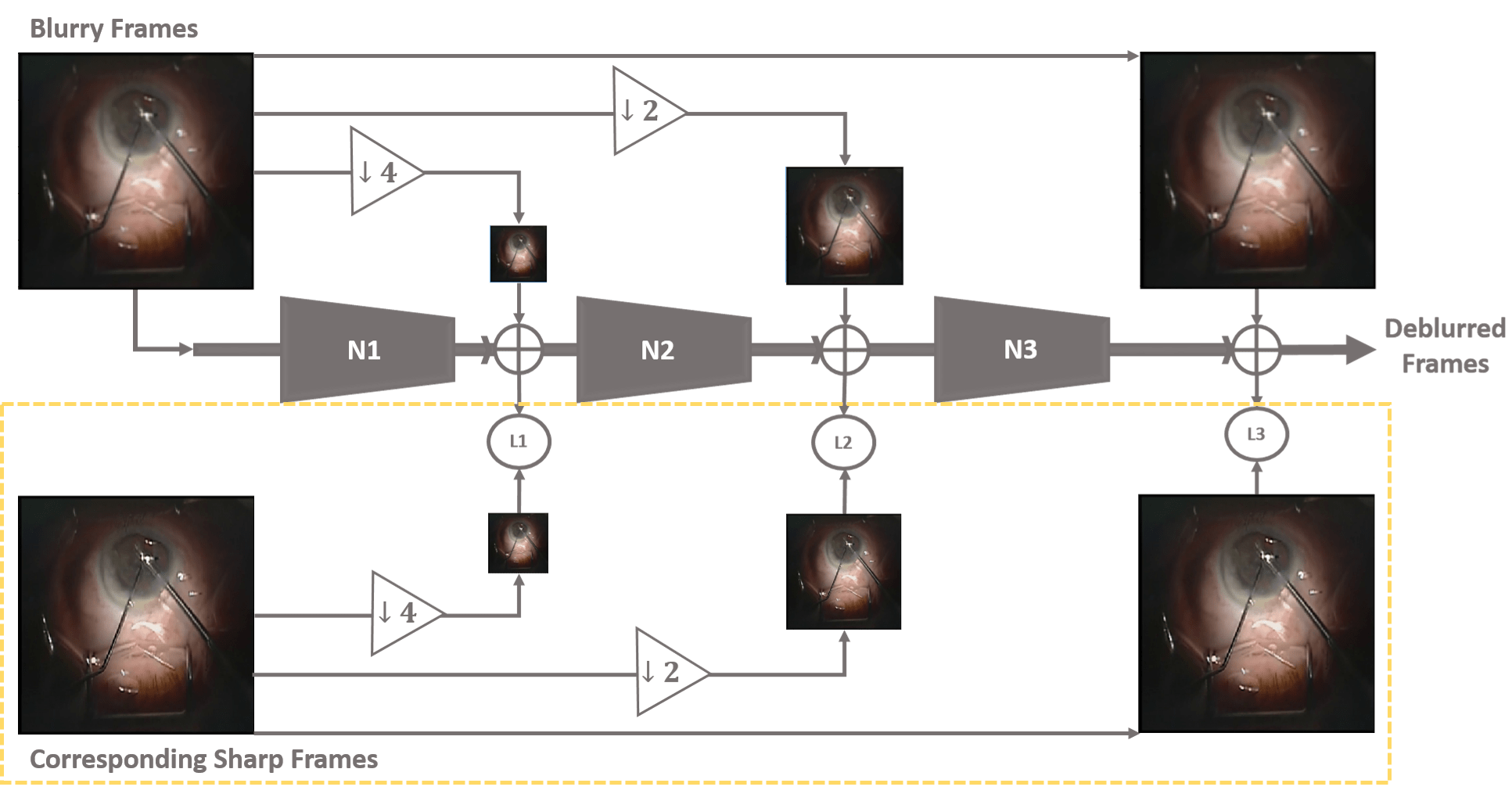}
\caption{The general architecture of the deblurring network, which is inspired by~\cite{MDW}.}
\label{schematic}
\end{figure*}
Inspired by the method suggested in~\cite{MDW}, we propose a multi-scale network in which we use the same number of filter response maps in consecutive layers, enabling the use of residual layers. 
However, our network has much fewer parameters compared to the network suggested in~\cite{MDW} (5.25 million vs. more than 24.5 million trainable parameters) and takes advantage of four residual layers. Besides the above differences, \textit{DRNet} (the proposed network) has been exploited to address defocus and Gaussian blur, while \textit{DeblurNet}~\cite{MDW} is designed to deal with motion blur.

\paragraph*{\textbf{Network Architecture.}}
Our proposed deconvolutional network is illustrated in Figure~\ref{schematic}. It is comprised of three sorts of convolutional layers: down-sampling convolutional layers which abstract the non-informative spatial information and subsequently reduce the features' spatial resolution, while providing more spatial support for each pixel in the down-sampled image; flattening convolutional layers, being responsible for extracting more complicated features from input features; and deconvolutional layers, which account for performing super-resolution on input features.  Each convolutional layer is followed by a batch normalization and a ReLU layer, except for the convolutional layer before the output layer of \textit{N1} (in which we concatenate the blurry image with the residual image). Padding the images by one pixel in all dimensions ensures to avoid the reduction of dimensions after applying flattening convolutional layers (convolutions with stride~1). 
The detailed specification of the proposed network is given in Table~\ref{NetArch}.
\begin{sidewaystable}
\renewcommand{\arraystretch}{1.3}
\caption{Configuration of the proposed deblurring network. The network consists of three sub-networks (SubNet). It includes some down-sampling (stride=2) and flattening (stride=1) convolutional (conv) layers as well as deconvolutional (deconv) layers. Except for the output layers, layers are followed by batch normalization (BN) and ReLU activation layers as operations (Ops). Furthermore, there are two skip connections in sub-network \textit{N1}, and one skip connection in each of the other sub-networks.}
\label{NetArch}
\centering
\resizebox{\textwidth}{!}{
\begin{tabu}{l| c c c c c c c | c c c c c | c c c c c}
\tabucline[1pt]{1-18}
SubNet & \multicolumn{7}{c|}{\textbf{N1}} & \multicolumn{5}{c|}{\textbf{N2}} & \multicolumn{5}{c}{\textbf{N3}}\\\hline
Layer & conv1 & conv2 & conv3 & conv4 & conv5 & conv6 & conv7 & conv8 & conv9 & conv10 & conv11 & \textbf{deconv1} & conv12 & conv13 & conv14 & conv15 & \textbf{deconv2}\\
Kernel& 11 & 7 & 7 & 7 & 3 & 3 & 3 & 5 & 5 & 5 & 5 & 5 & 5 & 5 & 5 & 5 & 5\\
OutCh & 128 & 128 & 128 & 128 & 128 & 128 & 3 & 128 & 128 & 128 & 128 & 3 & 128 & 128 & 128 & 128 & 3\\
Stride & 2 & 1 & 1 & 2 & 1 & 1 & 1 & 1 & 1 & 1 & 1 & 2 & 1 & 1 & 1 & 1 & 2\\
\multirow{2}{*}{Ops} & BN & BN & BN & BN & BN & BN & \multirow{2}{*}{$-$} & BN & BN & BN & BN & \multirow{2}{*}{$-$} & BN & BN & BN & BN & \multirow{2}{*}{$-$}\\
 & ReLU & ReLU & ReLU & ReLU & ReLU & ReLU &  & ReLU & ReLU & ReLU & ReLU & & ReLU & ReLU & ReLU & ReLU & \\
Skip & conv4 & $-$ & $-$ & conv7 & $-$ & $-$ & $-$ & deconv1 & $-$ & $-$ & $-$ & $-$ & deconv2 & $-$ & $-$ & $-$ & $-$ \\
\tabucline[1pt]{1-18}
\end{tabu}
}
\end{sidewaystable}

\paragraph*{\textbf{Implementation Details.}}
We experimentally found that the random normal initialization method of network parameters leads to the fastest convergence and is the most sophisticated one for the deblurring task. Hence, random normal initialization with a mean of zero and standard variation of $0.01$ is applied to all parameters. Adam activation~\cite{Adam} with default values of $\beta_1=0.9$, $\beta_2=0.999$,  $\epsilon=10^{-8}$ and $decay=0.0$ is applied as optimization. The learning rate is fixed to 0.0001 for the first 20 epochs and halved after each subsequent 20~epochs. The network is trained for 60 epochs. Finally, due to limitations of GPU memory, the batch size is set to~$4$.

\section{Experiments}
\label{Sec-ISBI: Experiments}
\paragraph*{\textbf{Dataset.}}
For evaluations, we use the Cataract-101 dataset~\cite{CAT101}, consisting of 101 cataract surgery videos with a resolution of $720\times540$ pixels (PAL standard). The videos  are compressed using H.264/AVC with a frame rate of $25fps$. 

\begin{figure*}[tbp!]
\centering
\includegraphics[width=1\textwidth]{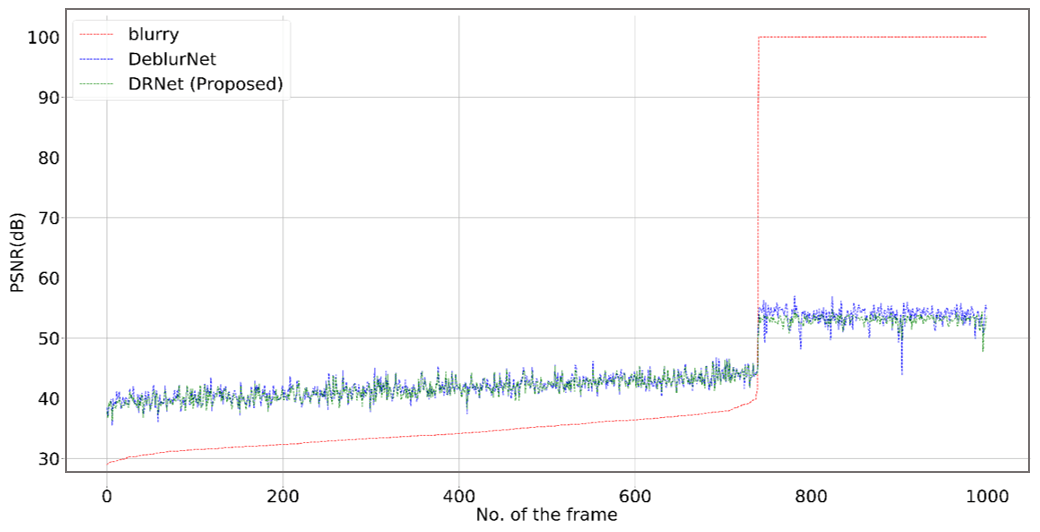}
\caption{Comparative results of the proposed network (\textit{DRNet}) and rival (\textit{DeblurNet}~\cite{MDW}) for $1000$ test frames (the frames are sorted by PSNR of input frames)}
\label{fig-ISBI:PSNR}
\end{figure*}

\paragraph*{\textbf{Training Details.}}
Our proposed neural network is implemented using Keras deep learning framework.
We have manually annotated the videos of our dataset and selected only naturally sharp frames. In order to provide the training set, a Gaussian filter with a randomly selected window size from $w\in\{1,3,5,7\}$ is applied to each sharp frame. In the case of $w=1$, no distortion will be applied. 
In fact, we feed the network with both blurry and sharp images in order to encourage it not to change the content of sharp images as far as possible.
\paragraph*{\textbf{Experimental Results.}}
To evaluate the achievable deblurring performance, we compute the PSNR values between naturally sharp frames and their artificially blurred and deblurred representatives. We also evaluate the performance of the \emph{DeblurNet} network~\cite{MDW} trained using the same configuration as our proposed method.
The results are given in Figure~\ref{fig-ISBI:PSNR} and reveal the following insights: (1) the deblurred versions of blurry frames have clearly better visual quality than their input (higher PSNR values); (2) deblurred versions of naturally sharp frames (right part in the figure) have high PSNR values, meaning that they are not significantly distorted%
\footnote{Images with  $PSNR\ge 50$ dB are perceived as visually not distinguishable from their originals.}%
; (3) our proposed model with less than a quarter of parameters provides smoother fluctuations and more stable results in comparison with the one proposed in~\cite{MDW}. Backed by the last observation, we argue that since our task is deblurring a dataset with very similar content in different frames, using a network with a large receptive field as in \textit{DeblurNet}~\cite{MDW} can lead to overfitting.

\begin{figure*}[bt!]
\centering
\includegraphics[width=1\textwidth]{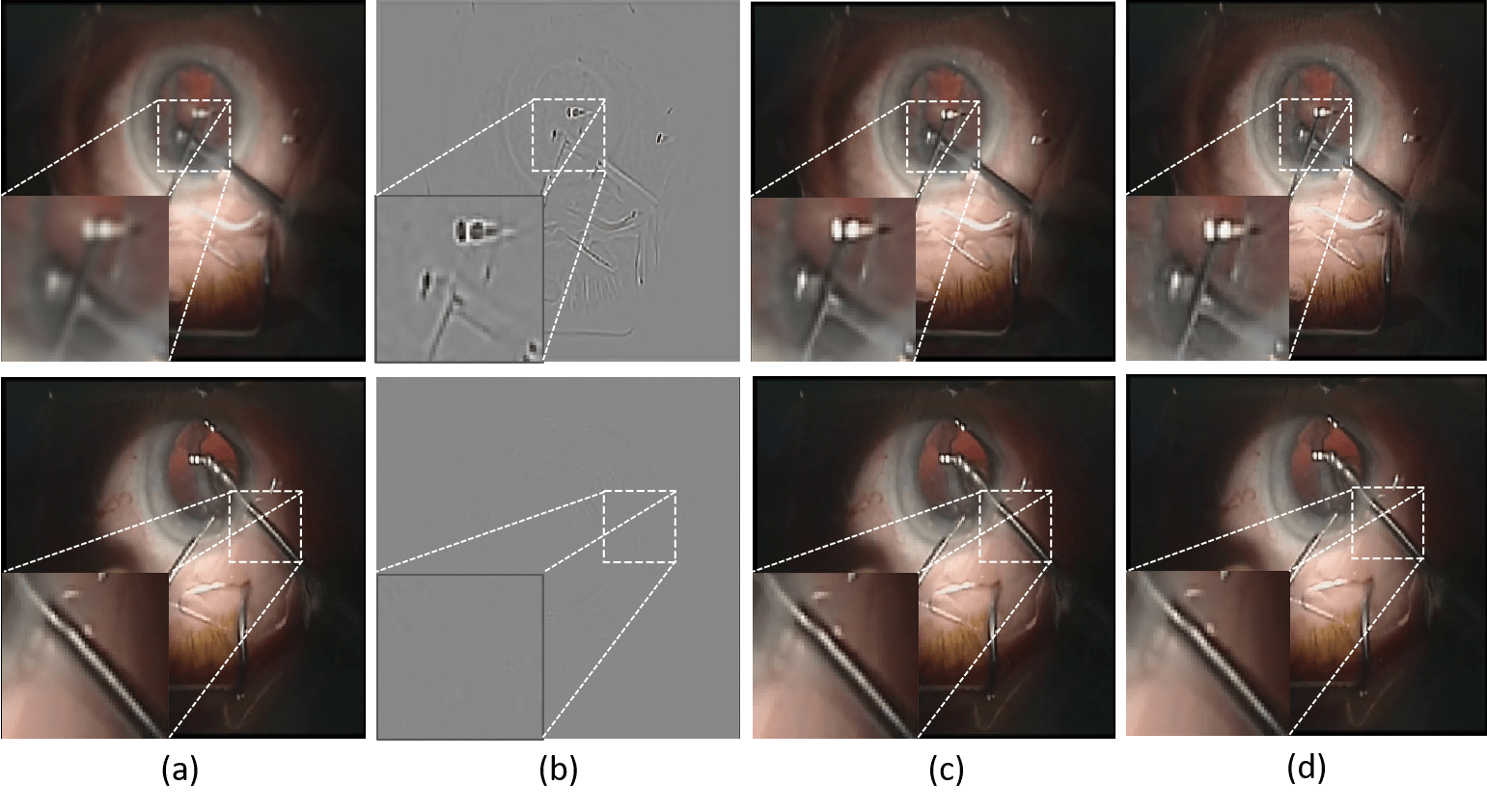}
\caption{Comparative results of \textit{DRNet}. (a) the input of the network (b) the residual frame estimated by the network (c) the output of the network (d) the naturally sharp image. The first row represents a blurry input frame, while the second row corresponds to a sharp image being fed to the network.}
\label{fig-ISBI:visualexamples1}
\end{figure*}

As perceived from Table~\ref{MPSNR}, the differences between mean PSNRs of our \textit{DRNet} network and \textit{DeblurNet}~\cite{MDW} are not significant.
\begin{table}[tbp!]
\caption{Mean PSNR of deblurred video frames ($w$~denotes the size of the Gaussian filter used to blur input frames).}
\label{MPSNR}
\centering
\begin{tabu}{l c c c c c c c c}
\tabucline[1pt]{1-17}
Method & $w=1$ && $w=3$ && $w=5$ && $w=7$\\\hline
\textit{DRNet} & 53.06 && 43.05 && 41.70 && 40.11\\
\textit{DeblurNet}~\cite{MDW} & 53.78 && 43.06 && 41.86 && 40.47 \\
\tabucline[1pt]{1-17}
\end{tabu}
\end{table}
Figure~\ref{fig-ISBI:visualexamples1} shows visual examples that further confirm the correct operation of the proposed network: the deblurred versions (c) of the input images (a) are perceived as almost identical to the naturally sharp images (d). In addition, norms of residual frames (b) are directly related to the amount of blur in input images.

\begin{figure}[bt!]
\centering
\includegraphics[width=0.8\textwidth]{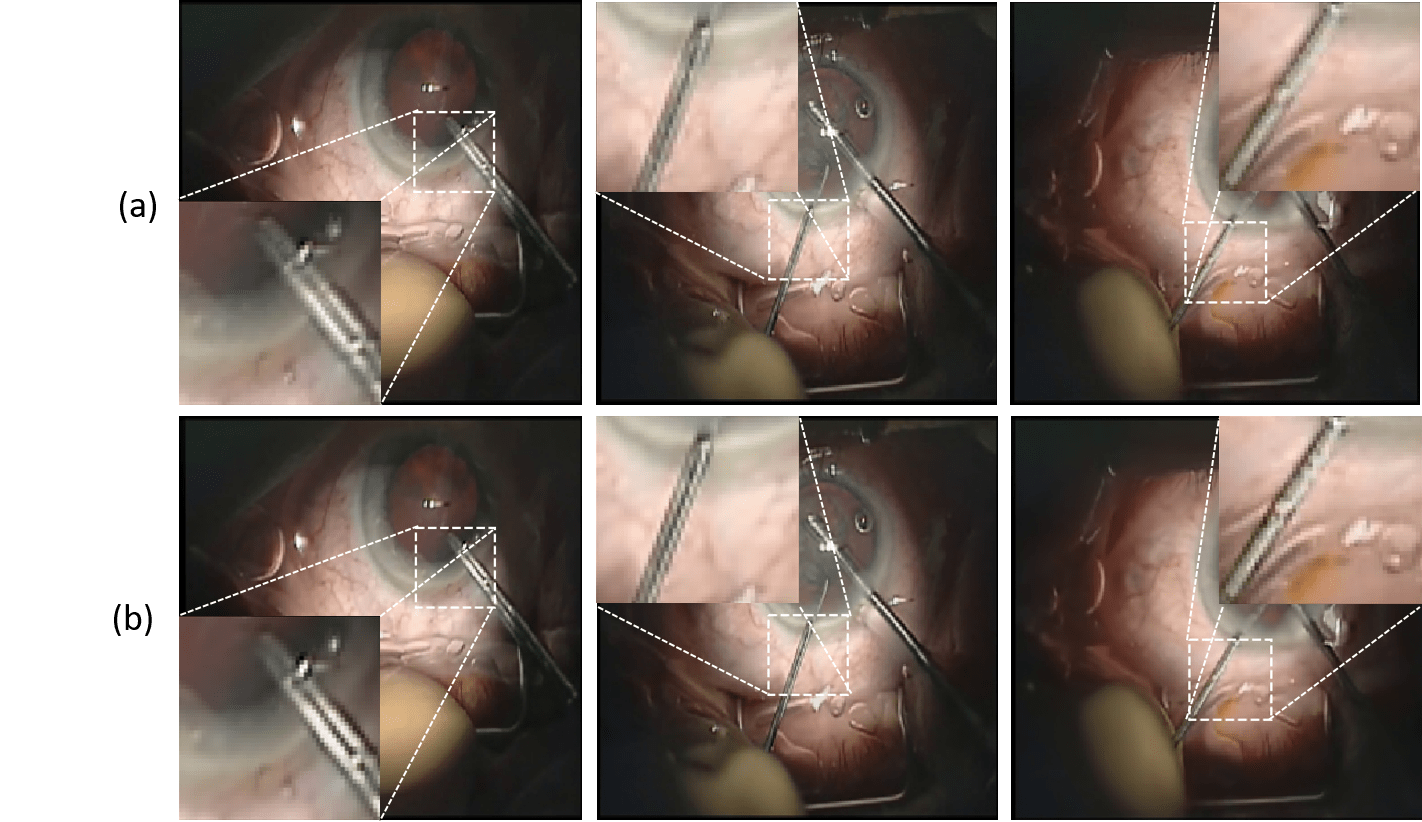}
\caption{Performance of \textit{DRNet} in case of naturally blurry frames. (a) Naturally blurry images (b) The corresponding output of the proposed network.}
\label{fig-ISBI:visualexamples2}
\end{figure}

Since we have trained the network with artificially blurred frames (random selection of Gaussian convolutions), we also want to evaluate whether the proposed network is able to deblur naturally blurred frames. For this purpose, we have manually selected blurry frames 
from the dataset and created visual examples that we qualitatively assess in Figure~\ref{fig-ISBI:visualexamples2}. We can conclude from the figure (which is representative for many examples we have inspected), that the network is also able to deblur such naturally blurred frames (caused by an unfocused camera), validating our assumption that defocus degradation can be simulated by Gaussian filtering.  
\section{Discussion}
\label{Sec-ISBI: Discussion}
The distortion resulting from undesired blur in cataract surgery videos can lead to a substantial drop in the effectiveness of recognition tasks, such as instrument or phase recognition. This may be explained by the fact that noisy data will make it harder for machine learning algorithms to learn the underlying patterns. Cataract surgery video deblurring can, therefore, boost the accuracy of machine-learning-based recognition approaches~\cite{IoB}. Moreover, deblurring will enhance the visual quality of such videos and make them more favorable for clinical documentation or teaching. In this chapter, we have proposed a novel multi-scale residual deconvolutional network to reduce distortion in cataract surgery videos caused by a defocused camera. The proposed network is trained to estimate a residual frame which after adding to the blurry frame will result in reducing Gaussian or defocus blur being inflicted on the underlying sharp frame. Experimental results verify the capability of our proposed \textit{DRNet} network in handling defocus blur. Even though \textit{DRNet} originates from \textit{DeblurNet}, it has much lower complexity in terms of parameters, leading to dependency to less annotations and less tendency to overfit when it comes to surgical videos. Moreover, it has demonstrated more reliable results thanks to residual layers.

Future work could include generalizing the proposed method to other types of surgical videos or even to general deblurring problems.

\chapter{Relevance Detection via Spatio-Temporal Action Localization \label{Chapter:Relevance-Detection}}

\chapterintro{
	To optimize the training procedure with the video content, the surgeons require an automatic relevance detection approach. In addition to relevance-based retrieval, these results can be further used for skill assessment and irregularity detection in cataract surgery videos. In this chapter, a three-module framework is proposed to detect and classify the relevant phase segments in cataract videos. Taking advantage of an idle frame recognition network, the video is divided into idle and action segments. To boost the performance in relevance detection, the cornea where the relevant surgical actions are conducted is detected in all frames using Mask R-CNN. The spatiotemporally localized segments containing higher-resolution information about the pupil texture and actions, and complementary temporal information from the same phase are fed into the relevance detection module. This module consists of four parallel recurrent CNNs being responsible to detect four relevant phases that have been defined with medical experts. The results will then be integrated to classify the action phases as irrelevant or one of four relevant phases. Experimental results reveal that the proposed approach outperforms static CNNs and different configurations of feature-based and end-to-end recurrent networks.
}

{
	\singlespacing This chapter is an adapted version of:
	
	`` Ghamsarian, N., Taschwer, M., Putzgruber-Adamitsch, D., Sarny,
S., and Schoeffmann, K. Relevance detection in cataract surgery videos
by spatio- temporal action localization. In 2020 25th International Conference
on Pattern Recognition (ICPR) (2021), pp. 10720–10727.''

}

\section{Introduction}
\label{sec-ICPR: Introduction}

\begin{figure*}[!bh]
    \centering
    \includegraphics[width=1\textwidth]{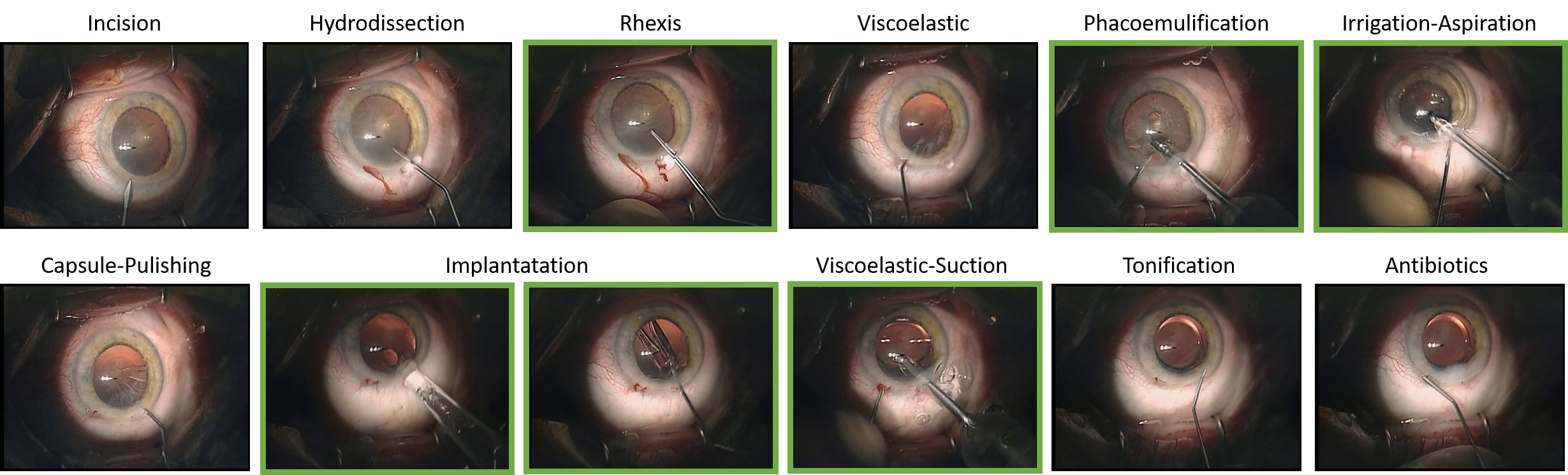}
    \caption{Sample frames of action phases in cataract surgery. Medically relevant phases are illustrated with green borders.}
    \label{fig-ICPR: samples}
\end{figure*}

To systematically study the relevant surgical phases and investigate the irregularities in cataract surgeries, the sole videos are not sufficient. Dealing with tens of thousands of videos to find particular relevant phases and irregularities is burdensome and time-consuming. Hence, the surgeons require a surgical video exploration system which can shorten the surgical training curve (\ie reducing the training time by enabling fast search) and subsequently result in improved overall surgical outcomes. One of the fundamental components of such a system is an automatic phase segmentation and classification tool~\cite{RBE}. 

A cataract surgery video regularly consists of eleven action phases: incision, hydrodissection, rhexis, viscoelastic, phacoemulsification, irrigation-aspiration, capsule polishing, lens implantation, viscoelastic-suction, tonification, and antibiotics. However, not all of the aforementioned phases are equally relevant to clinicians. They consider only \textit{rhexis}, \textit{phacoemulsification}, \textit{irrigation-aspiration} with \textit{viscoelastic-suction}, and \textit{lens implantation} as important from a medical perspective. The intraoperative complications resulting from these phases are reported to have a higher rate compared to that of the irrelevant phases~\cite{ICR2018}. Hence, detecting the aforementioned relevant phases in cataract surgery videos is of prime concern. 

Figure~\ref{fig-ICPR: samples} displays sample frames from relevant and irrelevant phases in cataract surgery videos. Designing an approach to detect and classify the relevant phases in these videos with the frame-wise temporal resolution is quite challenging due to several reasons: (i) These videos may contain defocus blur due to manual adjustment of the camera focus~\cite{DCS}. (ii) Unconscious eye movements and fast motion of the instrument lead to motion blur and subsequently dilution of the discriminative and salient spatial segments.
(iii) As shown in Figure~\ref{fig-ICPR: samples}, the instruments that are regarded as the major difference between relevant phases are highly similar in some phases. This similarity can result in a narrow inter-class deviation in a trained classifier. (iv) The stored videos do not contain any metadata to be used as side information. 

In this chapter, we propose a novel deep-learning-based approach to detect the relevant phases in cataract surgery videos. Our main contributions are listed as follows:
\begin{enumerate}
    \item We present a broad comparison between many different neural network architectures for relevant phase detection in cataract surgery videos, including static CNNs, feature-based CNN-RNNs, and end-to-end CNN-RNNs. This comparison enables confidently scaling up the best approach to various types of surgeries and different datasets. 
    \item We propose a novel framework for relevance detection in cataract surgery videos using cooperative localized spatio-temporal features.
    \item To enable utilizing complementary temporal information for relevance detection, idle frame recognition is proposed to temporally localize the distinct action phases in cataract surgery videos. Besides, using a state-of-the-art semantic segmentation approach, the cornea region in each frame is extracted to localize the spatial content in each action phase. In this way, we avoid inputting the substantial redundant and misleading information to the network as well as providing higher resolution for the relevant spatial content.
    \item Together with this work, we publish a dataset containing the training and testing videos with their corresponding annotations. This public dataset will allow direct comparison to our results. 
    \item The experimental results confirm the superiority of the proposed approach over static, feature-based recurrent, and end-to-end recurrent CNNs. 
\end{enumerate}

In Section~\ref{sec-ICPR: Methodology}, we describe the shortcomings of existing approaches and give a brief explanation of \textit{Mask R-CNN}. We then delineate the proposed relevance detection framework based on action localization termed as \textit{LocalPhase}. The experimental settings are explained in Section~\ref{sec-ICPR: Experimental Setup} and experimental results are presented in section~\ref{sec-ICPR: Experimental Results}. The paper is finally concluded in Section~\ref{sec-ICPR: Conclusion}.

\section{Methodology}
\label{sec-ICPR: Methodology}
\subsection{Shortcomings of Existing Approaches}
Despite using the state-of-the-art baselines and showing good performance in phase recognition, the existing approaches suffer from several flaws, which are discussed as follows: 

\paragraph{(I)} Cataract surgery videos which contain irregularities in the succession of phases are of major importance for clinicians. Hence, it is expected that the trained model can recognize the phases in case of irregularities in the order and duration of them on a frame-level basis. Such a network should not be trained on the time-related and neighboring-phase-related information. Otherwise, the network memorizes the succession of phases and the relative time index of each phase in regular surgeries. Consequently, the network will fail to accurately infer the phases in irregular surgeries~\footnote{As a concrete example, a rarely occurred irregularity in cataract surgery videos is \textit{hard-lens} condition where the surgeons have to perform \textit{phacoemulification} phase in two stages. In the second stage, \textit{phacoemulsification} is performed after \textit{irrigation-aspiration}, while in a regular cataract surgery, the \textit{irrigation-aspiration} phase is always followed by \textit{capsule-polishing} and \textit{viscuelastic}.}. 

\paragraph{(II)} Previous methods in phase recognition either suppose that each video is initially segmented into different actions~\cite{SGC} or perform action recognition with low temporal resolution~\cite{DPS}. However, these approaches are incapable of providing automatic frame-wise labeling.
Similarly, some methods assume that particular side information, such as tool presence signals, is available during training~\cite{RTM}. However, relying purely on surgical videos for workflow analysis is preferred, since (1) providing side information by the surgeons will impose burdens on their constraint schedule, and (2) RFID data or tool usage signals are not ubiquitous in regular hospitals.    

\paragraph{(III)} The existing RNN architectures such as DeepPhase~\cite{DPS}  are not end-to-end approaches and this may result in suboptimal performance. An end-to-end approach can enable correlated spatio-temporal feature learning between recurrent and spatial convolutional layers. DeepPhase also exploits the information from the previous states to infer the corresponding phase to the current state. Due to the high computational complexity of recurrent neural networks, however, the recurrent layer can be unrolled over a limited number of frames. Since some phases may span for several seconds, in the mentioned schemes, the input frames are sampled with a low frame-rate per second (3fps in DeepPhase). Consequently, these schemes are unable to infer the phases with high temporal resolution.

\paragraph{(IV)} To perform classification on barely separable data, more complicated features, and therefore deeper neural network architectures are required. However, a deep neural network entails more annotations and is more vulnerable to overfitting. To obtain higher accuracy with fewer image annotations, the networks are trained on low-resolution inputs. A serious defect of these approaches is the distortion of the relevant content during image downsampling. Regarding phase recognition in cataract surgery videos, these relevant contents include the instruments and cornea. The distortion inflicted by downsampling can negatively affect the classification performance.

\paragraph{(V)} Finally, due to class imbalance in some datasets~\cite{DPS}, the classification results cannot accurately reflect the performance of the trained networks.

\subsection{Mask R-CNN}
Mask Region-based CNN~\cite{MRCNN} (Mask R-CNN) is designed to perform instance segmentation by dividing it into two sub-problems: (1) object detection that is the process of localizing and classifying each object of interest, and (2) semantic segmentation that handles the delineation of the detected object at pixel-level. In the object detection module, a region proposal network (RPN) uses the low-resolution convolutional feature map (CFM) coming from a backbone network. RPN attaches nine different anchors centered around each feature vector in CFM. These anchors have three different aspect ratios to deal with different object types (horizontal, vertical, or squared) and three different scales to deal with scale variance (small, medium, and large-sized objects). The anchor properties along with the computed features corresponding to each anchor are used to decide the most fitted bounding box for each object of interest. In the semantic segmentation module, a fully convolutional network (FCN) is utilized to output instance segmentation results for each object of interest. Using the low-resolution detection of the object detection module, the FCN produces the masks with the same resolution as the original input of the network. 

\subsection{Proposed Approach}
Figure~\ref{fig-ICPR: Block_diagram} demonstrates the overview of the proposed framework which consists of three modules:

\begin{figure*}[!th]
    \centering
    \includegraphics[width=1\textwidth]{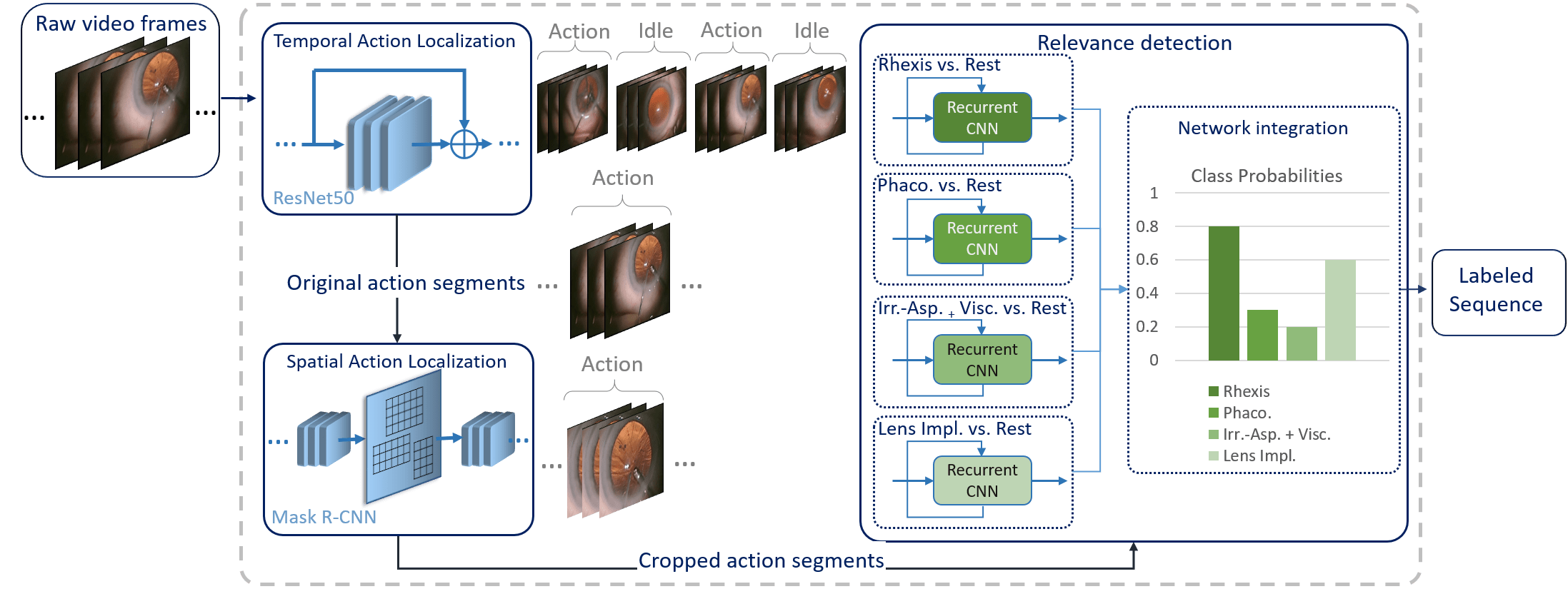}
    \caption{Block diagram of the proposed approach.}
    \label{fig-ICPR: Block_diagram}
\end{figure*}

\paragraph{\textit{Temporal action localization: }}
 Each action phase in cataract surgery is always delimited by two idle phases. An \textit{idle phase} refers to a temporal segment in a cataract surgery video where no instrument is visible inside the frame -- and accordingly, no surgical action is performed. Detecting the idle phases can enhance relevance detection results by enabling the use of complementary spatio-temporal information from the same action phase. We propose to use a static residual network to categorize the frames of cataract surgery into \textit{action} or \textit{idle} frames. This pre-processing step plays a crucial role in alleviating phase classification in the \textit{relevance detection} module. 

\paragraph{\textit{Spatial action localization:}}
The rationale behind spatial action localization is to mitigate the effect of the low-resolution input image on classification performance while retaining all discriminative and informative content for training. Since all the relevant phases in cataract surgery occur inside the cornea, higher resolution of the cornea can significantly boost the classification results. One way to provide a higher-resolution cornea for a network with a particular input size is to detect the cornea, crop the bounding box of the cornea, and use the cropped version instead of the original frame as the input of the network. In addition to providing high-resolution relevant content, this localization approach results in eliminating the redundant information that can cause network overfitting during training. We suggest using \textit{Mask R-CNN} as the state-of-the-art approach in instance-segmentation to detect the cornea and then resize the bounding box of the cornea to fit the input size of the relevance detection module.

\paragraph{\textit{Relevance Detection: }}
\begin{figure}[!t]
    \centering
    \includegraphics[width=0.8\columnwidth]{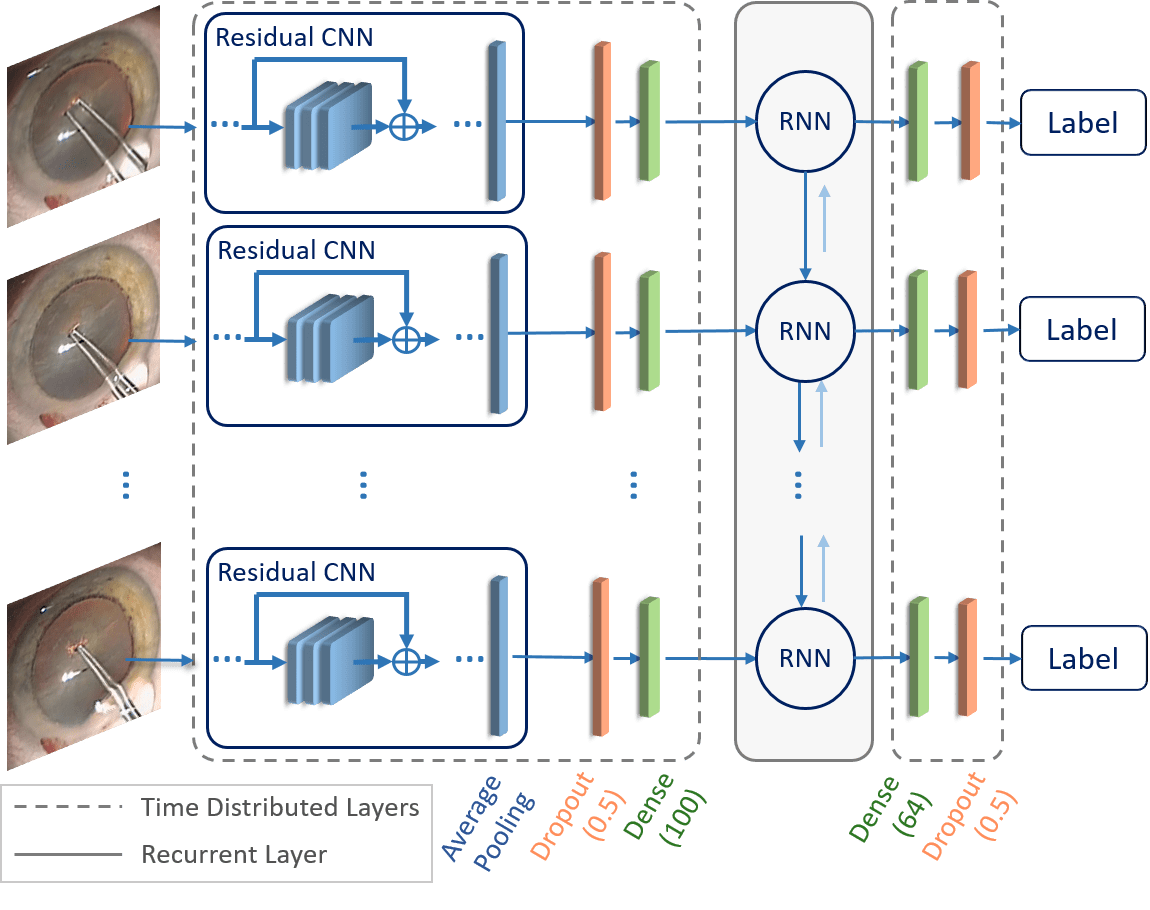}
    \caption{Schematic of the proposed CNN-RNNs for relevance detection.}
    \label{fig-ICPR: Recurrent_model}
\end{figure}
For this module, we propose a recurrent CNN to be trained on the spatiotemporally localized action segments of cataract videos. The network is provided with sequences of the relevance-based cropped frames that contain higher-resolution information for the cornea region and complementary temporal information from the neighboring frames. In an end-to-end training manner, the network can benefit from the seamless integration of spatio-temporal features. Moreover, using back-propagation through time, the recurrent layer encourages the spatial descriptors to learn the shared representations among the input frames while preventing them from learning exceptional irrelevant features. We have experimentally found out that integrated \textit{one-vs-rest} networks provide higher accuracy compared to multi-class classification networks. Thus we propose to train four parallel recurrent networks (Figure~\ref{fig-ICPR: Block_diagram}), each one being responsible for detecting one particular relevant phase.  Fig~\ref{fig-ICPR: Recurrent_model} demonstrates the configuration of the proposed network for relevance detection. The network contains a time distributed residual CNN (ResNet101 in our experiments) that outputs a sequence of spatial feature maps for the sequence of input frames. The network is trained one time per each relevant phase (to detect the relevant phase versus rest). The results of the four networks trained to detect four relevant phases are further integrated by using their output class probabilities in addition to the classification results. If an input is classified as the relevant phase in more than one network, the relevant phase with the highest probability is chosen as the corresponding class to that input.

\section{Experimental Settings}
\label{sec-ICPR: Experimental Setup}
\subsection{Alternative approaches}
For relevance detection, we have implemented and compared many approaches that can be categorized into (i) \textit{static convolutional networks}, (ii) \textit{feature-based CNN-RNN} -- here, we first train the CNN independently, then replace the output layer with recurrent layers (all the layers of CNN are frozen during training the RNN layers), and (iii) \textit{end-to-end CNN-RNN} -- in contrast with feature-based CNN-RNN, the CNN is not frozen in end-to-end training manner. 

To provide a fair comparison, all the networks are trained and tested on the same dataset being created based on the results of the \textit{temporal action localization} module. In the proposed approach, however, we further pass the dataset through the \textit{spatial action localization} module to prove the effectiveness of this module in enhancing the model accuracy.

\subsection{Dataset}

Together with clinicians from \textit{Klinikum Klagenfurt} (Austria), we recorded videos from 22 cataract surgeries and annotated medically relevant phases. The dataset (with all videos and annotations) is publicly released with this work in the following link: \href{https://ftp.itec.aau.at/datasets/ovid/relevant-cat-actions/}{https://ftp.itec.aau.at/datasets/ovid/relevant-cat-actions/}.

\paragraph{\textit{Temporal action localization: }}
For this step, all frames of 22 videos from the dataset are annotated and categorized as \textit{idle} or \textit{action} frames. We have used 18 randomly selected videos from the annotations for training and the remaining videos are used for testing. To prepare a balanced dataset for both training and testing stages, 500 idle and 500 action frames are uniformly sampled from each video, composing 9000 frames per class in the training set and 2000 frames per class in the testing set.

\paragraph{\textit{Spatial action localization: }}
The area of the \textit{cornea} in 262 frames from 11 cataract surgery videos is annotated for the cornea detection network. The network is trained using 90\% of the annotations and tested on the remaining 10\%.

\paragraph{\textit{Relevance detection: }}
For this module, all the action segments of 10 videos are annotated and categorized as \textit{rhexis}, \textit{phacoemulsification} (termed as \textit{Phaco.}), \textit{irrigation-aspiration} with \textit{viscoelastic-suction} (\textit{Irr.-Asp.+Visc.}), \textit{lens implantation} (\textit{Lens Impl.}), and the remaining content (\textit{Rest}). From these annotations, eight videos are randomly selected for training and two other videos are used for testing. Since recurrent CNNs require a sequence of images as input, we have created a video dataset using the annotated segments. Each annotated segment is decoded and 75 successive frames (three seconds) with a particular overlapping step are losslessly encoded as one input video. Due to different average duration of different relevant phases, we use a different overlapping step for short relevant phases to yield a balanced dataset. This overlapping step is one frame for the rhexis and implantation phase, and four frames for other phases. Afterward for each network, 2000 clips per class from the training videos and 400 clips per class from the test videos are uniformly sampled. This amounts to 4000 clips from eight videos as the training set and 800 clips from two other videos as the testing set.

\subsection{Neural Network Models} 
\paragraph{\textit{Temporal action localization: }}For idle-frame recognition, we have exploited ResNet50 and ResNet101~\cite{ResNet50} pre-trained on ImageNet~\cite{ImageNet}. Excluding the top of these networks, the average pooling layer is followed by a \textit{dropout} layer with its dropping probability being equal to 0.5. Next, a \textit{dense} layer with two output neurons and \textit{softmax} activation is added to the network to form the output layer. The classification performances of these networks are compared and the network with the best performance is used for later experiments.

\paragraph{\textit{Spatial action localization: }}For cornea detection, we utilize the \textit{Mask R-CNN} network~\cite{MRCNN, matterport_maskrcnn_2017}. We train the network on two different backbones (ResNet50 and ResNet101) and use the backbone with the best results to produce the cropped input for the relevance detection module.

\paragraph{\textit{Relevance detection: }}For static CNNs, we have used ResNet50 and ResNet101 with the same configuration as for the \textit{temporal action localization} module. The network with the best results is then used as the baseline for both feature-based and end-to-end recurrent networks. Figure~\ref{fig-ICPR: Recurrent_model} shows the shared schematics of CNN-RNNs. In feature-based models, the average-pooling layers of the trained static models are used as a backbone by freezing all of the CNN layers. In contrary to feature-based models, we train the static and recurrent layers of the end-to-end models simultaneously and by starting from the weights initialized from ImageNet. We have compared four different recurrent models: (1) CNN+LSTM in which the recurrent layer includes one LSTM layer, (2) CNN+GRU which contains a GRU layer, (3) CNN+BiLSTM that utilizes a bidirectional LSTM layer, and (4) CNN+BiGRU containing a bidirectional GRU layer. All of the recurrent layers contain five units.

\subsection{Neural Network Settings}
\paragraph{\textit{Temporal action localization: }}The SGD optimizer with $decay = 1e-6$ and $momentum = 0.9$ is set as the optimization function during training. The temporal action localization network is trained for 10 epochs with the initial learning rate $lr_1$ being set to $0.0005$.  The network is then fine-tuned for 10 epochs with $lr_2 = lr_1/5$, and 10 other epochs with $lr_3 = lr_1/10$. To avoid network overfitting, all the layers except for the last 20 layers are frozen during training. Also, \textit{categorical cross-entropy} is used as the loss function.

\paragraph{\textit{Spatial action localization: }} 
The \textit{Mask R-CNN} network pre-trained on the COCO dataset~\cite{COCO} is fine-tuned in an end-to-end manner starting with learning rate being equal to $0.001$. The network is trained for 50 epochs; the initial learning rate is divided by 2, 10, 20 and 100 after epochs 10, 20, 30, and 40, respectively.

\paragraph{\textit{Relevance detection: }}
Table~\ref{Tab-ICPR: hyperparameters} details the settings of hyper-parameters for the proposed relevance detection approach and the rival neural networks simulated to evaluate the effectiveness of the proposed approach. We have performed extensive hyperparameter optimizations to achieve a feasible speed in training while preventing overfitting during training. We came up with different numbers of frozen layers and different learning rates. For instance, the initial learning rate for static networks is set to $10^{-4}$. This learning rate is divided by 2, 10, and 20 after 2, 20, and 30 epochs respectively. Besides, we use the same settings for the SGD optimizer in this module as for the \textit{temporal action localization} module. 
\begin{sidewaystable}
\renewcommand{\arraystretch}{1}
\caption{Training hyperparameters and specification of the proposed and alternative \textit{relevance detection} approaches.}
\label{Tab-ICPR: hyperparameters}
\centering
\resizebox{\columnwidth}{!}{%
\begin{tabular}{ l l l ccccc }
\specialrule{.12em}{.05em}{.05em} 
\multirow{2}{*}{Model} & \multirow{2}{*}{specification} & \multirow{2}{*}{Name} &\multicolumn{5}{c}{Hyper parameters}\\\cdashline{4-8}[0.8pt/1pt] 
&&&Frozen-Layers&Optimizer&Epochs&Learning-Rate&Batch-Size\\\specialrule{.12em}{.05em}{.05em}
\multirow{3}{*}{Static}&ResNet50&CNN50&[1:-20]&&&&\\

&ResNet101&CNN101&[1:-10]&SGD&40&$lr = 10^{-4}, \frac{lr}{2}\mid_{2}, \frac{lr}{10}\mid_{20}, \frac{lr}{20}\mid_{30}$&16\\

&ResNet152&CNN152&[1:-10]&&&&\\\hline

\multirow{4}{*}{\begin{tabular}{@{}l@{}}Recurrent\\(feature-based)\\ResNet50 backbone\end{tabular}}&GRU&GRU-FB&\multirow{4}{*}{CNN}&\multirow{4}{*}{SGD}&\multirow{4}{*}{15}&\multirow{4}{*}{$lr = 10^{-4}, \frac{lr}{2}\mid_{5}$}&\multirow{4}{*}{16}\\

&LSTM&LSTM-FB&&&&&\\

&Bidirectional GRU&BiGRU-FB&&&&&\\

&Bidirectional LSTM&BiLSTM-FB&&&&&\\\hline

\multirow{4}{*}{\begin{tabular}{@{}l@{}}Recurrent\\(end to end)\\ResNet50 backbone\end{tabular}}&GRU&GRU-E2E&\multirow{4}{*}{[1:-10]}&\multirow{4}{*}{SGD}&\multirow{4}{*}{15}&\multirow{4}{*}{$lr = 10^{-4}$}&\multirow{4}{*}{16}\\
&LSTM&LSTM-E2E&&&&&\\
&Bidirectional GRU&BiGRU-E2E&&&&&\\
&Bidirectional LSTM&BiLSTM-E2E&&&&&\\\hline

\multirow{2}{*}{\begin{tabular}{@{}l@{}}Recurrent\\
(end to end)\end{tabular}}&Bidirectional GRU&LocalPhase-G&\multirow{2}{*}{[1:-10]}&\multirow{2}{*}{SGD}&\multirow{2}{*}{15}&\multirow{2}{*}{$lr = 10^{-4}$}&\multirow{2}{*}{16}\\
&Bidirectional LSTM&LocalPhase-L&&&&&\\
\specialrule{.12em}{.05em}{.05em} 
\end{tabular}}
\end{sidewaystable}

Due to the high computational complexity of the end-to-end training approaches, the RNN layer should be just unrolled over a short segment (clip) instead of a complete video. In this study, we assume that the RNN layers can access up to five frames for decision. The sequence generator divides each input video into five temporal segments and randomly chooses one frame from each temporal segment. This random choosing of the frames encourages the network to learn the relationships in diverse temporal distances. It can also act as a temporal data augmentation technique.

\subsection{Data Augmentation Methods}
Data augmentation during training plays a vital role in preventing network overfitting as well as boosting the network performance in case of unseen data. Accordingly, the input frames to all networks are augmented during training using offline and online transformations. Table~\ref{Tab-ICPR:aug} lists the detailed descriptions of augmentation methods utilized for all networks. The listed transformations are selected based on either inherent or statistical features in the dataset. For instance, motion blur and Gaussian blur augmentations are chosen due to having harsh motion blur and defocus blur in our dataset. 
\begin{table}[tbh!]
\renewcommand{\arraystretch}{1}
\caption{Data augmentation methods applied to the classification and segmentation networks.}
\label{Tab-ICPR:aug}
\centering
\begin{tabu}{lcc}
\specialrule{.12em}{.05em}{.05em}
Augmentation Method&Property&Value\\\specialrule{.12em}{.05em}{.05em}
Brightness&Value range&[-50,50]\\
Gamma contrast&Gamma coefficient&[0.5,2]\\
Gaussian blur&Sigma&[0.0, 5.0]\\
Motion blur&Kernel size&9\\
Crop and pad&Percentage&[-0.25,0.25]\\
Affine&Scaling percentage&[0.5,1.5]\\
\specialrule{.12em}{.05em}{0.05em}
\end{tabu}
\end{table}
\subsection{Evaluation Metrics}
We report the performances of \textit{temporal action localization} and \textit{relevance detection} networks using the common classification metrics namely precision, recall, accuracy, and F1-score. For the \textit{spatial action localization} network, we evaluate the performance using average precision over recall values with different thresholds for Intersection-over-Union (IoU), as well as mean average precision (mAP) over IoU in the range of 0.5 to 0.95.
\begin{table}[!thp]
\renewcommand{\arraystretch}{1}
\caption{Instance detection and segmentation results of \textit{spatial action localization} module.}
\label{Tab-ICPR: instance-segmentation}
\centering
\begin{tabular}{ l c c c }
\specialrule{.12em}{.05em}{.05em} 
 Backbone &\multicolumn{3}{c}{Mask Segmentation}\\\specialrule{.12em}{.05em}{.05em} 
&$mAP_{80}$&$mAP_{85}$&$mAP$\\\cdashline{2-4}[0.6pt/1pt]
ResNet101&1.00&0.92&0.89\\
ResNet50&1.00&1.00&0.88\\
\specialrule{.12em}{.05em}{.05em} 
 &\multicolumn{3}{c}{Bounding-Box Segmentation}\\\specialrule{.12em}{.05em}{.05em}
&$mAP_{80}$&$mAP_{85}$&$mAP$\\\cdashline{2-4}[0.6pt/1pt]
ResNet101&1.00&1.00&0.95\\
ResNet50&1.00&1.00&0.94\\
\specialrule{.12em}{.05em}{.05em} 
\end{tabular}
\end{table}
\begin{figure}[!thpb]
    \centering
    \includegraphics[width=0.8\textwidth]{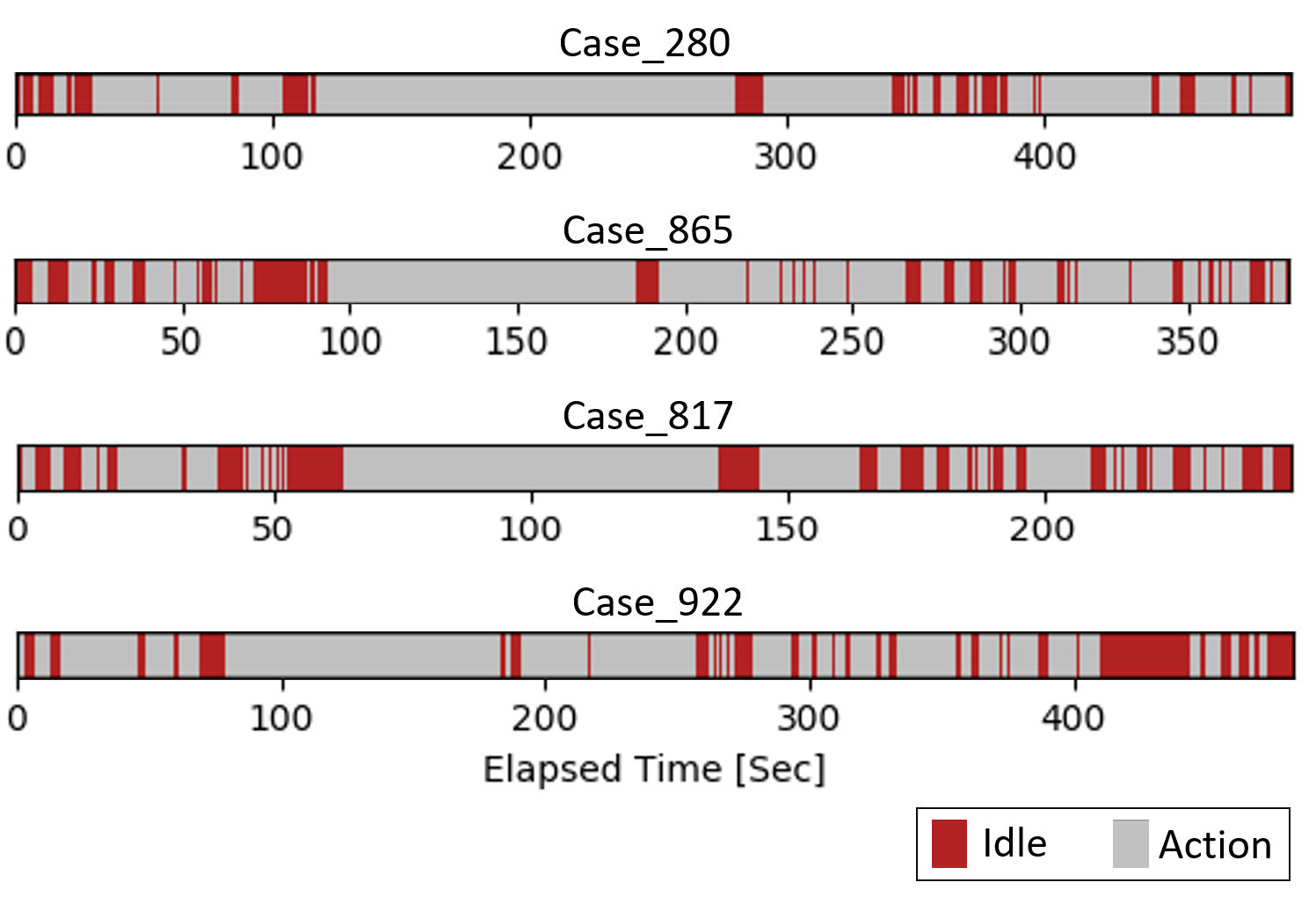}
    \caption{Pattern of \textit{temporal action localization} for four representative videos.}
    \label{fig-ICPR: Idle-pattern}
\end{figure}

\section{Experimental Results and discussion}
\label{sec-ICPR: Experimental Results}
\subsection{Temporal Action Localization}
\begin{table}[th!]
\renewcommand{\arraystretch}{1}
\caption{Precision, Recall, F1-Score, and accuracy of \textit{temporal action localization} module.}
\label{Tab-ICPR: Classification_report_Idle}
\centering
\begin{tabular}{lccccc}
\specialrule{.12em}{.05em}{.05em}
 Network & Class & Precision & Recall & F1-Score & Accuracy \\\specialrule{.12em}{.05em}{.05em}
 \multirow{3}{*}{ResNet50} & Action&  1.00 & 0.85 & 0.92 & \\
 & Idle & 0.87 & 1.00 & 0.93 &\\\cdashline{2-5}[0.6pt/1pt]
 & Macro avg &  0.93 & 0.92 & 0.92 & 0.92\\\hline
 \multirow{3}{*}{ResNet101} & Action & 0.99 & 0.88 & 0.93 &\\
 & Idle & 0.89 & 0.99 & 0.94 &\\\cdashline{2-5}[0.6pt/1pt]
 & Macro avg &  0.94 & 0.94 & 0.94 & 0.94\\\hline
 \multirow{3}{*}{ResNet152} & Action & 0.73 & 0.78 & 0.76 &\\
 & Idle & 0.77 & 0.72 & 0.74 &\\\cdashline{2-5}[0.6pt/1pt]
 & Macro avg & 0.75  & 0.75 & 0.75 & 0.75\\\specialrule{.12em}{.05em}{.05em}
 \multirow{3}{*}{DenseNet121
 } & Action & 0.99 & 0.87 & 0.93 &\\
 & Idle & 0.89 & 0.99 & 0.93 &\\\cdashline{2-5}[0.6pt/1pt]
 & Macro avg &  0.94 & 0.93 & 0.93 & 0.93\\\hline
 \multirow{3}{*}{DenseNet169
 } & Action & 0.96 & 0.90 & 0.93 &\\
 & Idle & 0.90 & 0.96 & 0.93 &\\\cdashline{2-5}[0.6pt/1pt]
 & Macro avg & 0.93 & 0.93 & 0.93 & 0.93\\\hline
 \multirow{3}{*}{DenseNet201} & Action & 0.98 & 0.88 & 0.93 &\\
 & Idle & 0.89& 0.99 & 0.93 &\\\cdashline{2-5}[0.6pt/1pt]
 & Macro avg & 0.94  & 0.93 & 0.93 & 0.93\\\specialrule{.12em}{.05em}{.05em}
 \multirow{3}{*}{VGG16
 } & Action & 0.98 & 0.95 & 0.96 &\\
 & Idle & 0.95 & 0.99 & 0.97 &\\\cdashline{2-5}[0.6pt/1pt]
 & Macro avg & 0.97 & 0.97 & 0.97 & 0.97\\\hline
 \multirow{3}{*}{VGG19
 } & Action & 0.99 & 0.96  & 0.97 &  \\
 & Idle & 0.96 & 0.99 & 0.97 &\\\cdashline{2-5}[0.6pt/1pt]
 & Macro avg & 0.97  & 0.97 & 0.97 & 0.97\\
\specialrule{.12em}{.05em}{.05em}
\end{tabular}
\end{table}

Table~\ref{Tab-ICPR: Classification_report_Idle} reports the main classification metrics for \textit{temporal action localization} using ResNet50 and ResNet101. As can be perceived, both models are fairly accurate and show a close performance, with the F1-score of ResNet101 being $1\%$ better than ResNet50. Thus we use ResNet101 for further experiments. The reason why we have a lower rate of recall for action phases is rooted in the inherent problems in the dataset. The harsh motion blur in some action frames distorts the instruments' spatial content and makes them even invisible for the human eye. To retrieve these wrong predictions, we use a temporal mean filter with a window size of 15 (around 0.5 seconds) as a post-processing step. Figure~\ref{fig-ICPR: Idle-pattern} illustrates the filtered \textit{temporal action localization} results for four representative cataract surgery videos.

\subsection{Spatial Action Localization}
The mask segmentation and bounding-box detection results for cornea tracking are presented in Table~\ref{Tab-ICPR: instance-segmentation}. It should be noted that the bounding-box segmentation results based on instance segmentation networks are much more accurate compared to that of the object detectors. This is the reason why we use \textit{Mask R-CNN}, although we just need the bounding-box of the cornea. The figures for bounding-box segmentation affirm that both networks can detect the cornea with at least $0.85\%$ IoU. Since the network trained with the ResNet101 backbone shows $1\%$ higher $mAP$ compared to that with ResNet50 backbone, this trained network will be used for further experiments.

\subsection{Relevance Detection}

\begin{sidewaystable}
\renewcommand{\arraystretch}{1.1}
\caption{Precision, Recall, and F1-Score of the proposed and alternative \textit{relevance detection} approaches.}
\label{Tab-ICPR: precision-recall2}
\centering
\resizebox{\columnwidth}{!}{%
\begin{tabular}{ l  cccccccccccc }
\specialrule{.12em}{.05em}{.05em} 
\multirow{2}{*}{Network}&\multicolumn{3}{c}{Rhexis}&\multicolumn{3}{c}{Phaco.}&\multicolumn{3}{c}{Lens Impl.}&\multicolumn{3}{c}{Irr.-Asp.+Visc.}\\\cdashline{2-13}[0.8pt/1pt] 
&Precision&Recall&F1-Score&Precision&Recall&F1-Score&Precision&Recall&F1-Score&Precision&Recall&F1-Score\\\hline
CNN50&0.81&0.81&0.81&0.88&0.85&0.85&0.83&0.82&0.82&0.74&0.55&0.45\\
CNN101&0.82&0.73&0.71&0.86&0.83&0.83&0.77&0.58&0.49&0.72&0.72&0.72\\
CNN152&0.95&0.95&0.95&0.76&0.71&0.69&0.81&0.74&0.72&0.69&0.63&0.60\\\hline
GRU-FB&0.99&0.99&0.99&0.93&0.92&0.92&\textbf{0.88}&0.85&0.84&0.80&0.73&0.71\\
LSTM-FB&0.95&0.95&0.95&0.90&0.90&0.90&0.85&0.80&0.79&0.77&0.77&0.77\\
BiGRU-FB&0.98&0.98&0.98&0.93&0.92&0.92&0.87&0.82&0.79&0.78&0.79&0.67\\
BiLSTM-FB&\textbf{1.00}&\textbf{1.00}&\textbf{1.00}&0.92&0.91&0.91&\textbf{0.88}&0.84&0.83&0.79&0.76&0.76\\\hline
GRU-E2E&0.98&0.98&0.98&0.92&0.91&0.91&0.83&0.75&0.74&0.78&0.66&0.63\\
LSTM-E2E&0.95&0.94&0.94&0.83&0.83&0.83&0.84&0.79&0.78&0.69&0.66&0.64\\
BiGRU-E2E&0.99&0.99&0.99&0.92&0.91&0.91&0.84&0.76&0.75&0.80&0.76&0.75\\
BiLSTM-E2E&\textbf{1.00}&\textbf{1.00}&\textbf{1.00}&0.93&0.93&0.93&0.80&0.67&0.63&0.74&0.71&0.70\\\hline
LocalPhase-G&0.99&0.98&0.98&0.95&0.94&0.94&0.86&0.85&0.85&\textbf{0.85}&0.80&0.80\\
LocalPhase-L&0.99&0.99&0.99&\textbf{0.96}&\textbf{0.96}&\textbf{0.96}&0.87&\textbf{0.86}&\textbf{0.86}&0.84&\textbf{0.83}&\textbf{0.83}\\
\specialrule{.12em}{.05em}{.05em} 
\end{tabular}}
\end{sidewaystable}

\begin{table}[thpb!]
\renewcommand{\arraystretch}{1.1}
\caption{Accuracy of the proposed and alternative \textit{relevance detection} approaches.}
\label{Tab-ICPR: accuracy}
\centering
\begin{tabular}{ lP{2cm}P{2cm}P{2cm}P{2cm} }
\specialrule{.12em}{.05em}{.05em} 
Network&Rhexis&Phaco.&Lens Impl.&Irr.-Asp.+Visc.\\\hline
CNN50&0.81&0.85&0.82&0.55\\
CNN101&0.73&0.83&0.58&0.72\\
CNN152&0.95&0.71&0.74&0.63\\\hline
GRU-FB&0.99&0.92&0.85&0.73\\
LSTM-FB&0.95&0.90&0.80&0.77\\
BiGRU-FB&0.98&0.92&0.82&0.69\\
BiLSTM-FB&\textbf{1.00}&0.91&0.84&0.76\\\hline
GRU-E2E&0.98&0.91&0.75&0.66\\
LSTM-E2E&0.94&0.83&0.79&0.66\\
BiGRU-E2E&0.99&0.91&0.76&0.76\\
BiLSTM-E2E&\textbf{1.00}&0.93&0.67&0.71\\\hline
LocalPhase-G&0.98&0.94&0.85&0.81\\
LocalPhase-L&0.99&\textbf{0.96}&\textbf{0.86}&\textbf{0.83}\\
\specialrule{.12em}{.05em}{.05em} 
\end{tabular}
\end{table}

The classification reports of the different static, feature-based recurrent, and end-to-end recurrent neural networks are listed in Table~\ref{Tab-ICPR: precision-recall2} and Table~\ref{Tab-ICPR: accuracy}. Considering the static CNNs (namely ResNet50, ResNet101, and ResNet152), we can see different behaviors of a network for different phases. This difference lies in the level of similarity between each target phase and other phases. For instance, the  \textit{Irr.-Asp.+Visc.} phase shares a lot of statistics and visual similarities with the \textit{Phaco.} phase. Since the frames corresponding to \textit{Phaco.} contain more feature variations, all static CNNs tend to classify \textit{Irr.-Asp.+Visc.} frames as \textit{Phaco.} This tension decreases by increasing the number of layers in the network, as it contributes to discriminating more complicated features. On the other hand, networks with more parameters are more prone to overfit during training on small datasets. In summary, ResNet50 and ResNet101 have shown the same level of accuracy on average. Thus we choose ResNet50 having fewer parameters as the baseline for the recurrent networks.

Thanks to the \textit{temporal action localization} module, the feature-based recurrent neural networks have shown noticeable enhancement in classification results, specifically for \textit{rhexis} and \textit{Phaco.} phase. Interestingly, the bidirectional LSTM network can retrieve 100\% of the frames corresponding to the \textit{rhexis} phase. In summation, it can be observed that all the different configurations of the feature-based recurrent networks outperform the static CNNs. 
Regarding the end-to-end training approaches, we can notice some drops in the classification results for \textit{Irr.-Asp.+Visc.} phase and \textit{Lens Impl.} phase. This drop can occur due to an insufficient number of training examples. The end-to-end training approaches are more vulnerable to overfitting due to their high degree of freedom. 

Both configurations of the proposed approach (namely bidirectional GRU and bidirectional LSTM) have achieved superior performance compared to the alternative approaches. Also, it can be perceived from the Table~\ref{Tab-ICPR: accuracy} that our models have the best accuracy in detecting the \textit{Irr.-Asp.+Visc.} that is the most challenging phase to retrieve. With completely identical configuration to BiGRU-E2E, LocalPhase-G which takes advantage of the \textit{spatial action localization} module has achieved more reliable results (3\% gain in F1-score for \textit{Phaco.} phase and 5\% percent gain in F1-score for \textit{Irr.-Asp.+Visc.} phase). Likewise, LocalPhase-L has achieved 3\%  and 13\% higher F1-score for the \textit{Phaco.} and \textit{Irr.-Asp.+Visc.} phase, respectively. These results reveal the influence of high-resolution relevant content on network training as well as the effect of redundant-information elimination on preventing network overfitting.


\section{Conclusion}
\label{sec-ICPR: Conclusion}
Today, considerable attention from the computer-assisted intervention community (CAI) is focused on enhancing the surgical outcomes and diminishing the potential clinical risks through context-aware assistant or training systems. The primary requirement of such a system is a surgical phase segmentation and recognition tool. In this chapter, we have proposed a novel framework for relevance detection in cataract surgery videos to address the shortcomings of the existing phase recognition approaches. Indeed, the proposed approach is designed to (i) work independently of any metadata or side information, (ii) provide relevance detection with a high temporal resolution, (iii) be able to detect the relevant phases notwithstanding the irregularities in the order or duration of the phases, and (iv) be less prone to overfitting in case of the small non-diverse training sets. To alleviate the network convergence and avoid network overfit on small training sets, we have proposed to localize the spatio-temporal segments of each action phase. A recurrent CNN is then utilized to take advantage of this complementary spatio-temporal information by simultaneous training of static and recurrent layers. Experimental results confirm that the networks trained on the relevant spatial regions are more robust against overfitting due to substantially less misleading content. Besides, we have presented the first systematic analysis of recurrent CNN frameworks for phase recognition in cataract surgeries that further confirms the superiority of the proposed approach. 

\chapter{Relevance-Based Compression \label{Chapter:Compression}}

\chapterintro{
	In this chapter, to facilitate cataract surgery video streaming and address storage limitation, we propose a relevance-based compression technique consisting of two modules: (\textit{i}) relevance detection, which uses neural networks for semantic segmentation and classification of the videos to detect relevant spatio-temporal information, and (\textit{ii}) content-adaptive compression, which restricts the amount of distortion applied to the relevant content while allocating less bitrate to irrelevant content. The proposed relevance-based compression framework is implemented considering five scenarios based on the definition of relevant information from the target audience's perspective. Experimental results demonstrate the capability of the proposed approach in relevance detection. We further show that the proposed approach can achieve high compression efficiency by abstracting substantial redundant information while retaining the high quality of the relevant content.
}

{
	\singlespacing This chapter is an adapted version of: 
	
	``Ghamsarian, N., Amirpourazarian, H., Timmerer, C., Taschwer,
M., and Schoffmann, K. ¨ Relevance-based compression of cataract surgery
videos using convolutional neural networks. In Proceedings of the 28th ACM
International Conference on Multimedia (New York, NY, USA, 2020), MM ’20,
Association for Computing Machinery, p. 3577–3585.''
}

\section{Introduction}
\label{Sec-ACMMM: Introduction}

In the field of ophthalmology, there is an ever-increasing demand to record videos from cataract surgeries. These videos are urgently required for teaching purposes -- a fact that is specifically important in ophthalmology, where operations are performed with a microscope, typically allowing for only one additional viewer (\eg a trainee surgeon). Due to documenting every single moment of the surgery, much better than any textual report would do, these videos are also used for other purposes such as forensics, surgical quality assessment, and post-operative case investigations. 

Recording videos of cataract surgery, however, requires huge storage space being generally not available in hospitals~\cite{IoCoP}. Assuming an exemplary mid-sized hospital with two operating rooms for ophthalmology, 20 surgeries per room with their duration being around seven minutes, we end up with more than 4.5 hours of videos per day. These videos are often stored with at least 720p resolution at 60 frames per second (fps) and 12 MBit/s \footnote{As we experienced in the hospital of our research partner}, amounting to a storage space of 630 MB per operation, \ie 12.3 GB per day for that hospital. As this can sum up to more than 3 TB per year, the recorded ophthalmic videos are often deleted after a few days or weeks. This is especially unfortunate as the recorded videos document every subtle detail, including irregularities and complications, which rarely happen but are particularly important for teaching. On top of that, these videos can be used for knowledge exchange among hospitals through a remote online exploration system to accelerate the training process and improve the surgical outcomes~\cite{RBE}. The remote exploration entails degraded visual quality resulting from the limited transmission bandwidth. The quality degradation in the relevant regions of video can distort the information being crucial to the surgeons. 
Hence, a domain-specific compression technique is needed to penalize the distortion in relevant regions of these videos to allow efficient utilization of this great source of information.    

\begin{figure}[!t]
    \centering
    \includegraphics[width=1\columnwidth]{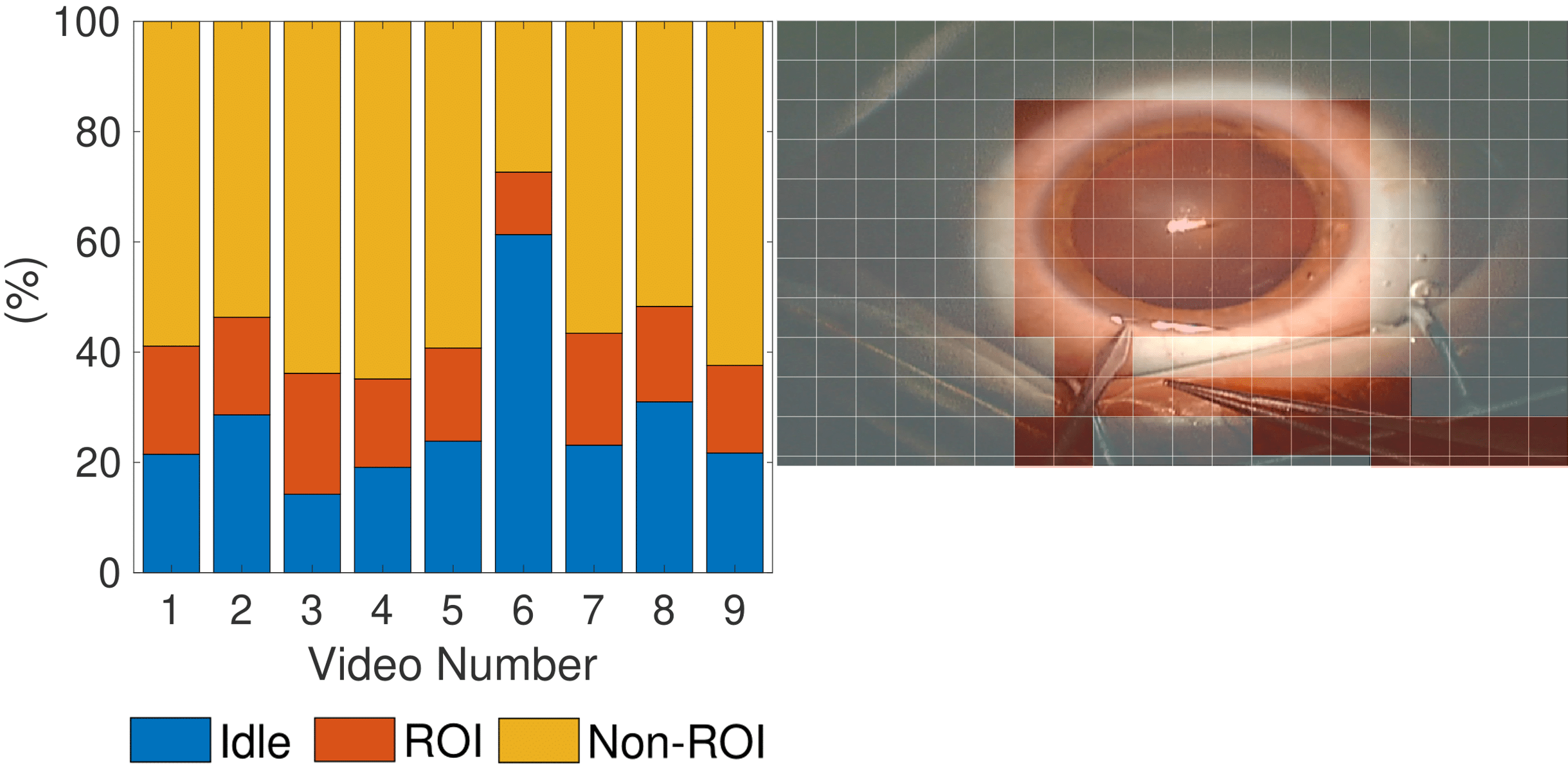}
    \caption{Left: the percentage of CTUs corresponding to relevant and irrelevant content in nine representative cataract surgery videos. Right: HEVC Coding Tree Units (CTUs) aligned with relevant (red) and irrelevant content.}
    \label{Fig-ICPR:CTUs}
\end{figure}
In this work, we investigate the achievable storage space gain with content-adaptive compression of cataract surgery videos using special domain knowledge and varying quantization parameters (QP) for different coding tree units (CTU) in HEVC. We utilize the fact that \textit{(I)} \textit{idle phases} (\ie temporal regions of the video where no instrument is visible) and \textit{(II)} \textit{all spatial content except the inner part of the eye (\ie inner circle of the cornea) and visible instruments} in \textit{action phases} (\ie phases in which a particular surgical action is conducted using at least one instrument) are not relevant for the target audience. Accordingly, we train two different kinds of CNNs, namely \textit{(a)} a static frame-based CNN to automatically detect \textit{idle frames}, and \textit{(b)} a region-based convolutional neural network (Mask R-CNN) to automatically locate the \textit{relevant spatial regions} (cornea and surgical instruments). The results from both CNNs are further utilized to penalize the distortion of relevant content by relevance-based encoding with \HEVC (HEVC) \cite{HEVC}. That is, the relevant content or region of interest (ROI) is compressed yielding high quality by using a default QP, while the irrelevant content is compressed with lower quality by using a higher QP. Figure~\ref{Fig-ICPR:CTUs}
~(right) visualizes the $64\times64$ pixels CTU structure of HEVC for an action frame, which contains two instruments (\textit{primary incision knife} on the left, and  \textit{stabilizing forceps} on the right). 
Only these two instruments and the inner part of the eye (\ie everything inside the cornea) is relevant for users. The percentage of CTUs corresponding to idle frames, non-ROI, and ROI regions in action frames for nine cataract surgery videos are represented in Figure~\ref{Fig-ICPR:CTUs}~(left). It can be noticed that the percentage of the relevant content in cataract surgery videos hovers around $20\%$. The rest of the frame can be encoded with stronger quantization, or even be removed entirely (\eg by pure black inpainting). Our main contributions are summarized as follows:
\begin{enumerate}
    \item We propose a relevance-based compression approach using surgical domain knowledge to compress cataract surgery videos. The proposed method integrates neural networks and HEVC to achieve high compression efficiency by applying more distortion to irrelevant content, while providing high quality for the relevant content.
    \item Considering the target audience of these videos -- which can be expert surgeons, trainees, or computer scientists (for the sake of computerized surgical workflow analysis) -- several definitions for relevant content are possible. Accordingly, we introduce a set of novel scenarios to compress cataract surgery videos to be proportionate to the users' demands.
    \item The compression results show a storage space gain of up to \textbf{63\%} when only slightly reducing the visual quality of the irrelevant content, and storage gain of up to \textbf{68\%} by removing some parts of the irrelevant content.
    \item The proposed relevance-based compression approach is generalizable to all sorts of surgical videos where the domain knowledge is required for the extraction of relevant content.
\end{enumerate}
To ensure reproducibility, the selected dataset (\ie video files with annotations) has been publicly released in the project website:\\ \href{http://ftp.itec.aau.at/datasets/ovid/CatRelevanceCompression/}{http://ftp.itec.aau.at/datasets/ovid/CatRelevanceCompression/}.

The remainder of the paper is organized as follows. 
In Section~\ref{Sec-ACMMM:relatedwork}, we position our approach in the literature by reviewing the related work on ROI-based compression and instance segmentation. 
We detail our method termed as RECOV (RElevance-based Compression of Ophthalmic Videos) in Section~\ref{Sec-ACMMM:Proposed Approach}.  
The experimental setup is presented in Section~\ref{Sec-ACMMM:Experimental Setup}. The relevance detection module of the proposed method is evaluated in Section~\ref{Sec-ACMMM:Relevant Detection Results}. We then evaluate the achievable compression efficiency using videos from nine cataract surgeries adopting content-adaptive HEVC in Section~\ref{Sec-ACMMM:Compression Results}. We finally conclude the paper in Section 7.

\section{Related Work}
\label{Sec-ACMMM:relatedwork}
\subsection{ROI-Based Compression}
\begin{figure*}[thbp!]
\centering
\includegraphics[width=1\textwidth]{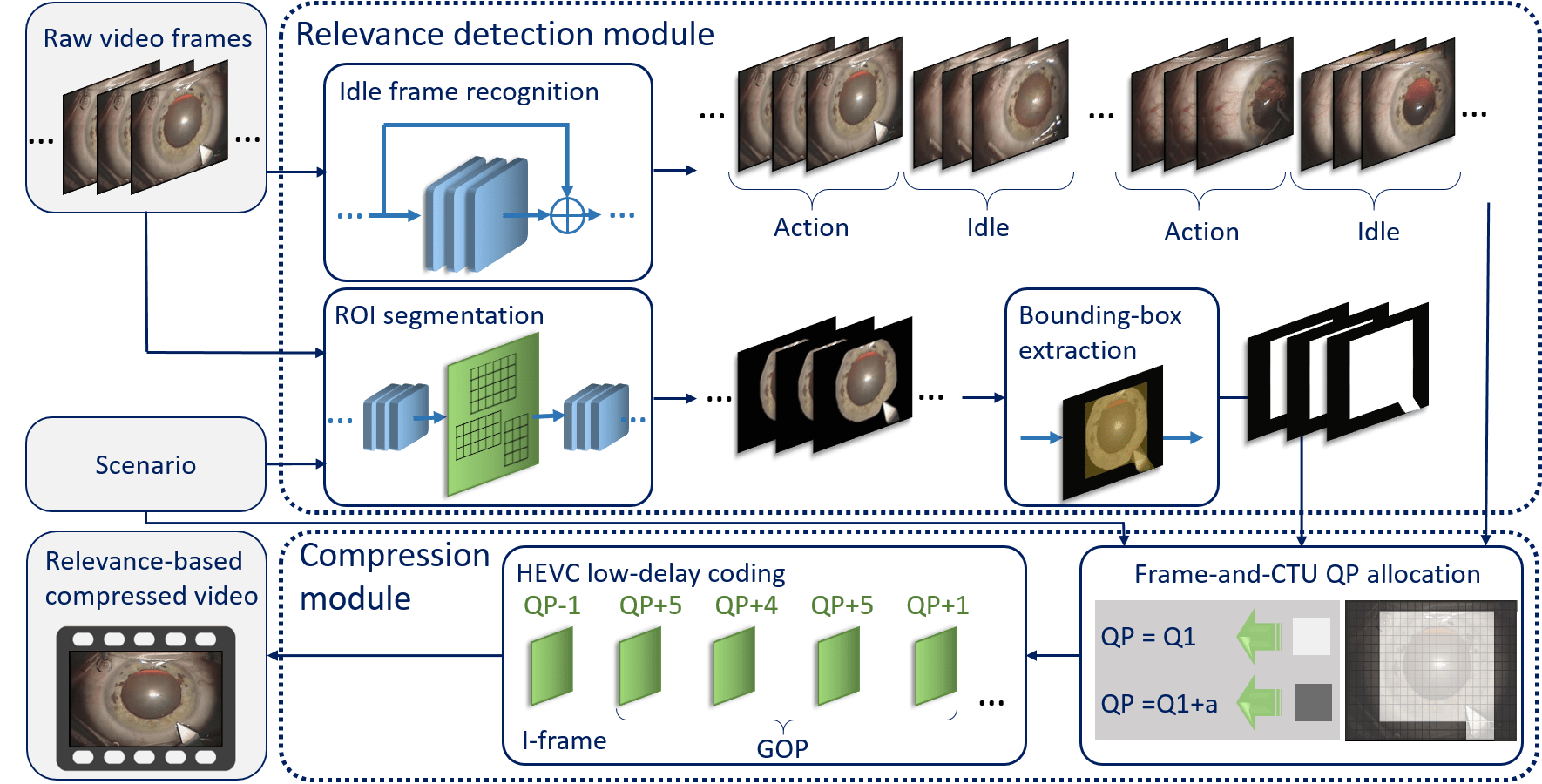}
\caption{Overview of the proposed content-adaptive cataract surgery video compression framework.}
\label{Fig-ICPR:block-diagram}
\end{figure*}
ROI-based compression can be modeled as a rate-distortion problem where ROI quality is of prime concern. 
It can be generally split into two problems: \textit{(i)} how to efficiently segment the ROI (\ie ROI detection), and \textit{(ii)} how to compress the ROI segment and background (\ie ROI encoding). ROI detection methods have experienced two stages of development. In the first generation, ROI prediction methods exploit hand-crafted features including \textit{(a)} \textit{local features} such as relation to the border regions~\cite{real-time} and color information~\cite{RAC, LDS}, \textit{(b)} \textit{global features} such as histogram and contrast, and \textit{(c)} \textit{dynamic features} such as motion-based saliency estimation~\cite{SA}. Some methods have used a combination of low-level hand-crafted features with CNN's outputs to improve the saliency detection results~\cite{VSMDF}.
For ROI encoding, traditional methods use entropy and Huffman coding~\cite{CTim}. Some methods adapted compression standards to be compatible with different ROI encoding strategies. The authors of~\cite{VAG} adapted Advanced Video Coding (AVC) to be capable of using different coding parameters assigned to different macroblocks. In~\cite{SA}, the rate-distortion optimization in AVC is further improved to avoid quality degradation in salient regions in the motion compensation phase. A more practical approach is to utilize image or video compression standards that support ROI encoding, such as JPEG2000~\cite{JPEG2000, NST} and HEVC~\cite{HEVC}. In~\cite{HPM, PHEVC}, an ROI encoding scheme in HEVC is proposed for a hierarchical perception-based model of faces. 
Some recent works propose joint ROI detection-encoding approaches. In~\cite{real-time}, the ROI is extracted during the inter-prediction phase of HEVC, using motion estimation results. Using encoder-decoder networks, the authors of~\cite{E2EROI} propose an ROI compression framework that performs image ROI detection and compression simultaneously. 

In the aforementioned methods, ROI prediction is generally performed using salient-object-detection approaches. Saliency detection approaches attempt to identify the regions in images or videos that grab the human visual system's attention.
Due to the characteristics of cataract surgery videos such as moving background with the same motion properties as ROI, saliency extraction methods are inefficient. In fact, the relevant segments in cataract surgery videos do not contain the saliency characteristics. Moreover, because of fairly similar visual content, hand-engineered features are incapable of achieving high accuracy in relevance-base segmentation. The neural-network-based instance segmentation methods have superseded the traditional approaches in object segmentation. Hence, we focus on instance segmentation using neural networks in the rest of this section to be further used in the proposed relevance-based compression approach. Besides, HEVC/H265~\cite{HEVC} is set to be the state-of-the-art in video coding, yielding superior compression efficiency compared to its predecessors. Thus, we utilize this codec to compress the cataract surgery videos using the detected relevant content.

\subsection{Instance Segmentation}
Since the rise of the Convolutional Neural Networks (CNNs), the ability of machines in recognition tasks is improving from the course level of image classification to the precise level of pixel classification. Before the rise of deep-learning-based semantic segmentation, successful methods relied on hand-crafted features. These features were used in traditional methods such as Boosting or Random Forests to classify the central pixel of each input patch, independently and regardless of the patch's relative spatial information~\cite{RF}. Conditional Random Fields were then employed to smooth these noisy predictions and improve the classification accuracy~\cite{CRF}.

The deep-learning-based instance segmentation approaches can be grouped into two categories: \textit{(i)} pixel-based and \textit{(ii)} region-based approaches. Pixel-based approaches aim at performing semantic segmentation by merging the pixel-wise predictions using clustering methods, whereas region-based methods aim at jointly solving object detection and segmentation. The methods in the first category use metric learning~\cite{LocAtt, MetLea, AssEmb}, boundary clustering~\cite{InstanceCut}, conditional random fields~\cite{Pixelwise}, watershed transform~\cite{Watershed}, \etc~Regarding the region-based approaches, Long \etal~\cite{FCN} proposed a method to convert classifier networks to Fully Convolutional Networks (FCN). The resulting networks apply a series of convolutional layers to the input image of arbitrary size and output a pixel-wise probability map of the same size per each semantic category. The major flaw of FCNs is that these networks are unable to distinguish individual object instances and consequently fail to address \textit{instance-aware semantic segmentation}. To provide instance-aware segmentation, some methods belonging to the second category apply object detectors such as \textit{R-FCN}~\cite{RFCN} and \textit{Faster R-CNN}~\cite{FasterRCNN} to detect the bounding box of the target object and predict the mask using the detected region. Building on top of \textit{Faster R-CNN}, \textit{Mask R-CNN}~\cite{MRCNN} performs object detection and instance-aware semantic segmentation simultaneously. For object detection, a feature proposal network (FPN) uses the convolutional feature map (CFM) from a backbone network. FPN attaches nine different anchors centered around each feature vector in CFM with different aspect ratios to deal with different object types and different scales to deal with scale variance. The anchor properties along with the computed features corresponding to each anchor are used to decide the most fitted bounding box for each object. For semantic segmentation, a fully convolutional network uses the low resolution detected objects to finally output instance segmentation results for each object with the same resolution as the original input of the network.
Due to fast training and exceptional performance, the Mask R-CNN network has become a solid baseline in instance segmentation~\cite{MaskScore, OrientedBB, MaskLab}.

\section{Proposed Approach}
\label{Sec-ACMMM:Proposed Approach}
Figure~\ref{Fig-ICPR:block-diagram} shows the block diagram of the proposed cataract surgery video compression approach. Our relevance-based compression approach consists of two modules: (1) \textit{relevance detection} that accounts for classifying the video frames into idle and action frames and classifying the pixels in action frames into ROI and non-ROI pixels, (2) a \textit{compression module} being responsible for compressing the video content using different QPs allocated to different CTUs and different frames. We generally consider two different QPs: one for the CTUs related to the relevant content ($OP_r$), and another one for CTUs related to the irrelevant content ($OP_i$).

Cataract surgery video normally consists of eleven action phases. Each action phase in cataract surgery is delimited by two idle phases, which do not contain any informative content and are regarded as irrelevant phases. To investigate this surgery, besides, not all the action phases are important~\footnote{Rhexis, phacoemulsification, irrigation-aspiration, and lens implantation are the only phases being important for clinicians.}. Since the actions and reactions in important phases do not occur outside of it, the cornea (\ie iris and pupil) is considered as the region-of-interest for cataract surgery. In case that video and image processing approaches such as instrument, phase, and action recognition are to be conducted on the cataract surgery videos, distortion on instruments should also be penalized.  
Everything outside of these regions is not relevant and can be heavily quantized or even be removed. Accordingly, five scenarios for relevant detection based on user preference (cataract surgeons or computer scientist) are proposed in this study:

\begin{itemize}[leftmargin=*]
    \item{\textbf{Scenario I:}} $Action$\\
    In this scenario, it is supposed that idle frames are the only irrelevant parts. Therefore, the whole action frames are considered as ROI.
    \item{\textbf{Scenario II:}} $Action \cap (cornea\cup instrument)$\\
    A primary goal of recording cataract surgeries is to employ advanced technology to analyze the surgeries to achieve superior surgical outputs.
    This profile is based on the assumption that the resulting videos might be exploited in computerized image processing, where the quality of frames in eye and instrument positions play a major role in recognition tasks, such as phase and action recognition. Thus, cornea and instruments in action frames are regarded as the ROI.
    \item{\textbf{Scenario III:}} $Action \cap cornea$ (simple)\\
    In exploration, teaching, and surgical documentation, just the cornea in action frames is regarded as the relevant content. Therefore, the smaller QP ($QP_r$) is allocated to the CTUs belonging to the cornea in action frames, whereas CTUs belonging to the non-ROI regions and idle frames are quantized with the higher QP ($QP_i$).
    \item{\textbf{Scenario IV:}} $Action \cap cornea$ (Luma preference)\\ 
    As chroma channels convey less important information, in this scenario we aim to compress $C_b$ and $C_r$ channels of irrelevant content with more distortion compared to the $Y$ channel. Therefore, the luma components of CTUs belonging to irrelevant content will be compressed by the same $QP_i$ as the previous scenario, while the chroma components will be compressed using $QP_i+\alpha$.
    \item{\textbf{Scenario V:}} $Action \cap cornea$ (removed background)\\
    In this scenario, we attempt to further reduce the required storage by completely removing the content of CTUs not belonging to the cornea segments in both idle and action frames. The cornea-related content in idle frames is kept (and compressed with $QP_i$) to avoid attention distraction during watching the videos.
    
\end{itemize}

\subsection{Relevance Detection Module}
\paragraph{Idle frame recognition: }As mentioned in Section~\ref{Sec-ACMMM: Introduction}, cataract surgery video can be divided into idle and action phases. An idle phase refers to a temporal segment in which no instrument is visible inside the frame. It is typically the moment where surgeons are changing the instruments. As shown in Figure~\ref{Fig-ICPR:CTUs}, usually around 20\% of the frames in a cataract surgery video belongs to the idle frames. As these frames are not relevant from a surgeons' point of view, such frames can be compressed with a higher quantization parameter (\ie resulting in lower quality). To achieve this goal, we suggest employing a classifier network to categorize the cataract surgery frames into idle and action frames. Due to the fast convergence of ResNet50 and ResNet101~\cite{ResNet50},  thanks to the residual layers, these networks will be trained and exploited for idle frame recognition. A mean filter of size 15 is then applied to the labels of consecutive frames to reduce the wrong predictions.

\paragraph{ROI segmentation: }Depending on the selected scenario, the region-of-interest in action frames is extracted using trained Mask R-CNN networks. For the first scenario, the whole action frame is regarded as the ROI. Therefore, ROI segmentation will not be applied in this scenario. In the other scenarios except for II, just the cornea is considered as the ROI. 

We experimentally discovered that the accuracy of mask segmentation drops when trained to jointly segment the instruments and eye. This is due to the fact that in some phases (such as \textit{phacoemulsification}), corresponding regions to the corrupted lens have visually similar properties to the instruments. Accordingly, we suggest training two separate networks to segment the cornea and instrument and combine the results for scenario II.

\paragraph{ROI segmentation: }{Bounding-box extraction:} 
 Since the surgical videos usually undergo particular machine-learning-based image and video processing approaches for effective exploration and investigation, it is crucial to preserve the high quality of relevant content\footnote{The distortion resulting from low-quality compression of the \textit{false negative} detections (\ie the relevant regions which are wrongly classified as irrelevant) can extremely degrade the performance of machine learning approaches in recognition tasks (\eg phase and action recognition). Notably, the diluted high-frequency features in relevant regions hamper the learning process in neural networks.}. To make sure that the relevant content is far less likely to be distorted, we suggest using bounding boxes instead of the masks.
For typical bounding box extraction, the top-left and bottom-right coordinates of the mask are extracted and used as the input for the compression module. Since the instruments are long and angled in relation to the bounding boxes, typical bounding box extraction may lead to bitrate wastage. A more optimized method to accommodate both demands (\ie less distortion for ROI and less bitrate wastage) is to extract an oriented bounding box using the segmentation results~\cite{OrientedBB}. The instrument is first rotated to be positioned vertically, the bounding box is extracted using the rotated instrument, and the extracted bounding box is reversely rotated to fit the actual position of the instrument.

\subsection{Compression Module}
\paragraph{Frame-and-CTU QP allocation: }In this stage, two QPs will be defined for the input video: the smaller one for the relevant content based on the selected scenario, and the larger one for the irrelevant content. For the first scenario, only a frame-based QP allocation will be conducted. The smaller QP ($Q_r$) is allocated to the action frames, while the greater QP ($QP_i = QP_r + \Delta Q$) will be assigned to the idle frames. For the other profiles, the smaller QP ($QP_r$) will be just allocated to the CTUs of the action frames in which there exists at least one pixel from the ROI. In scenario IV, the irrelevant content will be further distorted by assigning $QP_i+\alpha$ to the chroma channels of irrelevant content. These QP masks along with the raw video frames will be fed into the HEVC encoder for compression.

\paragraph{HEVC encoding: } 
In this stage, using the raw video frames and initially allocated QPs, the whole cataract surgery video will be encoded. Since the idle frames are always encoded with a higher QP compared to the action frames, the idle frames should not be exploited as the reference frame for the action frames. Otherwise, the distortion inflicted by the higher QP in idle frames can propagate in subsequent action frames and adversely affect their output quality. Thus, in addition to the first frame of the video, the first frame of each action phase is considered as I-frame. The remaining frames will be set to P-frames according to the HEVC Low-Delay-P (IPPP) configuration
~\cite{HEVC_CTC}. The reason why we encode the remaining frames as P-frame is the inherent temporal properties of the cataract surgery videos. The succession of idle and action phases with different durations makes it difficult to use an encoding profile including bidirectional frames without negatively affecting the visual quality in action frames.

According to the Low-Delay-P configuration~\cite{HEVC_CTC}, the initially assigned QPs are further adjusted to provide better compression efficiency as well as retaining the perceived visual quality. Except for the I-frames, the consecutive frames are split into GOPs (Group Of Pictures) consisting of four frames. As shown in Figure~\ref{Fig-ICPR:block-diagram}, the assigned QP to each CTU in I-frames is subtracted by one, and in successive frames of each GOP is added by 5, 4, 5, and 1, respectively. The video is then encoded using the offset QPs.

\section{Experimental Setup}
\label{Sec-ACMMM:Experimental Setup}

\subsection{Dataset} Without loss of generality, the released Cataract-101 dataset~\cite{CAT101} being collected in 2017-2018 at Klinikum Klagenfurt (Austria) is used to evaluate the proposed method. 

\paragraph{\textit{Classification Networks: }}
For idle frame recognition, all frames of 22 videos from the dataset are annotated and categorized as idle or action frame. From these annotations, 18 videos are randomly selected for training and remaining videos are used for testing. Subsequently, 500 idle and 500 action frames are uniformly sampled from each video, composing 9000 frames per class in the training set and 2000 frames per class in the testing set.

\paragraph{\textit{Segmentation Network: }}
We have annotated the cornea of 262 frames from 11 cataract surgery videos for the eye segmentation network, and the instruments of 216 frames from the same videos for the instrument segmentation network. We trained each network using 90\% of annotations and tested them using the remaining 10\%.

\paragraph{\textit{Relevance-Based Compression:}}
The complete sequences of nine videos (excluding the annotated videos for classification) are selected as representative videos (Figure~\ref{Fig-ICPR:CTUs}) to test the compression efficiency of the proposed relevance-based compression approach.

\subsection{Neural Network Models and Settings:}
\textit{Classification Networks: }For idle-frame recognition networks, we have utilized ResNet50 and ResNet101~\cite{ResNet50} pre-trained on ImageNet~\cite{ImageNet}. We use the average pooling layer of these networks. This layer is followed by a \textit{dropout} layer with its dropping probability being equal to 0.5, before feeding to a \textit{Dense} layer with two output neurons and \textit{Softmax} activation. As we experimentally found out that the networks perform better with Stochastic Gradient Decent (SGD) rather than Adam optimizer, SGD optimizer with $decay = 1e-6$ and $momentum = 0.9$ is used for training. We train the classification networks for 30 epochs with the initial learning rate being set to $0.0005$. The learning rate is divided by 5 after 10 epochs, and by 10 after 20 epochs. To avoid overfitting, the layers before the last 20 layers are frozen during training. Also, \textit{categorical cross-entropy} is used as the loss function. The classification performance of these two networks will be compared and the best network will be used for later experiments.

\paragraph{\textit{Segmentation Network: }}The Mask R-CNN network~\cite{MRCNN, matterport_maskrcnn_2017} as the state-of-the-art method in image segmentation is exploited for ROI segmentation. The network is trained on two different backbones (ResNet50 and ResNet101) and the backbone with the best results is employed in the relevant detection module. The pre-trained network on the COCO dataset~\cite{COCO} is fine-tuned in an end-to-end manner starting from $learning-rate = 0.001$. The network is trained for 50 epochs; the initial learning rate is divided by 2, 10, 20 and 100 after epochs 10, 20, 30, and 40 respectively.
\begin{table}[t!]
\renewcommand{\arraystretch}{1}
\caption{Data augmentation methods applied to the classification and segmentation networks.}
\label{tabACM:aug}
\centering
\begin{tabular}{lccc}
\specialrule{.12em}{.05em}{.05em}
\multirow{2}{*}{Network}&Augmentation&\multirow{2}{*}{Property}&\multirow{2}{*}{Value}\\
&Method&&\\ \specialrule{.12em}{.05em}{.05em}
\multirow{6}{*}{\rotatebox[origin=c]{90}{Classification}}&brightness&percentage&[0.5,1.5]\\
&rotation&degree&[-20,20]\\
&width shift&fraction of width&$\pm$0.1\\
&height shift&fraction of height&$\pm$0.1\\
&zoom&scaling percentage&[0.8,1.2]\\
&shear&intensity&0.15\\\hline
\multirow{6}{*}{\rotatebox[origin=c]{90}{Segmentation}}&brightness&value range&[-50,50]\\
&Gamma contrast&Gamma coefficient&[0.5,2]\\
&Gaussian blur&sigma&[0.0, 5.0]\\
&motion blur&kernel size&9\\
&crop and pad&percentage&[-0.25,0.25]\\
&affine&scaling percentage&[0.5,1.5]\\
\specialrule{.12em}{.05em}{0.05em}
\end{tabular}
\end{table}

\subsection{Data Augmentation Methods} 
To optimize the training procedure and subsequently boost the performance of the network as well as avoiding overfitting, the input images of both networks are augmented applying offline and online transformations. The detailed descriptions of augmentation methods exploited for both types of networks are listed in Table~\ref{tabACM:aug}. These augmentation methods are selected based on the inherent features in the dataset (and verified by several additional experiments). As a concrete example, the cataract surgery videos usually suffer from defocus blur due to manually adjusted focus~\cite{DRNet}. Moreover, unconscious eye movements and fast instrument motions result in severe motion blur in these videos. Some other data augmentation methods such as mirroring are not employed because of the statistical features of the videos in our specific medical domain (\eg the surgeons always insert the instruments from the same side).

\begin{table}[thpb!]
\renewcommand{\arraystretch}{1}
\caption{Classification report of \textit{Idle frame recognition}}
\label{Tab-ACMMM:Classification_report}
\centering
\begin{tabular}{l c c c c}
\specialrule{.12em}{.05em}{.05em}
 Network & Class & Precision & Recall & f1-score \\\specialrule{.12em}{.05em}{.05em}
 \multirow{3}{*}{ResNet50} & Action& 1.00 & 0.85 & 0.92 \\
 & Idle & 0.87 & 1.00 & 0.93 \\\cdashline{2-5}[0.6pt/1pt]
 & Macro avg &  0.93 & 0.92 & 0.92 \\\hline
 \multirow{3}{*}{ResNet101} & Action & 0.99 & 0.88 & 0.93 \\
 & Idle & 0.89 & 0.99 & 0.94 \\\cdashline{2-5}[0.6pt/1pt]
 & Macro avg &  0.94 & 0.94 & 0.94\\
\specialrule{.12em}{.05em}{.05em}
\end{tabular}
\end{table}

\subsection{Evaluation Metrics}
The performances of \textit{classification networks} are compared using the common classification metrics namely precision, recall, and f1-score. For the \textit{segmentation networks}, we report the performance using average precision over recall values with different Intersection-over-Union (IoU) thresholds, as well as mean average precision (mAP) with specific levels of IoU in the range of 0.5 to 0.95. 

To evaluate the compression efficiency, we compute the percentage of storage gain compared to regular compression (considering the whole content of the video as the relevant content). Moreover, we demonstrate the capability of the proposed method in yielding high-quality relevant content using Peak Signal to Noise Ratio (PSNR).

\section{Relevance Detection Results}
\label{Sec-ACMMM:Relevant Detection Results}

\begin{table*}[bthp!]
\renewcommand{\arraystretch}{1}
\caption{Instance detection and segmentation results of Mask R-CNN.}
\label{Tab-ACMMM:instance-segmentation}
\centering
\resizebox{\textwidth}{!}{
\begin{tabular}{ l c c c c c c c}
\specialrule{.12em}{.05em}{.05em} 
Target & Backbone &\multicolumn{3}{c}{Mask Segmentation}&\multicolumn{3}{c}{Bounding-Box Segmentation}\\\specialrule{.12em}{.05em}{.05em} 
&&$mAP_{80}$&$mAP_{85}$&$mAP$&$mAP_{80}$&$mAP_{85}$&$mAP$\\\cdashline{3-8}[0.6pt/1pt]
\multirow{2}{*}{Cornea}&ResNet101&1.00&0.92&0.89&1.00&1.00&0.95\\
&ResNet 50&1.00&1.00&0.88&1.00&1.00&0.94\\\hline
&&$mAP_{60}$&$mAP_{65}$&$mAP$&$mAP_{80}$&$mAP_{85}$&$mAP$\\\cdashline{3-8}[0.6pt/1pt]
\multirow{2}{*}{Instrument}&ResNet 101&0.77&0.65&0.41&1.00&1.00&0.89\\
&ResNet 50&0.58&0.49&0.29&0.64&0.26&0.65\\
\specialrule{.12em}{.05em}{.05em} 
\end{tabular}}
\end{table*}

Table~\ref{Tab-ACMMM:Classification_report} reports the main classification metrics for idle frame recognition using ResNet50 and ResNet101. Overall, it is evident that both networks are capable of discriminating idle frames from action frames with high accuracy. Additionally, both networks have shown quite similar results, with ResNet101 being a little bit more accurate. Looking first at the \textit{precision} results, ResNet50 is perfectly accurate in predicting the action frames, while having 13\% error in detecting the idle frames. ResNet101 has shown more balanced results in terms of precision per action and idle class. Comparing the \textit{recall} values, it can be perceived that ResNet101 is more successful in retrieving the action frames. The \textit{f1-score} is reported to establish a balance between \textit{precision} and \textit{recall} scores. Based on the \textit{f1-score}, ResNet101 shows 2\% higher performance on average. It is worth mentioning that the wrong prediction in the idle class usually occurs at the beginning and last frames of each action phase (\ie during insertion and removal of an instrument). This is because in these frames just a small part of instruments is visible and strong motion blur in some frames contributes to the network confusion between the lid holders and instruments. Since in this work the predicted idle frames are regarded as irrelevant and compressed using a larger QP, it is important to have as a few false positives as possible in the idle class. Accordingly, the trained ResNet101 is used in the idle-frame-recognition stage of the relevance detection module.

\begin{figure}[!t]
    \centering
    \includegraphics[width=0.75\columnwidth]{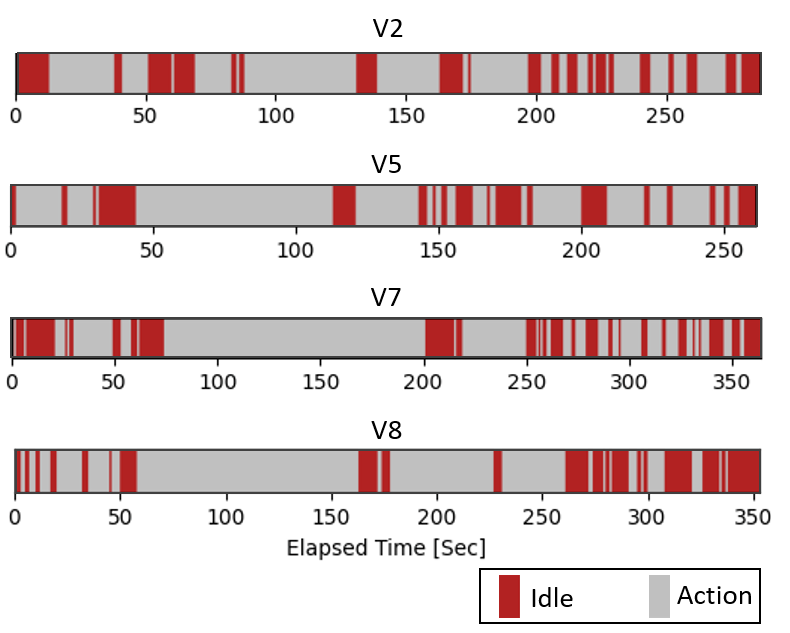}
    \caption{Pattern of idle frames for four videos out of nine representative videos.}
    \label{Fig-ICPR:Idle-pattern}
\end{figure}
\begin{figure*}[!th]
    \centering
    \includegraphics[width=1
    \textwidth]{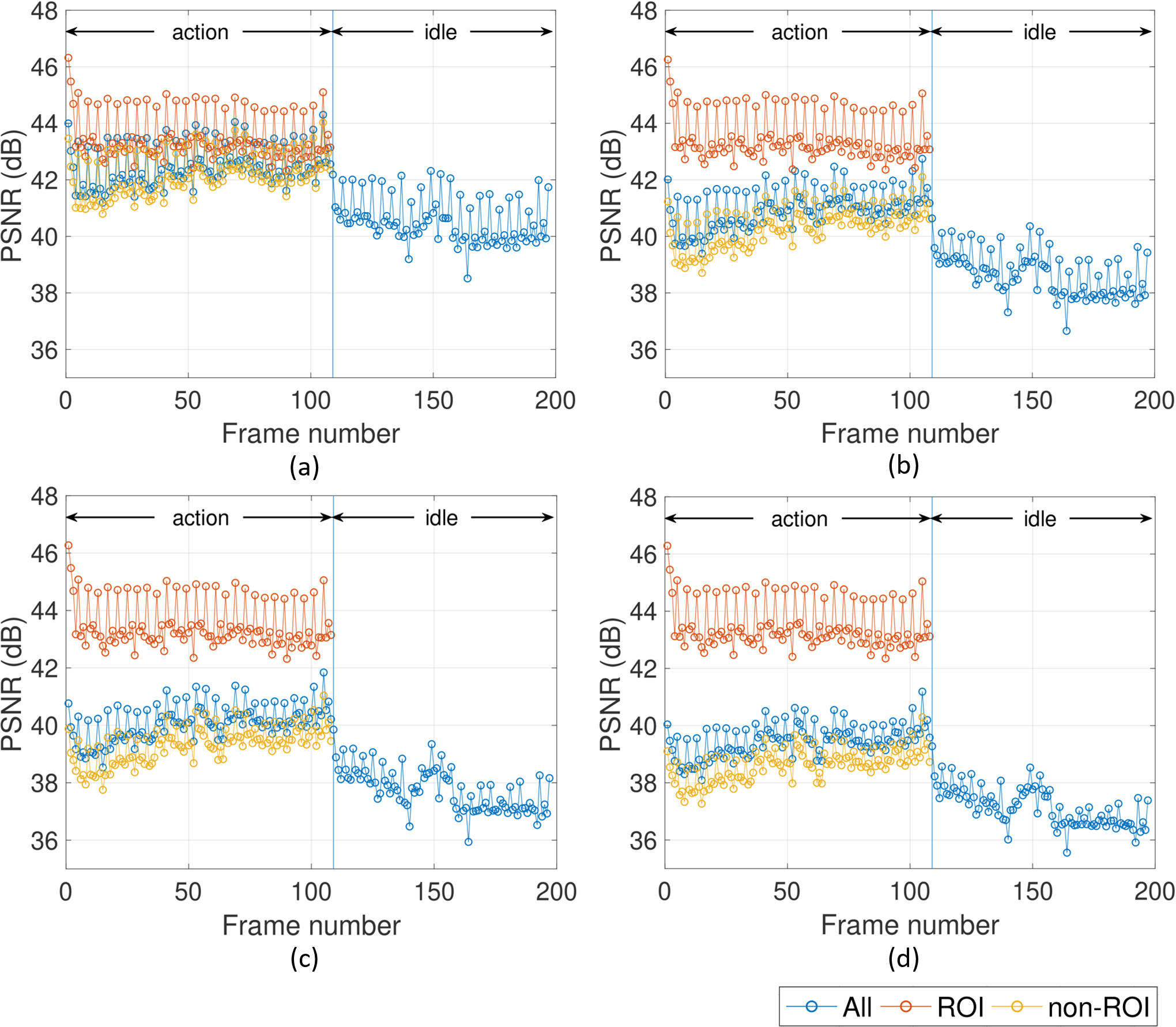}
    \caption{PSNR values for an exemplary segment of a cataract surgery video compressed using Scenario II with different QP differences. (a) $\Delta Q = 5$, (b) $\Delta Q = 10$, (c) $\Delta Q = 13$, (d) $\Delta Q = 15$.}
    \label{Fig-ICPR:PSNR1}
\end{figure*}
Figure~\ref{Fig-ICPR:Idle-pattern} illustrates the predicted patterns of idle frames in four cataract surgery videos out of the nine representative videos (described in Figure~\ref{Fig-ICPR:CTUs}) with different durations. These patterns are obtained after applying a mean filter of size 15 to the predictions of ResNet101. It can be seen that around 20\% of frames in cataract surgery videos are idle and consequently irrelevant from a medical perspective.

The cornea and instrument segmentation results of Mask R-CNN using two different backbones are listed in Table~\ref{Tab-ACMMM:instance-segmentation}. Considering cornea segmentation, we can observe fairly similar results from the networks with different backbones. Although, ResNet50 has shown better performance in mean average precision for IoU$\geq$80 compared to ResNet101 (1.00 versus 0.92). For the bounding boxes extracted using the mask segmentation results~\footnote{It should be noted that these bounding boxes are different from the output bounding boxes of object detection networks and much more accurate.}, the figures affirm that both networks are capable of detecting the cornea with IoU being at least equal to 0.85. Besides, the mean average precision over different IoU thresholds equals to 0.95 and 0.94 for ResNet101 and ResNet50 backbones, respectively. This affirms the reliability of our trained Mask R-CNN network in segmenting the cornea, which is the most relevant region in cataract surgery videos.

Regarding the instrument segmentation results, we can see lower accuracy in both mask and bounding box segmentation results compared to eye segmentation networks. This is based on the fact that the cornea is regularly located in the center of the frames, specifically in the action frames, where illumination is focused, while a part of the instruments is always located in the corners being usually dark. In fact, it is even hard for human eyes to distinguish the instruments' edges in dark and low-contrast regions. Fortunately, wrong ground-truth segmentations in these areas are not important and do not affect any machine-learning-based image processing approaches.

Since Mask R-CNN trained with ResNet101 as the backbone network provides much better results for both bounding-box and mask segmentation of instruments, this trained network will be used in the ROI segmentation stage. Also, due to slightly better performance on average, the trained network for cornea detection using ResNet101 as a backbone will be utilized for later experiments.

Finally, it should be noted that we are the first to address the problem of content-adaptive compression for videos in ophthalmology. This does not allow us to compare our results to any other work. However, since we release our dataset, any further work on this subject can be directly compared to our results.

\section{Compression Results}
\label{Sec-ACMMM:Compression Results}
Figure~\ref{Fig-ICPR:PSNR1} demonstrates the output PSNR for a segment of a cataract surgery video including four seconds of an action phase followed by four seconds of an idle phase. This segment is compressed using scenario III in which the whole idle frames, as well as outside of cornea in action frames, are regarded as irrelevant content. The CTUs belonging to the relevant content (cornea in action frames) are compressed using a small QP ($QP_r$), whereas the other CTUs are compressed using a larger QP ($QP_i = QP_r + \Delta Q$). In this study, $QP_r$ is fixed to 22, and $QP_i$ is increased from $QP_i = QP_r +5$ in Figure~\ref{Fig-ICPR:PSNR1} (a) to $QP_i = QP_r + 15$ in Figure~\ref{Fig-ICPR:PSNR1} (d). As can be seen in the figures, the PSNR of ROI is equivalent in all four situations -- hence, the relevant content always keeps high quality. Depending on the QP assigned to irrelevant content, the distortion in idle frames and non-ROI regions of action frames is gradually increased. The fluctuations in PSNR of consecutive frames are due to the HEVC low-delay encoding configuration that adapts the quantization parameter based on perceptive visual quality. These figures confirm the effectiveness of the proposed approach in preserving the high quality of the relevant content during the compression procedure.

\begin{figure}[!t]
    \centering
    \includegraphics[width=0.8 \columnwidth]{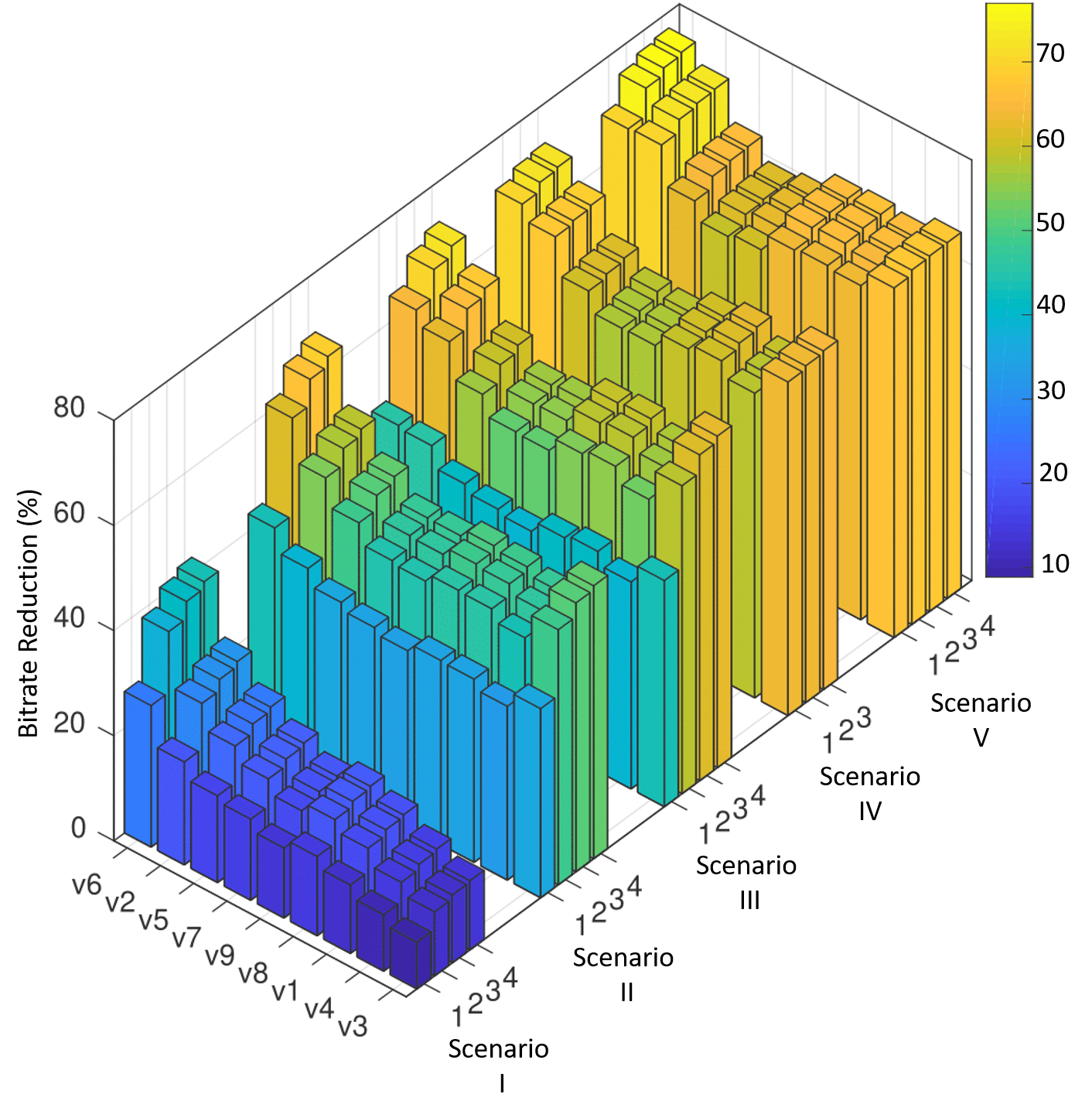}
    \caption{The percentage of bitrate reduction resulting from different scenarios and different QP differences for nine representative videos.}
    \label{Fig-ICPR:bitrate-reduction}
\end{figure}

The achievable bitrate reduction of our proposed method for nine representative cataract surgery videos is shown in Figure~\ref{Fig-ICPR:bitrate-reduction}. For all scenarios excluding Scenario IV, we have considered four different QPs for the irrelevant content: (1) $QP_i = QP_r + 5$, (2) $QP_i = QP_r + 10$, (3) $QP_i = QP_r + 13$, (4) $QP_i = QP_r + 15$.

To avoid attention-grabbing artifacts (\ie the blocky content in non-ROI regions resulting from a large QP that may distract the viewer's attention from the relevant content), we do not go further and choose $QP_r+15$ as the maximum QP for non-ROI. In Scenario IV, QP for the luma channel of the irrelevant content is fixed ($QP_i^{Y} = QP_r + 13$), and a different quantization parameter is applied to the chroma channels of irrelevant content: (1) $QP_i^{C_bC_r} = QP_i^{Y} + 3$, (2) $QP_i^{C_bC_r} = QP_i^{Y} + 5$, (3) $QP_i^{C_bC_r} = QP_i^{Y} + 7$.

\begin{figure}[!t]
    \centering
    \includegraphics[width=0.8\columnwidth]{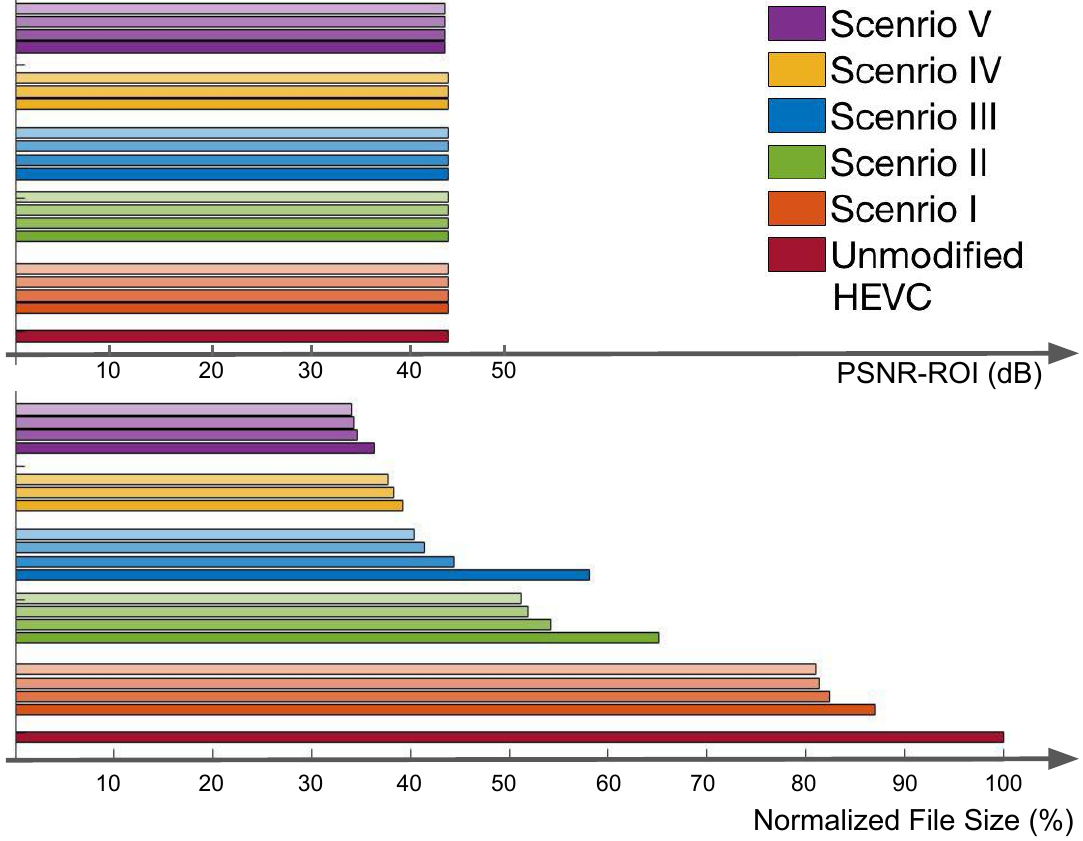}
    \caption{The PSNR of ROI and output size corresponding to an exemplary video compressed in different scenarios.}
    \label{Fig-ICPR:ROI-size}
\end{figure}

It can be perceived from Figure~\ref{Fig-ICPR:bitrate-reduction} that the storage space gain in each scenario is fairly close for different input videos. These results show that regardless of the input video, we can gain up to 23\% storage space for the first scenario to 68\% storage space for Scenario V, compared to the output bitrate of regular compression.
Furthermore, Figure~\ref{Fig-ICPR:ROI-size} shows the output size of a representative video as well as the PSNR of ROI after compression using different scenarios (the darker bars correspond to larger QPs for irrelevant content). It is evident that the proposed approach is capable of preserving the high quality of relevant regions while reducing the overall output bitrate. 

\paragraph{\textbf{Expert Review. }}We have conducted a qualitative study to verify the suitability of the relevance-based compressed videos with six cataract experts for the five defined scenarios. The surgeons are asked to rate the visual quality of the relevant content in compressed videos in comparison with the results of unmodified HEVC on a Likert-scale from 5 (strongly agree -- very suitable) to 1 (strongly disagree -- not suitable). For scenarios I, II, and IV, the average ratings (ave) were 5.0 with its standard deviation (std) being equal to $0.0$. In scenario III, ave $=3.66$ and std $=0.81$; and in scenario V, ave $=4.66$ and std $=0.51$. In can be inferred that for the first four scenarios, the clinicians consider the visual quality resulting from the proposed approach as almost equal to the original video, while they seem to consider the complete removal of irrelevant content in action frames as a little distracting.

\section{Conclusion}
\label{Sec-ACMMM:Conclusion}
To date, numerous investigations have sought efficient ROI-based compression. While these research efforts provide precious insights, scant attention has been devoted to ROI-based video compression. In particular, ROI-based video compression in case of dynamic backgrounds using domain knowledge has been undervalued. In this chapter, a relevance-based compression approach for cataract surgery videos (where relevant content is regarded as ROI) is proposed. The goal is to achieve high compression efficiency for irrelevant content as well as preserving the high quality of relevant content considering the target audience. To achieve this goal, the amount of distortion in different regions of videos should be proportionate to their level of relevance. Due to the inherent features of cataract surgery videos (dynamic background and close statistical properties of relevant and irrelevant content), common ROI detection approaches (\eg saliency detection approaches) are incompetent. Hence, we have proposed to use classification and semantic segmentation convolutional neural networks to detect the relevant content. These results are exploited to penalize the distortion in relevant content using CTU-wise QP allocation in HEVC. Experimental results confirm the capability of the proposed method in detecting and subsequently retaining the high quality of the relevant content, while gaining high compression efficiency by imposing more distortion on irrelevant content. The compression results show up to 63\% storage-saving by applying stronger quantization on spatio-temporal irrelevant content, and up to 68\% gain in storage space by removing the spatially irrelevant content. In contrast to the existing ROI-based video compression frameworks,  the proposed framework can be generalized to all sorts of videos, specifically surgical videos, notwithstanding the presence of dynamic backgrounds. By automatically segmenting out the irrelevant part of videos, high fidelity for ROI as well as high video compression efficiency is possible.
\chapter{Lens Irregularity Detection \label{Chapter:Irregularity-Detection}}

\chapterintro{
	 In this chapter, we propose a novel framework as the major step towards lens irregularity detection. In particular, we propose (I) an end-to-end recurrent neural network to recognize the lens-implantation phase and (II) a novel semantic segmentation network to segment the lens and pupil after the implantation phase. The phase recognition results reveal the effectiveness of the proposed surgical phase recognition approach. Moreover, the segmentation results confirm the proposed segmentation network's effectiveness compared to state-of-the-art rival approaches.
}

{
	\singlespacing This chapter is an adapted version of:
	
	``Ghamsarian, N., Taschwer, M., Putzgruber-Adamitsch, D., Sarny,
S., El-Shabrawi, Y., and Schoeffmann, K. LensID: A CNN-RNN-based
framework towards lens irregularity detection. In 24th International Conference
on Medical Image Computing \& Computer Assisted Interventions (MICCAI
2021) (2021), p. to appear.''
}

\section{Introduction}
\label{Sec-MICCAI: Introduction}

Over several years, there have been numerous advances in surgical techniques, tools, and instruments in ophthalmic surgeries. Such advances resulted in decreasing the risk of severe intraoperative and postoperative complications. Still, there are many ongoing research efforts to prevent the current implications during and after surgery. A critical issue in cataract surgery that has not yet been addressed is intraocular lens (IOL) dislocation. This complication leads to various human sight issues such as vision blur, double vision, or vision inference as observing the lens implant edges. Intraocular inflammation, corneal edema, and retinal detachment are some other consequences of lens relocation. Since patient monitoring after the surgery or discharge is not always possible, the surgeons seek ways to diagnose evidence of potential irregularities that can be investigated during the surgery. 

Recent studies show that particular intraocular lens characteristics can contribute to lens dislocation after the surgery~\cite{IIOL}. Moreover, the expert surgeons argue that there can be a direct relationship between the overall time of lens unfolding and the risk of lens relocation after the surgery. Some surgeons also hypothesize that severe lens instability during the surgery is a symptom of lens relocation. To discover the potential correlations between lens relocation and its possibly contributing factors, surgeons require a tool for systematic feature extraction. Indeed, an automatic approach is required for (i) detecting the lens implantation phase to determine the starting time for lens statistics' computation and (ii) segmenting the lens and pupil to compute the lens statistics over time. The irregularity-related statistics can afterward be extracted by tracking the lens's relative size (normalized by the pupil's size) and relative movements (by calibrating the pupil). 
Due to the dearth of computing power in the operation rooms, automatic phase detection and lens/pupil segmentation on the fly is not currently achievable. Alternatively, this analysis can be performed in a post hoc manner using recorded cataract surgery videos. The contributions of this work are: 
\begin{enumerate}
    \item We propose a novel CNN-RNN-based framework for evaluating lens unfolding delay and lens instability in cataract surgery videos. 
    \item We propose and evaluate a recurrent convolutional neural network architecture to detect the ``implantation phase'' in cataract surgery videos.
    \item We further propose a novel semantic segmentation network architecture termed as \textit{AdaptNet}\footnote[1]{The PyTorch implementation of AdaptNet is publicly available at https://github.com/Negin-Ghamsarian/AdaptNet-MICCAI2021.}, that can considerably improve the segmentation performance for the intraocular lens (and pupil) compared to ten rival state-of-the-art approaches.
    \item We introduce three datasets for phase recognition, pupil segmentation, and lens segmentation that are publicly released to support reproducibility and allow further investigations for lens irregularity detection\footnote[2]{http://ftp.itec.aau.at/datasets/ovid/LensID/}.
\end{enumerate}

\section{Methodology}
\label{Sec-MICCAI: Methodology}
\begin{figure}[!tb]
    \centering
    \includegraphics[width=1\textwidth]{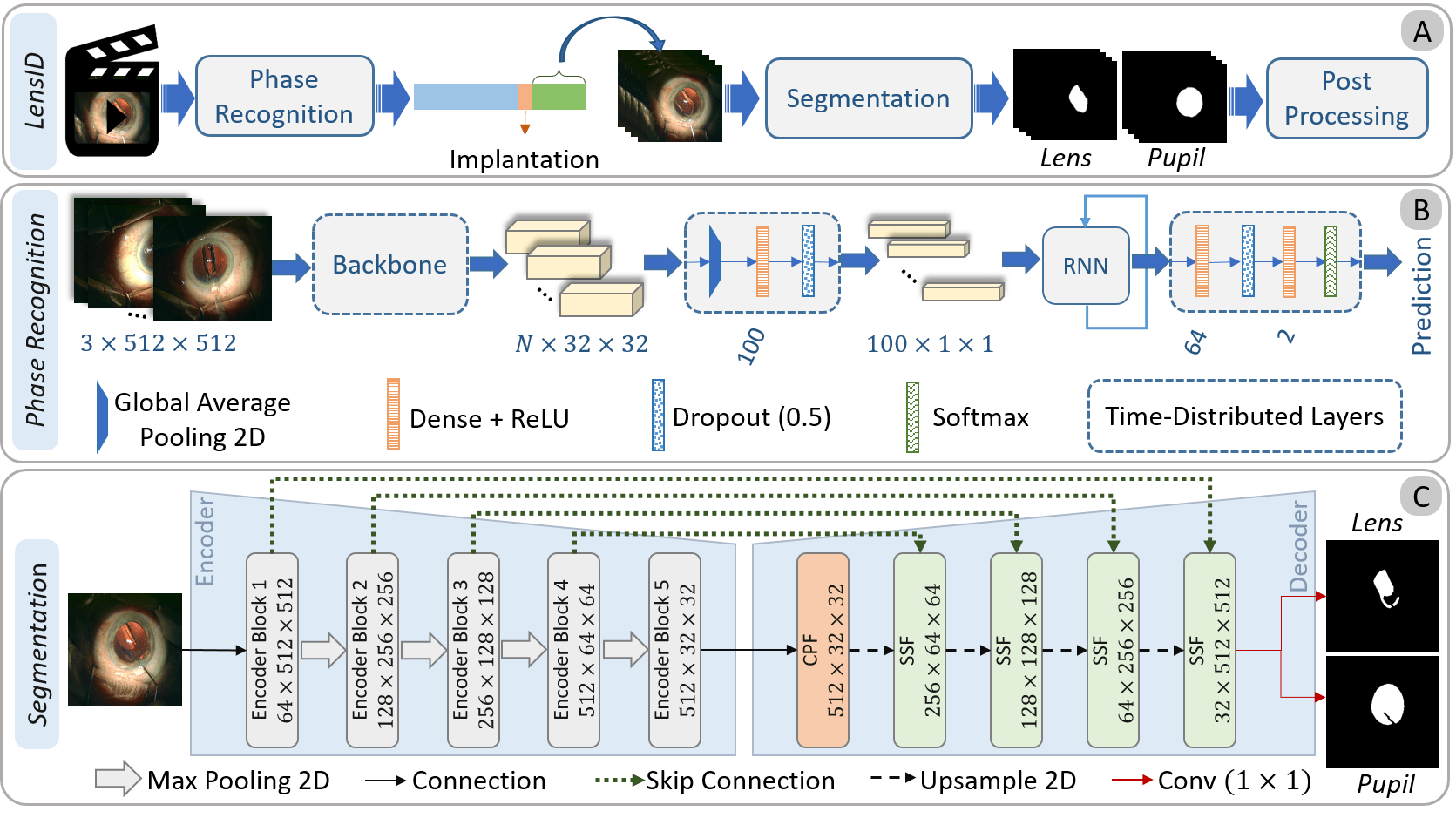}
    \caption{The block diagram of \textit{LensID} and the architecture of \textit{Phase Recognition} and \textit{Semantic Segmentation} networks.}
    \label{fig-MICCAI: BD1}
\end{figure}
Figure~\ref{fig-MICCAI: BD1} demonstrates the block diagram of \textit{LensID} and the network architecture of the phase recognition and segmentation steps. As the first step towards lens irregularity detection, we adopt a recurrent convolutional network (Figure~\ref{fig-MICCAI: BD1}-B) to detect the lens implantation phase (the temporal segment in which the lens implantation instrument is visible). We start segmenting the artificial lens and pupil exactly after the lens implantation phase using the proposed semantic segmentation network (Figure~\ref{fig-MICCAI: BD1}-C). The pupil and lens segmentation results undergo some post-processing approaches to compute lens instability and lens unfolding delay. More precisely, we draw the smallest convex polygon surrounding pupil's and lens' masks using binary morphological operations. For lens instability, we use the normalized distance between the lens and pupil centers. For lens unfolding, we track the lens' area over time, considering its relative position.
\paragraph{\textbf{Phase Recognition. }}
As shown in Figure~\ref{fig-MICCAI: BD1}-B, we use a pre-trained backbone followed by global average pooling to obtain a feature vector per each input frame. These features undergo a sequence of Dense, Dropout, and ReLU layers to extract higher-order semantic features. A recurrent layer with five units is then employed to improve the feature representation by taking advantage of temporal dependencies. These features are then fed into a sequence of layers to finally output the predicted class for each input frame.

\begin{figure}[!tb]
    \centering
    \includegraphics[width=1\textwidth]{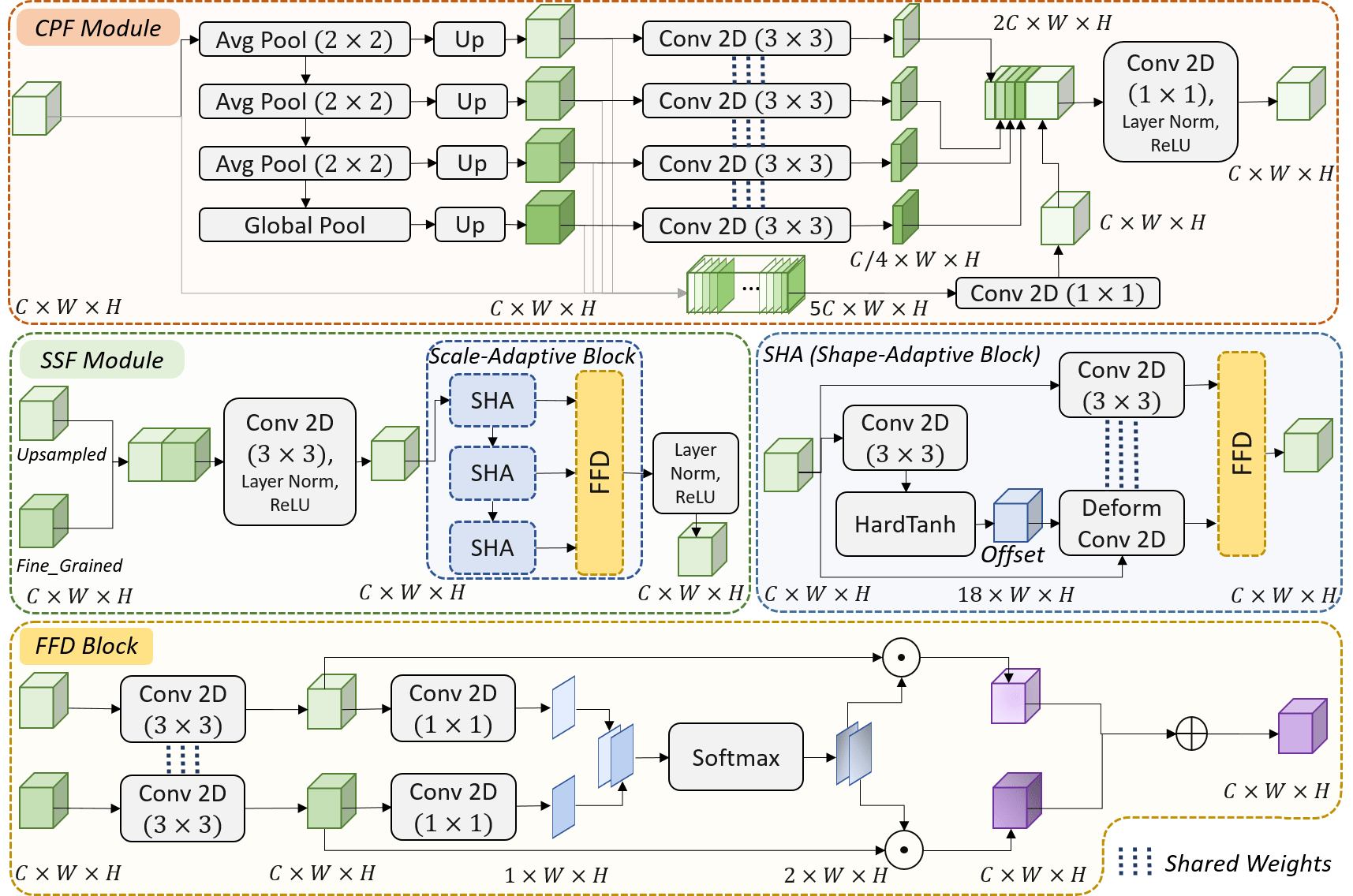}
    \caption{The detailed architecture of the \textit{CPF} and \textit{SFF} modules of AdaptNet.}
    \label{fig-MICCAI: AdaptNet}
\end{figure}

\paragraph{\textbf{Lens \& Pupil Segmentation. }}
In cataract surgery, a folded artificial lens is implanted inside the eye. The lens is transparent and inherits the pupil's color after implantation. Moreover, it is usually being unfolded very fast (sometimes with the help of an instrument). The transparency and unpredictable formation of this object, as well as occlusion, defocus blur, and motion blur~\cite{DCS}, make lens segmentation and tracking more challenging. Hence, we require a semantic segmentation network that can be adapted to the changes in the artificial lens's shape and scale.
We adopt a U-Net-based encoder-decoder architecture for the proposed semantic segmentation network termed as AdaptNet. AdaptNet consists of three main components: encoder, \textit{cascade pooling fusion (CPF)} module, and \textit{shape/scale-adaptive feature fusion (SSF)} module. We use the VGG16 network as the encoder network.
The encoder's output feature map is fed into the CPF module to enhance the feature representation using pyramid features. This feature map is then fed into a sequence of SSF modules, which decode low-resolution semantic features.

As shown in Figure~\ref{fig-MICCAI: AdaptNet}, the CPF module applies a sequence of three average pooling layers (with a stride of two pixels) followed by a global average pooling layer to the input features. The obtained feature maps are upsampled to the original size of the input and concatenated together with the input feature map in a depth-wise manner. Each group of five channels in the generated feature map undergoes a distinct convolution for intra-channel feature refinement (which is performed using a convolutional layer with $C$ groups). Besides, the upsampled features are mapped into a smaller channel space while extracting higher-order semantic features using convolutional layers with shared weights. The obtained features are concatenated with the intra-channel refined features and undergo a convolutional layer for inter-channel feature refinement.

The \textit{SSF} module starts with concatenating the upsampled semantic feature map with the fine-grained feature map coming from the encoder. The concatenated feature map is fed into a sequence of convolutional,  layer normalization, and ReLU layers for feature enhancement and dimensionality reduction. The resulting features are fed into the \textit{scale-adaptive block}, which aims to fuse the features coming from cascade convolutional blocks. This succession of convolutional layers with small filter sizes can factorize the large and computationally expensive receptive fields~\cite{MultiResUNet}. Moreover, the fusion of these successive feature maps can play the role of scale-awareness for the network. The \textit{shape-adaptive (SHA) block} is responsible for fusing the resulting feature maps of deformable and structured convolutions. At first, a convolutional layer followed by a hard tangent hyperbolic function is employed to produce the offsets for the deformable convolutional layer~\cite{DeformConv}. The input features are also fed into a regular convolutional layer that shares the weights with the deformable layer for structured-feature extraction. These features are then fused to induce the awareness of shape and deformation to the network. 

The \textit{feature fusion decision (FFD)} block inspired by CPFNet~\cite{CPFNet} accounts for determining the importance of each input feature map in improving semantic features. Figure~\ref{fig-MICCAI: AdaptNet} shows the \textit{FFD Block} in the case of two input branches. At first, shared convolutional layers are applied to the input feature maps to extract the shared semantic features. The resulting feature maps undergo shared pixel-wise convolutions to produce the pixel-wise attention maps. The concatenated attention maps are fed into a softmax activation layer for normalization. The obtained features are used as pixel-wise weights of the shared-semantic feature maps. The shape/scale adaptive features are computed as the sum of pixel-wise multiplications ($\odot$) between the normalized attention maps and their corresponding semantic feature maps.
\section{Experimental Setup}
\label{Sec-MICCAI: Experimental Settings}
We use three datasets for this study: (i)~a large dataset containing the annotations for the lens implantation phase versus the rest of phases from 100 videos of cataract surgery, (ii)~a dataset containing the lens segmentation of 401 frames from 27 videos (292 images from 21 videos for training, and 109 images from six videos for testing), and (iii)~a dataset containing the pupil segmentation of 189 frames from 16 videos (141 frames from 13 videos for training, and 48 frames from three videos for testing). Regarding the phase recognition dataset, since lens implantation is a very short phase (around four seconds) compared to the whole surgery (seven minutes on average), creating a balanced dataset that can cover the entire content of videos from the ``Rest'' class is quite challenging. Hence, we propose a video clip generator that can provide diverse training sequences for the recurrent neural network by employing stochastic functions. At first, 12 three-second video clips with overlapping frames are extracted from the implantation phase of each cataract surgery video. Besides, the video segments before and after the implantation phase are divided into eight and four video clips, respectively (these clips have different lengths depending on the length of the input video). Accordingly, we have a balanced dataset containing 2040 video clips from 85 videos for training and 360 video clips from the other 15 videos for testing. For each training example, the video generator uses a stochastic variable to randomly select a three-second clip from the input clip. We divide this clip into $N$ sub-clips, and $N$ stochastic variables are used to randomly select one frame per sub-clip (in our experiments, $N$ is set to five to reduce computational complexity and avoid network overfitting). 

For phase recognition, all networks are trained for 20 epochs. The initial learning rate for these networks is set to 0.0002 and 0.0004 for the networks with VGG19 and Resnet50 backbones, respectively, and halved after ten epochs. Since the segmentation networks used for evaluations have different depths, backbones, and the number of trainable parameters, all networks are trained with three different initial learning rates ($lr_0\in\{0.0005, 0.001, 0.002\}$). For each network, the results with the highest Dice coefficient are listed. All segmentation networks are trained for 30 epochs, and the learning rate is decreased by a factor of 0.8 in every other epoch.  To prevent overfitting and improve generalization performance, we have used motion blur, Gaussian blur, random contrast, random brightness, shift, scale, and rotation for data augmentation. The backbones of all networks evaluated for phase recognition and lens/pupil semantic segmentation are initialized with ImageNet~\cite{ImageNet} weights. The size of input images to all networks is set to $512\times 512\times 3$.
The loss function for the phase recognition network is set to \textit{Binary Cross Entropy}. For the semantic segmentation task, we adopt a loss function consisting of categorical cross entropy and logarithm of soft Dice coefficient as follows (in Eq.~\eqref{eq-MICCAI:loss}, \textit{CE} stands for \textit{Cross Entropy}, and $\mathcal{X}_{Pred}$ and $\mathcal{X}_{True}$ denote the predicted and ground-truth segmentation images, respectively. Besides, we use a Dice smoothing factor equal to 1, and set $\lambda=0.8$ in our experiments):
\begin{align}
\mathcal{L} = \lambda\times CE(\mathcal{X}_{Pred},\mathcal{X}_{True}) - (1-\lambda)\times \log_2 Dice(\mathcal{X}_{Pred},\mathcal{X}_{True})
\label{eq-MICCAI:loss}
\end{align}
To evaluate the performance of phase recognition networks, we use \textit{Precision}, \textit{Recall}, \textit{F1-Score}, and \textit{Accuracy}, which are the common classification metrics. The semantic segmentation performance is evaluated using \textit{Dice coefficient} and \textit{Intersection over Union (IoU)}. We compare the segmentation accuracy of the proposed approach (AdaptNet) with ten state-of-the-art approaches including UNet++ (and UNet++\slash DS)~\cite{UNet++}, MultiResUNet~\cite{MultiResUNet}, CPFNet~\cite{CPFNet}, dU-Net~\cite{dU-Net}, CE-Net~\cite{CE-Net}, FEDNet~\cite{FED-Net}, PSPNet~\cite{PSPNet}, SegNet~\cite{SegNet}, and U-Net~\cite{U-Net}. It should be mentioned that the rival approaches employ different backbone networks, loss functions (cross entropy or cross entropy log Dice), and upsampling methods (bilinear, transposed convolution, pixel-shuffling, or max unpooling).
\section{Experimental Results and Discussion}
\label{Sec-MICCAI: Experimental Results}
\begin{table}[t!]
\renewcommand{\arraystretch}{1}
\caption{Phase recognition results of the end-to-end recurrent convolutional networks.}
\label{Tab-MICCAI:RNN}
\centering
\resizebox{\textwidth}{!}{
\begin{tabular}{lP{1.25cm}P{1.25cm}P{1.25cm}P{1.25cm}P{1.25cm}P{1.25cm}P{1.25cm}P{1.25cm}}
\specialrule{.12em}{.05em}{.05em}
&\multicolumn{4}{c}{Backbone: VGG19}&\multicolumn{4}{c}{Backbone: ResNet50}\\\cmidrule(lr){2-5}\cmidrule(lr){6-9}
RNN&Precision&Recall&F1-Score&Accuracy&Precision&Recall&F1-Score&Accuracy\\\specialrule{.12em}{.05em}{.05em}
GRU&0.97&0.96&0.96&0.96&0.9&0.94&0.94&0.94\\
LSTM&0.98&0.98&0.98&0.98&0.96&0.96&0.96&0.96\\
BiGRU&0.97&0.96&0.96&0.96&0.95&0.95&0.95&0.95\\
\rowcolor{shadecolor} BiLSTM&1.00&1.00&1.00&1.00&0.98&0.98&0.98&0.98\\ \specialrule{.12em}{.05em}{.05em}
\end{tabular}}
\end{table}
Table~\ref{Tab-MICCAI:RNN} compares the classification reports of the proposed architecture for phase recognition considering two different backbone networks and four different recurrent layers. Thanks to the large training set and taking advantage of recurrent layers, all networks have shown superior performance in classifying the implantation phase versus other phases. However, the LSTM and bidirectional LSTM (BiLSTM) layers have shown better performance compared to GRU and BiGRU layers, respectively. Surprisingly, the network with a VGG19 backbone and BiLSTM layer has achieved 100\% accuracy in classifying the test clips extracted from the videos which are not used during training.
Figure~\ref{fig-MICCAI:Dice-IoU} compares the segmentation results (mean and standard deviation of IoU and Dice coefficient) of AdaptNet and ten rival state-of-the-art approaches. Overall, it can be perceived that AdaptNet, UNet++, UNet++\slash DS, and FEDNet have achieved the top four segmentation results. However, AdaptNet has achieved the highest mean IoU and Dice coefficient compared to the rival approaches. In particular, the proposed approach achieves 3.48\% improvement in mean IoU and 2.22\% improvement in mean Dice for lens segmentation compared to the best rival approach (UNet++). Moreover, the smaller standard deviation of IoU (10.56\% vs. 12.34\%) and Dice (8.56\% vs. 9.65\%) for AdaptNet compared to UNet++ confirms the reliability and effectiveness of the proposed architecture. For pupil segmentation, AdaptNet shows subtle improvement over the best rival approach (UNet++) regarding mean IoU and Dice while showing significant improvement regarding the standard deviation of IoU (1.91 vs. 4.05). Table~\ref{Tab-MICCAI:ablation} provides an ablation study of AdaptNet. We have listed the Dice and IoU percentage with two different learning rates by gradually adding the proposed modules and blocks (for lens segmentation). It can be perceived from the results that regardless of the learning rate, each distinctive module and block has a positive impact on segmentation performance. We cannot test the FFD block separately since it is bound with the SSF module.

\begin{figure*}[t!]
\begin{tabular}{c}
\begin{subfigure}{1\textwidth}\vspace{-0.5\baselineskip}
  \centering
  \begin{adjustbox}{width=1\textwidth}

\begin{tikzpicture}
\begin{axis}[
ybar=0pt,
axis on top,
bar width=0.38cm,
width=\textwidth,
height=.4\textwidth,
ylabel=Lens Segmentation,
enlarge x limits=0.05,
every axis x label/.append style={xshift=0.2cm},
symbolic x coords={U-Net, SegNet, PSPNet, FED-Net, CE-Net, dU-Net, CPFNet, MultiResUNet, UNet++\slash DS, UNet++, AdaptNet},
xtick=\empty,
nodes near coords align={vertical},
cycle list/Set2,
ymin=55,ymax=105,
ymajorgrids = true,
yminorgrids = true,
bar width=0.3cm,
nodes near coords,
minor ytick={70,90},
label style={font=\small},
tick label style={font=\small},
major x tick style = transparent,
x tick label style={rotate=45,anchor=east},
legend style={at={(0.5,+1.19)},
	            anchor=north,legend columns=6},
every node near coord/.append style={rotate=90, anchor=west, font=\scriptsize, color = black, opacity=1}	            
]
\addplot+[style={draw=black,solid,fill,opacity=0.6,
}, 
             error bars/.cd, 
             y dir=minus,y explicit]
             coordinates {
                  (U-Net,61.89) +- (0,20.93)
                  (SegNet,73.17) +- (0,12.62)
                  (PSPNet,71.40) +- (0,18.92)
                  (FED-Net,80.59) +- (0,11.53)
                  (CE-Net,70.56) +- (0,10.62)
                  (dU-Net,56.86) +- (0,32.22)
                  (CPFNet,75.38) +- (0,12.17)
                  (MultiResUNet,61.42) +- (0,19.91)
                  (UNet++\slash DS,82.32) +- (0,14.10)
                  (UNet++,83.61) +- (0,12.34)
                  (AdaptNet,87.09) +- (0,10.56)}; 
                  
\addplot+[style={draw=black,solid,fill,opacity=0.6,
}, 
             error bars/.cd, 
             y dir=minus,y explicit]
             coordinates {
                  (U-Net,73.86) +- (0,20.39)
                  (SegNet,83.75) +- (0,10.48)
                  (PSPNet,81.53) +- (0,16.63)
                  (FED-Net,88.68) +- (0,9.01)
                  (CE-Net,82.22) +- (0,8.51)
                  (dU-Net,65.81) +- (0,32.19)
                  (CPFNet,85.26) +- (0,10.28)
                  (MultiResUNet,73.88) +- (0,18.26)
                  (UNet++\slash DS,89.95) +- (0,10.63)
                  (UNet++,90.44) +- (0,9.65)
                  (AdaptNet,92.62) +- (0,8.56)}; 
\legend{IoU(\%),Dice(\%)}
\end{axis}
\end{tikzpicture}

\end{adjustbox}
\end{subfigure} \\
\begin{subfigure}{1\textwidth}\vspace{-0.2\baselineskip}
  \centering
  \begin{adjustbox}{width=1\textwidth}
  \begin{tikzpicture}
\begin{axis}[
ybar=0pt,
axis on top,
bar width=0.38cm,
width=\textwidth,
height=.4\textwidth,
ylabel=Pupil Segmentation,
enlarge x limits=0.05,
every axis x label/.append style={xshift=0.2cm},
symbolic x coords={U-Net, SegNet, PSPNet, FED-Net, CE-Net, dU-Net, CPFNet, MultiResUNet, UNet++\slash DS, UNet++, AdaptNet},
nodes near coords align={vertical},
cycle list/Set2,
ymin=60,ymax=110,
ymajorgrids = true,
yminorgrids = true,
bar width=0.3cm,
nodes near coords,
minor ytick={70,90},
ticklabel style = {font=\small},
major x tick style = transparent,
x tick label style={rotate=25,anchor=east},
legend style={at={(0.5,+1.22)},
	            anchor=north,legend columns=6},
every node near coord/.append style={rotate=90, anchor=west, font=\scriptsize, color = black, opacity=1}	            
]
\addplot+[style={draw=black,solid,fill,opacity=0.6,
}, 
             error bars/.cd, 
             y dir=minus,y explicit]
             coordinates {
                  (U-Net,83.51) +- (0,20.24)
                  (SegNet,84.24) +- (0,6.13)
                  (PSPNet,89.55) +- (0,9.43)
                  (FED-Net,94.43) +- (0,1.93)
                  (CE-Net,87.66) +- (0,5.94)
                  (dU-Net,75.34) +- (0,27.60)
                  (CPFNet,92.33) +- (0,2.98)
                  (MultiResUNet,63.16) +- (0,36.90)
                  (UNet++\slash DS,95.28) +- (0,6.13)
                  (UNet++,96.02) +- (0,4.05)
                  (AdaptNet,96.06) +- (0,1.91)}; 
                  
\addplot+[style={draw=black,solid,fill,opacity=0.6,
}, 
             error bars/.cd, 
             y dir=minus,y explicit]
             coordinates {
                  (U-Net,89.36) +- (0,15.07)
                  (SegNet,91.31) +- (0,4.20)
                  (PSPNet,94.18) +- (0,6.21)
                  (FED-Net,97.12) +- (0,1.03)
                  (CE-Net,93.32) +- (0,3.41)
                  (dU-Net,82.40) +- (0,22.62)
                  (CPFNet,95.99) +- (0,1.47)
                  (MultiResUNet,69.59) +- (0,34.93)
                  (UNet++\slash DS,97.53) +- (0,2.27)
                  (UNet++,97.96) +- (0,0.96)
                  (AdaptNet,97.98) +- (0,1.00)}; 
\end{axis}
\end{tikzpicture}

\end{adjustbox}
\end{subfigure}
\end{tabular}
\caption{Quantitative comparison of segmentation results for the proposed approach (AdaptNet) and rival approaches.}
\label{fig-MICCAI:Dice-IoU}
\end{figure*}
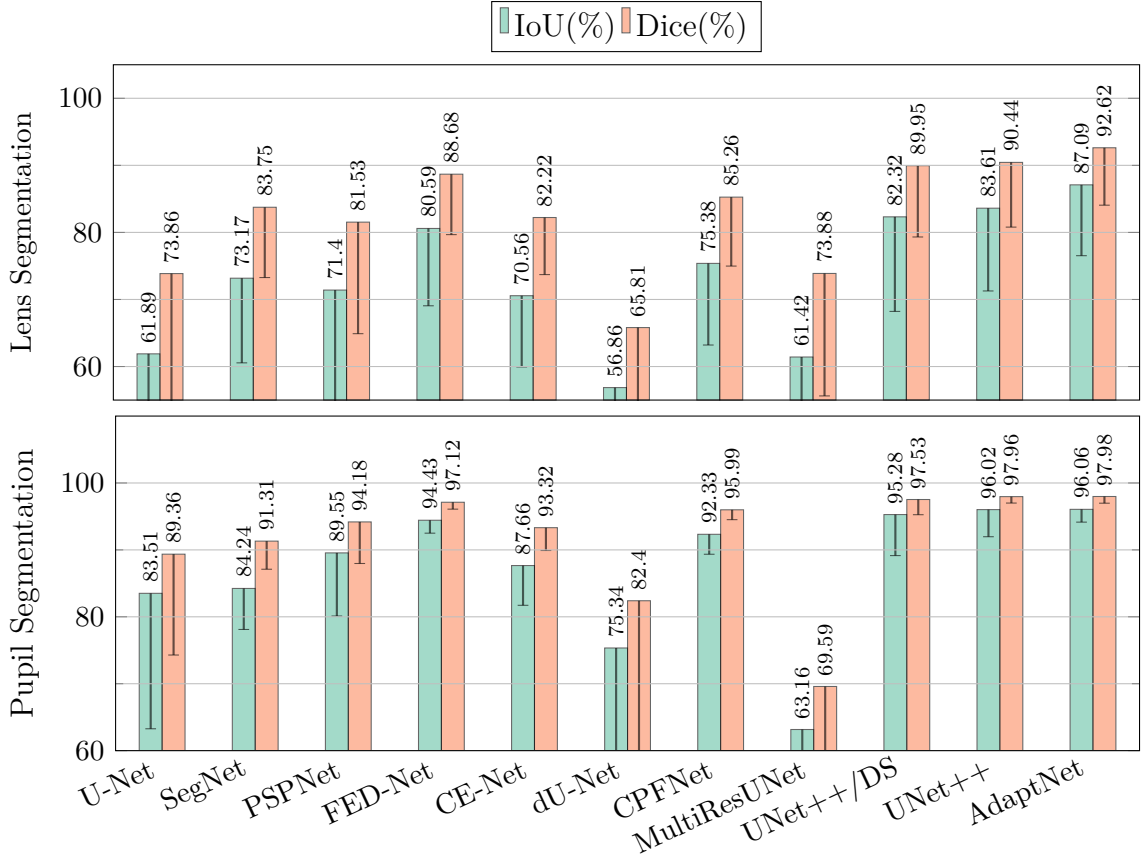
\begin{figure}[!tb]
    \centering
    \includegraphics[width=1\textwidth]{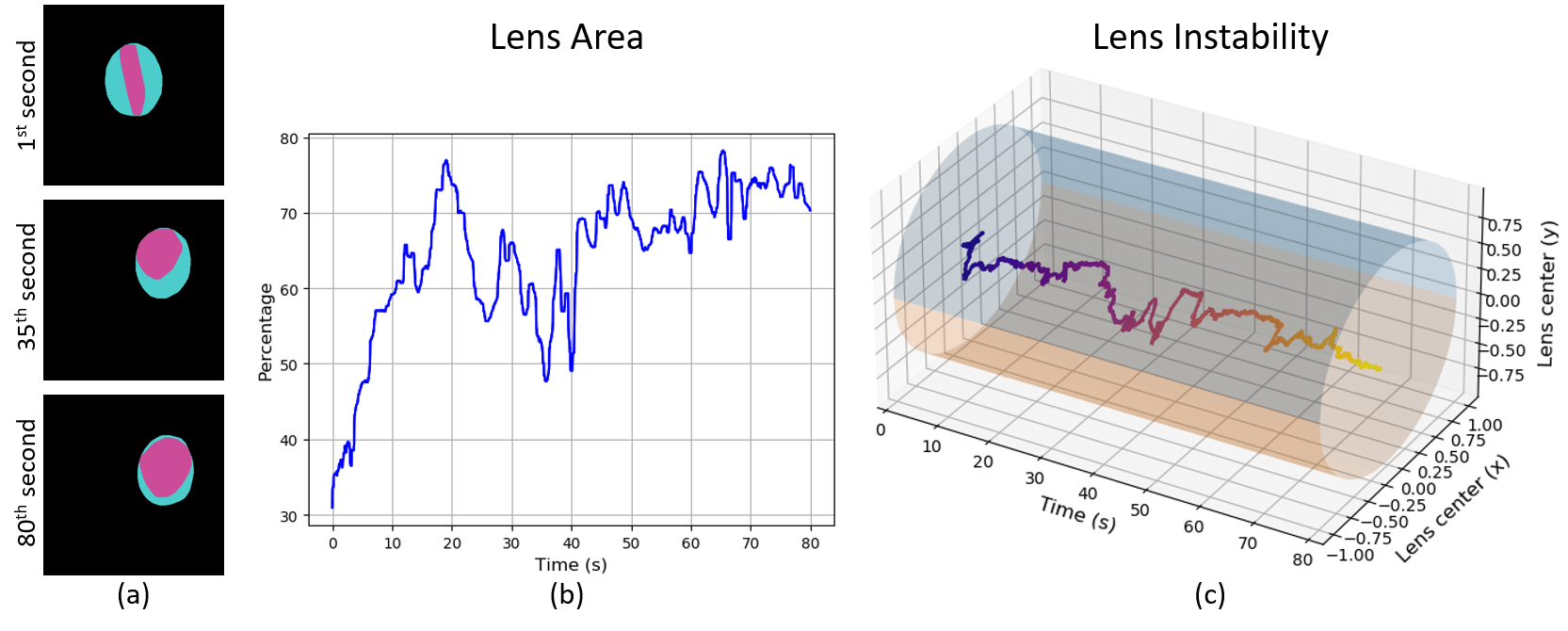}
    \caption{The lens statistics for one representative cataract surgery video.}
    \label{fig-MICCAI:ID}
\end{figure}

\begin{table}[t!]
\renewcommand{\arraystretch}{1}
\caption{Impact of different modules on the segmentation results of AdaptNet.}
\label{Tab-MICCAI:ablation}
\centering
\begin{tabular}{P{1.2cm} P{1.2cm} P{1.2cm} P{1.2cm} P{1.2cm} P{1.2cm} P{1.2cm} P{1.2cm}}
\specialrule{.12em}{.05em}{.05em}
\multicolumn{4}{c}{Components}&\multicolumn{2}{c}{lr = 0.001}&\multicolumn{2}{c}{lr = 0.002}\\\cmidrule(lr){1-4}\cmidrule(lr){5-6}\cmidrule(lr){7-8}
Baseline&SSF&SHA&CPF&IoU(\%)&Dice(\%)&IoU(\%)&Dice(\%)\\\specialrule{.12em}{.05em}{.05em}
\checkmark&\xmark&\xmark&\xmark&82.79&89.94&84.33&90.90\\
\checkmark&\checkmark&\xmark&\xmark&83.54&90.33&84.99&91.22\\
\checkmark&\checkmark&\checkmark&\xmark&84.76&91.12&86.34&92.17\\
\rowcolor{shadecolor} \checkmark&\checkmark&\checkmark&\checkmark&85.03&91.28&87.09&92.62\\
\specialrule{.12em}{.05em}{.05em}
\end{tabular}
\end{table}
Figure~\ref{fig-MICCAI:ID} shows the post-processed lens segments (pink) and pupil segments (cyan) from a representative video in different time slots (a), the relative lens area over time (b), and relative lens movements over time (c). Due to lens instability, a part of the lens is sometimes placed behind the iris, as shown in the segmentation results in the 35th second. Accordingly, the visible area of the lens can change independently of the unfolding state. Hence, the relative position of the lens should also be taken into account for lens unfolding delay computations. As can be perceived, the visible area of the lens is near maximum at 20 seconds after the implantation phase, and the lens is located nearly at the center of the pupil at this time. Therefore, the lens unfolding delay is 20 seconds in this case. However, the lens is quite unstable until 70 seconds after implantation. 
\section{Conclusion}
\label{Sec-MICCAI: Conclusion}
Lens irregularity detection is a highly relevant problem in ophthalmology, which can play a prominent role in predicting and preventing lens relocation after surgery. This study focuses on two significant steps towards lens irregularity detection: (i) ``lens implantation phase'' detection and (ii) lens/pupil segmentation. We propose an end-to-end recurrent convolutional network to detect the lens implantation phase. Moreover, we propose a novel semantic segmentation network termed as AdaptNet. The proposed approach can deal with severe deformations and scale variations in the intraocular lens by adaptively fusing sequential and parallel feature maps. Experimental results reveal the effectiveness of the proposed phase recognition and semantic segmentation networks.

\section*{Appendix 1}
Table~\ref{Tab-MICCAI:specification} lists the specifications of the rival state-of-the-art approaches used in our evaluations. In ``Upsampling'' column, ``Trans Conv'' stands for \textit{Transposed Convolution}.

\begin{table*}[h!]
\renewcommand{\arraystretch}{1}
\caption{Specifications of the proposed and rival segmentation approaches.}
\label{Tab-MICCAI:specification}
\centering
\resizebox{\textwidth}{!}{
\begin{tabular}{lccccc}
\specialrule{.12em}{.05em}{.05em}
Model & Backbone & Params & Upsampling & Reference & Year\\\specialrule{.12em}{.05em}{.05em}
UNet$++$ (\slash DS) &VGG16&24.24 M& Bilinear &~\cite{UNet++} & 2020\\
MultiResUNet &\xmark& 9.34 \enspace M& Trans Conv &~\cite{MultiResUNet} & 2020\\
CPFNet &ResNet34&34.66 M& Bilinear &~\cite{CPFNet} & 2020\\
dU-Net &\xmark &31.98 M&Trans Conv &~\cite{dU-Net} & 2020\\
CE-Net &ResNet34&29.90 M&Trans Con&~\cite{CE-Net} & 2019\\
FED-Net &ResNet50&59.52 M& Trans Conv \& PixelShuffle &~\cite{FED-Net} & 2019\\
PSPNet &ResNet50&22.26 M&Bilinear&~\cite{PSPNet} & 2017\\
SegNet &VGG16&14.71 M&Max Unpooling&~\cite{SegNet} & 2017\\
U-Net &\xmark&17.26 M& Bilinear &~\cite{U-Net} & 2015\\\cdashline{1-6}[0.8pt/1pt]
AdaptNet&VGG16 &23.61 M& Bilinear &\multicolumn{2}{c}{Proposed}\\
\specialrule{.12em}{.05em}{0.05em}
\end{tabular}}

\end{table*}

\begin{figure}[!ht]
\centering
\includegraphics[width=1\columnwidth]{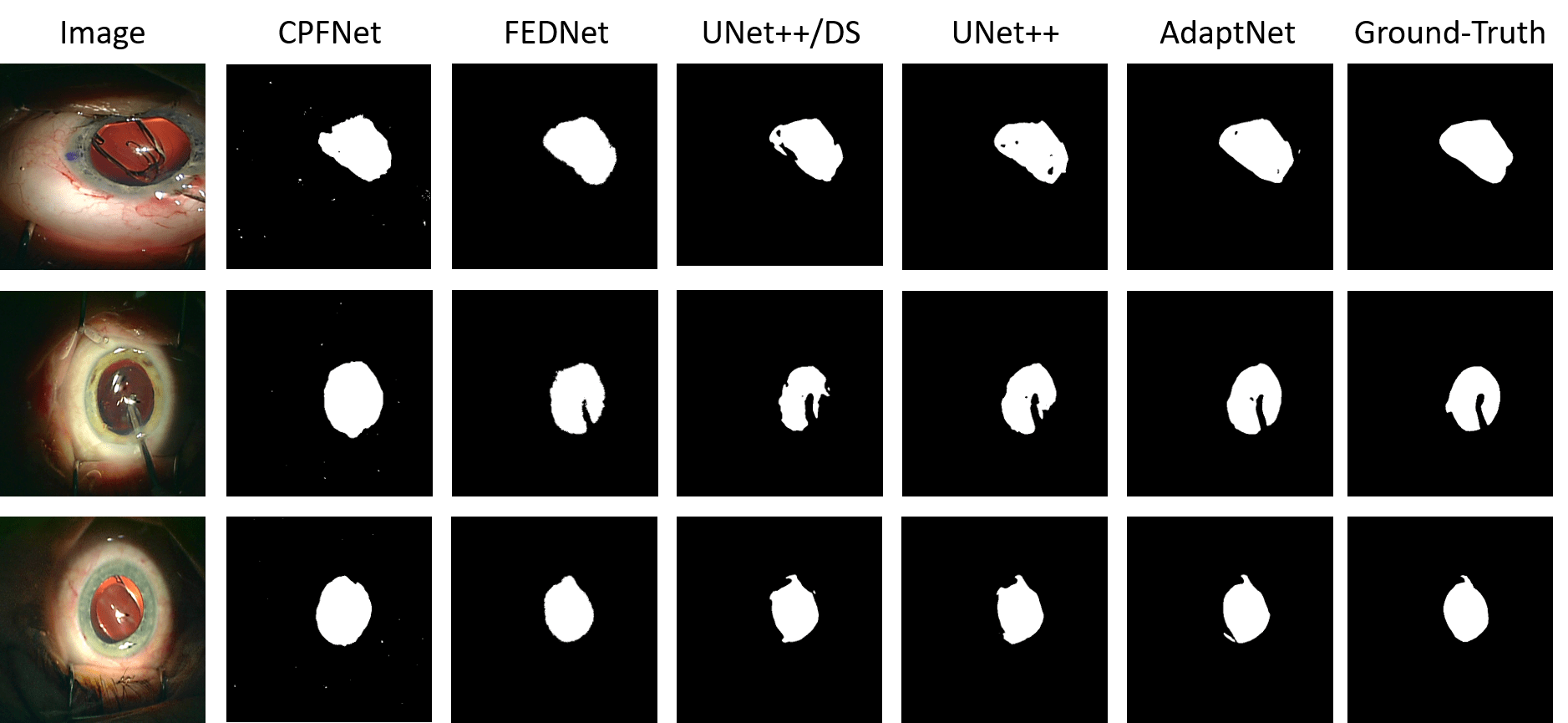}
\caption{Qualitative comparisons among the top five segmentation approaches.}
\label{fig-MICCAI: qualitative}
\end{figure}

Figure~\ref{fig-MICCAI: qualitative} presents qualitative comparisons among the top five approaches for lens segmentation in three representative frames. It can be perceived from the figure that AdaptNet can provide the most visually close segmentation results to the ground truth. Moreover, AdaptNet is more robust against lens deformations as it provides the most delineated predictions compared to the rival approaches.

\begin{figure}[!ht]
\centering
\includegraphics[width=0.9\columnwidth]{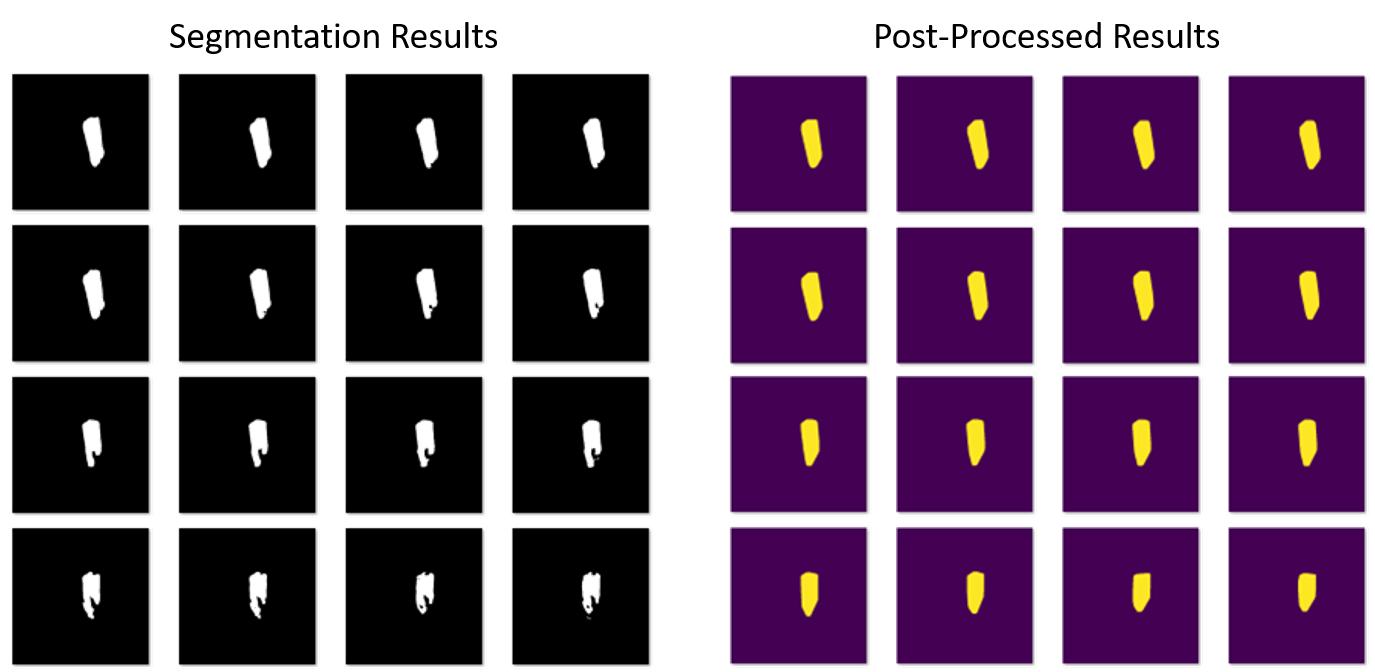}
\caption{Qualitative comparisons among the top five segmentation approaches.}
\label{fig-MICCAI: post-proc}
\end{figure}

Figure~\ref{fig-MICCAI: post-proc} demonstrates the effect of post-processing on segmentation results. We use three morphological operations to improve the semantic segmentation results: (i) opening (with the kernel size of $10\times 10$) to attach the separated regions due to instrument covering, (ii) closing (with the kernel size of $15\times 15$) to remove the distant wrong detections, and (iii) convex polygon. Since instruments usually cover a part of the pupil and intraocular lens during surgery, the segmentation results may contain some holes in the location of instruments. However, the occluded parts should be included in the lens and pupil area. Since the pupil is inherently a convex object, and the intraocular lens is usually convex during unfolding, we draw the smallest convex polygon around these objects to retrieve the occluded segments. For convex polygons, we used the ``Scipy ConvexHull'' function.

\section*{Appendix 2}
Herein, we highlight the application of the proposed LensID framework in comparing the statistical distributions of different intraocular lenses. In particular, we compare three types of statistics naming lens unfolding time (per seconds), lens instability (based on absolute lens' relative movements), and lens rotation (based on absolute lens' degree changes) for two brands of IoL labeled as ``NC1'' and ``XC1''.

\subsection{Lens Unfolding Delay}
\begin{figure}[!bt]
\centering
\includegraphics[width=1\columnwidth]{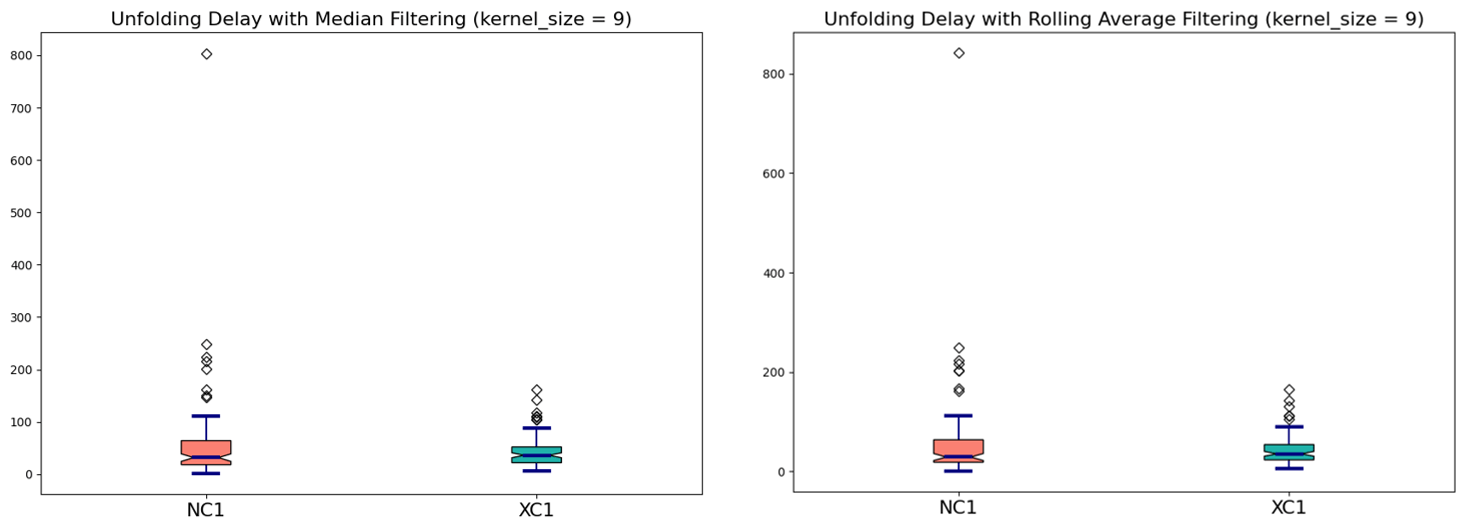}
\caption{Statistical comparison between the unfolding delay of NC1 and XC1 lenses.}
\label{fig-MICCAI: unfolding}
\end{figure}

Figure~\ref{fig-MICCAI: unfolding} compares the distribution of lens unfolding time for 100 surgeries with NC1 lenses versus 100 surgeries with XC1 lenses using boxplot. This figure reveals that the interquartile range (IQR) of XC1 lenses is more compact than that of NC1 lenses. Moreover, there are many outliers being far away from the median value in the distribution of NC1 lenses. These results suggest that XC1 lenses have a more predictable behavior compared to NC1 lenses.

\subsection{Lens Instability}
\begin{figure}[!tb]
\centering
\includegraphics[width=1\columnwidth]{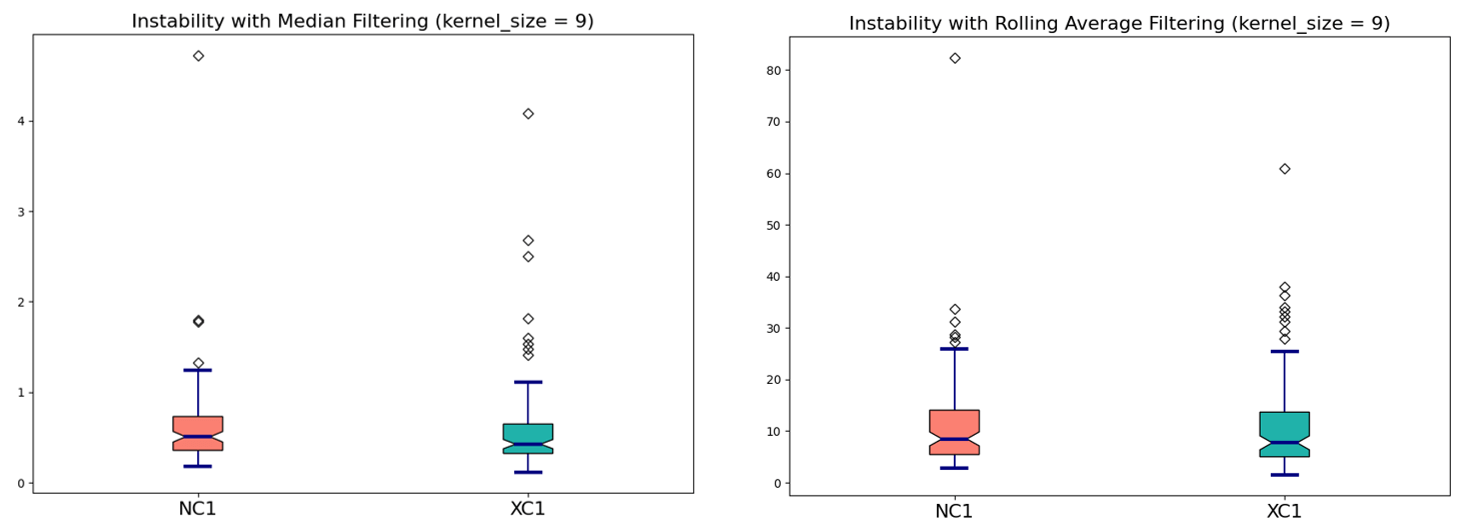}
\caption{Statistical comparison between the instability of NC1 and XC1 lenses.}
\label{fig-MICCAI: instability}
\end{figure}

Lens instability is computed based on the sum of the lens' absolute relative movements inside the pupil. As shown in Figure~\ref{fig-MICCAI: instability}, both NC1 and XC1 lenses offer a relatively close interquartile range (IQR). However, NC1 lenses have a larger value in the lower and upper whisker of lens instability distribution. Besides, the NC1 group has a very distant outlier in the instability distribution.

\begin{figure}[!tb]
\centering
\includegraphics[width=1\columnwidth]{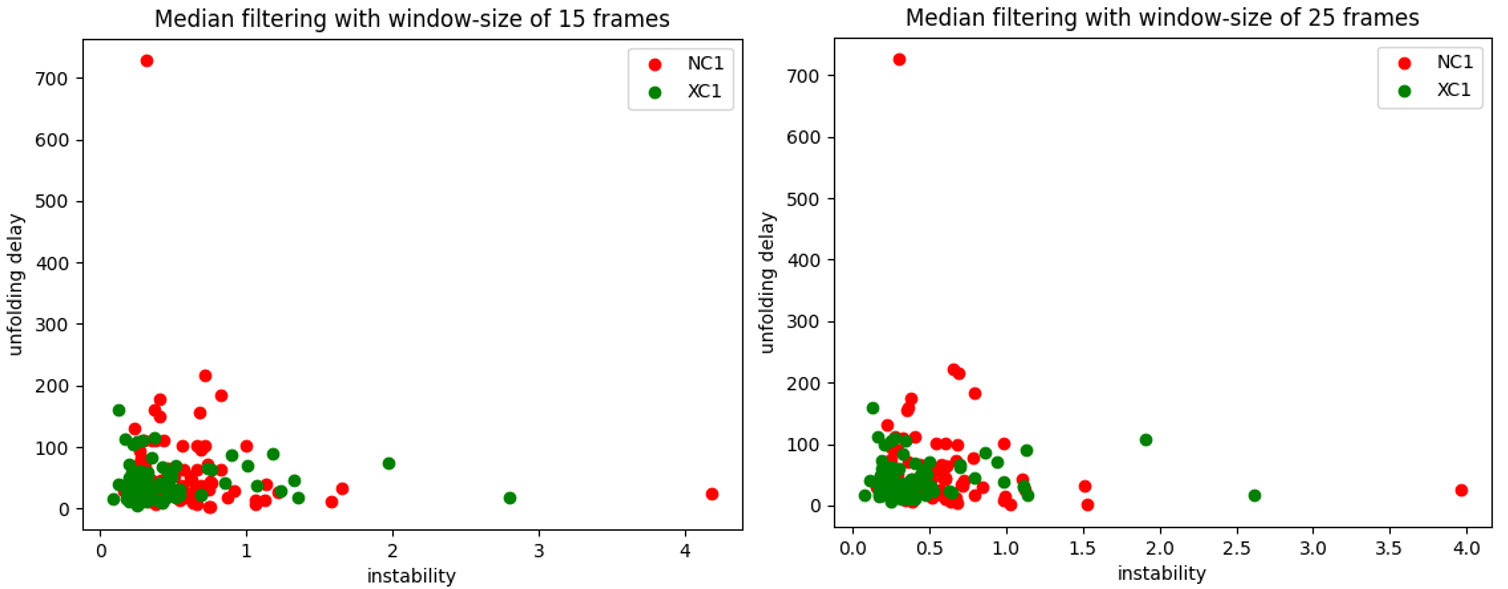}
\caption{Joint distribution of lens unfolding delay and lens instability for NC1 and XC1 lenses.}
\label{fig-MICCAI: unfolding-instability}
\end{figure}

Moving on to Figure~\ref{fig-MICCAI: unfolding-instability}, we can infer that there are some correlations between the lens unfolding time and lens instability. Indeed, NC1 and XC1 lenses' statistical distribution suggests that lenses with more unfolding delay are usually more unstable. Besides, the distribution of XC1 lenses is compact and focused near the origin, whereas the distribution of XC1 lenses is more scattered and farther from the origin. This suggests NC1 lenses have some irregular behavior.

\subsection{Lens Rotation}
\begin{figure}[!tb]
\centering
\includegraphics[width=1\columnwidth]{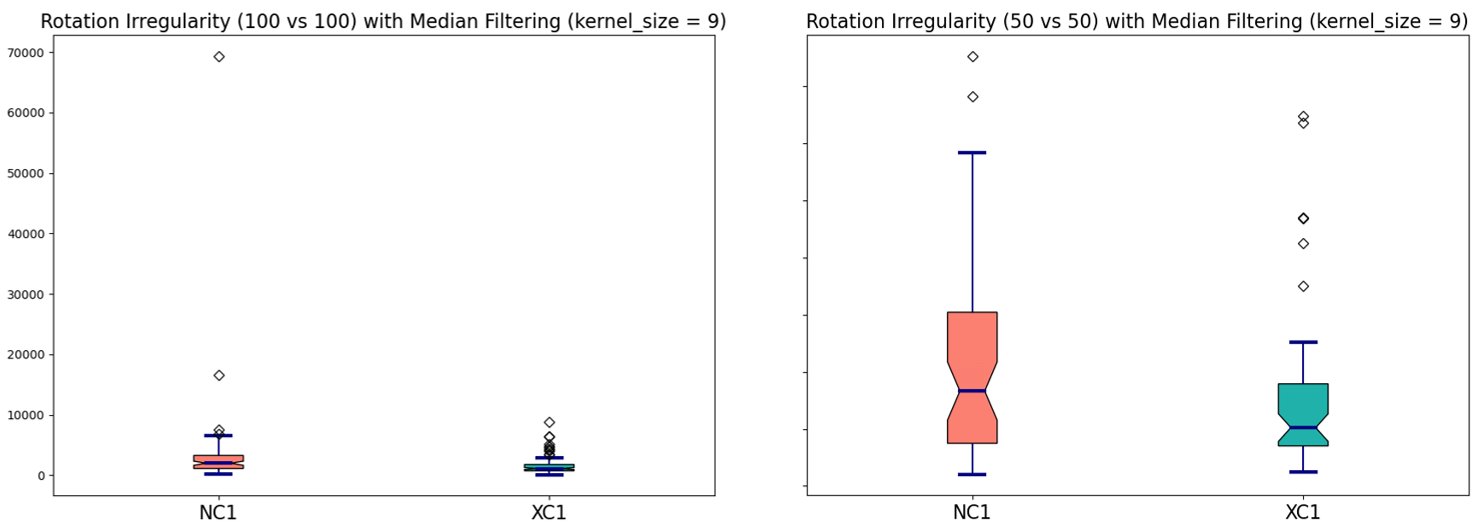}
\caption{Statistical comparison between the rotation of NC1 and XC1 lenses.}
\label{fig-MICCAI: unfolding-irregularity}
\end{figure}

For lens rotation computation, we need a pose estimation approach in addition to the LensID framework. We use Faster R-CNN to compute the degree of IoL when its hooks are visible. We start calculating the lens rotation statistics after lens unfolding. Then lens rotation is calculated based on the sum of absolute IoL degree changes over time. Figure~\ref{fig-MICCAI: unfolding-irregularity}-left compares the boxplots of rotation irregularity for NC1 vs. XC1 lenses (100 vs. 100). It is evident from the figure that the NC1 lens' distribution contains very distant outliers. For better visualization, we have also compared the distributions of 50 NC1 lenses with 50 XC1 lenses in Figure~\ref{fig-MICCAI: unfolding-irregularity}-right. It is evident from the boxplots that the NC1 lens' rotation distribution has a very high upper whisker being more than two times the upper whisker of the XC1 lens' rotation distribution. Besides, the NC1 lens's rotation distribution has a much larger median value and much wider interquartile range compared to the NC1 lens' rotation distribution. This suggests NC1 lenses have less predictable behavior compared to XC1 lenses.

\chapter{Semantic Segmentation using ReCal-Net \label{Chapter: Recal-Net}}

\chapterintro{
	In this chapter, we propose a novel convolutional module termed as \textit{ReCal} module, which can calibrate the feature maps by employing region intra-and-inter-dependencies and channel-region cross-dependencies. This calibration strategy can effectively enhance semantic representation by correlating different representations of the same semantic label, considering a multi-angle local view centering around each pixel. Thus the proposed module can deal with distant visual characteristics of unique objects as well as cross-similarities in the visual characteristics of different objects. Moreover, we propose a novel network architecture based on the proposed module termed as \textit{ReCal-Net}. Experimental results confirm the superiority of ReCal-Net compared to rival state-of-the-art approaches for all relevant objects in cataract surgery. Moreover, ablation studies reveal the effectiveness of the ReCal module in boosting semantic segmentation accuracy.
}

{
	\singlespacing This chapter is an adapted version of:

           ``Ghamsarian, N., Taschwer, M., Putzgruber-Adamitsch, D., Sarny,
S., El-Shabrawi, Y., and Schoeffmann, K. ReCal-Net: Joint Region-Channel-Wise Calibrated Network for Semantic Segmentation in Cataract Surgery Videos. Under Review.'' 
}

\section{Introduction}
\label{Sec-ICONIP: Introduction}

In recent years, a large body of research has been focused on computerized surgical workflow analysis in cataract surgery~\cite{EndoNet,RDCSV,RBCCSV,CataNet,MTRCN,RBCCSV}, with a majority of approaches relying on semantic segmentation. Hence, improving semantic segmentation accuracy in cataract surgery videos can play a leading role in the development of a reliable computerized clinical diagnosis or surgical analysis approach~\cite{BARNet,PAANet}.

Semantic segmentation of the relevant objects in cataract surgery videos is quite challenging due to (i) transparency of the intraocular lens, (ii) color and contextual variation of the pupil and iris, (iii) blunt edges of the iris, and (iv) severe motion blur and reflection distortion of the instruments. In this chapter, we propose a novel module for joint Region-channel-wise Calibration, termed as \textit{ReCal} module. The proposed module can simultaneously deal with the various segmentation challenges in cataract surgery videos. In particular, the ReCal module is able to (1) employ multi-angle pyramid features centered around each pixel position to deal with transparency, blunt edges, and motion blur, (2) employ cross region-channel dependencies to handle texture and color variation through interconnecting the distant feature vectors corresponding to the same object. The proposed module can be added on top of every convolutional layer without changing the output feature dimensions. Moreover, the ReCal module does not impose a significant number of trainable parameters on the network and thus can be used after several layers to calibrate their output feature maps. Besides, we propose a novel semantic segmentation network based on the ReCal module termed as \textit{ReCal-Net}. The experimental results show significant improvement in semantic segmentation of the relevant objects with ReCal-Net compared to the best-performing rival approach ($85.38\%$ compared to $83.32\%$ overall IoU (intersection over union) for ReCal-Net vs. UNet++). 

The rest of this work is organized as follows. 
In Section~\ref{Sec-ICONIP: Methodology}, we first discuss two convolutional blocks from which the proposed approach is inspired, and then delineate the proposed ReCal-Net and ReCal module. We detail the experimental settings in Section~\ref{Sec-ICONIP: Experimental Settings} and present the experimental results in Section~\ref{Sec-ICONIP: Experimental Results}. We finally conclude the paper in Section~\ref{Sec-ICONIP: Conclusion}.

\section{Methodology}
\label{Sec-ICONIP: Methodology}
\paragraph{\textbf{Notations. }}Everywhere in this chapter, we show convolutional layer with the kernel-size of $(m\times n)$, $P$ output channels, and $g$ groups as $*_{(m\times n)}^{P,g}$ (we consider the default dilation rate of 1 for this layer). Besides, we show average-pooling layer with a kernel-size of $(m\times n)$ and a stride of $s$ pixels as $\avsum_{(m\times n)}^{s}$, and global average pooling as $\avsum^{G}$.

\paragraph{\textbf{Feature Map Recalibration. }}The Squeeze-and-Excitation (SE) block~\cite{SAE} was proposed to model inter-channel dependencies through squeezing the spatial features into a channel descriptor, applying fully-connected layers, and rescaling the input feature map via multiplication. This low-complexity operation unit has proved to be effective, especially for semantic segmentation. However, the SE block does not consider pixel-wise features in recalibration. Accordingly, scSE block~\cite{SCSE} was proposed to exploit pixel-wise and channel-wise information concurrently. This block can be split into two parallel operations: (1) spatial squeeze and channel excitation, exactly the same as the SE block, and (2) channel squeeze and spatial excitation. The latter operation is conducted by applying a pixel-wise convolution with one output channel to the input feature map, followed by multiplication. The final feature maps of these two parallel computational units are then merged by selecting the maximum feature in each location. 

\begin{figure}[!tb]
    \centering
    \includegraphics[width=1\columnwidth]{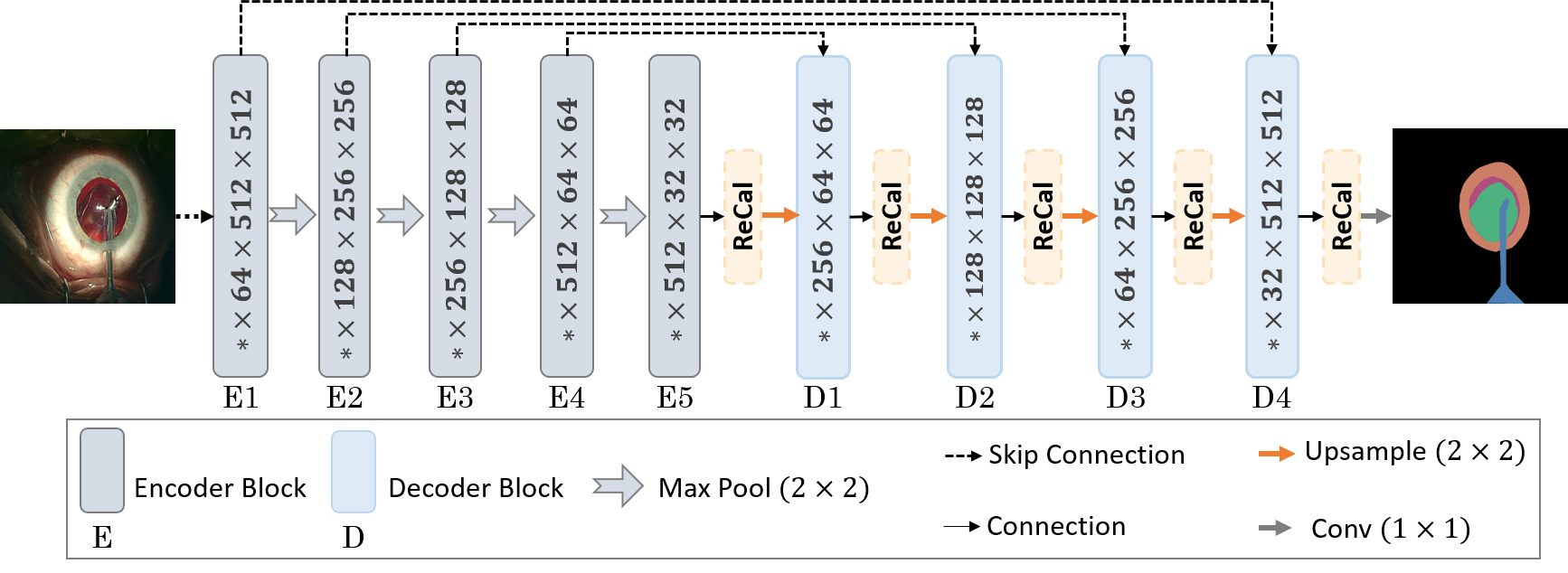}
    \caption{The overall architecture of ReCal-Net containing five ReCal blocks.}
    \label{fig-ICONIP: Block_diagram}
\end{figure}

\paragraph{\textbf{ReCal-Net. }}
Figure~\ref{fig-ICONIP: Block_diagram} depicts the architecture of ReCal-Net. Overall, the network consists of three types of blocks: (i) encoder blocks that transform low-level features to semantic features while compressing the spatial representation, (ii) decoder blocks that are responsible for improving the semantic features in higher resolutions by employing the symmetric low-level feature maps from the encoder blocks, (iii) and \textit{ReCal} modules that account for calibrating the semantic feature maps. We use the VGG16 network as the encoder network. The \textit{i}th encoder block ($\textrm{E}i, i\in\{1,2,3,4\}$) in Figure~\ref{fig-ICONIP: Block_diagram} correspond to all layers between the \textit{i-1}th and \textit{i}th max-pooling layers in the VGG16 network (max-pooling layers are indicated with gray arrows). The last encoder block ($\textrm{E}5$) corresponds to the layers between the last max-pooling layer and the average pooling layer. Each decoder block follows the same architecture of decoder blocks in U-Net~\cite{U-Net}, including two convolutional layers, each of which being followed by batch normalization and ReLU.

\begin{figure}[!tb]
    \centering
    \includegraphics[width=1\columnwidth]{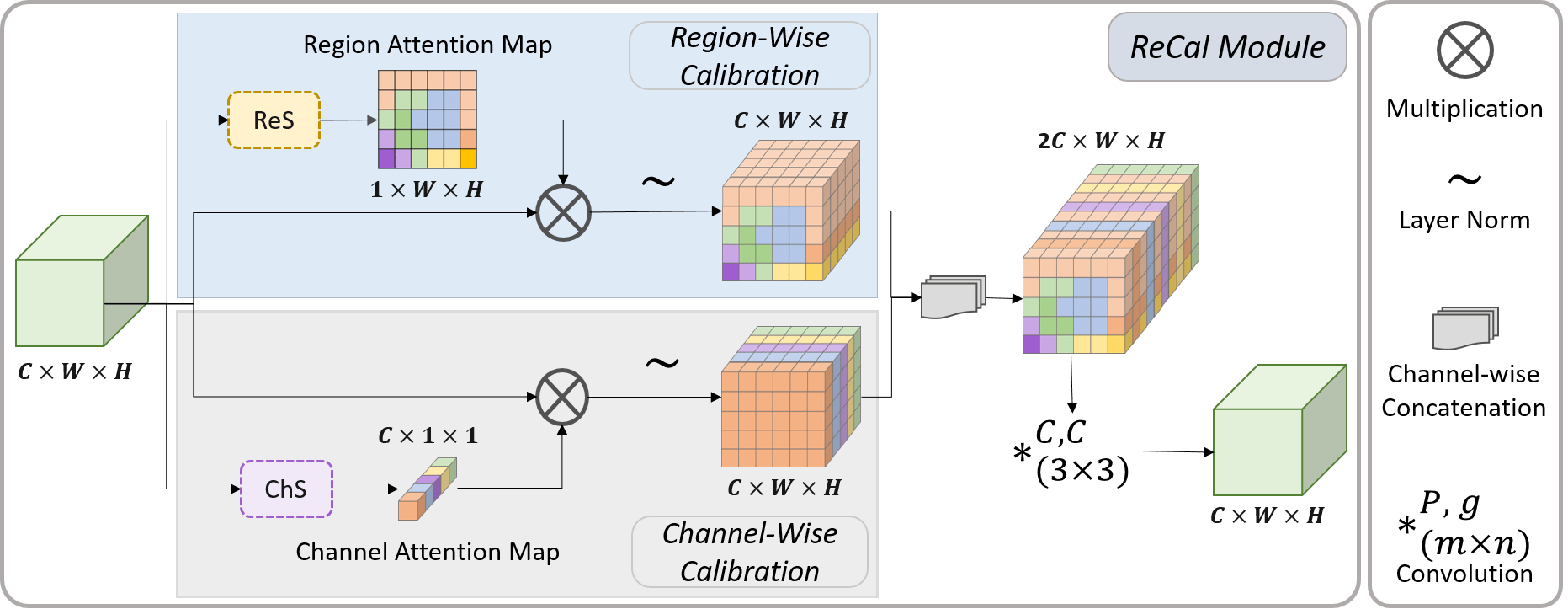}
    \caption{The detailed architecture of ReCal block containing regional squeeze block (ReS) and channel squeeze block (ChS).}
    \label{fig-ICONIP: ReCal Block}
\end{figure}
\paragraph{\textbf{ReCal Module. }}Despite the effectiveness of SE and scSE blocks in boosting feature representation, both fail to exploit region-wise dependencies. However, employing region-wise inter-dependencies and intra-dependencies can significantly enhance semantic segmentation performance. We propose a joint region-channel-wise calibration (ReCal) module to calibrate the feature maps based on joint region-wise and channel-wise dependencies. Figure~\ref{fig-ICONIP: ReCal Block} demonstrates the architecture of the proposed ReCal module inspired by~\cite{SAE,SCSE}. This module aims to reinforce a semantic representation considering inter-channel dependencies, inter-region and intra-region dependencies, and channel-region cross-dependencies. The input feature map of ReCal module $\mathcal{F}_{In}\in \mathbb{R}^{C\times W\times H}$ is first fed into two parallel blocks: (1) the Region-wise Squeeze block (\textit{ReS}), and (2) the Channel-wise Squeeze block (\textit{ChS}). Afterward, the region-wise and channel-wise calibrated features ($\mathcal{F}_{Re} \in \mathbb{R}^{C\times W\times H}$ and $\mathcal{F}_{Ch} \in \mathbb{R}^{C\times W\times H}$) are obtained by multiplying ($\otimes$) the input feature map to the region-attention map and channel-attention map, respectively, followed by the layer normalization function. In this stage, each particular channel $\mathcal{F}_{In}(C_j)\in \mathbb{R}^{W\times H}$ in the input feature map of a ReCal module has corresponding region-wise and channel-wise calibrated channels ($\mathcal{F}_{Re}(C_j)\in \mathbb{R}^{W\times H}$ and $\mathcal{F}_{Ch}(C_j)\in \mathbb{R}^{W\times H}$). To enable the utilization of cross-dependencies between the region-wise and channel-wise calibrated features, we concatenate these two feature maps in a depth-wise manner. Indeed,  the concatenated feature map ($\mathcal{F}_{Concat}$) for each $p\in [1,C]$, $x\in [1,W]$, and $y\in [1,H]$ can be formulated as~\eqref{eq-ICONIP: concat}.  

\begin{equation}
        \begin{cases}
            &\mathcal{F}_{Concat}(2p,x,y) =\mathcal{F}_{Re}(p,x,y) \\
            &\mathcal{F}_{Concat}(2p-1,x,y)
            =\mathcal{F}_{Ch}(p,x,y)
        \end{cases}
        \label{eq-ICONIP: concat}
    \end{equation}

The cross-dependency between region-wise and channel-wise calibrated features is computed using a convolutional layer with $C$ groups. More concretely, every two consecutive channels in the concatenated feature map undergo a distinct convolution with a kernel-size of $(3\times 3)$. This convolutional layer considers the local contextual features around each pixel (a $3\times 3$ window around each pixel) to determine the contribution of each of region-wise and channel-wise calibrated features in the output features. Using a kernel size greater than one unit allows jointly considering inter-region dependencies. 

\begin{figure}[!tb]
    \centering
    \includegraphics[width=1\columnwidth]{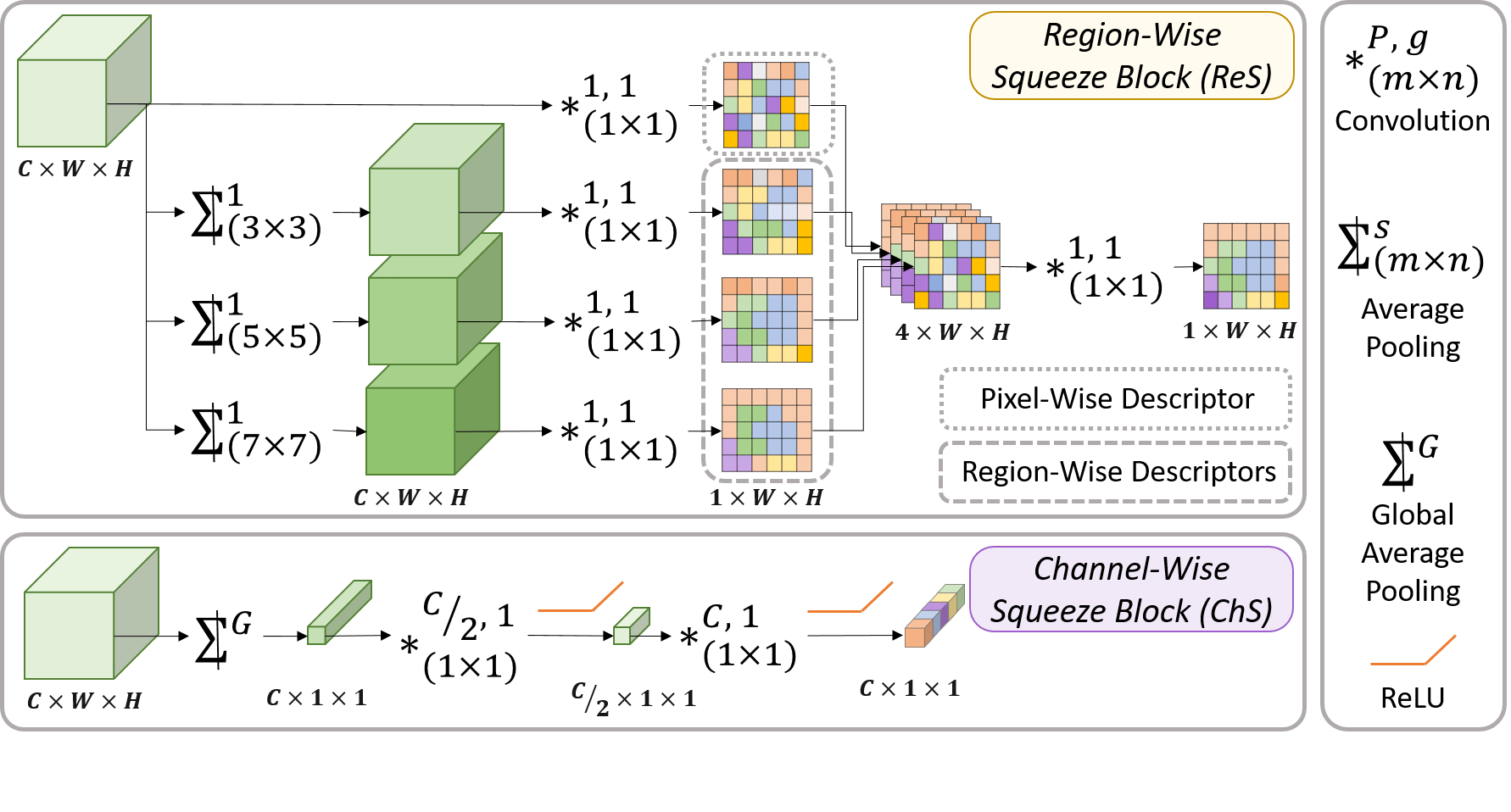}
    \caption{Demonstration of regional squeeze block (ReS) and channel squeeze block (CS).}
    \label{fig-ICONIP: Subblocks}
\end{figure}

\paragraph{\textbf{Region-Wise Squeeze Block. }}Figure~\ref{fig-ICONIP: Subblocks} details the architecture of the ReS block, which is responsible for providing the region attention map. The region attention map is obtained by taking advantage of multi-angle local content based on narrow to wider views around each distinct pixel in the input feature map. We model multi-angle local features using average pooling layers with different kernel sizes and the stride of one pixel. The average pooling layers do not any number of impose trainable parameters on the network and thus ease using the ReS block and ReCal module in multiple locations. Besides, the stride of one pixel in the average pooling layer can stimulate a local view centered around each distinctive pixel. We use three average pooling layers with kernel-sizes of $(3\times 3)$, $(5\times 5)$, and $(7\times 7)$, followed by pixel-wise convolutions with one output channel ($*_{(1\times 1)}^{1,1}$) to obtain the region-wise descriptors. In parallel, the input feature map undergoes another convolutional layer to obtain the pixel-wise descriptor. The local features can indicate if some particular features (could be similar or dissimilar to the centering pixel) exist in its neighborhood, and how large is the neighborhood of each pixel containing particular features. The four attention maps are then concatenated and fed into a convolutional layer ($*_{(1\times 1)}^{1,1}$) that is responsible for determining the contribution of each spatial descriptor in the final region-wise attention map.

\paragraph{\textbf{Channel-Wise Squeeze Block. }}For ChS Block, we follow a similar scheme as in~\cite{SAE}. At first, we apply global average pooling ($\avsum^{G}$) on the input convolutional feature map. Afterward, we form a bottleneck via a pixel-wise convolution with $C/r$ output channels ($*_{(1\times 1)}^{\sfrac{C}{r},1}$) followed by ReLU non-linearity. The scaling parameter $r$ can curbs the computational complexity. Besides, it can act as a smoothing factor that can yield a better-generalized model by preventing the network from learning outliers . In experiments, we set $r=2$ as it is proved to have the best performance~\cite{SCSE}. Finally, another pixel-wise convolution with $C$ output channels ($*_{(1\times 1)}^{C,1}$) followed by ReLU non-linearity is used to provide the channel attention map.

\paragraph{\textbf{Module Complexity. }} Suppose we have an intermediate layer in the network with convolutional response map $\mathcal{X} \in \mathbb{R}^{C\times H\times W}$. Adding a ReCal module on top of this layer with its scaling parameter being equal to 2, amounts to ``$C^2+22C+4$'' additional trainable weights. More specifically, each convolutional layer $*_{(m\times n)}^{P,g}$ applied to $C$ input channels amounts to $\sfrac{((m\times n)\times C\times P)}{g}$ trainable weights. Accordingly, we need ``$4C+4$'' weights for the ReS block, ``$C^2$'' weights for the ChS block, and ``$18C$'' weights for the last convolution operation of the ReCal module. In our proposed architecture, adding five ReCal modules on convolutional feature maps with 512, 256, 128, 64, and 32 channels sums up to $371 K$ additional weights, and only $21 K$ more trainable parameters compared to the SE block~\cite{SAE} and scSE block~\cite{SCSE}.

\section{Experimental Settings}
\label{Sec-ICONIP: Experimental Settings}
\paragraph{\textbf{Datasets. }}
We use four datasets in this study. The iris dataset is created by annotating the cornea and pupil from 14 cataract surgery videos using ``supervisely'' platform. The iris annotations are then obtained by subtracting the convex-hull of the pupil segment from the cornea segment. This dataset contains 124 frames from 12 videos for training and 23 frames from two videos for testing. For lens and pupil segmentation, we employ the two public datasets of the LensID framework~\cite{LensID}, containing the annotation of the intraocular lens and pupil. The lens dataset consists of lens annotation in 401 frames sampled from 27 videos. From these annotations, 292 frames from 21 videos are used for training, and 109 frames from the remaining six videos are used for testing. The pupil segmentation dataset contains 189 frames from 16 videos. The training set consists of 141 frames from 13 videos, and the testing set contains 48 frames from three remaining videos. For instrument segmentation, we use the instrument annotations of the CaDIS dataset~\cite{CaDIS}. We use 3190 frames from 18 videos for training and 459 frames from three other videos for testing.

\paragraph{\textbf{Rival Approaches. }}
Table~\ref{Tab-ICONIP:specification} lists the specifications of the rival state-of-the-art approaches used in our evaluations. In ``Upsampling'' column, ``Trans Conv'' stands for \textit{Transposed Convolution}. To enable direct comparison between the ReCal module and scSE block, we have formed scSE-Net by replacing the ReCal modules in ReCal-Net with scSE modules. Indeed, the baseline of both approaches are the same, and the only difference is the use of scSE blocks in scSE-Net at the position of ReCal modules in ReCal-Net.

\begin{table*}[t!]
\renewcommand{\arraystretch}{1}
\caption{Specifications of the proposed and rival segmentation approaches.}
\label{Tab-ICONIP:specification}
\centering
\resizebox{\textwidth}{!}{
\begin{tabular}{lccccc}
\specialrule{.12em}{.05em}{.05em}
Model & Backbone & Params & Upsampling & Reference & Year\\\specialrule{.12em}{.05em}{.05em}
UNet$++$ (\slash DS) &VGG16&24.24 M& Bilinear &~\cite{UNet++} & 2020\\
MultiResUNet &\xmark& 9.34 \enspace M& Trans Conv &~\cite{MultiResUNet} & 2020\\
BARNet&ResNet34&24.90 M & Bilinear &~\cite{BARNet} & 2020\\
PAANet &ResNet34& 22.43 M & Trans Conv \& Bilinear & \cite{PAANet} & 2020 \\
CPFNet &ResNet34&34.66 M& Bilinear &~\cite{CPFNet} & 2020\\
dU-Net &\xmark &31.98 M& Trans Conv &~\cite{dU-Net} & 2020\\
CE-Net &ResNet34&29.90 M&Trans Conv&~\cite{CE-Net} & 2019\\
scSE-Net &VGG16&22.90 M&Bilinear&~\cite{SCSE} & 2019\\
U-Net &\xmark &17.26 M& Bilinear &~\cite{U-Net} & 2015\\\cdashline{1-6}[0.8pt/1pt]
ReCal-Net&VGG16 &22.92 M& Bilinear &\multicolumn{2}{c}{Proposed}\\
\specialrule{.12em}{.05em}{0.05em}
\end{tabular}
}
\end{table*}

\paragraph{\textbf{Data Augmentation Methods. }}We use the Albumentations~\cite{Albumentations} library for image and mask augmentation during training. Considering the inherent features of the relevant objects and problems of the recorded videos~\cite{DCS}, we apply motion blur, median blur, brightness and contrast change, shifting, scaling, and rotation for augmentation. We use the same augmentation pipeline for the proposed and rival approaches.

\paragraph{\textbf{Neural Network Settings. }}
We initialize the parameters of backbones for the proposed and rival approaches (in case of having a backbone) with ImageNet~\cite{ImageNet} training weights. We set the input size of all networks to ($3\times 512\times 512$). 

\paragraph{\textbf{Training Settings. }}
During training with all networks, a threshold of $0.1$ is applied for gradient clipping. This strategy can prevent the gradient from exploding and result in a more appropriate behavior during learning in the case of irregularities in the loss landscape. Considering the different depths and connections of the proposed and rival approaches, all networks are trained with two different initial learning rates ($lr\in\{0.005,0.002\}$) for 30 epochs with SGD optimizer. The learning scheduler decreases the learning rate every other epoch with a factor of $0.8$. We list the results with the highest IoU for each network.

\paragraph{\textbf{Loss Function. }}
To provide a fair comparison, we use the same loss function for all networks. The loss function is set to a weighted sum of binary cross-entropy ($BCE$) and the logarithm of soft Dice coefficient as follows. 

\begin{equation}
\begin{aligned}
    \mathcal{L} = &(\lambda)\times BCE(\mathcal{X}_{true}(i,j),\mathcal{X}_{pred}(i,j))\\
    &-(1-\lambda)\times (\log \frac{2\sum \mathcal{X}_{true}\odot \mathcal{X}_{pred}+\sigma}{\sum \mathcal{X}_{true} + \sum \mathcal{X}_{pred}+ \sigma})
\end{aligned}
\label{eq-ICONIP: loss}
\end{equation}
Soft Dice refers to the dice coefficient computed directly based on predicted probabilities rather than the predicted binary masks after thresholding. In~\eqref{eq-ICONIP: loss}, $\mathcal{X}_{true}$ refers to the ground truth mask, $\mathcal{X}_{pred}$ refers to the predicted mask, $\odot$ refers to Hadamard product (element-wise multiplication), and $\sigma$ refers to the smoothing factor. In experiments, we set $\lambda = 0.8$ and $\sigma = 1$.

Soft Dice refers to the dice coefficient computed directly based on predicted probabilities rather than the predicted binary masks after thresholding. In~\eqref{eq-ICONIP: loss}, $\mathcal{X}_{true}$ refers to the ground truth mask, $\mathcal{X}_{pred}$ refers to the predicted mask, $\odot$ refers to Hadamard product (element-wise multiplication), and $\sigma$ refers to the smoothing factor. In experiments, we set $\lambda = 0.8$ and $\sigma = 1$.

\begin{table}[t!]
\renewcommand{\arraystretch}{0.9}
\caption{Quantitative comparisons among the semantic segmentation results of Recal-Net and rival approaches based on IoU(\%).}
\label{ICONIP_Tab-ICONIP:IoU}
\centering
\resizebox{\textwidth}{!}{
\begin{tabular}{lP{2.3cm}P{2.3cm}P{2.3cm}P{2.3cm}P{2.3cm}}
\specialrule{.12em}{.05em}{.05em}
Network&Lens&Pupil&Iris&Instruments&Overall\\\specialrule{.12em}{.05em}{.05em}
\rowcolor{shadecolor}U-Net&61.89 \scriptsize{$\pm 20.93$}&83.51 \scriptsize{$\pm 20.24$}&65.89 \scriptsize{$\pm 16.93$}&60.78 \scriptsize{$\pm 26.04$}&68.01 \scriptsize{$\pm 21.03$}\\
CE-Net&78.51 \scriptsize{$\pm 11.56$}& 92.07 \scriptsize{$\pm \enspace 4.24$}&71.74 \scriptsize{$\pm \enspace 6.19$}& 69.44 \scriptsize{$\pm 17.94$}& 77.94 \scriptsize{$\pm \enspace 9.98$}\\
\rowcolor{shadecolor}dU-Net&60.39 \scriptsize{$\pm 29.36$}&68.03 \scriptsize{$\pm 35.95$}&70.21 \scriptsize{$\pm 12.97$}&61.24 \scriptsize{$\pm 27.64$}&64.96 \scriptsize{$\pm 26.48$}\\
scSE-Net&86.04 \scriptsize{$\pm 11.36$}&96.13 \scriptsize{$\pm \enspace 2.10$}&78.58 \scriptsize{$\pm \enspace 9.61$}& 71.03 \scriptsize{$\pm 23.25$}& 82.94 \scriptsize{$\pm 11.58$}\\
\rowcolor{shadecolor}CPFNet&80.65 \scriptsize{$\pm 12.16$}&93.76 \scriptsize{$\pm \enspace 2.87$}&77.93 \scriptsize{$\pm \enspace 5.42$}& 69.46 \scriptsize{$\pm 17.88$}& 80.45 \scriptsize{$\pm \enspace 9.58$}\\
BARNet& 80.23 \scriptsize{$\pm 14.57$}& 93.64 \scriptsize{$\pm \enspace 4.11$}& 75.80 \scriptsize{$\pm \enspace 8.68$}&69.76 \scriptsize{$\pm 21.29$}&79.86 \scriptsize{$\pm 12.16$}\\
\rowcolor{shadecolor}PAANet& 80.30 \scriptsize{$\pm 11.73$}& 94.35 \scriptsize{$\pm \enspace 3.88$}& 75.73 \scriptsize{$\pm 11.67$}&68.01 \scriptsize{$\pm 22.29$}&79.59 \scriptsize{$\pm 12.39$}\\
MultiResUNet&61.42 \scriptsize{$\pm 19.91$}&76.46 \scriptsize{$\pm 29.43$}&49.99 \scriptsize{$\pm 28.73$}&61.01 \scriptsize{$\pm 26.94$}&62.22 \scriptsize{$\pm 26.25$}\\
\rowcolor{shadecolor}UNet++/DS&84.53 \scriptsize{$\pm 13.42$}&96.18 \scriptsize{$\pm \enspace 2.62$}&74.01 \scriptsize{$\pm 13.13$}&65.99 \scriptsize{$\pm 25.66$}&79.42 \scriptsize{$\pm 14.75$}\\
UNet++&85.74 \scriptsize{$\pm 11.16$}&96.50 \scriptsize{$\pm \enspace 1.51$}&81.98 \scriptsize{$\pm \enspace 6.96$}&69.07 \scriptsize{$\pm 23.89$}&83.32 \scriptsize{$\pm 10.88$}\\
\rowcolor{shadecolor}ReCal-Net&\textbf{87.94} \scriptsize{$\pm \textbf{10.72}$}&\textbf{96.58} \scriptsize{$\pm \enspace \textbf{1.30}$}&\textbf{85.13} \scriptsize{$\pm \enspace \textbf{3.98}$}& \textbf{71.89} \scriptsize{$\pm \textbf{19.93}$}&\textbf{85.38} \scriptsize{$\pm \enspace \textbf{8.98}$}\\
 \specialrule{.12em}{.05em}{.05em}
\end{tabular}
}
\end{table}
\begin{table}[t!]
\renewcommand{\arraystretch}{0.9}
\caption{Quantitative comparisons among the semantic segmentation results of Recal-Net and rival approaches based on Dice(\%).}
\label{ICONIP_Tab-ICONIP:Dice}
\centering
\resizebox{\textwidth}{!}{
\begin{tabular}{lP{2.3cm}P{2.3cm}P{2.3cm}P{2.3cm}P{2.3cm}}
\specialrule{.12em}{.05em}{.05em}
Network&Lens&Pupil&Iris&Instruments&Overall\\\specialrule{.12em}{.05em}{.05em}
\rowcolor{shadecolor}U-Net&73.86 \scriptsize{$\pm 20.39$}&89.36 \scriptsize{$\pm 15.07$}&78.12 \scriptsize{$\pm 13.01$}&71.50 \scriptsize{$\pm 25.77$}&78.21 \scriptsize{$\pm 18.56$}\\
CE-Net& 87.32 \scriptsize{$\pm \enspace 9.98$}&95.81 \scriptsize{$\pm \enspace 2.39$}&83.39 \scriptsize{$\pm \enspace 4.25$}&80.30 \scriptsize{$\pm 15.97$}& 86.70 \scriptsize{$\pm \enspace 8.15$}\\
\rowcolor{shadecolor}dU-Net&69.99 \scriptsize{$\pm 29.40$}&73.72 \scriptsize{$\pm 34.24$}&81.76 \scriptsize{$\pm \enspace 9.73$}&71.30 \scriptsize{$\pm 27.62$}&74.19 \scriptsize{$\pm 25.24$}\\
scSE-Net&91.95 \scriptsize{$\pm \enspace 9.14$}&98.01 \scriptsize{$\pm \enspace 1.10$}&87.66 \scriptsize{$\pm \enspace 6.35$}& 80.18 \scriptsize{$\pm 21.49$}& 89.45 \scriptsize{$\pm \enspace 9.52$}\\
\rowcolor{shadecolor}CPFNet&88.61 \scriptsize{$\pm 10.20$}&96.76 \scriptsize{$\pm \enspace 1.53$}&87.48 \scriptsize{$\pm \enspace 3.60$}& 80.33 \scriptsize{$\pm 15.85$}& 88.29 \scriptsize{$\pm \enspace 7.79$}\\
BARNet& 88.16 \scriptsize{$\pm 10.87$}& 96.66 \scriptsize{$\pm \enspace 2.30$}& 85.95 \scriptsize{$\pm \enspace 5.73$}&79.72 \scriptsize{$\pm 19.95$}&87.62 \scriptsize{$\pm \enspace 9.71$}\\
\rowcolor{shadecolor}PAANet& 88.46 \scriptsize{$\pm \enspace 9.59$}& 97.05 \scriptsize{$\pm \enspace 2.16$}& 85.62 \scriptsize{$\pm \enspace 8.50$}&78.15 \scriptsize{$\pm 21.51$}&87.32 \scriptsize{$\pm 10.44$}\\
MultiResUNet&73.88 \scriptsize{$\pm 18.26$}&82.45 \scriptsize{$\pm 25.49$}& 61.78 \scriptsize{$\pm 25.96$}&71.35 \scriptsize{$\pm 26.88$}&72.36 \scriptsize{$\pm 24.14$}\\
\rowcolor{shadecolor}UNet++/DS&90.80 \scriptsize{$\pm 11.41$}&98.03 \scriptsize{$\pm \enspace 1.41$}&84.38 \scriptsize{$\pm \enspace 9.06$}&75.64 \scriptsize{$\pm 25.38$}&87.21 \scriptsize{$\pm 11.81$}\\
UNet++&91.80 \scriptsize{$\pm 11.16$}&98.26 \scriptsize{$\pm \enspace 0.79$}&89.93 \scriptsize{$\pm \enspace 4.51$}&78.54 \scriptsize{$\pm 22.76$}&89.63 \scriptsize{$\pm \enspace 9.80$}\\
\rowcolor{shadecolor}ReCal-Net&\textbf{93.09} \scriptsize{$\pm \enspace \textbf{8.56}$}&\textbf{98.26} \scriptsize{$\pm \enspace \textbf{0.68}$}&\textbf{91.91} \scriptsize{$\pm \enspace \textbf{2.47}$}& \textbf{81.62} \scriptsize{$\pm \textbf{17.75}$}&\textbf{91.22} \scriptsize{$\pm \enspace \textbf{7.36}$}\\
 \specialrule{.12em}{.05em}{.05em}
\end{tabular}
}
\end{table}

\section{Experimental Results}
\label{Sec-ICONIP: Experimental Results}
Table~\ref{ICONIP_Tab-ICONIP:IoU} and Table~\ref{ICONIP_Tab-ICONIP:Dice} compare the segmentation performance of ReCal-Net and ten rival state-of-the-art approaches based on the average and standard deviation of IoU and Dice coefficient, respectively~\footnote{The ``Overall'' column in Table~\ref{ICONIP_Tab-ICONIP:IoU} and Table~\ref{ICONIP_Tab-ICONIP:Dice} is the mean of the other four values.}. Overall, ReCal-Net, UNet++, scSE-Net, and CPFNet have shown the top four segmentation results. Moreover, the experimental results reveal that ReCal-Net has achieved the highest average IoU and Dice coefficient for all relevant objects compared to state-of-the-art approaches. Considering the IoU report, ReCal-Net has gained considerable enhancement in segmentation performance compared to the second-best approach in lens segmentation ($87.94\%$ vs. $86.04\%$ for scSE-Net) and iris segmentation ($85.13\%$ vs. $81.98\%$ for UNet++). Having only 21k more trainable parameters than scSE-Net ($0.08\%$ additive trainable parameters), ReCal-Net has achieved $8.3\%$ relative improvement in iris segmentation, $2.9\%$ relative improvement in instrument segmentation, and $2.2\%$ relative improvement in lens segmentation in comparison with scSE-Net. 
Regarding the Dice coefficient, ReCal-Net and UNet++ show very similar performance in pupil segmentation. However, with 1.32M fewer parameters than UNet++ as the second-best approach, ReCal-Net shows $1.7\%$ relative improvement in overall Dice coefficient ($91.22\%$ vs. $89.63\%$). Surprisingly, replacing the scSE blocks with the ReCal modules results in $4.25\%$ higher Dice coefficient for iris segmentation and $1.44\%$ higher Dice coefficient for instrument segmentation.

\begin{figure}[!tb]
    \centering
    \includegraphics[width=1\columnwidth]{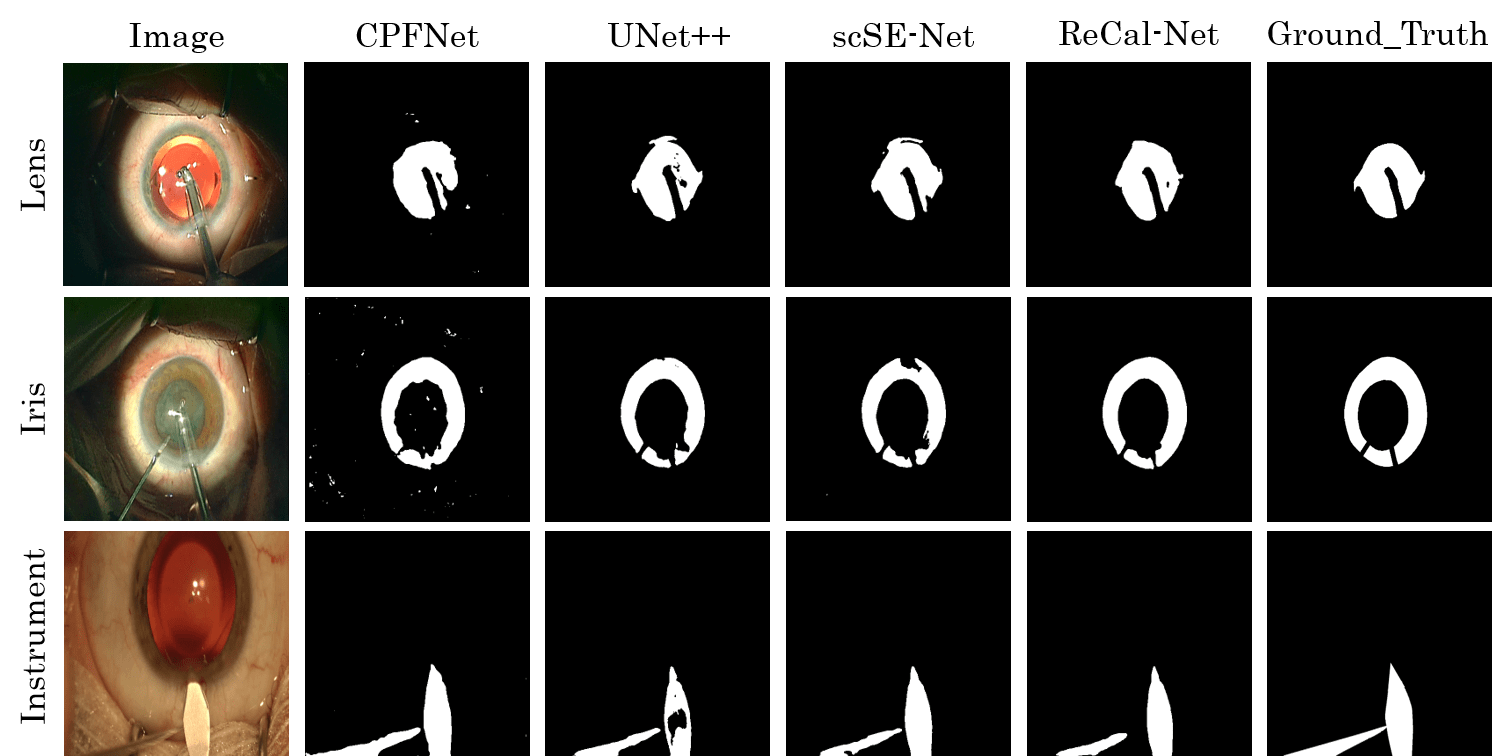}
    \caption{Qualitative comparisons among the top four segmentation approaches.}
    \label{fig-ICONIP: qualitative}
\end{figure}

Figure~\ref{fig-ICONIP: qualitative} provides qualitative comparisons among the top four segmentation approaches for lens, iris, and instrument segmentation. Comparing the visual segmentation results of ReCal-Net and scSE-Net further highlights the effectiveness of region-wise and cross channel-region calibration in boosting semantic segmentation performance.

Table~\ref{ICONIP_Tab-ICONIP:ablation} reports the ablation study by comparing the segmentation performance of the baseline approach with ReCal-Net considering two different learning rates. The baseline approach refers to the network obtained after removing all ReCal modules of ReCal-Net in Figure~\ref{fig-ICONIP: Block_diagram}. These results approve of the ReCal module's effectiveness regardless of the learning rate.

\begin{figure}[!tb]
    \centering
    \includegraphics[width=1\columnwidth]{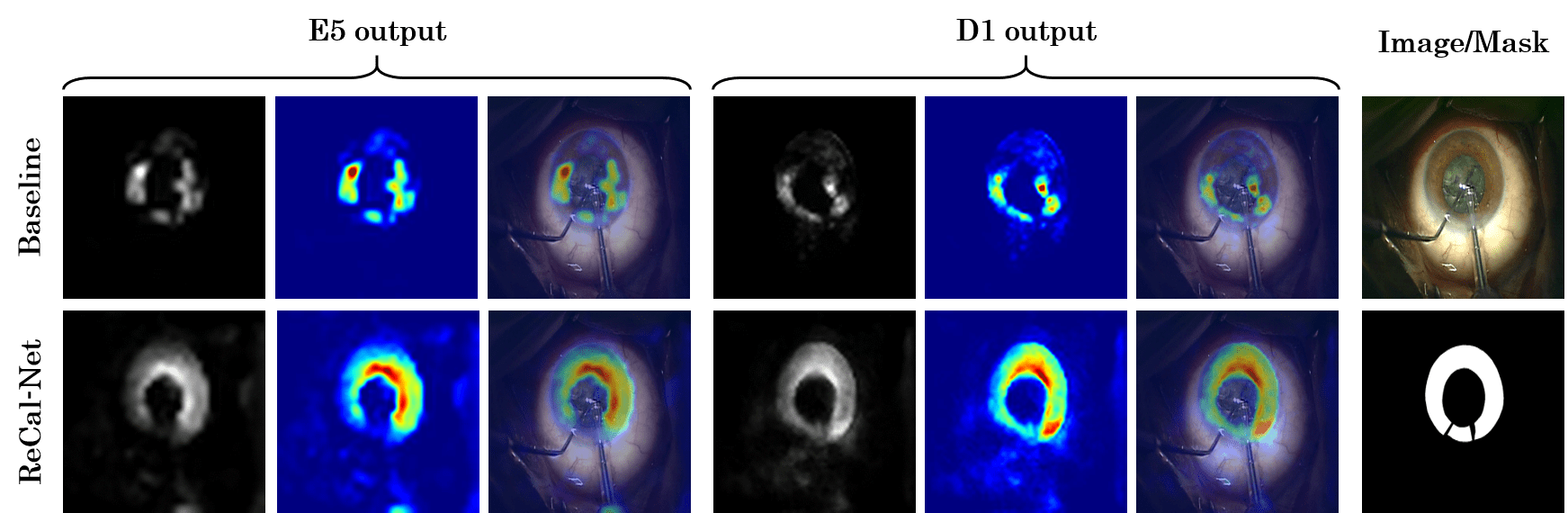}
    \caption{Visualizations of the intermediate outputs in the baseline approach and ReCal-Net based on  class activation maps~\cite{Score-CAM}. For each output, the figures from left to right represent the gray-scale activation maps, heatmaps, and heatmaps on images.}
    \label{fig-ICONIP: score-cam}
\end{figure}

To further investigate the impact of the ReCal modules on segmentation performance, we have visualized two intermediate filter response maps for iris segmentation in Figure~\ref{fig-ICONIP: score-cam}. The E5 output corresponds to the filter response map of the last encoder block, and the D1 output corresponds to the filter response map of the first decoder block (see Figure~\ref{fig-ICONIP: Block_diagram}). A comparison between the filter response maps of the baseline and ReCal-Net in the same locations indicated the positive impact of the ReCal modules on the network's semantic discrimination capability. Indeed, employing the correlations between the pixel-wise, region-wise, and channel-wise descriptors can reinforce the network's semantic interpretation.

\begin{table}[t!]
\caption{Impact of adding ReCal modules on the segmentation accuracy based on IoU(\%).}
\label{ICONIP_Tab-ICONIP:ablation}
\centering
\begin{tabular}{P{2cm} P{2cm} P{2cm} P{2cm} P{2cm}}
\specialrule{.12em}{.05em}{.05em}
Learning Rate & Network & Lens & Iris & Instrument\\
\specialrule{.12em}{.05em}{.05em}
\multirow{2}{*}{0.002} & Baseline & 84.83 \scriptsize{$\pm 11.62$} & 81.49 \scriptsize{$\pm \enspace 6.82$}& 70.04 \scriptsize{$\pm 23.94$}\\
& ReCal-Net & 85.77 \scriptsize{$\pm 12.33$}& 83.29 \scriptsize{$\pm \enspace 5.82$}& 71.89 \scriptsize{$\pm 19.93$}\\\hline
\multirow{2}{*}{0.005} & Baseline & 86.13 \scriptsize{$\pm 11.63$}& 81.00 \scriptsize{$\pm \enspace 8.06$}& 67.16 \scriptsize{$\pm 24.67$}\\
& ReCal-Net & 87.94 \scriptsize{$\pm 10.72$} &85.13 \scriptsize{$\pm \enspace 3.98$}& 70.43 \scriptsize{$\pm 21.17$}\\
\specialrule{.12em}{.05em}{.05em}
\end{tabular}
\end{table}

\section{Conclusion}
\label{Sec-ICONIP: Conclusion}
This chapter presents a novel convolutional module, termed as ReCal module, that can adaptively calibrate feature maps considering pixel-wise, region-wise, and channel-wise descriptors. The ReCal module can effectively correlate intra-region information and cross-channel-region information to deal with severe contextual variations in the same semantic labels and contextual similarities between different semantic labels. The proposed region-channel recalibration module is a very light-weight computational unit that can be applied to any feature map $\mathcal{X} \in \mathbb{R}^{C\times H\times W}$ and output a recalibrated feature map $\mathcal{Y} \in \mathbb{R}^{C\times H\times W}$. Moreover, we have proposed a novel network architecture based on the ReCal module for semantic segmentation in cataract surgery videos, termed as ReCal-Net. The experimental evaluations confirm the effectiveness of the proposed ReCal module and ReCal-Net in dealing with various segmentation challenges in cataract surgery. The proposed ReCal module and ReCal-Net can be adopted for various medical image segmentation and general semantic segmentation problems.
\chapter{Semantic Segmentation using DeepPyram \label{Chapter:DeepPyram}}

\chapterintro{
	Semantic segmentation in cataract surgery has a wide range of applications contributing to surgical outcome enhancement and clinical risk reduction. However, the varying issues in segmenting the different relevant instances make the designation of a unique network quite challenging. This chapter proposes a semantic segmentation network termed as DeepPyram that can achieve superior performance in segmenting relevant objects in cataract surgery videos with varying issues. This superiority mainly originates from three modules: (i) Pyramid View Fusion, which provides a varying-angle global view of the surrounding region centering at each pixel position in the input convolutional feature map; (ii) Deformable Pyramid Reception, which enables a wide deformable receptive field that can adapt to geometric transformations in the object of interest; and (iii) Pyramid Loss that adaptively supervises multi-scale semantic feature maps. These modules can effectively boost semantic segmentation performance, especially in the case of transparency, deformability, scalability, and blunt edges in objects.
The proposed approach is evaluated using four datasets of cataract surgery for objects with different contextual features and compared with thirteen state-of-the-art segmentation networks. The experimental results confirm that DeepPyram outperforms the rival approaches without imposing additional trainable parameters. Our comprehensive ablation study further proves the effectiveness of the proposed modules.
}

{
	\singlespacing This chapter is an adapted version of:

``Ghamsarian, N., Taschwer, and Schoeffmann, K. DeepPyram: Enabling Pyramid View and Deformable Pyramid Reception for Semantic Segmentation in Cataract Surgery Videos. Under Review.'' 
}

\section{Introduction}
\label{sec-TMI:introduction}
\begin{figure}[!tb]
    \centering
    \includegraphics[width=1\columnwidth]{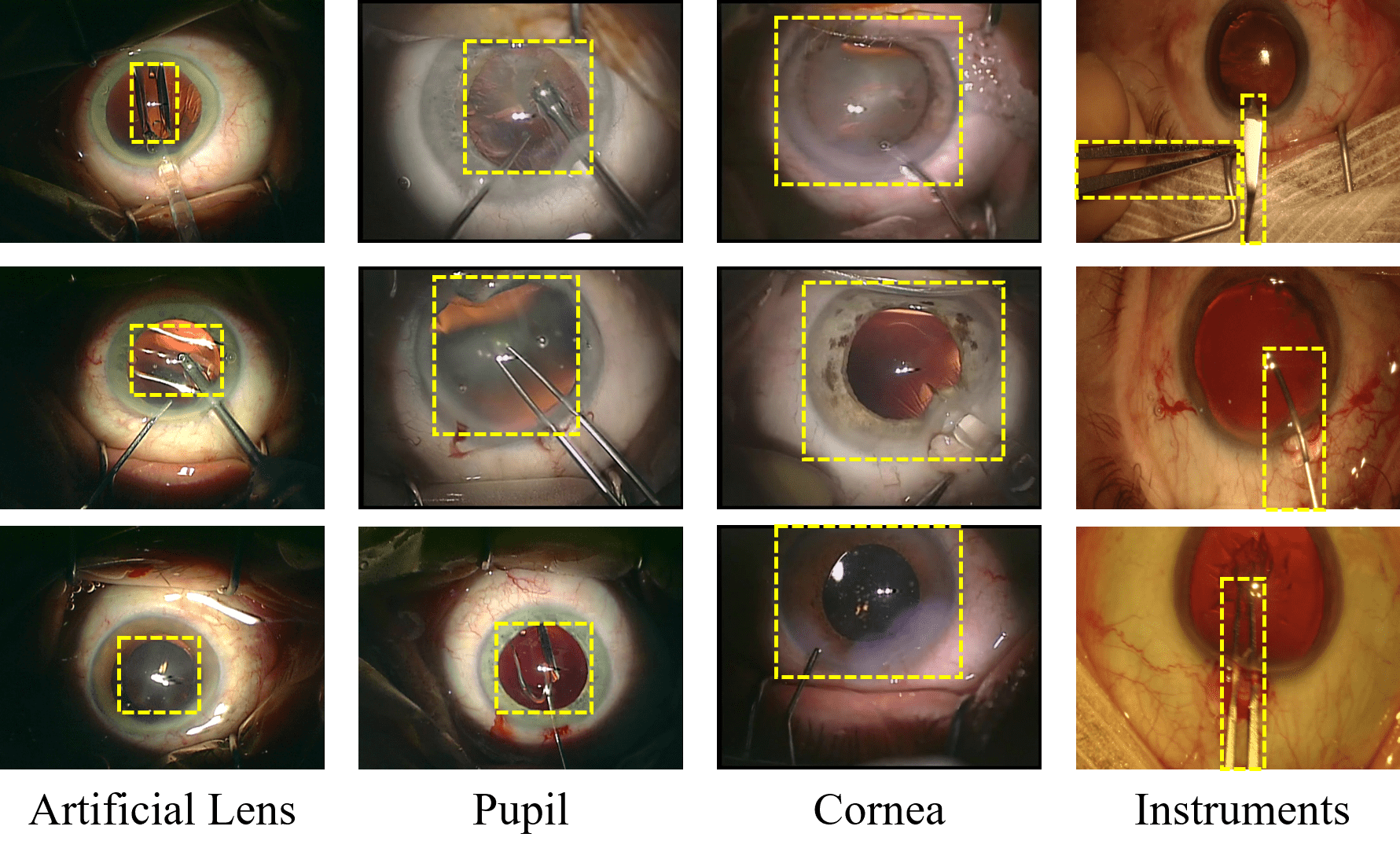}
    \caption{Semantic Segmentation difficulties for different relevant objects in cataract surgery videos.}
    \label{Fig-TMI: Problems}
\end{figure}

Semantic segmentation plays a prominent role in computerized surgical workflow analysis. Especially in cataract surgery, where workflow analysis can highly contribute to the reduction of intra-operative and post-operative complications\cite{RBE}, semantic segmentation is of great importance. Cataract refers to the eye's natural lens having become cloudy and causing vision deterioration. Cataract surgery is the procedure of restoring clear eye vision via cataract removal followed by artificial lens implantation. This surgery is the most common ophthalmic surgery and one of the most frequent surgical procedures worldwide~\cite{JFCS}. Semantic segmentation in cataract surgery videos has several applications ranging from phase and action recognition~\cite{RDCSV, DPS}, irregularity detection (pupillary reaction, lens rotation, lens instability, and lens unfolding delay detection), objective skill assessment, relevance-based compression\cite{RBCCSV}, and so forth~\cite{IoLP, RTIT, RTSB, NoR, MTU}. Accordingly, there exist four different relevant objects in videos from cataract surgery, namely Intraocular Lens, Pupil, Cornea, and Instruments.  The diversity of features of different relevant objects in cataract surgery imposes a challenge on optimal neural network architecture designation. More concretely, a semantic segmentation network is required that can simultaneously deal with (I) deformability and transparency in case of the artificial lens, (II) color and texture variation in case of the pupil, (III) blunt edges in case of the cornea, and (IV) harsh motion blur degradation, reflection distortion, and scale variation in case of the instruments (Figure~\ref{Fig-TMI: Problems}).

\begin{figure}[!tb]
    \centering
    \includegraphics[width=0.6\columnwidth]{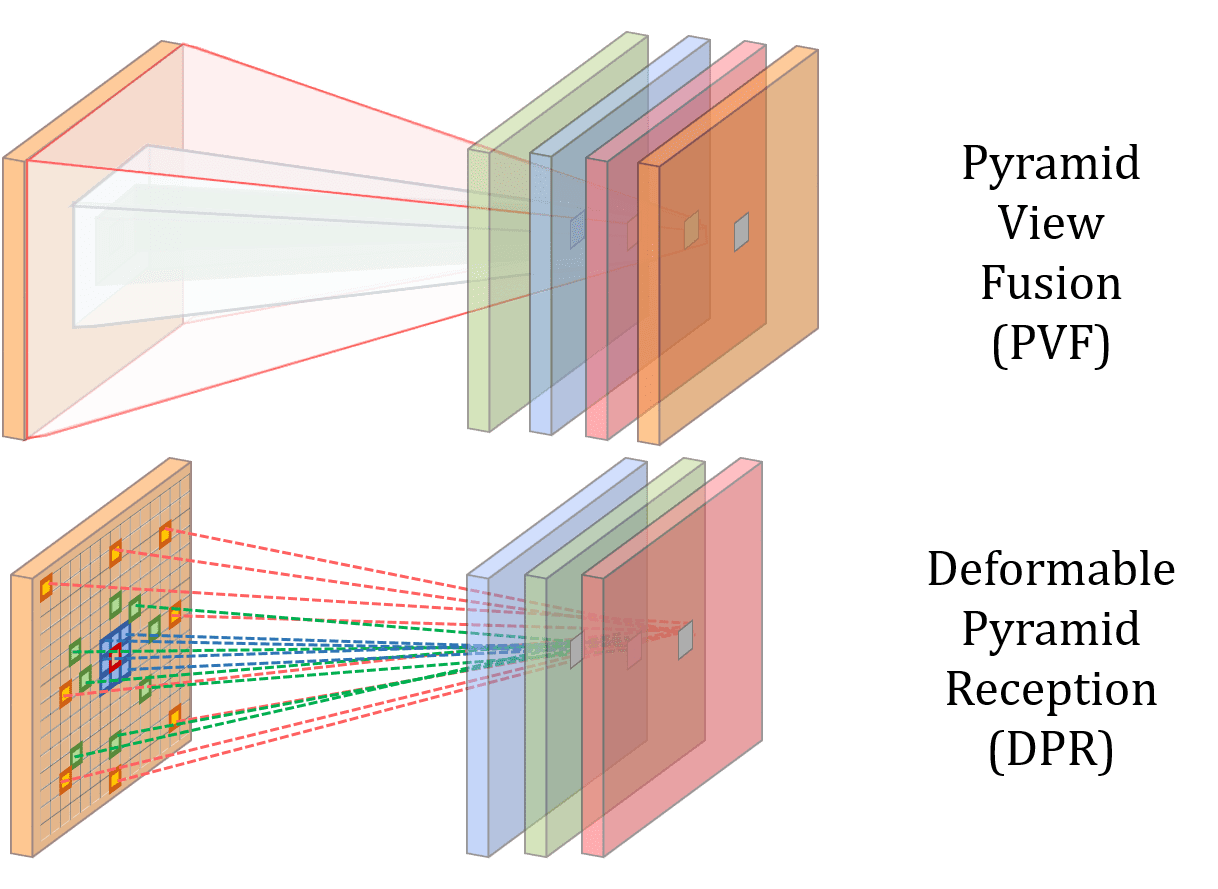}
    \caption{Two major operations in DeepPyram.}
    \label{Fig-TMI: DeepPyram}
\end{figure}
\begin{sidewaysfigure}
    \centering
    \includegraphics[width=1\textwidth]{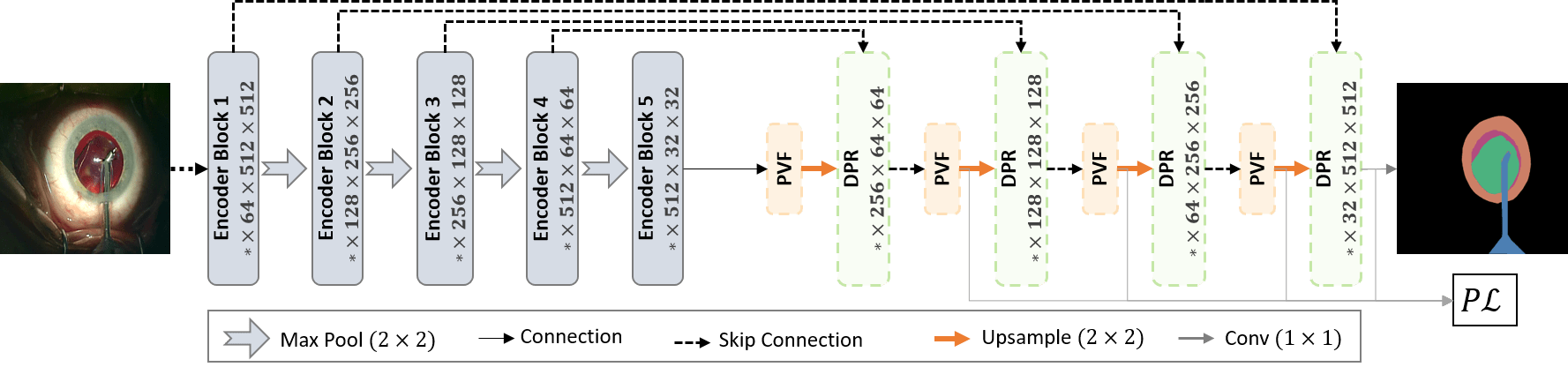}
    \caption{The overall architecture of the proposed DeepPyram network. It contains Pyramid View Fusion (PVF), Deformable Pyramid Reception (DPR), and Pyramid Loss ($P\mathcal{L}$) modules.}
    \label{Fig-TMI: Block_diagram}
\end{sidewaysfigure}

Thischapter presents a U-Net-based CNN for semantic segmentation that can adaptively capture the semantic information in cataract surgery videos.
The proposed network, termed as DeepPyram, mainly consists of three modules: (i) Pyramid View Fusion (PVF) module enabling a varying-angle surrounding view of feature map for each pixel position, (ii) Deformable Pyramid Reception (DPR) module being responsible for performing shape-wise feature extraction on the input convolutional feature map  (Figure~\ref{Fig-TMI: DeepPyram}), and (iii) Pyramid Loss ($P\mathcal{L}$) module that directly supervises the multi-scale semantic feature maps. We have provided a comprehensive study to compare the performance of DeepPyram with thirteen rival state-of-the-art approaches for relevant-instance segmentation in cataract surgery. The experimental results affirm the superiority of DeepPyram, especially in the case of scalable and deformable transparent objects. 

The rest of the paper is organized as follows. In Section~\ref{Sec-TMI: relatedwork}, we position our approach in the literature by reviewing the state-of-the-art semantic segmentation approaches. We delineate the proposed network (DeepPyram) in Section~\ref{sec-TMI: Methodology}. We describe the experimental settings in Section~\ref{sec-TMI: Experimental Settings} and analyze the experimental results in Section~\ref{sec-TMI: Experimental Results}. We also provide an ablation study on DeepPyram in Section~\ref{sec-TMI: Experimental Results} and conclude the paper in Section~\ref{sec-TMI: Conclusion}.

\begin{sidewaysfigure}
    \centering
    \includegraphics[width=0.9\textwidth]{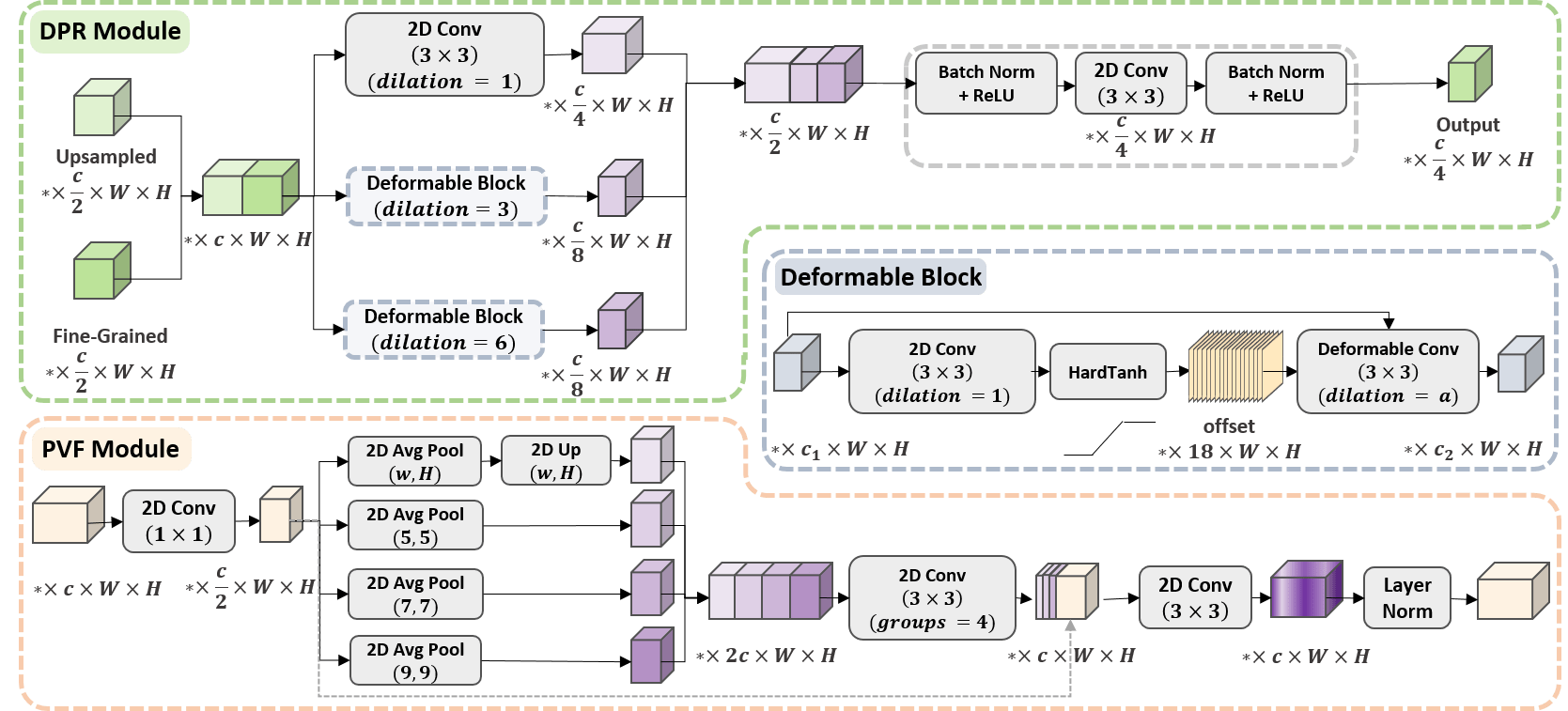}
    \caption{The detailed architecture of the Deformable Pyramid Reception (DPR) and Pyramid View Fusion (PVF) modules.}
    \label{Fig-TMI: DPR}
\end{sidewaysfigure}

\cite{Inf-Net} for deep supervision
\section{Related Work}
\label{Sec-TMI: relatedwork}

This section briefly reviews U-Net-based approaches, state-of-the-art semantic segmentation approaches related to attention and fusion modules, and multi-branch supervision. 

\paragraph{\textit{U-Net-based Networks. }}
U-Net~\cite{U-Net} was initially proposed for medical image segmentation and achieved succeeding performance being attributed to its skip connections. In the encoder side of this encoder-decoder network, low-level features are combined and transformed into semantic information with low resolution. The decoder network improves these low-resolution semantic features and converts them to semantic segmentation results with the same resolution as the input image. The role of skip connections is to transmit the fine-grained low-level feature maps from the encoder to the decoder layers. The coarse-grained semantic feature maps from the decoder and fine-grained low-level feature maps from the encoder are accumulated and undergo convolutional operations. This accumulation technique helps the decoder retrieve the object of interest's high-resolution features to provide delineated semantic segmentation results. Many U-Net-based architectures have been proposed over the past five years to improve the segmentation accuracy and address different flaws and restrictions in the previous architectures~\cite{SegNet, FED-Net, RAUNet, CE-Net, MultiResUNet, dU-Net, PAANet, BARNet, CPFNet, UNet++}. 
In SegNet~\cite{SegNet}, non-linear upsampling in a decoder's layer is performed using the stored max-pooling indices in its symmetrical encoder's layer. The sparse upsampled feature maps are then converted to dense feature maps using convolutional layers. 
In another work, dU-Net~\cite{dU-Net} replaces the classic convolutional operations in U-Net with deformable convolutions~\cite{DeformConv} to deal with shape variations.
Full-Resolution Residual Networks (FRRN) adopt a two-stream architecture, namely the residual and the pooling stream~\cite{FRRN}. The residual stream preserves the full resolution of the input image. The pooling stream employs consecutive full-resolution residual units (FRRU) to add semantic information to the residual stream gradually.  
Inspired by FRNN, Jue~\etal~\cite{MRRC} developed two types of multiple resolution residually connected networks (MRRN) for lung tumor segmentation. The two versions of MRNN, namely incremental-MRNN and dense-MRNN, aim to fuse varying-level semantic feature maps with different resolutions to deal with size variance in tumors.
It is argued that the optimal depth of a U-Net architecture for different datasets is different. UNet++~\cite{UNet++} as an ensemble of varying-depth UNets is proposed to address this depth optimization problem.
Hejie~\etal~\cite{Fused-UNet++} employe UNet++ as their baseline and add dense connections between the blocks with matching-resolution feature maps for pulmonary vessel segmentation.

\paragraph{\textit{Attention Modules. }}
Attention mechanisms can be broadly described as the techniques to guide the network's computational resources (\ie the convolutional operations) towards the most determinative features in the input feature map. Such mechanisms have been especially proven to be gainful in the case of semantic segmentation.  
Inspired by Squeeze-and-Excitation block~\cite{SAE}, the SegSE block~\cite{AFRR} and scSE block~\cite{SCSE} aim to utilize inter-channel dependencies and recalibrate the channels spatially by applying fully connected operations on the globally pooled feature maps. 
RAUNet~\cite{RAUNet} includes an attention module to merge the multi-level feature maps from the encoder and decoder using global average pooling. BARNet~\cite{BARNet} adopts a bilinear-attention module to extract the cross semantic dependencies between the different channels of a convolutional feature map. This module is specially designed to enhance the segmentation accuracy in the case of illumination and scale variation for surgical instruments. 
PAANET~\cite{PAANet} uses a double attention module (DAM) to model semantic dependencies between channels and spatial positions in the convolutional feature map.

\paragraph{\textit{Fusion Modules. }}
Fusion modules can be characterized as modules designed to improve semantic representation via combining several feature maps. The input feature maps could range from varying-level semantic features to the features coming from parallel operations.
PSPNet~\cite{PSPNet} adopts a pyramid pooling module (PPM) containing parallel sub-region average pooling layers followed by upsampling to fuse the multi-scale sub-region representations. This module has shown significant improvement and is frequently used in semantic segmentation architectures~\cite{3D-PPN}. 
Atrous spatial pyramid pooling (ASPP)~\cite{DeepLab, DeepLabv3+} was proposed to deal with objects' scale variance as an efficient alternative to Share-net~\cite{AtS}. Indeed, the ASPP module aggregates multi-scale features extracted using parallel varying-rate dilated convolutions and obviates the need to propagate and aggregate the features of multi-scale inputs. This module is employed in many segmentation approaches due to its effectiveness in capturing multi-scale contextual features~\cite{MSCA, AFPNet, 3D-ASPP}. 

Autofocus Convolutional Layer~\cite{Autofocus} uses a novel approach to fuse resulting feature maps of dilated convolutions adaptively. CPFNet~\cite{CPFNet} uses another fusion approach for scale-aware feature extraction. MultiResUNet~\cite{MultiResUNet} and Dilated MultiResUNet~\cite{DilatedMultiResUNet} factorize the large and computationally expensive receptive fields into a fusion of successive small receptive fields.

\paragraph{\textit{Multi-Branch Supervision. }}
The idea of deep supervision was initially proposed by Chen-Yu~\etal~\cite{DSN}. The authors proved that introducing a classifier (SVM or Softmax) on top of hidden layers can improve the learning process and minimize classification failure. 
This idea is simultaneously adopted by GoogleNet~\cite{GoogleNet} to facilitate gradient flow in deep neural network architectures.
The auxiliary loss in PSPNet~\cite{PSPNet} follows the same strategy and guides one feature map in the encoder network to reinforce learning discriminative features in shallower layers.
The architecture of DensNets~\cite{DCCN} implicitly enables such deep supervision. Multi-branch supervision approaches are frequently used for improving semantic segmentation. Qi~\etal~\cite{3DDSN} suggest directly supervising multi-resolution feature maps of the encoder network by adding deconvolutional layers followed by a Softmax activation layer on top of them. Zhu~\etal~\cite{DSPS} adopt five deep supervision modules for the multi-scale feature maps in the encoder network and three deep supervision modules in the decoder network. In each module, the input feature map is upsampled to its original version and undergoes a deconvolutional layer which extracts semantic segmentation results.
The nested architecture of UNet++~\cite{UNet++} inherently provides such multi-depth semantic feature maps with original resolution. 

\section{Methodology}
\label{sec-TMI: Methodology}

\subsection{Overview}
Figure~\ref{Fig-TMI: Block_diagram} depicts the architecture of the proposed network. Overall, the network consists of a contracting path and an expanding path. The contracting path is responsible for converting low-level to semantic features. The expanding path accounts for performing super-resolution on the coarse-grained semantic feature maps and improving the segmentation accuracy by taking advantage of the symmetric\footnote{Symmetric feature maps are the feature maps with the same spatial resolution.} fine-grained feature maps. The encoder network in the baseline approach is VGG16, pretrained on ImageNet. The decoder network consists of three modules: (i) Pyramid View Fusion (PVF), which induces a large-scale view with progressive angles, (ii) Deformable Pyramid Reception (DPR), which enables a large, sparse, and learnable receptive field, which can sample from up to seven pixels far from each pixel position in the input convolutional feature map, and (iii) Pyramid Loss ($P\mathcal{L}$), which is responsible for the direct supervision of multi-scale semantic feature maps. In the following subsections, we detail each of the proposed modules.

\subsection{Pyramid View Fusion (PVF)} By this module, we aim to stimulate a neural network deduction process analogous to the human visual system. Due to non-uniformly distributed photoreceptors on the retina, the perceived region with high resolution by the human visual system is up to $2\degree-5\degree$ of visual angle centering around the gaze~\cite{Af}. Correspondingly, we infer that the human eye recognizes the semantic information considering not only the internal object's content but also the relative information between the object and the surrounding area. The pyramid view fusion (PVF) module's role is to reinforce the feeling of such relative information at every distinct pixel position. One way to exploit such relative features is to apply convolutional operations with large receptive fields. However, increasing the receptive field's size is not recommended due to imposing huge additional trainable parameters, and consequently (i) escalating the risk of overfitting and (ii) increasing the requirement to more annotations. Alternatively, we use average pooling for fusing the multi-angle local information in our novel attention mechanism. At first, as shown at the bottom of Figure~\ref{Fig-TMI: DPR}, a bottleneck is formed by employing a convolutional layer with a kernel size of one to curb the computational complexity. After this dimensionality reduction stage, the convolutional feature map is fed into four parallel branches. The first branch is a global average pooling layer followed by upsampling. The other three branches include average pooling layers with progressive filter sizes and the stride of one. Using a one-pixel stride is specifically essential for obtaining pixel-wise centralized pyramid view in contrast with region-wise pyramid attention in PSPNet~\cite{PSPNet}. The output feature maps are then concatenated and fed into a convolutional layer with four groups. This layer is responsible for extracting inter-channel dependencies during dimensionality reduction. A regular convolutional layer is then applied to extract joint intra-channel and inter-channel dependencies before being fed into a layer-normalization function.

\paragraph{\textit{\textbf{Discussion. }}}There are three significant differences between the PVF module in DeepPyram and pyramid pooling module in PSPNet~\cite{PSPNet}: (i) All average pooling layers in the PVF module use a stride of one pixel, whereas the average pooling layers in PSPNet adopt different strides of 3, 4, and 6 pixels. The pyramid pooling module separates the input feature map into varying-size sub-regions and plays the role of object detection as region proposal networks (RPNs~\cite{RPN}). However, the stride of one pixel in the PVF module is used to capture the subtle changes in pyramid information that is especially important for segmenting the narrow regions of objects such as instruments and the artificial lens hooks\footnote{Precise segmentation of the artificial lens hooks is crucial for lens rotation and irregularity detection.}. These subtle differences can be diluted after fusing the pyramid information in the pyramid pooling module in PSPNet. (ii) In contrast with PSPNet, the PVF module applies three functions to the concatenated features to capture high-level contextual features: (1) a group convolution function to fuse the features obtained from the average pooling filters independently (as in PSPNet), (2) a convolutional operation being responsible for combining the multi-angle local features to reinforce the semantic features corresponding to each target label, and (3) a layer normalization function that accelerates training and avoids overfitting to the color and contrast information in case of few annotations.  

\subsection{Deformable Pyramid Reception (DPR)}
Figure~\ref{Fig-TMI: DPR} (top) demonstrates the architecture of the deformable pyramid reception (DPR) module. At first, the fine-grained feature map from the encoder and coarse-grained semantic feature map from the previous layer are concatenated. Afterward, these features are fed into three parallel branches: a regular $3\times 3$ convolution and two deformable blocks with different dilation rates. These layers together cover a learnable sparse receptive field of size $15\times 15$\footnote{The structured $3\times 3$ filter covers up to one pixel far from the central pixel). The deformable filter with $dilation=3$ covers an area from two to four pixels far away from each central pixel. The second deformable convolution with $dilation=6$ covers an area from five to seven pixels far away from each central pixel. Therefore, these layers together form a sparse filter of size $15\times 15$ pixels. This sparse kernel can be better seen in Figure~\ref{Fig-TMI: DeepPyram}.}. The output feature maps are then concatenated before undergoing a sequence of regular layers for higher-order feature extraction and dimensionality reduction.

\paragraph{\textit{\textbf{Deformable Block. }}}
Dilated convolutions can implicitly enlarge the receptive field without imposing additional trainable parameters. Dilated convolutional layers can recognize each pixel position's semantic label based on its cross-dependencies with varying-distance surrounding pixels. These layers are exploited in many architectures for semantic segmentation~\cite{MSCA, DeepLab, AFPNet}. Due to the inflexible rectangle shape of the receptive field in regular convolutional layers, however,  the feature extraction procedure cannot be adapted to the target semantic label. Dilated deformable convolutional layers can effectively support the object's geometric transformations in terms of scale and shape. As shown in the \textit{Deformable Block} in Figure~\ref{Fig-TMI: DPR}, a regular convolutional layer is applied to the input feature map to compute the offset field for deformable convolution. The offset field provides two values per element in the convolutional filter (horizontal and vertical offsets). Accordingly, the number of offset field's output channels for a kernel of size $3\times 3$ is equal to 18. Inspired by dU-Net~\cite{dU-Net}, the convolutional layer for the offset field is followed by an activation function. We use the hard tangent hyperbolic function (HardTanh), which is computationally cheap, to clip the offset values in the range of $[-1,1]$. The deformable convolutional layer uses the learned offset values along with the convolutional feature map with a predetermined dilation rate to extract object-adaptive features.

The output feature map ($y$) for each pixel position ($p_0$) and the receptive field ($\mathcal{RF}$) for a regular 2D convolution with a $3\times 3$ filter and dilation rate of one can be defined as:

\begin{equation}
    y(p_o)=\sum_{p_i\in \mathcal{RF}_1}{x(p_0+p_i).w(p_i)}
\end{equation}
\begin{equation}
    \mathcal{RF}_1=\{(-1,-1),(-1,0), ..., (1,0),(1,1)\}
\end{equation}

Where $x$ denotes the input convolutional feature map and $w$ refers to the weights of the convolutional kernel. In a dilated 2D convolution with a dilation rate of $\alpha$, the receptive field can be defined as $\mathcal{RF}_{\alpha}=\alpha \times \mathcal{RF}_1$. Although the sampling locations in a dilated receptive field have a greater distance with the central pixel, they follow a firm structure. In a deformable dilated convolution with a dilation rate of $\alpha$, the sampling locations of the receptive field are dependent to the local contextual features. In the proposed deformable block, the sampling location for the $i$th element of the receptive field and the input pixel $p_0$ can be formulated as:

\begin{equation}
\begin{aligned}
   \mathcal{RF}_{def,\alpha}[i,p_{0}]= \mathcal{RF}_\alpha[i]+f(\sum_{p_j\in \mathcal{RF}_1}{x(p_0+p_j).\hat{w}(p_j)})
\end{aligned}   
\label{eq-TMI: RF3}
\end{equation}

In~\eqref{eq-TMI: RF3}, $f$ denotes the activation function, which is the tangent hyperbolic function in our case, and $\hat{w}$ refers to the weights of the offset filter. This learnable receptive field can be adapted to every distinct pixel in the convolutional feature map and allows the convolutional layer to extract more informative semantic features compared to the regular convolution.
\begin{figure}[!tb]
    \centering
    \includegraphics[width=0.9\columnwidth]{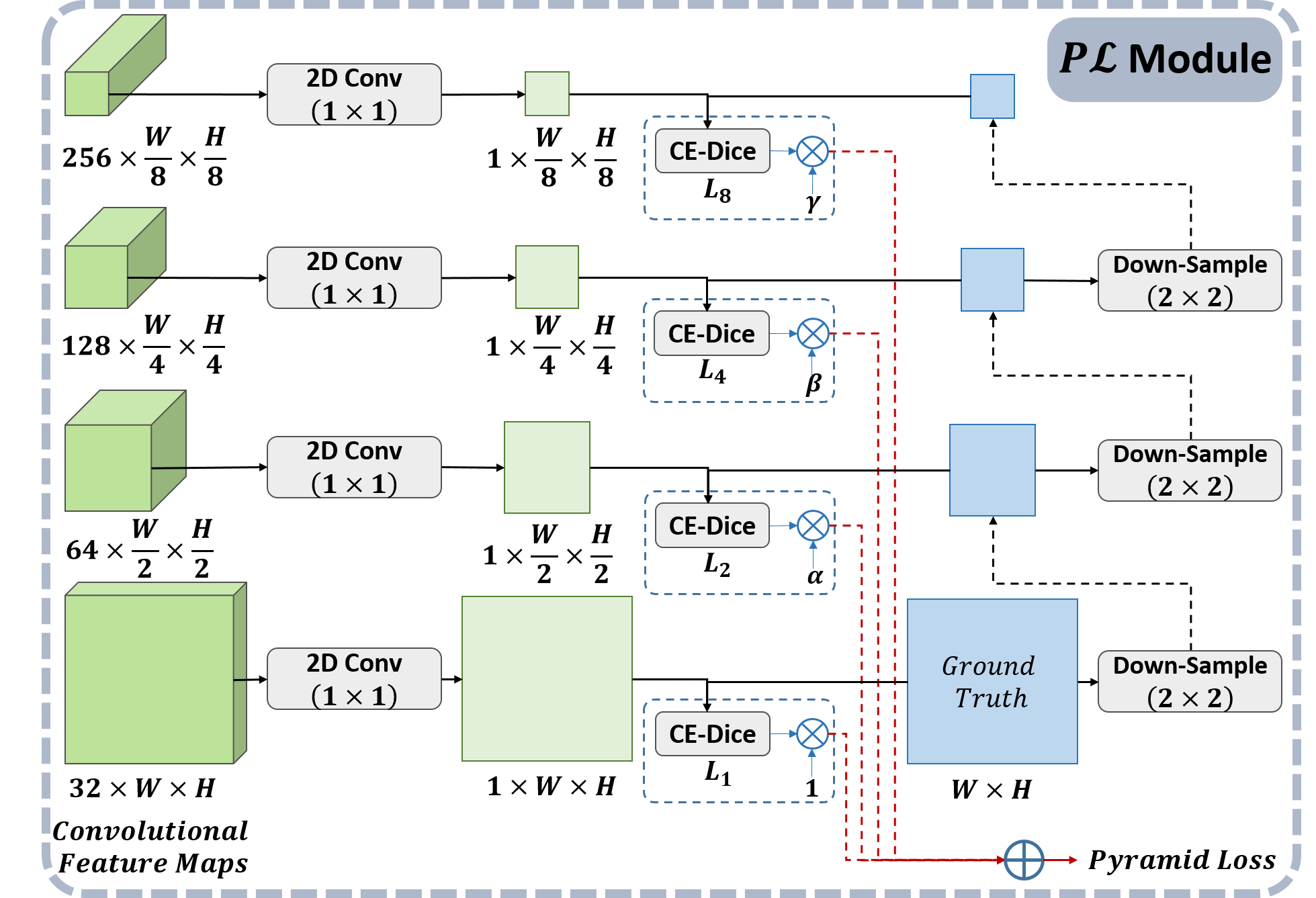}
    \caption{Demonstration of the \textit{Pyramid Loss} module.}
    \label{Fig-TMI:PL}
\end{figure}

\paragraph{\textit{\textbf{Discussion. }}}deformable convolutions are usually used for video object detection, tracking, and segmentation since a deformable layer needs offsets. These offsets are usually provided by optical flow computation, subtracting consecutive frames, or substracting the corresponding feature maps of consecutive frames (as in MaskProp~\cite{MaskProp} and MF-TAPNet~\cite{ITP2019}). Since we use videos recorded in usual and non-laboratory conditions, we have many problems in the video dataset such as defocus blur and harsh motion blur (due to fast movements of eye and motions of the instruments~\cite{DCS}). Using temporal information for semantic segmentation (as in MaskProp~\cite{MaskProp}) may lead to error accumulation and less precise results. Moreover, video object segmentation requires much more annotations which is an additional burden on the expert surgeons' time. For the DPR module, we propose (1) a deformable block to apply deformable convolutions based on learned offsets from static information, (2) a combination of deformable and static filters which can not only capture the edge-sensitive information in the case of sharp edges (as in pupil and lens), but also can cover a large area (up to a size of $15\times 15$ pixels) to deal with color and scale variations, blunt edges in case of cornea, and reflections in case of instruments.

\subsection{Pyramid Loss (P$\mathcal{L}$)}
The role of this module is to directly supervise the multi-scale semantic feature maps on the decoder's side. As shown in Fig~\ref{Fig-TMI:PL}, in order to enable direct supervision, a fully connected layer is formed using a pixel-wise convolution operation. The output feature map presents the semantic segmentation results with the same resolution as the input feature map. To compute the loss for varying-scale outputs, we downscale the ground-truth masks using nearest-neighbor downsampling for multi-class segmentation and max-pooling for binary segmentation. The overall loss is defined as:

\begin{equation}
    P\mathcal{L} = \mathcal{L}_1 + \alpha \mathcal{L}_2 + \beta \mathcal{L}_4 + \gamma \mathcal{L}_8
\end{equation}

Where $\alpha$, $\beta$, and $\gamma$ are predetermined weights in the range of $[0,1]$ (In the experiments, we have set $\alpha = 0.75$, $\beta=0.5$, and $\gamma = 0.25$). Besides, $\mathcal{L}_i$ denotes the loss of output mask segmentation result with the resolution of $(1/i)$ compared to the input resolution.

\paragraph{\textit{\textbf{Discussion. }}}The auxiliary losses for deep semantic segmentation that are proposed to date have two different purposes: (1) Many auxiliary losses aim to guide one or more feature maps in the encoder subnetwork to prevent gradient vanishing due to using a deep architecture (for instance ResNet50) as the backbone. (2) Some auxiliary losses attempt to improve the segmentation accuracy by directly guiding different layers of the encoder or decoder network. In the second case, the loss is computed by performing super-resolution through interpolating the output feature map (to obtain a feature map with the same resolution as the ground truth) and comparing it to the ground truth.  In contrast, the $P\mathcal{L}$ module directly compares each feature map to the downsampled version of the ground truth. This strategy is more time-efficient and computationally less expensive compared to the previously proposed auxiliary losses. Moreover, in contrast with state-of-the-art approaches, each loss branch in the $P\mathcal{L}$ module only consists of $Conv2D + Softmax$ to keep the number of trainable parameters as few as possible.

\begin{sidewaystable}[thpb!]
\renewcommand{\arraystretch}{1}
\caption{Specifications of the proposed and rival approaches. In the ``loss" column, ``CE" and ``CE-Dice" stand for \textit{Cross Entropy} and \textit{Cross Entropy Log Dice}. In ``Upsampling" column, ``Trans Conv" stands for \textit{Transposed Convolution}.}

\label{Tab-TMI:specification}
\centering
\begin{tabular}{lccccccc}
\specialrule{.12em}{.05em}{.05em}
Model & Backbone & Params & Loss & Upsampling & Target & Year & Reference\\\specialrule{.12em}{.05em}{.05em}
UNet$++$~&VGG16&24.24 M& CE-Dice & Bilinear & Medical Images & 2020 & ~\cite{UNet++}\\
UNet++\slash DS &VGG16& 24.24 M & CE-Dice & Bilinear & Medical Images & 2020 & ~\cite{UNet++}\\
CPFNet & ResNet34 &34.66 M& CE-Dice & Bilinear & Medical Images & 2020 & \cite{CPFNet}\\
BARNet&ResNet34&24.90 M& CE-Dice & Bilinear & Surgical
Instruments  & 2020 & ~\cite{BARNet}\\
PAANet &ResNet34& 22.43M & CE-Dice & Trans Conv \& Bilinear & Surgical Instruments & 2020 & \cite{PAANet}\\
dU-Net&\xmark &31.98 M& CE-Dice &Trans Conv & Blood Cells & 2020 & ~\cite{dU-Net}\\
MultiResUNet &\xmark& 9.34 M& CE & Trans Conv & Medical Images & 2020 & ~\cite{MultiResUNet}\\
CE-Net &ResNet34&29.9 M&CE&Trans Conv & Medical Images & 2019 & \cite{CE-Net}\\
RAUNet&ResNet34&22.14 M&CE-Dice&Trans Conv& Cataract  Surgical  Instruments & 2019 &~\cite{RAUNet}\\
FED-Net&ResNet50&59.52 M&CE-Dice& Trans Conv \& PixelShuffle & Liver  Lesion & 2019 &~\cite{FED-Net}\\
PSPNet&ResNet50&22.26 M&CE&Bilinear& Scene & 2017 &~\cite{PSPNet}\\
SegNet&VGG16&14.71 M&CE&Max Unpooling& Scene & 2017 &~\cite{SegNet}\\
U-Net&\xmark &17.26 M&CE& Bilinear & Medical Images & 2015 &~\cite{U-Net}\\\cdashline{1-8}[0.8pt/1pt]
DeepPyram&VGG16 &23.62 M&CE-Dice& Bilinear & Cataract Surgery &\multicolumn{2}{c}{Proposed Approach}\\
\specialrule{.12em}{.05em}{0.05em}
\end{tabular}
\end{sidewaystable}
\section{Experimental Setup}
\label{sec-TMI: Experimental Settings}

\paragraph{\textit{\textbf{Datasets. }}}
We have used four datasets with varying instance features to provide extensive evaluations for the proposed and rival approaches. The two public datasets include the ``Cornea''~\cite{RBCCSV} and ``Instruments''~\cite{CaDIS} mask annotations. Additionally, we have prepared a customized dataset for ``Intraocular Lens'' and ``Pupil'' pixel-wise segmentation\footnote{The dataset will be publicly released with the acceptance of the paper.}. The number of training and testing images for the aforementioned objects are 178:84, 3190:459, 141:48, and 141:48, respectively. All training and testing images are sampled from distinctive videos to meet real-world conditions. Our annotations are performed using the ``supervise.ly'' platform\footnote{https://supervise.ly/} based on the guidelines from cataract surgeons.

\paragraph{\textit{\textbf{Rival Approaches. }}}
Table~\ref{Tab-TMI:specification} details the specifications of the proposed approach and rival approaches employed in our experiments. To provide a fair comparison, we adopt our improved version of PSPNet, featuring a decoder designed similarly to U-Net (with four sequences of double-convolution blocks). Besides, our version of du-Net has the same number of filter-response-maps as for U-Net.

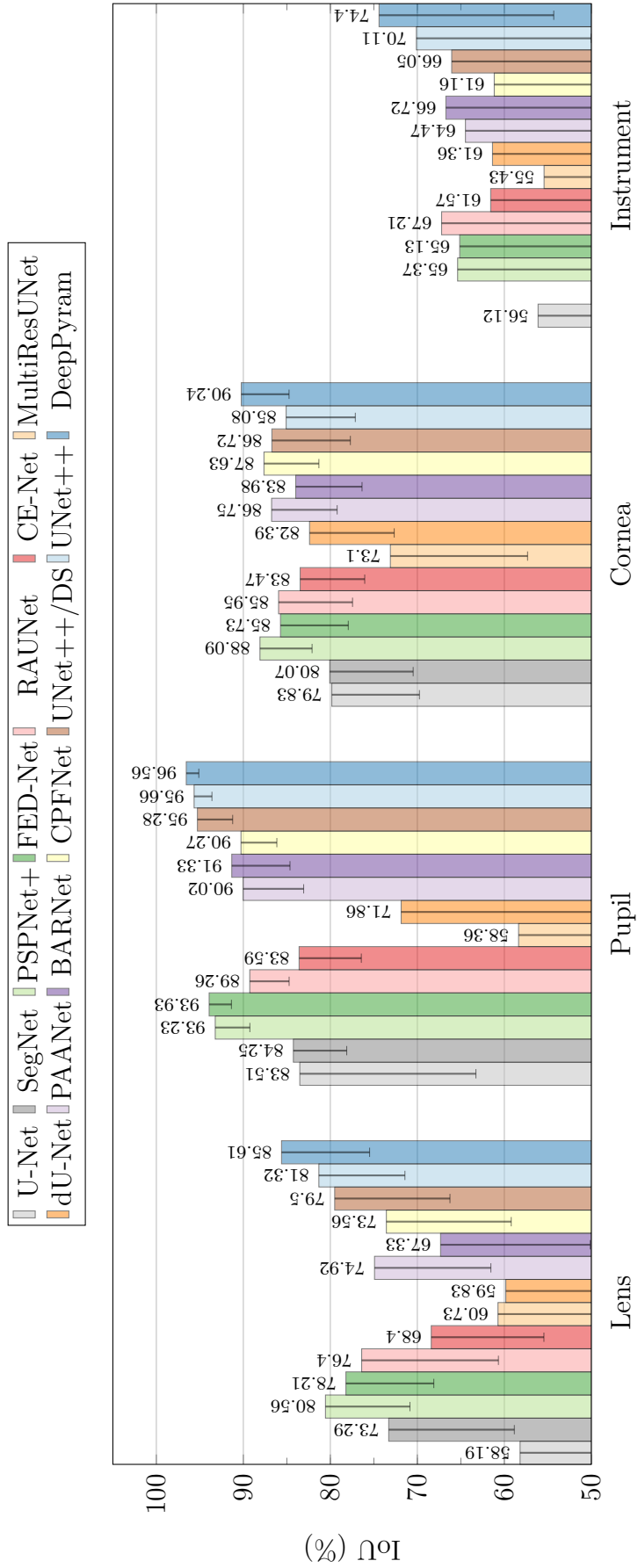
\begin{sidewaysfigure}
\begin{adjustbox}{width=1\textwidth}

\begin{tikzpicture}
    \begin{axis}[
            ybar=0pt,
            bar width=0.35cm,
            width=\textwidth,
            height=.37\textwidth,
            legend style={at={(0.5,1.22
            )},
                anchor=north,legend columns=7},
            symbolic x coords={Lens,Pupil,Cornea, Instrument},
            xtick=data,
            nodes near coords align={vertical},
            ymin=50,ymax=105,
            ylabel= IoU (\%),
            enlarge x limits={abs=7*\pgfplotbarwidth},
            minor ytick={65,70,75,80,85},
            major x tick style = transparent,
            ymajorgrids = true,
            cycle list/Paired,
            nodes near coords,
            every node near coord/.append style={rotate=90, anchor=west, font=\scriptsize, color = black, opacity=1}
        ]
        \addplot+[style={draw=black,fill=lightgray,opacity=0.5}, 
             error bars/.cd, 
             y dir=minus,y explicit]
             coordinates {
                  (Lens,58.19) +- (0, 30.35)
                  (Pupil,83.51) +- (0.0, 20.24)
                  (Cornea,79.83) +- (0.0, 10.07)
                  (Instrument,56.12) +- (0.0, 27.54)};

        \addplot+[style={draw=black,solid,fill=gray,opacity=0.5}, 
             error bars/.cd, 
             y dir=minus,y explicit]
             coordinates {
                  (Lens,73.29) +- (0.0, 14.44)
                  (Pupil,84.25) +- (0.0, 6.13)
                  (Cornea,80.07) +- (0.0, 9.59)
                  (Instrument,0) +- (0.0, 0)};            
        \addplot+[style={draw=black,solid,fill,opacity=0.5}, 
             error bars/.cd, 
             y dir=minus,y explicit]
             coordinates {
                  (Lens,80.56) +- (0.0, 9.71)
                  (Pupil,93.23) +- (0.0, 3.98)
                  (Cornea,88.09) +- (0.0, 5.99)
                  (Instrument,65.37) +- (0.0, 21.39)};         
                  
        \addplot+[style={draw=black,solid,fill,opacity=0.5}, 
             error bars/.cd, 
             y dir=minus,y explicit]
             coordinates {
                  (Lens,78.21) +- (0.0, 10.10)
                  (Pupil,93.93) +- (0.0, 2.56)
                  (Cornea,85.73) +- (0.0, 7.79)
                  (Instrument,65.13) +- (0.0, 22.66)};         
                  
        \addplot+[style={draw=black,solid,fill,opacity=0.5}, 
             error bars/.cd, 
             y dir=minus,y explicit]
             coordinates {
                  (Lens,76.40) +- (0.0, 15.72)
                  (Pupil,89.26) +- (0.0, 4.52)
                  (Cornea,85.95) +- (0.0, 8.50)
                  (Instrument,67.21) +- (0.0, 20.84)};
                  
        \addplot+[style={draw=black,solid,fill,opacity=0.5}, 
             error bars/.cd, 
             y dir=minus,y explicit]
             coordinates {
                  (Lens,68.40) +- (0.0, 12.95)
                  (Pupil,83.59) +- (0.0, 7.15)
                  (Cornea,83.47) +- (0.0, 7.42)
                  (Instrument,61.57) +- (0.0, 16.77)};

        \addplot+[style={draw=black,solid,fill,opacity=0.5}, 
             error bars/.cd, 
             y dir=minus,y explicit]
             coordinates {
                  (Lens,60.73) +- (0.0, 25.84)
                  (Pupil,58.36) +- (0.0, 33.80)
                  (Cornea,73.10) +- (0.0, 15.77)
                  (Instrument,55.43) +- (0.0, 28.44)};

        \addplot+[style={draw=black,solid,fill,opacity=0.5}, 
             error bars/.cd, 
             y dir=minus,y explicit]
             coordinates {
                  (Lens,59.83) +- (0.0, 29.41)
                  (Pupil,71.86) +- (0.0, 28.88)
                  (Cornea,82.39) +- (0.0, 9.73)
                  (Instrument,61.36) +- (0.0, 27.21)};   
                  
        \addplot+[style={draw=black,solid,fill,opacity=0.5}, 
             error bars/.cd, 
             y dir=minus,y explicit]
             coordinates {
                  (Lens,74.92) +- (0.0, 13.37)
                  (Pupil,90.02) +- (0.0, 6.95)
                  (Cornea,86.75) +- (0.0, 7.53)
                  (Instrument,64.47) +- (0.0, 23.19)};

        \addplot+[style={draw=black,solid,fill,opacity=0.5},  
             error bars/.cd, 
             y dir=minus,y explicit]
             coordinates {
                  (Lens,67.33) +- (0.0, 17.20)
                  (Pupil,91.33) +- (0.0, 6.69)
                  (Cornea,83.98) +- (0.0, 7.62)
                  (Instrument,66.72) +- (0.0, 22.91)};   
                  
        \addplot+[style={draw=black,solid,fill,opacity=0.5}, 
             error bars/.cd, 
             y dir=minus,y explicit]
             coordinates {
                  (Lens,73.56) +- (0.0, 14.35)
                  (Pupil,90.27) +- (0.0, 4.12)
                  (Cornea,87.63) +- (0.0, 6.31)
                  (Instrument,61.16) +- (0.0, 20.12)};          
                  
        \addplot+[style={draw=black,solid,fill,opacity=0.5},  
             error bars/.cd, 
             y dir=minus,y explicit]
             coordinates {
                  (Lens,79.50) +- (0.0, 13.25)
                  (Pupil,95.28) +- (0.0, 4.05)
                  (Cornea,86.72) +- (0.0, 9.02)
                  (Instrument,66.05) +- (0.0, 24.91)};          
                  
        \addplot+[style={draw=black,solid,fill,opacity=0.5},  
             error bars/.cd, 
             y dir=minus,y explicit]
             coordinates {
                  (Lens,81.32) +- (0.0, 9.89)
                  (Pupil,95.66) +- (0.0, 2.05)
                  (Cornea,85.08) +- (0.0,7.95)
                  (Instrument,70.11) +- (0.0, 20.07)};            
                  
        \addplot+[style={draw=black,solid,fill,opacity=0.5}, 
             error bars/.cd, 
             y dir=minus,y explicit]
             coordinates {
                  (Lens,85.61) +- (0.0, 10.12)
                  (Pupil,96.56) +- (0.0, 1.44)
                  (Cornea,90.24) +- (0.0, 5.49)
                  (Instrument,74.40) +- (0.0, 20.08)}; 
                  
        \legend{U-Net, SegNet, PSPNet+, FED-Net, RAUNet, CE-Net, MultiResUNet, dU-Net, PAANet, BARNet, CPFNet, UNet++\slash DS, UNet++, DeepPyram}
    \end{axis}
\end{tikzpicture}

\end{adjustbox}
\caption{Quantitative comparisons among DeepPyram and rival approaches based on average and standard deviation of IoU.}
\label{Fig-TMI:IoU}
\end{sidewaysfigure}

\paragraph{\textit{\textbf{Data Augmentation Methods. }}}
Data augmentation is a vital step during training, which prevents network overfitting and boosts the network performance in the case of unseen data. Accordingly, the training images for all evaluations undergo various augmentation approaches before being fed into the network.  We chose the transformations considering the inherent and statistical features of datasets. For instance, we use motion blur transformation to encourage the network to deal with harsh motion blur regularly occurring in cataract surgery videos~\cite{DCS}.  Table~\ref{Tab-TMI:aug} details the adopted augmentation pipeline.

\begin{table}[th!]
\renewcommand{\arraystretch}{1}
\caption{Augmentation Pipeline.}
\label{Tab-TMI:aug}
\centering
\begin{tabu}{lcc}
\specialrule{.12em}{.05em}{.05em}
Augmentation Method&Property&Value\\\specialrule{.12em}{.05em}{.05em}
Brightness~\& Contrast&Factor Range&(-0.2,0.2)\\
Shift~\& Scale&Percentage&10\%\\
Rotate&Degree Range&[-10,10]\\
Motion Blur&Kernel-Size Range &(3,7)\\
\specialrule{.12em}{.05em}{0.05em}
\end{tabu}
\end{table}

\paragraph{\textit{\textbf{Neural Network Settings. }}}
As listed in Table~\ref{Tab-TMI:specification}, U-Net, MultiResUNet, and dU-Net do not adopt a pretrained backbone. For the other approaches, the weights of the backbone are initialized with ImageNet~\cite{ImageNet} training weights.
The input size of all models is set to $3\times 512\times 512$. In the evaluation and testing stages of UNet++/DS and DeepPyram, we disregard the additional output branches (branches related to auxiliary loss functions) and only consider the master output branch.

\begin{sidewaystable}[thpb!]
\renewcommand{\arraystretch}{1}
\caption{Impact of different modules on the segmentation results (IoU\% and Dice\%) of DeepPyram.}
\label{Tab-TMI:ablation2}
\centering

\begin{tabular}{m{0.5cm} m{0.5cm} m{0.5cm} ccccccccccc}
\specialrule{.12em}{.05em}{.05em}
\multicolumn{3}{c}{Modules}&&\multicolumn{2}{c}{Lens}&\multicolumn{2}{c}{Pupil}&\multicolumn{2}{c}{Cornea}&\multicolumn{2}{c}{Instrument}&\multicolumn{2}{c}{Overall}\\\cmidrule(lr){1-3}\cmidrule(lr){5-6}\cmidrule(lr){7-8}\cmidrule(lr){9-10}\cmidrule(lr){11-12}\cmidrule(lr){13-14}
PVF&DPR&$P\mathcal{L}$&Params&IoU(\%)&Dice(\%)&IoU(\%)&Dice(\%)&IoU(\%)&Dice(\%)&IoU(\%)&Dice(\%)&IoU(\%)&Dice(\%)\\\specialrule{.12em}{.05em}{.05em}
\xmark&\xmark&\xmark&22.55 M&82.98&90.44&95.13&97.48&86.02&92.28&69.82&79.05&83.49&89.81\\
\checkmark&\xmark&\xmark&22.99 M&83.73&90.79&96.04&97.95&88.43&93.77&72.58&81.84&85.19&91.09\\
\xmark&\checkmark&\xmark&23.17 M&81.85&89.58&95.32&97.59&86.43&92.55&71.57&80.60&83.79&90.08\\
\checkmark&\checkmark&\xmark&23.62 M&83.85&90.89&95.70&97.79&89.36&94.29&72.76&82.00&85.42&91.24\\
\checkmark&\checkmark&\checkmark&23.62 M&\textbf{85.84}&\textbf{91.98}&\textbf{96.56}&\textbf{98.24}&\textbf{90.24}&\textbf{94.77}&\textbf{74.40}&\textbf{83.30}&\textbf{86.76}&\textbf{92.07}\\ \specialrule{.12em}{.05em}{.05em}
\end{tabular}

\label{Tab-TMI:modules}
\end{sidewaystable}

\paragraph{\textit{\textbf{Training Settings. }}}
Due to the different depth and connections of the proposed and rival approaches, all networks are trained with three different initial learning rates ($lr\in\{0.0005,0.0002,0.001\}$), and the results with the highest IoU for each network are listed. The learning rate is scheduled to decrease every two epochs with the factor of $0.8$. In all different evaluations, the networks are trained end-to-end and for 30 epochs. We use a threshold of $0.1$ for gradient clipping during training\footnote{Gradient clipping is used to clip the error derivatives during back-propagation to prevent gradient explosion.}.

\paragraph{\textit{\textbf{Loss Function. }}}
The \textit{cross entropy log dice} loss, which is used during training, is a weighted sum of binary cross-entropy ($BCE$) and the logarithm of soft Dice coefficient as follows:

\begin{equation}
\begin{aligned}
    \mathcal{L} = &(\lambda)\times BCE(\mathcal{X}_{true}(i,j),\mathcal{X}_{pred}(i,j))\\
    &-(1-\lambda)\times (\log \frac{2\sum \mathcal{X}_{true}\odot \mathcal{X}_{pred}+\sigma}{\sum \mathcal{X}_{true} + \sum \mathcal{X}_{pred}+ \sigma})
\end{aligned}
\label{eq: loss}
\end{equation}

Where $\mathcal{X}_{true}$ denote the ground truth binary mask, and $\mathcal{X}_{pred}$ denote the predicted mask ($0\leq \mathcal{X}_{pred}(i,j) \leq 1$). The parameter $\lambda \in [0,1]$ is set to $0.8$ in our experiments, and $\odot$ refers to Hadamard product (element-wise multiplication). Besides, the parameter $s$ is the Laplacian smoothing factor which is added to (i) prevent division by zero, and (ii) avoid overfitting (in experiments, $\sigma = 1$).
\paragraph{\textit{\textbf{Evaluation Metrics. }}}
The Jaccard metric (Intersection-over-Union -- IoU) and the Dice Coefficient (F1-score) are regarded as the major semantic segmentation indicators. Accordingly, we evaluate the proposed and rival approaches using average IoU and dice. In order to enable a broader analysis of the networks' performance, the standard deviation of IoU, and minimum and maximum of dice coefficient over all of the testing images are additionally compared.

\paragraph{\textit{\textbf{Ablation Study Settings. }}}
To evaluate the effectiveness of different modules, we have implemented another segmentation network, excluding all the proposed modules. This network has the same backbone as our baseline (VGG16). However, the PVF module is completely removed from the network. Besides, the DPR module is replaced with a sequence of two convolutional layers, each of which is followed by a batch normalization layer and a ReLU layer. The connections between the encoder and decoder remain the same as DeepPyram. Besides, DeepPyram++ in ablation studies is the nested version of DeepPyram implemented based on UNet++.

\section{Experimental Results}
\label{sec-TMI: Experimental Results}

\begin{figure*}[ht]
\begin{tabular}{cc}
\begin{subfigure}{.48\textwidth}\vspace{-0.5\baselineskip}
  \centering
\pgfplotstableread{
x y y-min y-max
{U-Net} 67.91 0 94.76
{SegNet} 83.67 19.15 92.47
{PSPNet+}  88.89 67.67 96.89
{FEDNet}  87.38 61.99 95.85
{RAUNet} 85.34 0 94.79
{CE-Net} 80.43 29.83 93.04
{MultiResUNet} 71.62 4.94 94.93
{dU-Net} 69.46 0 95.83
{PAANet} 84.83 15.64 94.77
{BARNet} 78.85 0 92.82
{CPFNet} 83.74 8.91 95.20
{UNet++\slash DS} 87.85 53.36 96.25
{UNet++}  89.34 64.85 96.87
{DeepPyram}  91.98 55.56 98.07

}{\differanser}
\begin{adjustbox}{height=0.2\textheight}
\begin{tikzpicture}[scale=1.3] 
\begin{axis} [
width  = 0.85\textwidth,
height = 8cm,
symbolic x coords={{U-Net},{SegNet},{PSPNet+},{FEDNet},{RAUNet},{CE-Net},{MultiResUNet},{dU-Net}, {PAANet},{BARNet},{CPFNet},{UNet++\slash DS},{UNet++},{DeepPyram}},
minor ytick={5,10,15,20,25},
yminorgrids,
xtick=data,
ticklabel style = {font=\tiny},
x tick label style={rotate=45,anchor=east},
legend style={at={(0.5,1.25)},anchor=north,legend columns=-1},
ymin=70,ymax=100,
height=.5\textwidth,
ylabel= Dice (\%),
label style={font=\tiny},
tick label style={font=\tiny},
extra y ticks=91.98,
extra y tick labels={},
extra y tick style={
ymajorgrids=true,
ytick style={/pgfplots/major tick length=0pt,
},
grid style={BurntOrange,dashed,/pgfplots/on layer=axis foreground,},
},
]

\addplot+[ForestGreen, very thick, forget plot,only marks,forget plot] 
plot[very thick, error bars/.cd, y dir=plus, y explicit]
table[x=x,y=y,y error expr=\thisrow{y-max}-\thisrow{y}] {\differanser};

\addplot+[red, very thick, only marks,xticklabels=\empty,forget plot] 
plot[very thick, error bars/.cd, y dir=minus, y explicit]
table[x=x,y=y,y error expr=\thisrow{y}-\thisrow{y-min}] {\differanser};

\addplot[only marks,mark=*,mark options={fill=BurntOrange,draw=BurntOrange,very thick}] 
table[x=x,y expr=\thisrow{y}] {\differanser};

\addplot[only marks,mark=square*,color=LimeGreen, mark options={scale=0.8}] 
table[x=x,y expr=\thisrow{y-max}] {\differanser};

\addplot[only marks,mark=square*,color=BrickRed, mark options={scale=0.8}] 
table[x=x,y expr=\thisrow{y-min}] {\differanser};

\end{axis} 
\end{tikzpicture}
\end{adjustbox}
\vspace{-1\baselineskip}
\caption{Lens}
\label{fig:sub-first}
\end{subfigure} &

\begin{subfigure}{.48\textwidth}\vspace{-0.5\baselineskip}
  \centering
\pgfplotstableread{
x y y-min y-max
{U-Net} 89.36 37.62 99.22
{SegNet} 91.31 66.04 95.60
{PSPNet+}  96.45 86.98 98.74
{FEDNet}  96.85 92.42 98.90
{RAUNet} 94.26 83.91 97.34
{CE-Net} 90.89 81.41 97.51
{MultiResUNet} 66.80 2.64 98.68
{dU-Net} 79.53 10.38 98.15
{PAANet} 94.59 73.79 98.31
{BARNet} 95.32 73.75 98.60
{CPFNet} 94.83 87.88 98.54
{UNet++\slash DS} 97.53 87.07 99.32
{UNet++}  97.77 95.08 99.22
{DeepPyram}  98.24 95.28 99.20
}{\differanser}
\begin{adjustbox}{height=0.2\textheight}
\begin{tikzpicture}[scale=1.3] 
\begin{axis} [
width  = 0.85\textwidth,
height = 8cm,
symbolic x coords={{U-Net},{SegNet},{PSPNet+},{FEDNet},{RAUNet},{CE-Net},{MultiResUNet},{dU-Net}, {PAANet},{BARNet},{CPFNet},{UNet++\slash DS},{UNet++},{DeepPyram}},
minor ytick={5,10,15,20,25},
yminorgrids,
xtick=data,
ticklabel style = {font=\tiny},
x tick label style={rotate=45,anchor=east},
legend style={at={(0.5,1.25)},anchor=north,legend columns=-1},
ymin=70,ymax=100,
height=.5\textwidth,
ylabel= Dice (\%),
label style={font=\tiny},
tick label style={font=\tiny},
extra y ticks=98.24,
extra y tick labels={},
extra y tick style={
ymajorgrids=true,
ytick style={/pgfplots/major tick length=0pt,
},
grid style={BurntOrange,dashed,/pgfplots/on layer=axis foreground,},
},
]

\addplot+[ForestGreen, very thick, forget plot,only marks,forget plot] 
plot[very thick, error bars/.cd, y dir=plus, y explicit]
table[x=x,y=y,y error expr=\thisrow{y-max}-\thisrow{y}] {\differanser};

\addplot+[red, very thick, only marks,xticklabels=\empty,forget plot] 
plot[very thick, error bars/.cd, y dir=minus, y explicit]
table[x=x,y=y,y error expr=\thisrow{y}-\thisrow{y-min}] {\differanser};

\addplot[only marks,mark=*,mark options={fill=BurntOrange,draw=BurntOrange,very thick}] 
table[x=x,y expr=\thisrow{y}] {\differanser};

\addplot[only marks,mark=square*,color=LimeGreen, mark options={scale=0.8}] 
table[x=x,y expr=\thisrow{y-max}] {\differanser};

\addplot[only marks,mark=square*,color=BrickRed, mark options={scale=0.8}] 
table[x=x,y expr=\thisrow{y-min}] {\differanser};

\end{axis} 
\end{tikzpicture}
\end{adjustbox}
\vspace{-1\baselineskip}
\caption{Pupil}
\label{fig:sub-first}
\end{subfigure}\\
\begin{subfigure}{0.48\textwidth}\vspace{-0.5\baselineskip}
  \centering
\pgfplotstableread{

x y y-min y-max
{U-Net} 86.20 59.59 96.78
{SegNet} 88.60 70.14 96.10
{PSPNet+}  93.55 81.90 97.86
{FEDNet}  92.10 63.48 97.30
{RAUNet} 92.18 53.21 97.98
{CE-Net} 90.85 77.39 96.97
{MultiResUNet} 83.40 39.51 96.38
{dU-Net} 90.00 65.38 97.36
{PAANet} 92.71 71.63 98.01
{BARNet} 91.09 73.17 97.93
{CPFNet} 93.28 76.78 97.67
{UNet++\slash DS} 92.57 45.19 97.50
{UNet++}  91.72 70.31 97.89
{DeepPyram}  94.63 81.37 98.40

}{\differanser}
\begin{adjustbox}{height=0.2\textheight}
\begin{tikzpicture}[scale=1.3] 
\begin{axis} [
width  = 0.85\textwidth,
height = 8cm,
symbolic x coords={{U-Net},{SegNet},{PSPNet+},{FEDNet},{RAUNet},{CE-Net},{MultiResUNet},{dU-Net}, {PAANet},{BARNet},{CPFNet},{UNet++\slash DS},{UNet++},{DeepPyram}},
minor ytick={5,10,15,20,25},
yminorgrids,
xtick=data,
ticklabel style = {font=\tiny},
x tick label style={rotate=45,anchor=east},
legend style={at={(0.5,1.25)},anchor=north,legend columns=-1},
ymin=70,ymax=100,
height=.5\textwidth,
ylabel= Dice (\%),
label style={font=\tiny},
tick label style={font=\tiny},
extra y ticks=94.63,
extra y tick labels={},
extra y tick style={
ymajorgrids=true,
ytick style={/pgfplots/major tick length=0pt,
},
grid style={BurntOrange,dashed,/pgfplots/on layer=axis foreground,},
},
]

\addplot+[ForestGreen, very thick, forget plot,only marks,forget plot] 
plot[very thick, error bars/.cd, y dir=plus, y explicit]
table[x=x,y=y,y error expr=\thisrow{y-max}-\thisrow{y}] {\differanser};

\addplot+[red, very thick, only marks,xticklabels=\empty,forget plot] 
plot[very thick, error bars/.cd, y dir=minus, y explicit]
table[x=x,y=y,y error expr=\thisrow{y}-\thisrow{y-min}] {\differanser};

\addplot[only marks,mark=*,mark options={fill=BurntOrange,draw=BurntOrange,very thick}] 
table[x=x,y expr=\thisrow{y}] {\differanser};

\addplot[only marks,mark=square*,color=LimeGreen, mark options={scale=0.8}] 
table[x=x,y expr=\thisrow{y-max}] {\differanser};

\addplot[only marks,mark=square*,color=BrickRed, mark options={scale=0.8}] 
table[x=x,y expr=\thisrow{y-min}] {\differanser};

\end{axis} 
\end{tikzpicture}
\end{adjustbox}
\vspace{-1\baselineskip}
\caption{Cornea}
\label{fig:sub-first}
\end{subfigure} &
\begin{subfigure}{.48\textwidth}\vspace{-0.5\baselineskip}
  \centering
\pgfplotstableread{
x y y-min y-max
{U-Net} 67.02 0 98.13
{SegNet} 0 0 0
{PSPNet+}  76.47 0 96.30
{FEDNet}  76.11 0 96.87
{RAUNet} 77.99 0 97.13
{CE-Net} 74.64 1.94 94.15
{MultiResUNet} 66.07 0 97.62
{dU-Net} 71.55 0 97.61
{PAANet} 75.24 0 97.88
{BARNet} 77.14 0 97.50
{CPFNet} 73.51 0 94.95
{UNet++\slash DS} 75.91 0 97.78
{UNet++}  79.56 0 97.81
{DeepPyram}  83.30 0 98.19

}{\differanser}
\begin{adjustbox}{height=0.2\textheight}
\begin{tikzpicture}[scale=1.3] 
\begin{axis} [
width  = 0.85\textwidth,
height = 8cm,
symbolic x coords={{U-Net},{SegNet},{PSPNet+},{FEDNet},{RAUNet},{CE-Net},{MultiResUNet},{dU-Net}, {PAANet},{BARNet},{CPFNet},{UNet++\slash DS},{UNet++},{DeepPyram}},
minor ytick={5,10,15,20,25},
yminorgrids,
xtick=data,
ticklabel style = {font=\tiny},
x tick label style={rotate=45,anchor=east},
legend style={at={(0.5,1.25)},anchor=north,legend columns=-1},
ymin=70,ymax=100,
height=.5\textwidth,
ylabel= Dice (\%),
label style={font=\tiny},
tick label style={font=\tiny},
extra y ticks=83.30,
extra y tick labels={},
extra y tick style={
ymajorgrids=true,
ytick style={/pgfplots/major tick length=0pt,
},
grid style={BurntOrange,dashed,/pgfplots/on layer=axis foreground,},
},
]

\addplot+[ForestGreen, very thick, forget plot,only marks,forget plot] 
plot[very thick, error bars/.cd, y dir=plus, y explicit]
table[x=x,y=y,y error expr=\thisrow{y-max}-\thisrow{y}] {\differanser};

\addplot+[red, very thick, only marks,xticklabels=\empty,forget plot] 
plot[very thick, error bars/.cd, y dir=minus, y explicit]
table[x=x,y=y,y error expr=\thisrow{y}-\thisrow{y-min}] {\differanser};

\addplot[only marks,mark=*,mark options={fill=BurntOrange,draw=BurntOrange,very thick}] 
table[x=x,y expr=\thisrow{y}] {\differanser};

\addplot[only marks,mark=square*,color=LimeGreen, mark options={scale=0.8}] 
table[x=x,y expr=\thisrow{y-max}] {\differanser};

\addplot[only marks,mark=square*,color=BrickRed, mark options={scale=0.8}] 
table[x=x,y expr=\thisrow{y-min}] {\differanser};

\end{axis} 

\end{tikzpicture}
\end{adjustbox}
\vspace{-1\baselineskip}
\caption{Instruments}
\label{fig:sub-first}
\end{subfigure}

\end{tabular}
\caption{Quantitative comparison of segmentation results for the proposed (DeepPyram) and rival architectures (some minimum and average values are not visible due to y-axis clipping).}
\label{Fig-TMI:dice}
\end{figure*}
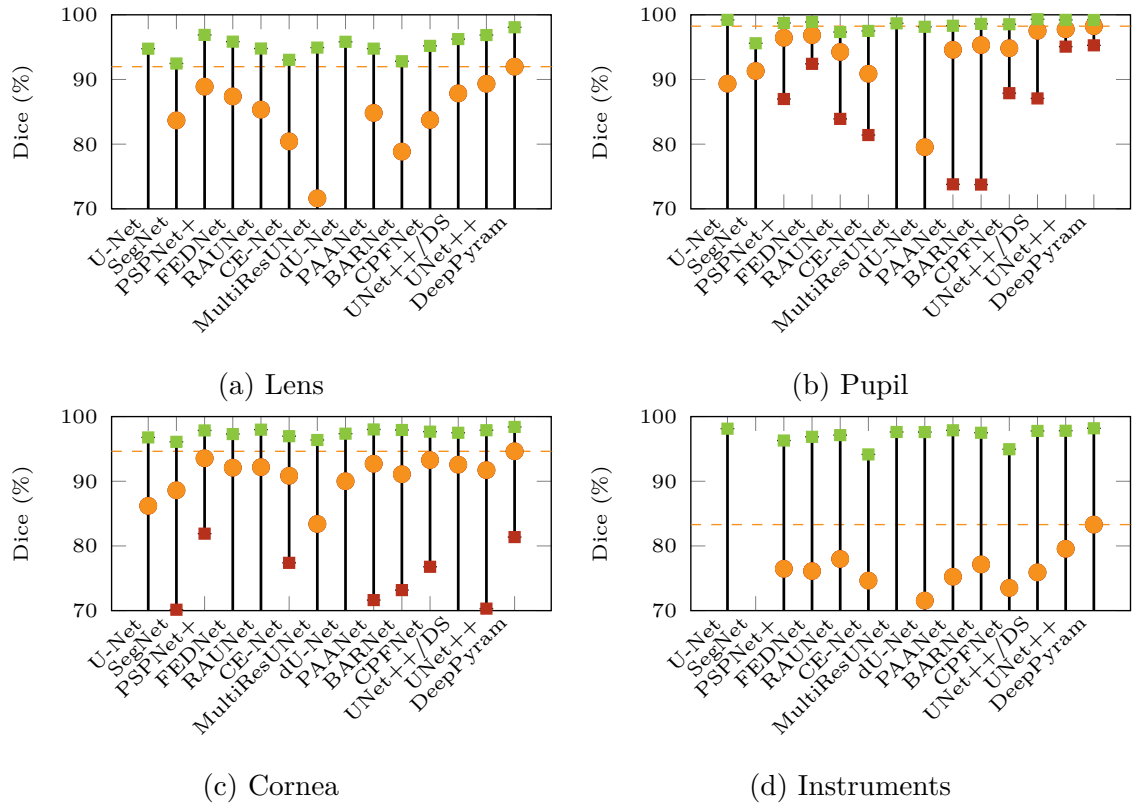
\subsection{Relevant Object Segmentation}
Figure~\ref{Fig-TMI:IoU} compares the resulting IoU of DeepPyram and thirteen rival approaches\footnote{SegNet did not converge during training for instrument segmentation with different initial learning rates.}. Overall, the rival approaches have shown a different level of performance for each of the four relevant objects.
Based on the mean IoU, the best four segmentation approaches for each of the four relevant objects are listed in descending order below:
\begin{itemize}
    \item \textit{Lens:} DeepPyram, Unet++, PSPNet+, UNet++/DS
    \item \textit{Pupil:} DeepPyram, Unet++, UNet++/DS, FEDNet
    \item \textit{Cornea:} DeepPyram, PSPNet+, CPFNet, UNet++/DS
    \item \textit{Instrument:} DeepPyram, Unet++, RAUNet, BARNet
\end{itemize}
Accordingly, DeepPyram, Unet++, and PSPNet+ contribute to the top three segmentation results for the relevant objects in cataract surgery videos. However, DeepPyram shows considerable improvement in segmentation accuracy compared to the second-best approach in each class. Specifically, DeepPyram has achieved more than $4\%$ improvement in lens segmentation ($85.61\%$ vs. $81.32\%$) and more than $4\%$ improvement in instrument segmentation ($74.40\%$ vs. $70.11\%$) compared to UNet++ as the second-best approach. Moreover, DeepPyram appears to be the most reliable approach considering the smallest standard deviation compared to the rival approaches. This significant improvement is attributed to the PVF, DPR, and $P\mathcal{L}$ modules. 

As shown in Fig~\ref{Fig-TMI:dice}, DeepPyram has achieved the highest dice coefficient compared to the rival approaches for the lens, pupil, cornea, and instrument segmentation. Moreover, DeepPyram is the most reliable segmentation approach based on achieving the highest minimum dice percentage.

\begin{figure}[!tb]
    \centering
    \includegraphics[width=0.53\textwidth]{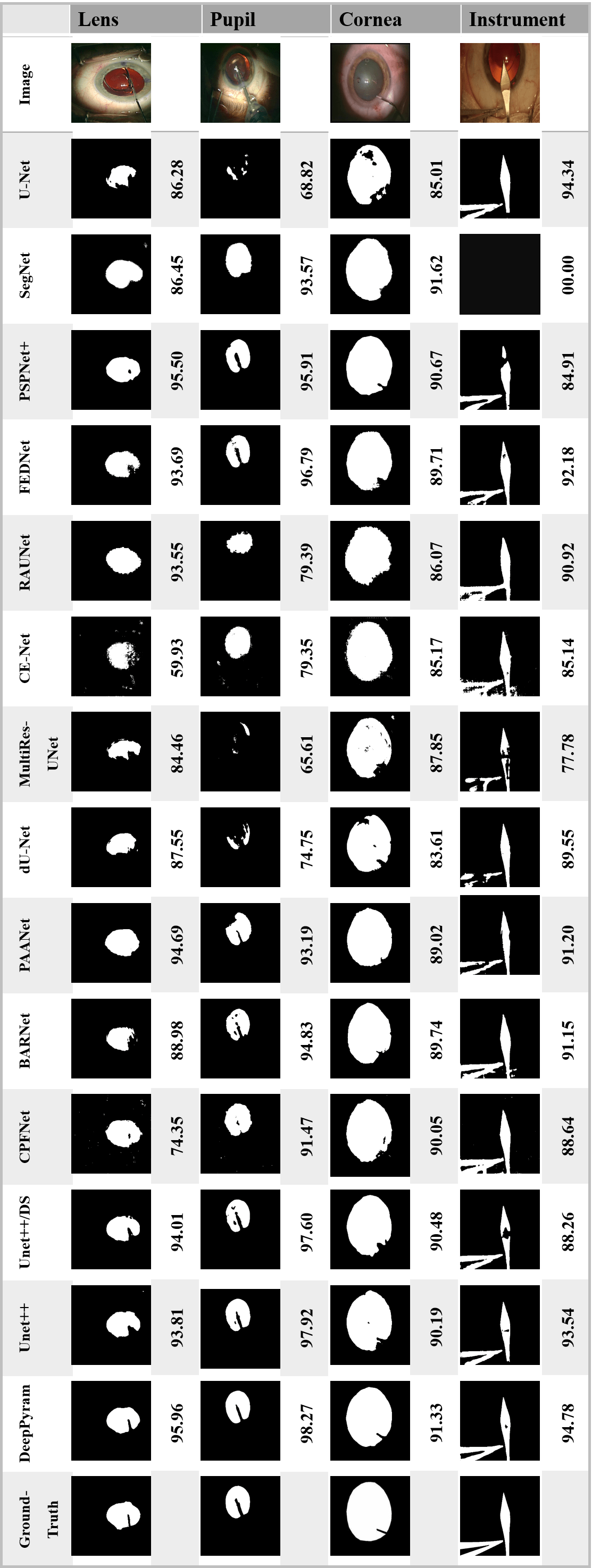}
    \caption{Qualitative comparisons among DeepPyram and the rival approaches for the relevant objects in cataract surgery videos (the numbers denote the Dice(\%) coefficient for each detection).}
    \label{Fig-TMI:Vis}
\end{figure}

Figure~\ref{Fig-TMI:Vis} further affirms the effectiveness of DeepPyram in enhancing the segmentation results. Taking advantage of the pyramid view provided by the PVF module, DeepPyram can handle reflection and brightness variation in instruments, blunt edges in the cornea, color and texture variation in the pupil, as well as transparency in the lens. Furthermore, powering by deformable pyramid reception, DeepPyram can tackle scale variations in instruments and blunt edges in the cornea. In particular, we can perceive from Figure~\ref{Fig-TMI:Vis} that DeepPyram shows much less distortion in the region of edges (especially in the case of the cornea. Furthermore, based on these qualitative experiments, DeepPyram shows much better precision and recall in the narrow regions for segmenting the instruments and other relevant objects in the case of occlusion by the instruments.

\subsection{Ablation Study}

Table~\ref{Tab-TMI:modules} validates the effectiveness of the proposed modules in segmentation enhancement. The PVF module can notably enhance the performance for cornea and instrument segmentation ($2.41\%$ and $2.76\%$ improvement in IoU, respectively). This improvement is due to the ability of the PVF module to provide a global view of varying-size sub-regions centering around each distinctive spatial position. Such a global view can reinforce semantic representation in the regions corresponding to blunt edges and reflections. Due to scale variance in instruments, the DPR module can effectively boost the segmentation performance for instruments. The addition of $P\mathcal{L}$ module results in the improvement of IoU for all relevant segments, especially lens segmentation (around $2\%$ improvement) and Instrument ($1.64\%$ improvement). The combination of PVF, DPR, and $P\mathcal{L}$ modules can contribute to $4.58\%$ improvement in instrument segmentation and $4.22\%$ improvement in cornea segmentation (based on IoU\%). These modules have improved the IoU for the lens and pupil by $2.85\%$ and $1.43\%$, respectively.

Overall\footnote{The ``Overall'' column in Table~\ref{Tab-TMI:modules} is the mean of the other four average values.}, the addition of different modules of DeepPyram has led to considerable improvement of segmentation performance ($3.27\%$ improvement in IoU) on average compared to the baseline approach. The PVF module can provide varying-angle global information centering around each pixel in the convolutional feature map to support relative information access. The DPR module enables large-field content-adaptive reception while minimizing the additive trainable parameters. 

It should be noted that even our baseline approach has much better performance compared to some rival approaches. We argue that the fusion modules adopted in some rival approaches may lead to the dilution of discriminative semantic information.

\begin{figure*}[tb!]
    \centering
    \includegraphics[width=1\textwidth]{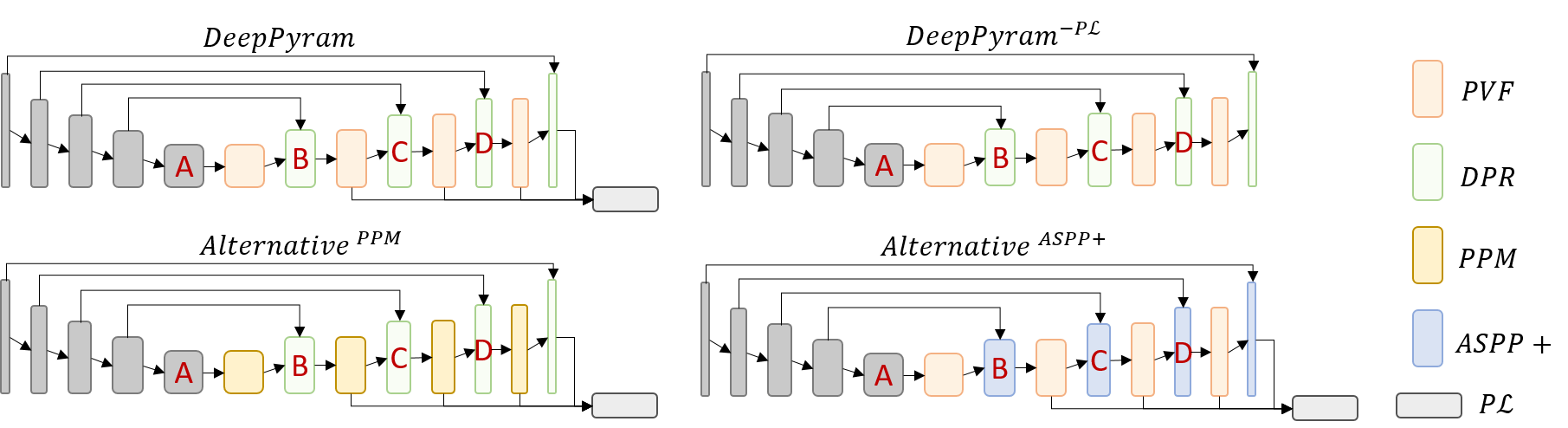}
    \caption{The overall architecture of DeepPyram compared to its three alternatives. The locations A, B, C, and D in each architecture correspond to the four modules for which we visualize the feature representations in Figure~\ref{Fig-TMI: replace_module_vis}.}
    \label{Fig-TMI: replace_module}
\end{figure*}
el{Fig-TMI:IoU}

\begin{sidewaysfigure}
    \centering
    \includegraphics[width=1\textwidth]{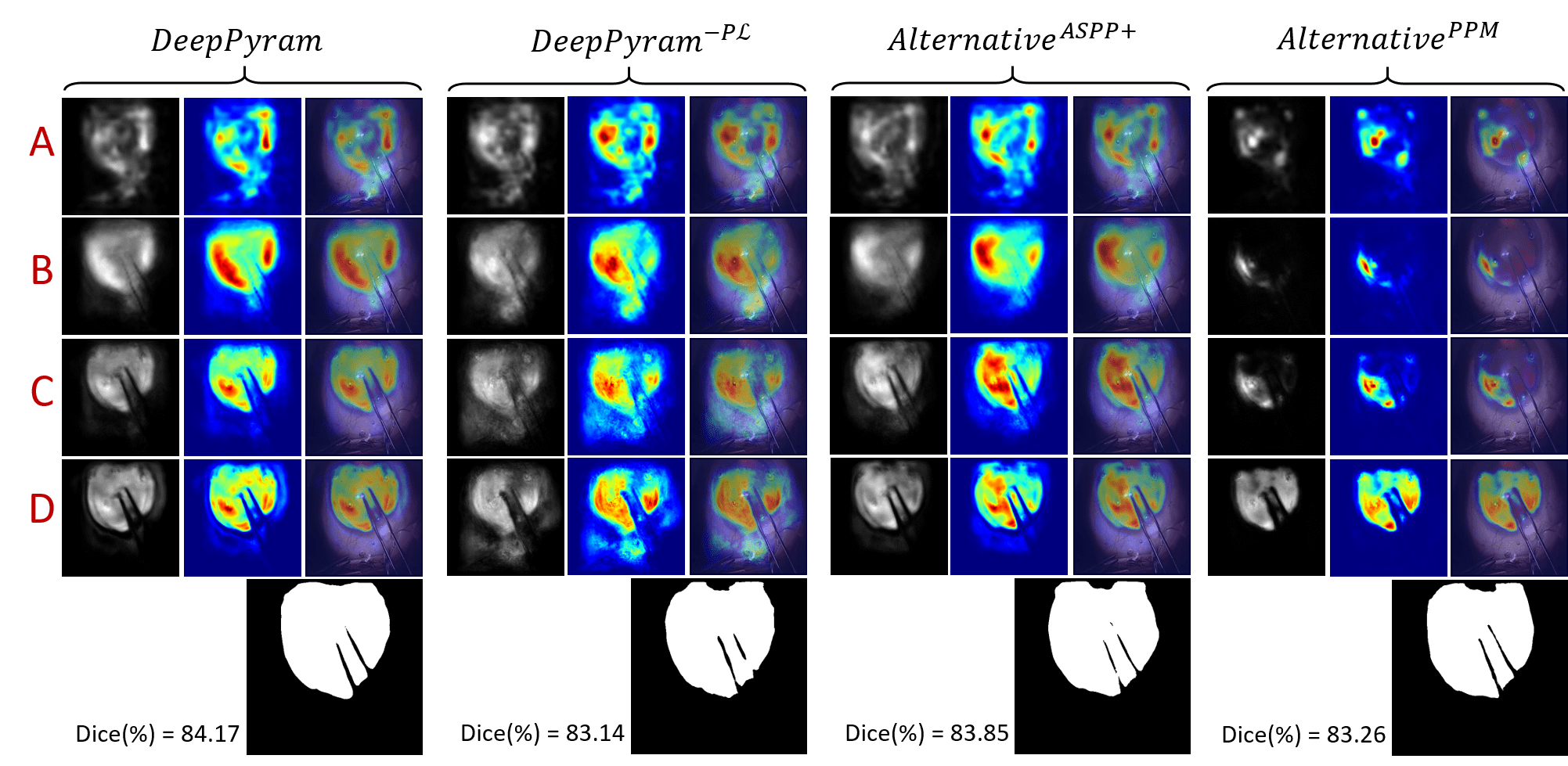}
    \caption{Visualization of the effect of the proposed and alternative modules based on class activation maps~\cite{Score-CAM} using the network architectures demonstrated in Figure~\ref{Fig-TMI: replace_module}. For each approach, the figures from left to right represent the gray-scale activation maps, heatmaps, and heatmaps on images.}
    \label{Fig-TMI: replace_module_vis}
\end{sidewaysfigure} 

\begin{sidewaystable}
\renewcommand{\arraystretch}{1}
\caption{Impact of alternative modules on the segmentation results (IoU\% and Dice\%) of DeepPyram.}
\label{tab:ablation}
\centering
\begin{tabular}{lccccccccccc}
\specialrule{.12em}{.05em}{.05em}%
&&\multicolumn{2}{c}{Lens}&\multicolumn{2}{c}{Pupil}&\multicolumn{2}{c}{Cornea}&\multicolumn{2}{c}{Instrument}&\multicolumn{2}{c}{Overall}\\\cmidrule(lr){3-4}\cmidrule(lr){5-6}\cmidrule(lr){7-8}\cmidrule(lr){9-10}\cmidrule(lr){11-12}
Network&Params&IoU(\%)&Dice(\%)&IoU(\%)&Dice(\%)&IoU(\%)&Dice(\%)&IoU(\%)&Dice(\%)&IoU(\%)&Dice(\%)\\\specialrule{.12em}{.05em}{.05em}%
$Alternative^{ASPP+}$&22.99 M&85.02&91.51&96.41&98.17&88.83&94.00&\textbf{74.81}&\textbf{83.66}&86.26&91.83\\
$Alternative^{PPM}$&23.44 M&83.40&90.66&95.70&97.79&87.44&92.88&72.51&81.03&84.76&90.59\\
DeepPyram&23.62 M&\textbf{85.84}&\textbf{91.98}&\textbf{96.56}&\textbf{98.24}&\textbf{90.24}&\textbf{94.77}&74.40&83.30&\textbf{86.76}&\textbf{92.07}\\\specialrule{.12em}{.05em}{.05em}%
\end{tabular}
\label{tab:alternative}
\vspace{8mm}
\renewcommand{\arraystretch}{1}
\caption{Impact of different backbones and combinations on the segmentation results (IoU\% and Dice\%) of DeepPyram.}
\label{tab:ablation}
\centering
\begin{tabular}{llccccccccccc}
\specialrule{.12em}{.05em}{.05em}%
&&&\multicolumn{2}{c}{Lens}&\multicolumn{2}{c}{Pupil}&\multicolumn{2}{c}{Cornea}&\multicolumn{2}{c}{Instrument}&\multicolumn{2}{c}{Overall}\\\cmidrule(lr){4-5}\cmidrule(lr){6-7}\cmidrule(lr){8-9}\cmidrule(lr){10-11}\cmidrule(lr){12-13}
&&Params&IoU(\%)&Dice(\%)&IoU(\%)&Dice(\%)&IoU(\%)&Dice(\%)&IoU(\%)&Dice(\%)&IoU(\%)&Dice(\%)\\\specialrule{.12em}{.05em}{.05em}%
\multicolumn{1}{ c|  }{\parbox[t]{1mm}{\multirow{4}{*}{\rotatebox[origin=c]{90}{Backbone}}}}&ResNet50&85.52 M&81.71&89.54&95.09&97.46&89.32&94.23&71.79&80.73&84.48&90.49\\
\multicolumn{1}{ c|  }{}&ResNet34&25.77 M&82.77&90.22&95.06&97.45&88.68&93.84&72.58&80.99&84.77&90.62\\
\multicolumn{1}{ c|  }{}&VGG19&28.93 M&85.33&91.66&96.36&98.14&88.77&93.93&\textbf{74.70}&\textbf{83.49}&86.29&91.80\\
\multicolumn{1}{ c|  }{}&VGG16&23.62 M&\textbf{85.84}&\textbf{91.98}&\textbf{96.56}&\textbf{98.24}&\textbf{90.24}&\textbf{94.77}&74.40&83.30&\textbf{86.76}&\textbf{92.07}\\
\hline\hline
\multicolumn{2}{ c }{DeepPyram++}&28.48 M&84.83&91.54&96.30&98.11&89.48&94.34&74.64&83.34&86.31&91.83\\\specialrule{.12em}{.05em}{.05em}%
\end{tabular}
\label{tab:backbone}
\vspace{8mm}
\renewcommand{\arraystretch}{1}
\caption{Impact of different super-resolution functions on the segmentation results (IoU\% and Dice\%) of DeepPyram.}
\label{tab:ablation}
\centering
\begin{tabular}{lccccccccccc}
\specialrule{.12em}{.05em}{.05em}%
&&\multicolumn{2}{c}{Lens}&\multicolumn{2}{c}{Pupil}&\multicolumn{2}{c}{Cornea}&\multicolumn{2}{c}{Instrument}&\multicolumn{2}{c}{Overall}\\\cmidrule(lr){3-4}\cmidrule(lr){5-6}\cmidrule(lr){7-8}\cmidrule(lr){9-10}\cmidrule(lr){11-12}
Upsampling&Params&IoU(\%)&Dice(\%)&IoU(\%)&Dice(\%)&IoU(\%)&Dice(\%)&IoU(\%)&Dice(\%)&IoU(\%)&Dice(\%)\\\specialrule{.12em}{.05em}{.05em}%
Trans Conv&25.01 M&84.37&91.22&96.01&97.96&88.80&93.97&\textbf{75.16}&\textbf{83.71}&86.08&91.71\\
PixelShuffle&36.15 M&84.31&91.00&96.49&98.20&89.17&94.15&74.55&83.41&86.13&91.69\\
Bilinear&23.62 M&\textbf{85.84}&\textbf{91.98}&\textbf{96.56}&\textbf{98.24}&\textbf{90.24}&\textbf{94.77}&74.40&83.30&\textbf{86.76}&\textbf{92.07}\\\specialrule{.12em}{.05em}{.05em}%

\end{tabular}
\label{tab:upsampling}
\end{sidewaystable}

\section{Comparisons with Alternative Modules} 
Herein, we compare the effectiveness of the proposed modules with our enhanced version of the ASPP module~\cite{DeepLab} and PPM~\cite{PSPNet}. More concretely, we replace the PVF module with PPM (referring to as $Alternative^{PPM}$) and DPR module with our improved version of the ASPP module (referring to as $Alternative^{ASPP+}$). Figure~\ref{Fig-TMI: replace_module} demonstrates the architecture of DeepPyram versus the alternative networks. The difference between the modules ASPP+ and ASPP lies in the filter size of the convolutional layers. In the ASPP module, there are four parallel convolutional layers: a pixel-wise convolutional layer and three dilated convolutions with different dilation rates. In ASPP+, the pixel-wise convolution is replaced with a $3\times 3$ convolutional layer that effectively enhances the segmentation performance. Instead of three dilated convolutions in ASPP+, we use two dilated convolutions with the same dilation rates as in the DPR module. Moreover, the parallel feature-maps in the ASPP module are fused using a pixel-wise convolution, whereas the ASPP+ module adopts a kernel-size of $3\times 3$ to boost the segmentation performance.   Besides, we have removed the $P\mathcal{L}$ module and referred to it as $DeepPyram^{-P\mathcal{L}}$, to qualitatively compare its performance with DeepPyram. As illustrated in Figure~\ref{Fig-TMI: replace_module}, location ``A'' corresponds to the output of the last encoder's layer in the bottleneck. locations ``B-D'' are the outputs of three modules in the same locations of the decoder networks in DeepPyram and the three alternative networks. Figure~\ref{Fig-TMI: replace_module_vis} compares the class activation maps corresponding to these four locations in DeepPyram and alternative approaches for cornea segmentation in a representative image~\footnote{These activation maps are obtained using Score-CAM~\cite{Score-CAM} visualization approach.}. A comparison between the activation maps of DeepPyram and $DeepPyram^{-P\mathcal{L}}$ indicates how negatively removing the $P\mathcal{L}$ module affects the discrimination ability in different semantic levels. It is evident that the activation map of block ``C'' in DeepPyram is even more concrete compared to the activation map of block ``D'' (which is in a higher semantic level) in $DeepPyram^{-P\mathcal{L}}$. We can infer that the $P\mathcal{L}$ module can effectively reinforce the semantic representations in different semantic levels of the network. The effect of pixel-wise global view (PVF module) versus region-wise global view (PPM) can be inferred by comparing the activation maps of DeepPyram and $Alternative^{PPM}$. The activation maps of $Alternative^{PPM}$ are impaired and distorted in different regions, especially in the lower semantic layers. The activation maps of $Alternative^{ASPP+}$ compared to DeepPyram confirm that replacing the deformable convolutions with regular convolutions can negatively affect semantic representation in narrow regions and the object borders.

\section{Effect of Different Backbones and Nested Architecture}

We have evaluated the achievable segmentation accuracy using different backbone networks to decide which backbone performs the best, considering the trade-off between the number of trainable parameters and the Dice percentage. As listed in Table~\ref{tab:backbone}, the experimental results show that the higher-depth networks such as ResNet50 cannot improve the segmentation accuracy. In contrast, VGG16 with the fewest number of trainable parameters has achieved the best segmentation performance. Moreover, VGG19, having around 5.3M parameters more than VGG16, performs just slightly better than the baseline backbone in instrument segmentation. In Table~\ref{tab:backbone}, DeepPyram++ is the nested version of DeepPyram \footnote{UNet++ uses the encoder-decoder architecture of U-Net as baseline, and adds additional convolutional layers between different encoder's and decoder's layers to form a nested multi-depth architecture. In DeepPyram++, we replace the the encoder-decoder baseline of UNet++ (which is U-Net) with our proposed DeepPyram network to see if these additional layers and connections can improve the segmentation performance of DeepPyram.} (with the same connections as in UNet++). This nested architecture shows around a one percent drop in IoU percentage on average. 

\section{Effect of Different Super-resolution Functions}

Table~\ref{tab:upsampling} compares the effect of three different super-resolution functions on the segmentation performance, including transposed convolution, Pixel-Shuffle~\cite{PixelShuffle}, and bilinear upsampling. Overall, the network with bilinear upsampling function with the fewest parameters has achieved the best performance among the networks with different upsampling functions. Besides, the results reveal that the bilinear upsampling function has the best performance in segmenting all relevant objects except for the instruments.

\section{Conclusion}
\label{sec-TMI: Conclusion}
In recent years, considerable attention has been devoted to computerized surgical workflow analysis for various applications such as action recognition, irregularity detection, objective skill assessment, and so forth. A reliable relevant-instance-segmentation approach is a prerequisite for a majority of these applications. In this chapter, we have proposed a novel network architecture for semantic segmentation in cataract surgery videos. The proposed architecture takes advantage of three modules, namely ``Pyramid View Fusion'', ``Deformable Pyramid Reception'', and ``Pyramid Loss'',  to simultaneously deal with different challenges. These challenges include: (i) geometric transformations such as scale variation and deformability, (ii) blur degradation and blunt edges, and (iii) transparency, and texture and color variation. Experimental results have shown the effectiveness of the proposed network architecture (DeepPyram) in retrieving the object information in all mentioned situations. DeepPyram stands in the first position for cornea, pupil, lens, and instrument segmentation compared to all rival approaches. The proposed architecture can also be adopted for various other medical image segmentation and general semantic segmentation problems.

\chapter{Self-Supervised Pretraining for Semantic Segmentation \label{Chapter:SSL}}

\chapterintro{
	The performance of a supervised-deep-learning approach is heavily reliant on annotations. This key demand is hard to meet, especially in the case of semantic segmentation. In the medical domain, where domain knowledge is a prerequisite, providing adequate annotations is even more expensive and challenging. This chapter proposes a novel self-supervised learning approach based on contrastive learning for semantic segmentation in surgical videos. In particular, the proposed method selects and augments a pair of temporally close frames with moderate-to-harsh frequency-based and region-based augmenters. The contrastive loss encourages a close representation for the original and augmented versions while forcing distant representations between the temporally close frames. We exploit a set of progressive discrimination impediments to avoid network under-fitting due to a complex learning task in the beginning and learning suspension due to task easiness in later epochs.
}

{
	\singlespacing This chapter is an adapted version of:
	
	``Ghamsarian, N., Taschwer, M., Putzgruber-Adamitsch, D., Sarny,
S., El-Shabrawi, Y., and Schoeffmann, K. Self-Supervised Progressive Representation Learning For Semantic Segmentation in Surgical Videos. ''
}

\section{Introduction}
\label{chapter-AAAI: Introduction}

A major contributing factor in the performance of supervised deep-learning approaches is a large labeled dataset. This requirement cannot usually be met in the case of medical image analysis since it requires expert knowledge and is consequently costly. This dearth of annotations is more severe in medical image segmentation since pixel-wise annotation is a time-extensive procedure. On the other hand, the common pre-trained backbones cannot provide optimal initialization for medical image analysis due to the large gap between the statistical distributions and semantic characteristics of the natural and medical images. During the past few years, many techniques have been developed to alleviate annotations' requirements and negate the dearth of annotations. These techniques include but are not limited to (i) data augmentation approaches such as affine and random transformations, mixup, and generative adversarial networks, (ii) semi-supervised learning, and (iii) self-supervised learning. Self-supervised learning is regarded as an effective technique to mitigate the negative impact of inadequate annotations. 

Self-supervised learning refers to the methods employed to encourage the network to learn semantic features from unlabeled data. This objective is met in two ways: (1) using the inherent labels in data such as spatial and temporal characteristics, and (2) forcing the network to solve a game with the input dataset. 

Despite the success of state-of-the-art approaches in alleviating the requirement for annotations, we argue that one critical aspect has not yet been explored. That is, how to stimulate a human-like semantic feature extraction through self-supervised learning. In this chapter, we propose a novel self-supervised learning framework for surgical videos. We aim to encourage learning the object's shape and configurations regardless of independent local characteristics and encapsulated rich statistics to bridge the gap between human and network interpretation. In particular, we propose:
\begin{enumerate}
    \item A novel progressive-learning strategy by providing easy-to-hard learning tasks for the network,
    \item A novel global contrastive strategy to encourage feature extraction and semantic feature deduction analogous to the human visual system (HVS),
    \item Two novel region-based contrastive strategies to reinforce the learning of local representations being advantageous for the segmentation tasks.
\end{enumerate}

We argue that the proposed self-supervised learning approach can prevent capturing unnecessary local information and consequently yield better generalization capability. We compare the achievable segmentation performance with and without pre-training with the proposed approach using different amounts of annotations. 
Finally, we assess the impact of two modules in self-supervised learning compared to the baseline network.

\section{Related Work}
\label{chapter-AAAI: Related Work}
The self-supervised learning approaches can be categorized into (i) pretext-task-based approaches and (ii) contrastive learning approaches. 

\subsection{Pretext Task}
In the pretext approaches, we withhold some visual characteristics of the input data and encourage the network to predict them. This objective can be achieved in two ways: regression or classification. The regression-based pretext tasks include Context restoration through context-based pixel prediction or inpainting~\cite{FLbI}, colorization~\cite{Colorization}, cross-channel prediction using split-brain autoencoders~\cite{Zhang_2017_CVPR}, and motion segmentation~\cite{MotionSegmentation}. 
Orientation degree prediction~\cite{PIR}, multi-task learning~\cite{MTSSVL,ConRe,SSMIA}, spatial position prediction~\cite{CMR,BPR}, Discriminating between surrogate classes~\cite{DUFL}, and transformation type prediction~\cite{SSSFL} are some examples of classification-based pretext tasks. Another work~\cite{UVRLCP} proposed to encourage representation learning through predicting the relative position of neighboring patches. 
Solving a Jigsaw puzzle~\cite{Jigsaw} and Rubik cube~\cite{Rubik,RubikCubePlus}, and non-parametric instance discrimination~\cite{Instance-Discrimination} are other variants of representation learning through classification-based pretext tasks. 

Regarding self-supervised learning from video sequences, many approaches have been recently proposed based on video frame ordering~\cite{URLSS, Shuffle-and-Learn}, video clip order prediction~\cite{Clip-Order}, predicting arrow of time~\cite{Arrow-of-Time}, wrong order prediction~\cite{Odd-One-Out}, object tracking~\cite{Objects-Move}, and space-time puzzle solving~\cite{Video-Jigsaw, Space-Time-Cubic-Puzzles}.

\subsection{Contrastive Learning}
Contrastive learning can be broadly described as the methods employed to reinforce a semantically close representation for similar pairs and distant representation for dissimilar pairs~\cite{CL-GLF}. This goal can be met using mutual information maximization~\cite{InfoMax,LRMMI}, transformation-invariant instance representation learning~\cite{SimCLR, SSLPIR}, momentum contrast~\cite{Momentum-Contrast}, multiview coding~\cite{Multiview},  and contrastive predictive coding~\cite{Contrastive-Predictive}.

Contrastive learning from videos is performed via tracked patches versus random patches from videos~\cite{Tracked-vs-Random}, temporal cycle-consistency~\cite{Temporal-Cycle-Consistency}, and multiple viewpoints correspondance~\cite{Multiple-Viewpoints}.

\subsection{Shortcomings of State-of-The-Art Approaches}
Despite the outstanding performance of state-of-the-art self-supervised learning approaches, scant attention has been devoted to self-supervised learning for semantic segmentation in surgical videos.
Some approaches exploit temporal-coherence in surgical videos with contrastive and ranking loss~\cite{TCBSSL}. However, since surgical videos usually contain many repetitive actions, the sequence sorting task can confuse the network and avoid semantic representation learning.

Besides, the general self-supervised learning approaches are not capable of reinforcing region-wise semantic interpretation. For instance, rotation degrees in surgical videos can be learned through simple concepts such as instrument orientation. Hence, rotation prediction seems to be an effortless task for the network to learn. Moreover, rotation prediction is not a suitable method for learning the representation of circular, deformable, and orientable objects. In surgical videos, however, we have many relevant deformable and circular objects. Examples of cataract surgery are: (i) the implanted artificial lens as a deformable object, (ii) pupil, cornea, and iris as circular objects.

Moreover, learning to classify particular transformations such as inpainting and rotation might encourage the network to look for easy clues instead of learning shape-wise representations. Indeed, an optimal self-supervised learning approach for semantic segmentation should encourage the network to not fire on the local regions independently but on cross-region dependencies.

\section{Methodology}
\label{chapter-AAAI: Methodology}

Inspired by the recent advancements in self-supervised contrastive learning, we propose a self-supervised progressive representation learning for semantic segmentation in surgical videos termed VidSeg-SSL. The proposed framework encourages the network to learn the semantic features based on a contrastive loss derived from the latent representations of close frames and their augmented versions. The VidSeg-SSL framework can be flexibly used for every neural network architecture and every video dataset.
\subsection{Notations}
Everywhere in this chapter:
\begin{itemize}
    \item $||\cdot||$ denotes the Euclidean norm. 
    \item $|\cdot|$ denotes the absolute value function.
    \item $\# A$ for each arbitrary matrix or vector $A$, denotes the cardinality of $A$.
    \item We define a uniformly random selection function   $\mathcal{R}and(K, [x_1, x_2])$ that outputs an array of $K$ non-repetitive integer numbers between $x_1$ and $x_2$. Besides, $A[k]$ for each $A$ and $k$ refers to the $k$th element of $A$.
\end{itemize}

\subsection{Self-Supervised Learning Algorithm}
\begin{figure}[!tb]
    \centering
    \includegraphics[width=1\columnwidth]{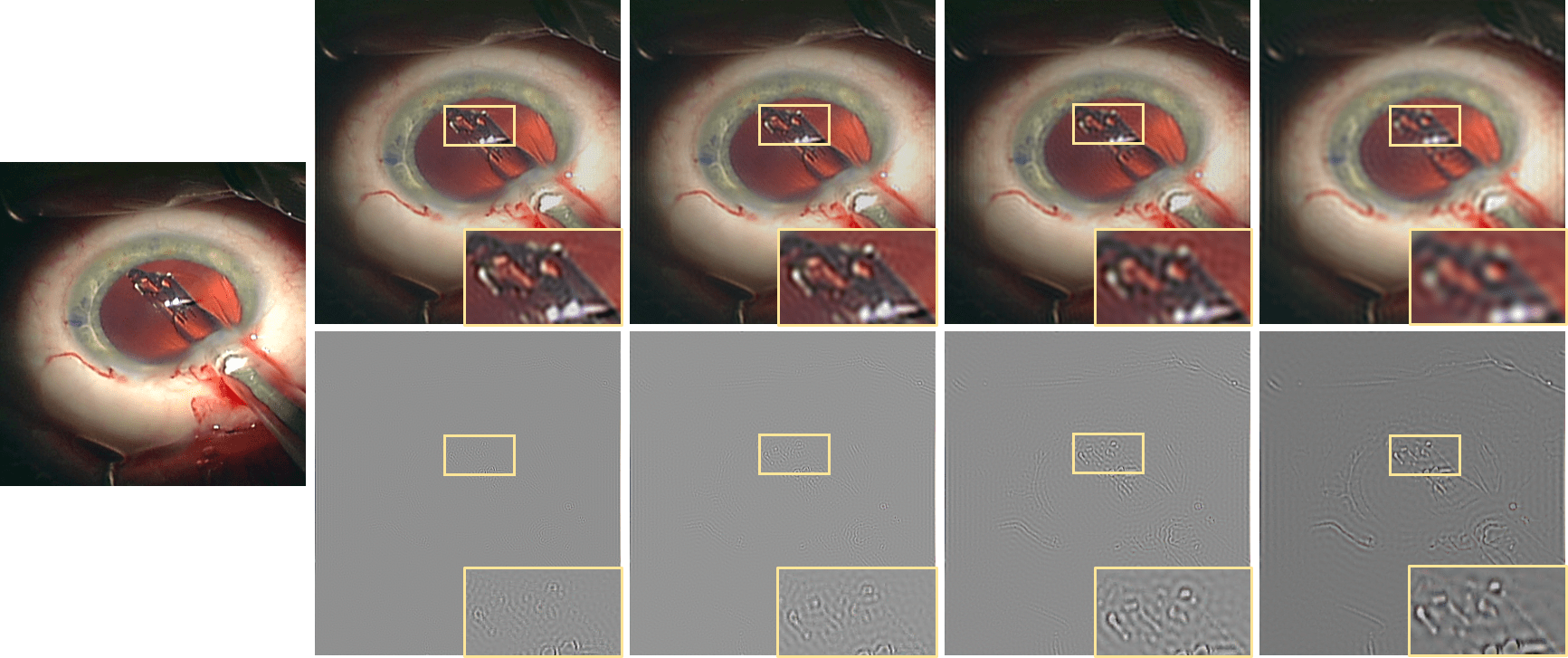}
    \caption{Effect of removing high-frequency components on the visual characteristics of images~\cite{HFComp}.}
    \label{fig: HF-Removal}
\end{figure}

As the first step, we need to create the training set for self-supervised learning. For this purpose, we extract non-overlapping ten-second clips with the temporal resolution of 10 frames per second. Accordingly, each video clip $V_i$ in our dataset has the dimension of $100\times 3 \times W\times H$.

Supposing we are given a video-clip dataset $\mathcal{D} =\{V_1, V_2, ... , V_{\#\mathcal{D}}\}$ with $V_i\in \mathbb{R}^{100\times 3 \times W\times H}$, a stochastic function $\mathcal{F}_t$ is employed to randomly select a training clip $V_t$ as follows:

\begin{equation} 
V_t = \mathcal{D}[\mathcal{F}_t], \mathcal{F}_t = \mathcal{R}and(1,[1, \#\mathcal{D}]) 
\end{equation}

We formulate the selected video-clip as a sequence of its frames as follows:
\begin{equation} 
V_t = (\mathrm{I}_1, \mathrm{I}_2, ... , \mathrm{I}_{100})
\end{equation} 
A second stochastic function $\mathcal{F}_e$ is used to randomly select an exemplary frame $\mathrm{I}_e$ from 100 possible frames in the selected video as follows:

\begin{equation} 
\mathrm{I}_e=V_t[\mathcal{F}_e], \mathcal{F}_e= \mathcal{R}and(1,[1,100]) 
\label{eq: Ie}
\end{equation}

Depending on the pre-determined frame-distance threshold $\mathrm{T}_{dist}$ and the index of exemplary frame $\mathrm{ind}(\mathrm{I}_e)$, the third stochastic function $\mathcal{F}_a$ selects an adversarial frame $\mathrm{I}_a$ as follows:

\begin{equation}
\begin{split} 
\mathrm{I}_a = V_t[\mathcal{F}_a], \mathcal{F}_a = \mathcal{R}and(1,[ max(0,\mathrm{ind}(\mathrm{I}_e)-\mathrm{T}_{dist}),  min(50,\mathrm{ind}(\mathrm{I}_e)+\mathrm{T}_{dist})] 
\end{split} 
\label{eq: Ia}
\end{equation}

In order to be sure that the sampled pair of frames $\mathrm{I}_e$ and $\mathrm{I}_a$ have enough visual discriminative features, we subtract these frames ($\mathrm{I}_{diff} = |\mathrm{I}_e - \mathrm{I}_e|$) and search for the elements with large absolute values in the difference matrix $\mathrm{I}_{diff}$ based on a pre-determined pixel-wise difference threshold $\mathrm{T}_{diff}$ as follows:

\begin{equation}
C(\mathrm{I}_{diff}) = \{ x \in \mathrm{I}_{diff} : x > \mathrm{T}_{diff} \}
\label{eq: frame-diff}
\end{equation}

We check the suitability of the two frames by counting the number of large values in the difference frame as follows:

\begin{equation}
    S(\mathrm{I}_{diff})= 
\begin{cases}
    1,& \text{if } \frac{\#C(\mathrm{I}_{diff})}{\#\mathrm{I}_{diff}}\geq 0.1\\
    0,              & \text{otherwise}
\end{cases}
\label{eq: frame-sutability}
\end{equation}

We pass the frames to the next step if $S(\mathrm{I}_{diff})=1$. Otherwise, we through the sampled frames away, choose an alternative clip, and repeat sampling until the sampled pair satisfy the mentioned condition. Afterwards, exemplary frame and adversarial frame undergo gradual transformations to produce contrastive pairs with different strategies:

\begin{equation}
    \begin{split}
      \hat{\mathrm{I}_e} = &\mathcal{T}(\mathrm{I}_e)\\
      \hat{\mathrm{I}_a} = &\mathcal{T}(\mathrm{I}_a)
    \end{split}
    \label{eq: transform}
\end{equation}

We explain our proposed novel strategies in the following three subsections.

In contrastive learning, the objective is to increase (i) the similarity between the representation of the exemplar image $\mathrm{I}_e$ and its transformed version $\hat{\mathrm{I}_e}$, and (ii) cross dissimilarities between all versions of exemplar and adversarial frames. To this end, we use a contrastive loss based on cosine similarity between the representations of exemplary and adversarial frames and their transformations as follows~\cite{SFCLVR}.

\begin{figure}[!tb]
    \centering
    \includegraphics[width=0.4\columnwidth]{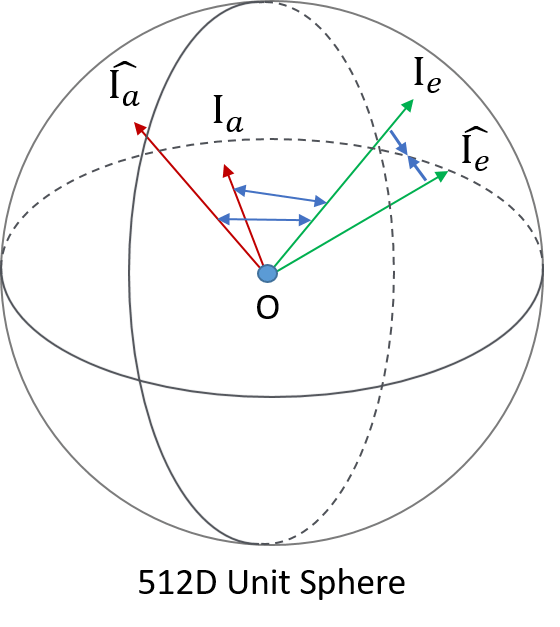}
    \caption{The proposed contrastive learning strategy.}
    \label{fig: 512D-Unit-Sphere}
\end{figure}

\begin{equation} 
\begin{split}
    \mathcal{L}(E(\mathrm{I}_e), E(\mathrm{I}_a)) = -\log \frac{exp(sim(E(\mathrm{I}_e),E(\hat{\mathrm{I}_e}))/\tau)}{\sum_{\mathrm{I}_1, \mathrm{I}_2} exp(sim(E(\mathrm{I}_1),E(\mathrm{I}_2))/\tau)}
\end{split}
\label{eq: contrastive-loss}
\end{equation}    

In \eqref{eq: contrastive-loss}, $\mathrm{I}_1 \in \{\mathrm{I}_e,\hat{\mathrm{I}_e}\}$, ${\mathrm{I}_2} \in \{\mathrm{I}_a,\hat{\mathrm{I}_a}\}$, and $\tau$ is the pre-determined scaling parameter. The function $E(\cdot)$ refers to the representation of the encoder network $e(\cdot)$ appended by a projection head $p(\cdot)$ as $E(x) = p(e(x))$. The projection head is used to provide a lower-dimensionality representation as well as allowing the encoder network $e(\cdot)$ to maintain transformation-related features~\cite{SFCLVR}. Beside, $sim(x,y)$ refers to the cosine similarity between the vectors $x$ and $y$ and is formulated as $sim(x,y) = (x^Ty)/max(\parallel x\parallel \parallel y\parallel, \epsilon)$ where $\epsilon$ is a very small value to avoid division by zero.

\begin{algorithm}[tb]
\caption{VidSeg-SSL's learning algorithm}
\label{alg:algorithm}
\textbf{Input}: batch size $N$, Network $E(\cdot)$, epoch-adaptive transformation $\mathcal{T}$, Video clip dataset $\mathcal{D}$.\\
\textbf{Parameter}: frame-distance threshold $\mathrm{T}_{dist}$, pixel-wise difference threshold $\mathrm{T}_{diff}$, \\
\textbf{Output}: pre-trained encoder $e(\cdot)$.
\begin{algorithmic}[1] 
\STATE Let $t=0$.
\FOR{sampled video-clip batch}
\FOR{training clip $V_t$ in the video-clip batch $b_V$ ($b_V = \{V_i|i\in [1,\# \mathcal{D}\} , \#b_V = N$)}
\STATE Select the exemplary frame $\mathrm{I}_e$ using eq.~\eqref{eq: Ie}.
\STATE Select the adversarial frame $\mathrm{I}_a$ using eq.~\eqref{eq: Ia}.
\STATE Compute the frame differences $C(\mathrm{I}_{diff})$ using eq.~\eqref{eq: frame-diff}.
\IF {$S(\mathrm{I}_{diff})=0$ (eq.~\eqref{eq: frame-sutability})}
\STATE Remove $V_t$ from the video-clip batch $b_V$ and add an alternative video clip to the batch.
\STATE Repeat the frame selection step (go to line 3).
\ENDIF
\STATE Transform $\mathrm{I}_e$ and $\mathrm{I}_a$ to obtain $\hat{\mathrm{I}_e}$ and $\hat{\mathrm{I}_a}$ using eq.~\eqref{eq: transform}.
\STATE Compute the latent representations of the exemplary and adversarial frames and their transformations using feed-forward:\\
$E(\mathrm{I}_e)$, $E(\mathrm{I}_a)$, $E(\hat{\mathrm{I}_e)}$, $E(\hat{\mathrm{I}_a})$.
\STATE Compute the contrastive loss for the current samples using eq.~\eqref{eq: contrastive-loss}.
\STATE Set $l[t] = \mathcal{L}(E(\mathrm{I}_e), E(\mathrm{I}_a))$.
\ENDFOR
\STATE Compute the overall loss for the batch:\\
$L=\frac{1}{N}\sum_{t=1}^N l[t]$.
\STATE Update $E(\cdot)$ including the encoder $e(\cdot)$ and projection the head $p(\cdot)$ using the overall batch loss $L$.
\ENDFOR
\STATE Remove the projection head from $E(\cdot)$ to obtain the encoder network $e(\cdot)$.
\STATE \textbf{return} the pretrained network $e(\cdot)$.
\end{algorithmic}
\end{algorithm}

The threshold $\mathrm{T}_a$ is reset to its highest value at the beginning of each training strategy and is gradually decreases to increase the task difficulty.

\subsection{Strategy 1: contrastive learning based on high-frequency component removal}
In this stage, we aim to encourage the network to output a very close representation for each examplar frame $\mathrm{I}_e$ and its transformed version $\hat{\mathrm{I}_e}$ obtained via removing the high-frequency components. In contrast, we reinforce distant representations for the examplar and adversarial frames. The removal of high-frequency components does not affect the semantic information perceived by the human visual system. Accordingly, we reinforce a neural network deduction of semantic information analogous to the human visual system.

The high-frequency-removal transformation function $\mathcal{T}_{HF}^{r}$ uses a pre-determined parameter $r$ to remove the high-frequency components outside of the circle with radius $r$ at the center of the fast Fourier transform (FFT) version of the exemplary and adversarial frames. Then we obtain:

\begin{align*} 
\hat{\mathrm{I}_e} = \mathcal{T}_{HF}^{r}(\mathrm{I}_e)\\
\hat{\mathrm{I}_a} = \mathcal{T}_{HF}^{r}(\mathrm{I}_a)
\end{align*}

The radial is scheduled to gradually decrease in order to provide a more complicated task for the network. However, we consider a pre-determined minimum radial to avoid the removal of semantically relevant components.

\paragraph{\textit{Discussion: }} With this strategy, we aim to address the misalignment between the semantic interpretation of CNNs and the human visual system (HVS). While HVS determines the label purely based on semantic information, the convolutional neural networks unintentionally learn joint high-frequency and semantic correlations between the visual signals and their corresponding labels.  
\subsection{Strategy 2: contrastive learning based on block-wise augmentation}
In this stage, we further change the frames transformed by Strategy 1 to increase the difficulty of the task and as a result, encourage the network to extract more discriminative information from the video frames. We use a set of non-geometric transformations (such as Gaussian and motion blur, brightness, and contrast transformations) for block-wise augmentation. More concretely, the examplar and adversarial frames are split into $k\times k$ blocks. Supposing $k=4$, the frames after high-frequency removal can be shown as the sequence of their blocks as follows:

\begin{align*} 
\mathcal{T}_{HF}^{r}(\mathrm{I}_e) = (b_{e1}, b_{e2}, ..., b_{e15}, b_{e16})\\
\mathcal{T}_{HF}^{r}(\mathrm{I}_a) = (b_{a1}, b_{a2}, ..., b_{a15}, b_{a16})
\end{align*}

Each block is distinctively augmented using two stochastic functions $\mathcal{F}_{b}$ and $\mathcal{F}_{Aug}$. The function $\mathcal{F}_{b}$ randomly selects $n$ blocks in the current frame to be augmented. For instance, supposing $n=2$,  $\mathcal{F}_{b}(\mathrm{I}_e) =\{b_{e1}, b_{e15}\}$, and $\mathcal{F}_{b}(\mathrm{I}_a) =\{b_{e2}, b_{e16}\}$, the frames $\hat{\mathrm{I}_e}$ and $\hat{\mathrm{I}_a}$ can be shown as:

\begin{align*} 
\hat{\mathrm{I}_e} = \mathcal{F}_{Aug}(\mathcal{T}_{HF}^{r}(\mathrm{I}_e), \mathcal{F}_{b}(\mathrm{I}_e)) = (\hat{b_{e1}}, b_{e2}, ..., \hat{b_{e15}}, b_{e16})\\
\hat{\mathrm{I}_a} = \mathcal{F}_{Aug}(\mathcal{T}_{HF}^{r}(\mathrm{I}_a), \mathcal{F}_{b}(\mathrm{I}_a)) = (b_{a1}, \hat{b_{a2}}, ..., b_{a15}, \hat{b_{a16}})
\end{align*}

The function $\mathcal{F}_{Aug}(a,b)$ transforms the input frame $a$ by augmenting the blocks in the array $b$ using a random selection of augmentation for each block independently of the transformations applied to other blocks. The decomposition size ($k$) progressively decreases, and the number of augmented blocks $n$  progressively increases during training to provide a more complex task for the network.

\begin{figure}[!tb]
    \centering
    \includegraphics[width=1\columnwidth]{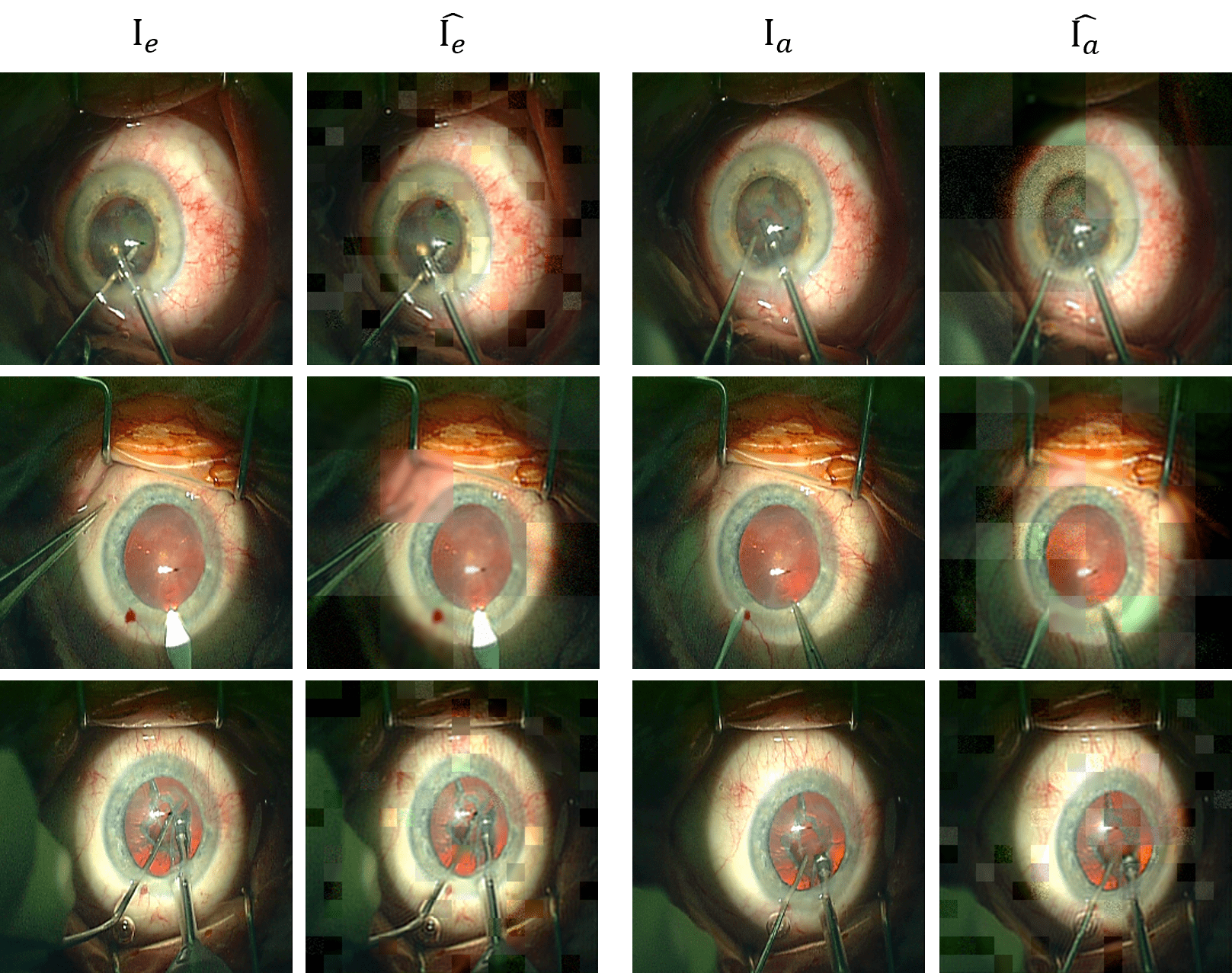}
    \caption{The training quadruple images generated with Strategy 2.}
    \label{fig: Strategy2}
\end{figure}

\paragraph{\textit{Discussion: }} As shown in Figure~\ref{fig: Strategy2}, block-wise augmentation produces many irrelevant edges and thus encourages the network to learn the relevant edges to maximize the representation agreements between the positive pairs.

\subsection{Strategy 3 : contrastive learning based on deformable-block-wise augmentation}

\begin{figure}[!tb]
    \centering
    \includegraphics[width=1\columnwidth]{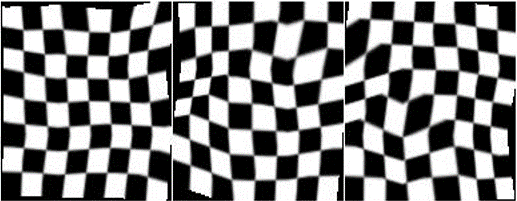}
    \caption{Deformation of blocks' borders via piece-wise affine transformation\\ (picture from https://imgaug.readthedocs.io/).}
    \label{fig: piece-wise-affine}
\end{figure}

\begin{figure}[!thb]
    \centering
    \includegraphics[width=1\columnwidth]{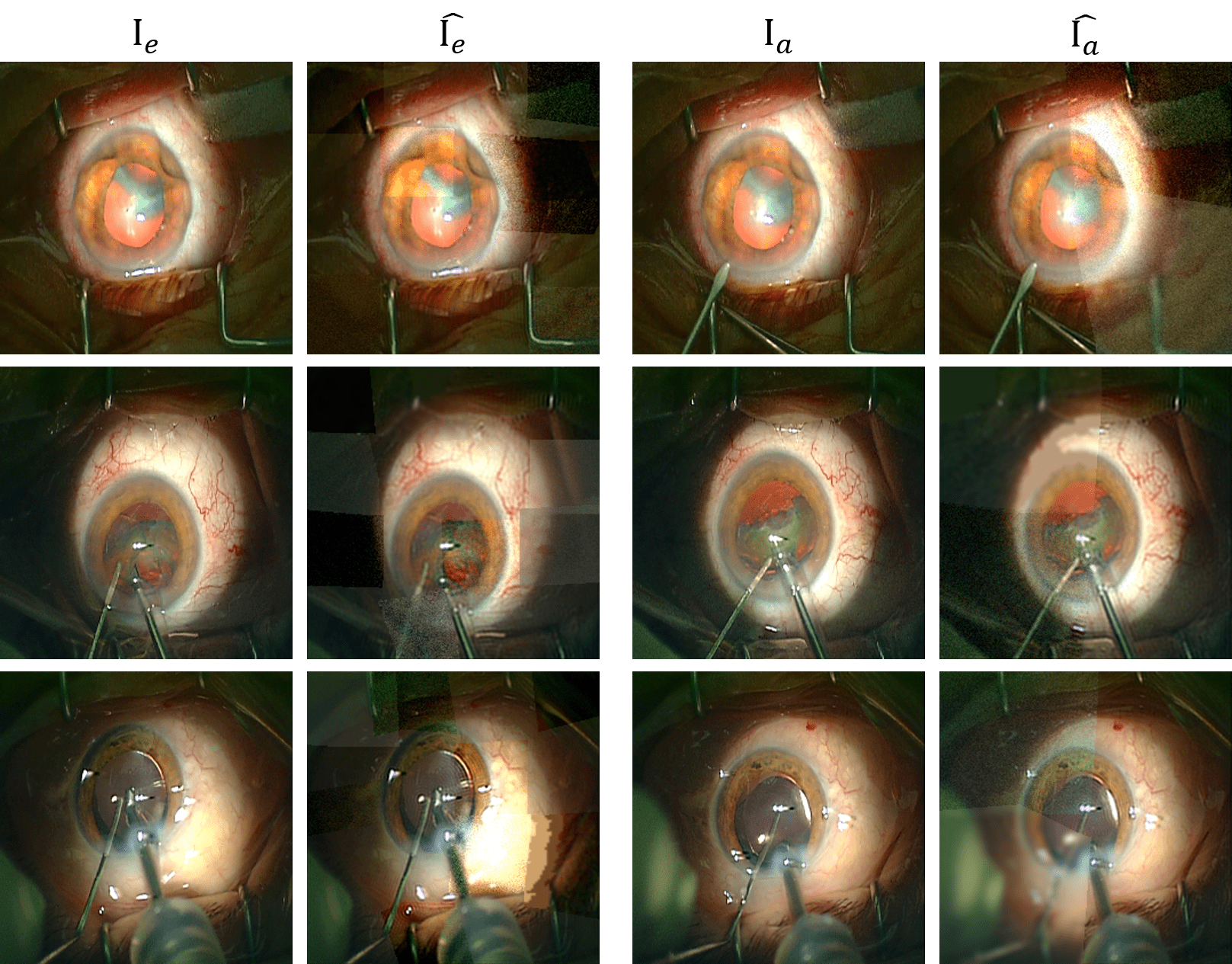}
    \caption{The training quadruple images generated with Strategy 3.}
    \label{fig: Strategy3}
\end{figure}

Block-wise augmentation leads to straight edges in the images. As a result, a network trained on such augmented images may learn to disregard the straight edges after some epochs, leading to stopping learning the underlying semantic representation. Hence, we propose to impede the contrastive learning task via augmenting deformable blocks. To produce such deformations, we take advantage of piece-wise affine transformations (Figure~\ref{fig: piece-wise-affine}). After splitting the image into $k\times k$ blocks and selecting the blocks to be augmented, we form a mask image per block with the white pixels corresponding to the selected 
block's pixels. We then apply the affine transformation on the mask to obtain a deformed block. By multiplying the obtained mask to the original image, we obtain the pixels corresponding to the deformable block that should be augmented. Similar to Strategy 2, we apply a random selection of pre-determined (non-geometric) transformations on the selected images. The augmented image is finally obtained by replacing the pixels in the original image with non-zero pixels in the augmented masks for all selected blocks. As shown in Figure~\ref{fig: Strategy3}, deformable-block-wise augmentation leads to some irrelevant curved edges that are more difficult than the straight lines in strategy 2 to distinguish from the relevant edges.

\section{Experimental Settings}
\label{chapter-AAAI: Experimental Settings}
In this chapter, we evaluate and compare the performance of representation learning frameworks using four evaluation setups as follows:
\begin{itemize}
    \item \textbf{Setup 1.} To evaluate the learned representations for classification, many approaches freeze the self-supervised pre-trained encoder, train a linear classifier added on top of the encoder network, and evaluate the whole network ~\cite{RSSV,LRMMI,SimCLR}. We use the same approach in this setup by adding the decoder network on top of the frozen pre-trained encoder, training the whole network for semantic segmentation. We also use this approach as the baseline for other setups.
    \item \textbf{Setup 2.} In this setup, we do not freeze the encoder network and fine-tune the whole network. The results reveal how the pre-trained weights affect the speed and performance of learning.
    \item \textbf{Setup 3.} We adopt a semi-supervised learning approach for semantic segmentation to see which of the self-supervised learning approaches leads to less outlier and more centralized semantic segmentation interpretation. In particular, we want to figure out which SSL approach can prevent the network from biasing to unexpected features within the dataset.
    \item \textbf{Setup 4.} As the main reason behind self-supervised learning, we assess the performance of a pre-trained network with supervised learning on a dataset such as ImageNet with the pre-trained networks using self-supervised learning. More concretely, in this setup, we evaluate the percentage of annotations required after self-supervised learning to achieve the same semantic segmentation performance.
    \item \textbf{Setup 5.} As an ablation study, we assess the effect of three modules based on average pooling on SSL performance.
\end{itemize}

\section{Conclusion}
\label{chapter-AAAI: Conclusion}
Considering that supervised learning based on manual annotation is time-consuming, expensive, prone to human error, and subject to distribution shift, self-supervised representation learning has witnessed considerable attention in recent years.
The proposed framework does not impose any constraints on the choice of the neural network architecture and video dataset. The proposed representation learning approach can initialize the networks for semantic segmentation tasks and other various downstream tasks.

\chapter{Concluding Remarks}
\label{chap:thesisconclusion}

\chapterintro{
	This chapter concludes the contributions of this thesis and discusses the open-ended research questions regarding deep-learning-assisted analysis of cataract surgery videos.
}

The recent technological advancements in robotic and minimally invasive surgeries have enabled recording and collecting surgical videos. Analyzing such videos can provide valuable insights to enhance post-surgical care, speed up surgical training, discover unexplored symptoms of surgical complications, and provide real-time guidance for OR planning. Indeed, computerized surgical video analysis can offer solutions to numerous demands of the modern operating rooms.

This thesis investigates novel frameworks and neural network architectures to address the significant research questions about the computerized analysis of cataract surgery videos. In particular, this thesis focuses on four critical subjects in computerized analysis of cataract surgery videos: (1) phase recognition, (2) semantic segmentation, (3) adaptive compression, and (4) irregularity detection.

\section{Contributions of This Dissertation}

\paragraph{\textit{Providing a powerful basis for high-level tasks. }}Phase recognition and semantic segmentation are two integral parts of many surgical video analysis approaches. Hence, accurate semantic segmentation and phase recognition can play a critical role in subsequent downstream tasks such as objective skill assessment, relevance detection, irregularity detection, adaptive compression, etc. To fulfill these requirements, this thesis proposes (i) novel frameworks, recurrent convolutional networks, and training strategies to improve phase recognition accuracy upon state-of-the-art approaches, and (ii) several novel convolutional modules to deal with diverse semantic segmentation challenges in cataract surgery videos.

\paragraph{\textit{Relevance-Based Compression. }}High visual quality of relevant content is a key factor in the usefulness of cataract surgery videos. This pre-condition notwithstanding, compression is a necessity for real-time streaming and efficient storage of cataract surgery videos. To accommodate both demands, we propose relevance-based compression of cataract surgery videos considering different scenarios for the relevant content.

\paragraph{\textit{Irregularity Detection. }}Deep neural networks are powerful means for performing large-scale evaluations to detect the symptoms and reason intra-operative irregularities and post-operative complications in surgical videos. We employ the capacity of deep neural networks and propose the first framework for automatic lens irregularity detection in cataract surgery videos.

\paragraph{\textit{Self-supervised pre-training. }}To alleviate the requirement for manual annotations, we propose novel self-supervised learning strategies. These strategies especially focus on reinforcing high-level semantic interpretation of raw video frames being determinative for semantic segmentation.

\section{Future Work}
Despite the recent advancements in computerized surgical video analysis, two critical aspects are mainly unexplored: (1) facilitating the use of 3D CNNs via self-supervised pre-training and (2) enhancing the generalization performance in semantic segmentation via domain adaptation.

\subsection{Self-Supervised Pre-Training of 3D CNNs}
Although recurrent CNNs have shown superior performance in phase recognition and relevance detection, they may not present a satisfactory performance in action recognition and skill assessment. This is due to the fact that actions involve intertwined spatio-temporal features, the property that recurrent CNNs cannot efficiently capture. On the other hand, action recognition and skill assessment via relevant object segmentation and motion trajectories appear to be suboptimal since (i) providing the supervisory signal for semantic segmentation is more time-consuming and expensive compared to action annotation, and (ii) action recognition and skill assessment using a sequence of steps may decrease the time and computation efficiency.

Three-dimensional CNNs can effectively capture joint spatio-temporal features associated with a particular action or skill level. However, employing 3D CNNs involves its specific challenges. Unlike 2D CNNs, for which many large-scale datasets for pre-training exist (examples are ImageNet~\cite{ImageNet}, COCO~\cite{COCO}, and Cityscapes~\cite{Cityscapes}), no large-scale dataset for pre-training the 3D CNNs exists. As a result, training 3D CNNs involves starting from scratch with random parameter initialization. Comparing to a pre-trained network, starting from random weights requires substantially more supervisory signals, imposing higher costs and more burden for further annotations.

As mentioned in chapter~\ref{Chapter:SSL}, self-supervised learning from raw video frames can provide optimal initial states for the neural networks and negate the lack of large-scale supervisory signal. Accordingly, using domain-specific (cataract surgery features) and problem-specific (action recognition or skill assessment) strategies for self-supervised learning can make ground on exploiting 3D CNNs.

\subsection{Domain Adaptation for Semantic Segmentation}
The rapid technological advancements have resulted in continuous changes in image and video capturing tools, lighting conditions, and compression standards that in turn lead to large distribution shifts from previous to new datasets. This distribution shift exists not only between different dataset generations but also between the contemporary datasets captured with different cameras and in different conditions. Such domain mismatch between the datasets cripples the performance of models trained on one dataset when being tested on the other datasets.

Domain mismatch necessitates unjustifiable human effort and cost for new annotations per dataset, a condition which is very hard to meet, especially in the case of pixel-level annotation for semantic segmentation. Domain adaptation suggests techniques to cut down or bypass the requirement for annotations in new domains. In the last couple of years, many attentions have been focused on domain adaptation~\cite{BLDASS-2019,ASIAD-2019,LTIR-2020,URMA-2021,CSFU-2021,MSDA-2021}. However, domain adaptation for semantic segmentation in cataract surgery videos taking advantage of domain-specific knowledge (unique characteristics of cataract surgery videos) is an important subject that is negated.

\cleardoublepage
\phantomsection
\addcontentsline{toc}{chapter}{Bibliography}
{
	\singlespacing
	\bibliographystyle{acm}
	\bibliography{bibtex.bib}
}

\end{document}